\documentclass[fleqn,10pt]{wlscirep}
\usepackage[export]{adjustbox}%
\usepackage{amsfonts}       %
\usepackage{amsmath}
\usepackage{amsthm}
\usepackage{bm}
\usepackage{caption}
\usepackage{comment}
\usepackage[T1]{fontenc}
\usepackage[utf8]{inputenc}
\usepackage{hyperref}
\usepackage{multirow}
\usepackage[square,sort,comma,numbers]{natbib}
\usepackage{placeins}
\usepackage{rotating}
\usepackage{tikz}
\usepackage{skak}
\usepackage{svg}
\usepackage{subcaption}

\usetikzlibrary{arrows,arrows.meta,decorations.shapes,decorations.pathreplacing,fit,positioning,trees,quotes}

\def\zerobf{\mathbf{0}}
\def\onebf{\mathbf{1}}
\def\c{\mathbf{c}}
\def\f{\mathbf{f}}
\def\p{\mathbf{p}}

\def\w{\mathbf{w}}

\def\z{\mathbf{z}}

\def\F{\mathbf{F}}

\def\Z{\mathbf{Z}}
\def\Ebb{\mathbb{E}}
\def\Rbb{\mathbb{R}}
\def\btheta{\bm{\theta}}
\def\bOmega{\bm{\Omega}}

\title{Acquisition of Chess Knowledge in AlphaZero}

\author[1,+]{Thomas McGrath}
\author[2,+]{Andrei Kapishnikov}
\author[1]{Nenad Toma\v{s}ev}
\author[2]{Adam Pearce}
\author[1]{Demis Hassabis}
\author[2]{Been Kim}
\author[1]{Ulrich Paquet}
\author[3]{Vladimir Kramnik}
\affil[1]{DeepMind}
\affil[2]{Google Brain}
\affil[3]{World Chess Champion, 2000--2007\footnote{Classical World Chess Champion (2000--2006); FIDE and Undisputed World Chess Champion (2006--2007).}}

\affil[+]{these authors contributed equally to this work}

\begin{abstract}
What is learned by sophisticated neural network agents such as AlphaZero?
This question is of both scientific and practical interest. If the representations of strong neural networks bear no resemblance to human concepts, our ability to understand faithful explanations of their decisions will be restricted, ultimately limiting what we can achieve with neural network interpretability.
In this work we provide evidence that human knowledge is acquired by the AlphaZero neural network as it trains on the game of chess.
By probing for a broad range of human chess concepts we show
when and where
these concepts are represented
in the AlphaZero network.
We also provide a behavioural analysis focusing on opening play, including qualitative analysis from chess Grandmaster Vladimir Kramnik. Finally, we carry out a preliminary investigation looking at the low-level details of AlphaZero's representations, and make the resulting behavioural and representational analyses available online.
\end{abstract}
\begin{document}

\flushbottom
\maketitle

\thispagestyle{empty}

\section{Introduction}

Machine learning systems are commonly believed to learn opaque, uninterpretable representations that have little in common with human understanding of the domain they are trained on. Recently, however, empirical evidence has suggested that at least in some cases neural networks learn human-understandable representations. Notable examples include single neurons in a classifier corresponding to mountain tops or snow~\cite{bau2020understanding}, syntactic information in language models~\cite{manning2020emergent}, surprisingly sophisticated conceptual representations in a multimodal network~\cite{goh2021multimodal} emerging from the alignment of visual and textual data, and the wide range of findings in the the growing field of ``BERTology''\cite{rogers2020primer}. Given these findings, it is natural to ask: do they generalise? 
Should we expect other capable deep learning systems to have similarly meaningful representations? If they do not, then our ability to provide explanations which reflect the computational processes of the model, rather than some alternative rationalisation, will be severely limited.

Although these examples of human-understandable activations are compelling evidence, there are reasons they may not apply more widely. Perhaps these networks only learn human-understandable representations because they are exposed solely to human-generated data, and (at least in the case of classifiers) have human concepts imposed on them via the choice of classification categories. 
Alternatively, their relative simplicity may make them easier to interpret -- none of the examples given earlier exceed human performance.
In order to further test the emergence of human-understandable concepts we need to find a domain where remarkable performance emerges without the use of human-labelled data. 
AlphaZero~\cite{Silver1140} meets both of these requirements: it is trained via self-play, and so has never been exposed to human data, and it performs at a superhuman level in all three domains it was trained on (Chess, Go, and Shogi)
when the network is paired with Monte Carlo tree search.
In this paper we investigate AlphaZero's representations, and their relation to human concepts in chess. Studying the AlphaZero network an important frontier in our understanding of strong neural networks: if we can find human-understandable concepts here, it is likely we will find them elsewhere.

The AlphaZero network provides a rare opportunity for us to study a system that performs at a superhuman level in a complex domain. The ability to interpret AI systems is particularly valuable when these systems  exceed human performance~\cite{andrulis2020domainlevel} on the tasks they were optimized for, as the systems could uncover previously unknown patterns that could prove to be useful beyond the systems themselves. Understanding such complex systems is likely to involve a substantial amount of analytical effort, so it is desirable to first understand whether the system's representations have any correspondence with human concepts. If no such correspondence exists, then further investigations are unlikely to pay off, whereas if they can be found then a deeper investigation may well uncover more. Although the results we show in this paper are far from a complete understanding of the AlphaZero system, we believe that they show strong evidence for the existence of human-understandable concepts of surprising complexity within AlphaZero's neural network.
This lays a solid foundation for further investigation of the AlphaZero network.

\subsection{Our approach}

Our approach to investigating the acquisition of chess knowledge in AlphaZero is a three-pronged one: we probe whether human chess concepts are linearly decodable from the neural network's internal layers;
we examine changing behaviours over the course of full training runs;
we investigate layers and activations directly.

\paragraph{Probe for concepts} Our first priority is to discover whether AlphaZero's internal representations (the activations of its neural network) can be related to human chess concepts. If human concepts can be easily predicted from the network's internal representation, then we should expect deeper investigations to reveal even more information, whereas if there is no relation to human concepts then it is likely that AlphaZero's internal computations will remain opaque after further study. Much of our work therefore focuses on concept-based methods~\cite{bau2017network,kim2018interpretability,melis2018towards, ghorbani2019towards,koh2020concept,chen2020concept}. Concept-based methods involve the detection of human concepts from network activations on a large dataset of inputs.
Because chess has been so extensively theorised -- and much of this theory has been instantiated in chess engine evaluation functions -- we have a broad range of human-defined concepts covering a wide range of complexity, from the existence of a passed pawn, through higher-level concepts such as mobility, all the way to a full position evaluation.
The probing process is automated, so we can probe for every concept, at every block and over many checkpoints during self-play training. This allows us to build up a picture of what is learned, when it was learned during training, and where in the network it is computed. Concept-based methods are far from the only approach to understanding neural network computations, and we review other approaches in Section~\ref{sec:related_work}.

\paragraph{Study behavioural changes} After studying how internal representations change over time, it is natural to investigate how these changing representations give rise to changing behaviours.
During the course of training, some moves are preferred over others in the same position,
and this preference evolves with training steps.
When AlphaZero operates without Monte Carlo Tree Search (MCTS), behavioural changes are simply changes in its prior move selection probability (a complete description of the AlphaZero network and training protocol is given in Section~\ref{s:alphazero_net_and_training}).
We study these behavioural changes by measuring changes in move probability on a curated set of chess positions and compare the evolution of play during self-play training to the evolution of move choices in top-level human play.
Our analysis primarily focuses on openings as these are most heavily theorised, and have made a dataset of these positions with both human and AlphaZero play data \href{https://storage.googleapis.com/uncertainty-over-space/alphachess/index.html}{available online}.

\paragraph{Investigate activations directly} Finally, having established that many human concepts can be predicted from AlphaZero's activations after training, we begin to investigate these activations directly. We use the established technique of non-negative matrix factorisation (NMF)~\cite{paatero1994positive, lee1999learning, olah2018building} to decompose AlphaZero's representations into multiple factors. This approach provides information that is not tied to pre-existing human concepts, providing a complementary view on what is being computed by the AlphaZero network. We make a dataset of matrix factorisations
\href{https://storage.googleapis.com/uncertainty-over-space/alphachess/index.html}{available online}.
We also investigate a new approach: measuring covariance directly between single-neuron activations and inputs.
This method is essentially providing a combination of input features
whose presence is most correlated
with the activation of a given neuron.

\subsection{Summary of results}
In this paper we consider the progression of AlphaZero's neural network
from initialization until the end of training.
We examine chess concepts as they emerge in the network, determine when they were learned in training and where they are computed in the network, and assess their impact on the aggregate value assessment.
Our contributions are an improved understanding of: 1) Encoding of human knowledge;
2) Acquisition of knowledge during training;
3) Reinterpreting the value function via the encoded chess concepts;
4) Comparison of AlphaZero's evolution to that of human history;
5) Evolution of AlphaZero's candidate move preferences;
6) Proof of concept towards unsupervised concept discovery.%

\paragraph{Many human concepts can be found in the AlphaZero network}
In Section~\ref{sec:encoding} we demonstrate that the AlphaZero network's internal learned representation of the chess board can be used to reliably reconstruct many human chess concepts. We adopt the approach of using concept activation vectors (CAV)~\cite{kim2018interpretability} by training sparse linear probes for a wide range of concepts. Our definition of concepts is given in Section~\ref{ss:concepts_defn} and our probing methodology is described in Section~\ref{sec:concept-regression}. We discuss the strengths and weaknesses of our approach in Section~\ref{ss:probing-challenges}. We show that many concepts can be regressed more accurately from internal representations than from the network input, indicating that relevant information is being computed by the AlphaZero network. We also show that, while AlphaZero's knowledge of chess seems to be tightly correlated with our concept probes, there is indeed a difference between them, as the reconstruction tends to be incomplete. An analysis of prediction errors in Section~\ref{sec:residuals} shows features in common between positions with high prediction error. 

\paragraph{A detailed picture of knowledge acquisition during training}
By using our concept probing methodology we can measure the emergence of relevant information over the course of training, and at every layer in the network. This allows us to produce what we refer to as \textbf{what-when-where plots}:
\emph{what} concept is learned \emph{when} in training time \emph{where} in the network.
What-when-where plots are plots of concept regression accuracy across training time and network depth.
We show results for a range of concepts in Section~\ref{sec:encoding}, with the remainder in the Supplementary Material. We repeat this analysis for a second training run, demonstrating remarkable consistency in the development of concepts. We also find that many concepts emerge at approximately the same point early in training, at which point AlphaZero's move selection also changes rapidly and dramatically. We discuss this rapid change in Section~\ref{sec:explosion}.

\paragraph{Use of concepts; Relative concept value}
In Section~\ref{sec:value-regression} we focus on the evolution of AlphaZero's value function over time. We again use a concept-based approach: we attempt to predict the output of the value function from sets of human concepts. Studying the evolution of concept weights over the course of training gives us a picture of how AlphaZero's behaviour is related to high-level human chess concepts, which we view as a surrogate for its `style'. We find that early in training AlphaZero focuses primarily on material, with more complex and subtle concepts such as king safety and mobility emerging as important predictors of the value function only relatively late in training.

\paragraph{Comparison to historical human play}
When looking at the evolution of chess understanding within a system like AlphaZero, one has to wonder whether there is such a thing as a natural progression of knowledge, a way of developing an understanding of the game that is specific to the game itself, rather than being purely arbitrary and left to chance -- or is it a mix of both? We investigate this question in Sections~\ref{sec:human-history} and~\ref{sec:explosion} by comparing AlphaZero training to human history and across multiple training runs respectively. Our analysis shows that there are both similarities as well as differences, and that AlphaZero doesn't recapitulate human history: there are notable differences in how human play has developed, but striking similarities in self-play policy. We also present a qualitative assessment of differences in playstyle over the course of training.

\paragraph{Unsupervised knowledge discovery}

In Section \ref{sec:intermediate_computations} we present preliminary findings from using unsupervised methods to inspect AlphaZero's representations directly. Section~\ref{ss:nmf} uses non-negative matrix factorisation to decompose activations at each layer into multiple factors, allowing us to identify factors relating to move selection. In Section~\ref{sec:activations} we compute the covariance between activations and inputs, providing a new view on the relationship between activations in early layers and network representations.

\section{Past and related work}\label{sec:related_work}
Our work relates two areas: machine learning explainability/interpretability (which we refer to simply as interpretability for conciseness), and chess as a testing ground for AI research. We use interpretability tools as experimental instruments that help us learn about the way a given model (in this case AlphaZero) operates on a specific domain. In this section we review prior work on neural network interpretability (Section~\ref{ss:interpretability}) and the role of chess as a proving ground for AI (Section~\ref{ss:ai_chess_history}).

\subsection{Interpretability}\label{ss:interpretability}
Neural network interpretability is a broad research area~\cite{arrieta2020explainable}, with different notions of interpretability covering a range of use cases~\cite{DoshiKim2017Interpretability, lipton2018mythos}. There are two distinct approaches to explaining neural networks: (1) build an inherently interpretable model or (2) generate post-hoc explanations given an already trained model. While (1) is an important approach for high stakes domains, it may not be always feasible: sometimes (as in this case) we already have a model we wish to understand, and the highest-performance models may often not be the ones with the most inherently interpretable architectures. In this case we must rely on post-hoc interpretability methods. Because of our interest in relating network activations to human concepts our work primarily uses concept-based explanations (Section~\ref{sec:encoding}), although we also explore approaches that focus directly on network activations in Section~\ref{sec:intermediate_computations}.

\paragraph{Concept-based interpretability}
Concept-based interpretability methods build on the idea of network probing~\cite{alain2016understanding}, and use human understandable concepts to explain neural network decisions in terms that users can understand~\cite{bau2017network,kim2018interpretability,melis2018towards,ghorbani2019towards, koh2020concept}. If these users are domain experts, they can specify complex domain-specific concepts, allowing them to probe network operation and understand network decisions in greater detail. Concept-based explanations have been successfully used in complex scientific and medical domains, as seen in ~\cite{clough2019global,graziani2018regression,sprague2018interpretable,mincu2021concept,bouchacourt2019educe,yeche2019ubs,sreedharan2020bridging,inproceedings}. Chess poses similar challenges and opportunities for interpretability research: complex concepts developed over centuries often cannot be simply expressed using a set of board positions/features and encode deep domain knowledge.
However, because of the rich history of chess as a domain for AI research, many of these concepts are already expressed (at least in a simplified, easy-to-compute form) inside modern chess engines, giving a natural starting point for concept-based investigations.

\paragraph{Other post-hoc interpretability methods}
A widely-used method for generating post-hoc explanations is to learn weights for each input feature (e.g., pixel for images) indicating how `important' each feature is to the prediction.
The definition of importance varies in papers depending on the goal of explanations:
some use game theoretic credit assignment approaches~\cite{lundberg2017unified}, while others derive methods from a set of axioms~\cite{sundararajan2017axiomatic} or use simpler approaches of perturbation~\cite{FongV17,selvaraju2016grad} or optimization of desiderata~\cite{DabkowskiGal2017RealB}.
While many of these methods are shown to be useful for end-tasks, more recent studies called some of these methods into question in terms of their validity~\cite{adebayo2018sanity, kindermans2017reliability}, robustness~\cite{alvarez2018robustness} or their formulation~\cite{suraj20}. For example, ~\cite{adebayo2018sanity} showed that explanations from a trained and an untrained model are visually indistinguishable and ~\cite{juliusDebugging} also hinted that the explanations do not seem to be useful for debugging typical ML model problems.

An alternative approach to post-hoc interpretability focuses on obtaining a low-level mechanistic understanding of a given neural network, understanding the precise algorithms implemented at each layer and the representations they give rise to. Early work in this approach focused on understanding the representations at intermediate layers of vision networks using matrix factorisation and feature visualisation approaches~\cite{olah2018building, carter2019activation}. More recent work has focused on representations in multimodal networks~\cite{goh2021multimodal}, as well as circuit-level understanding of neural network algorithms~\cite{olah2020zoom, cammarata2020thread}. The fact that these approaches are successful indicates some degree of alignment between learned representations and human-understandable concepts. This remarkable fact (which is also implicitly used by concept-based approaches) is probed in further detail by `network dissection' approaches~\cite{bau2017network, bau2020understanding}. Network dissection searches for interpretable hidden units in intermediate layers of neural networks, and has been used to quantify the conditions under which they emerge.

\paragraph{Challenges for post-hoc interpretability methods}
It is worth noting that post-hoc interpretation (feature or concept-based) may lack fidelity relative to inherently interpretable model:
the alignment between the model's reasoning and the explanation. For concept-based explanation, the alignment between the representation and a human's mental model of the concept remains an open problem~\cite{cai19,chen2020concept,pitfalls21}.
Our research is no exception: we conduct probing-based approach that can measure only a proximate correlation (between human concepts and the representation), and not causation in any forms. Understanding the causal relationship between concepts and behaviour is a challenging problem which has received relatively little attention. The current best solution requires a generative model for inputs which can be conditioned on specific concept values (simulating the do-operator)~\cite{goyal2019explaining}.

\paragraph{Explainability in reinforcement learning}
Recently, there has been also increasing interest in improving explainability for RL methods~\cite{heuillet2020explainability}. These types of systems pose some unique challenges, due to the complexity of both the environments in which they operate, as well as the complexity of many agent architectures. Some approaches aim to help design transparent and explainable RL models, while others aim to help improve the post-hoc explainability of existing systems. Representation learning in RL agents~\cite{raffin2019decoupling, raffin2018srl, 8852042, traore2019discorl, doncieux2020dream, openendedlearning} can be beneficial when used to learn low-dimensional representations for states, policies and actions, hoping to meaningfully capture the variations in agent's environment and disentangle the contributing factors. Symbolic approaches have been proposed for representing objects and relations and better utilisation of either background or acquired knowledge in RL agents~\cite{sreedharan2020tldr, garnelo2016deep, garcez2018symbolic, zambaldi2018relational}. 

Learning the explanations alongside the agent policy is another promising option when it is possible to define useful auxiliary goals and/or codify the knowledge of the environment. Structural causal models have been used in learning the action influence models~\cite{madumal2019explainable}, to answer counterfactual questions about the actions taken. Reward difference explanations~\cite{reward_decomposition} have shown to be helpful in understanding why certain actions are being taken instead of others. Hierarchical reinforcement learning~\cite{vanseijen2017hybrid} and sub-task decomposition~\cite{6630669} can be seen as more interpretable, by introducing structure in the action space, where high-level goals are divided into sub-goals, and the corresponding sets of actions can therefore be more easily interpreted. Yet, this involves a fairly specialized approach to agent building, and is not as applicable to understanding pre-existing systems. In contrast, our own work presented in this paper is of a post-hoc nature, given that we were trying to better understand AlphaZero as a fixed, pre-existing RL system that has demonstrated a high level of performance across multiple domains. In terms of prior approaches for post-hoc explainability of RL systems, most work to date has involved relying on saliency maps in the input space~\cite{Selvaraju_2019, greydanus2018visualizing, mundhenk2020efficient}. An interesting approach to understanding temporally-extended behaviour in agents with recurrent neural networks is to extract finite-state models of an agent's recurrent state~\cite{koul2018learning}. Agent behavior can also be analysed by trying to identify interesting points in behavioral trajectories~\cite{zahavy2016graying, Sequeira_2020} and highlight the key decisions. A similar concept-based analysis of AlphaZero trained on the game of Hex was carried out contemporaneously with our research~\cite{forde2022concepts}. This analysis studied both representation and use of concepts.

\subsection{Chess and AI research}\label{ss:ai_chess_history}

The game of chess was part of the narrative of AI research since the early days of computing, and it is growing into a
testing ground for AI interpretability.

\paragraph{Chess as a model system for AI research}
Chess has for a long time been a ``Drosophila'' of AI research~\cite{Kasparov1087}; a model game used to prototype novel approaches and models of cognition. Although many of the chess engines that play at levels beyond human ability do so by `brute force'
--- leveraging an enormous number of position evaluations to compensate for a lack of intuition ---
more recent neural network engines such as AlphaZero~\cite{Silver1140} and Lc0~\cite{leela} achieve superhuman strength while evaluating far fewer positions. Although neural network chess engines still evaluate many more positions than a human player would consider, their evaluation functions are learned rather than hardcoded. This raises the possibility that we could learn more about the game of chess by studying neural networks that play it well. The combination of extensive human knowledge, an ultimately verifiable `truth' in the position,
and involving play that is partly intuition and partly move-by-move calculation mean that the role of chess in AI research could be reinvigorated by recasting it at the forefront of AI explainability research. This sentiment has been voiced by former world champion Gary Kasparov~\cite{Kasparov1087, kasparov_explainable}, while recalling the match with DeepBlue, and looking forward towards AlphaZero and the importance of understanding what makes each version play better. AlphaZero has been used to help prototype new variants of chess~\cite{az_variants_preprint}, under a number of atomic rule alterations proposed by grandmaster Vladimir Kramnik, a former chess world champion.

\paragraph{Chess as a testing ground for interpretability}
A number of approaches have been proposed recently that aim to explain the play of chess engines in human-understandable terms.
Chess explanations patterns that are organized in a tree based on their complexity have been introduced in~\cite{10.1007/3-540-60364-6_40} for an educational chess system and further considered in~\cite{HaCohenKerner1995LearningSF}.
Focused feature saliency~\cite{puri2020explain} that balances specificity and relevance has been applied to chess-playing agents to produce saliency maps
that highlight the pieces of highest importance for playing the selected move.
The authors performed a user study involving chess players
that have an ELO rating between 1600 and 2000
These players were randomly shown chess tactics puzzles with saliency maps derived via different methods, and the authors report a significant difference in accuracy and time invested depending on the method. It is worth noting that interest in saliency maps goes beyond the understanding of machines and that human attention over the board during chess games has been analysed to predict saliency with respect to chess players' pattern recognition and line calculation~\cite{10.1145/3314111.3319827}. Even AI systems themselves have been used to model human behavior in chess, and better capture the playing style of different players~\cite{mcilroyyoung2021learning} and players of different strength~\cite{10.1145/3394486.3403219}. In~\cite{Czech_2020}, the authors show activation maps for a deep neural network trained to play the Crazyhouse chess variant at a level beyond human ability.

Natural language processing has been used for automatically generating move-by-move commentary of chess games based on social media forum data~\cite{jhamtani2018acl}. Another recent application of NLP focused instead on training the evaluation function based on the sentiment of free-text chess comments~\cite{kamlish2019sentimate}. A question answering dataset for chess has been proposed in~\cite{Cirik2015ChessQ}, as a benchmark for improving specific types of deep learning approaches, requiring the models to demonstrate knowledge of basic positional concepts on the board. In terms of conceptual chess understanding, DecodeChess~\cite{decodechess} is maybe the most well-known example of an application of using chess concepts to provide human-understandable explanations of critical factors in each position. Concepts are suggested to be relevant, in this context, only if they provably affect the course of the game, based on an extensive search tree generated by a chess engine. In our work, we have focused on understanding the AlphaZero neural network that produces the value assessment and the candidate moves, rather than the computations generated via Monte Carlo tree search (MCTS).
In doing so, we are trying to capture the ``intuitive'' aspect of chess play, rather than determine the truth in the position. When it comes to understanding the deep calculations performed by AlphaZero via MCTS, it would be interesting to consider an approach like the one in DecodeChess in the future, although this falls outside of the scope of our current study.

\section{AlphaZero: Network structure and training}\label{s:alphazero_net_and_training}

AlphaZero \cite{Silver1140} comprises of two components, a deep neural network that computes a policy and value estimate from a state, and Monte Carlo tree search (MCTS) that uses the neural network 
to repeatedly evaluate states and update 
its action selection rule.
A `state' is a position and possibly a history of preceding positions, along with ancillary information such as castling rights,
and is represented as a real-valued vector $\z^0 \in \Rbb^{d_0}$.
The purpose of superscript zero is to indicate that $\z^0$ is an input to the network, or the network's
representation after layer \emph{zero}.
The neural network
\begin{equation} \label{eq:az-network}
\p, v = f_{\btheta}(\z^0)    
\end{equation}
predicts two quantities that are learned from training data games:
It predicts the expected outcome of the game $v$
from the current position,
as well as a probability distribution $\p$ on the next move.
Both are used in MCTS, and are referred to as the `value head' and the `policy head' in the AlphaZero network in Figure \ref{fig:alphazero-network}.

Starting with a neural network with randomly initialized parameters $\btheta$, the AlphaZero network is trained
from data that is generated as the system repeatedly plays against itself.
A buffered queue of self-play games is generated, which serves as training data for the neural network. Because self-play involves search with MCTS, the games are of slightly better quality than those in the training data buffer. Better self-play games are added to the queue while earlier self-play games are dropped.
Concurrently, gradient descent 
is used to minimize the difference between the network's current predictions and a position's
played move and game outcome,
with positions taken from the self-play game queue.

\subsection{AlphaZero neural network}

\begin{figure}[t]
\centering
\tikzset{decorate sep/.style 2 args=
{decorate,decoration={shape backgrounds,shape=circle,shape size=#1,shape sep=#2}}}

\begin{tikzpicture}[every edge quotes/.append style={auto, text=black}]

\begin{scope}
\pgfmathsetmacro{\cubex}{1}
\pgfmathsetmacro{\cubey}{1}
\pgfmathsetmacro{\cubez}{1}
\draw [draw=black, every edge/.append style={draw=gray, densely dashed, opacity=.5}, fill=white]
(0,0,0) coordinate (o) -- ++(-\cubex,0,0) coordinate (a) -- ++(0,-\cubey,0) coordinate (b) edge coordinate [pos=1] (g) ++(0,0,-\cubez)  -- ++(\cubex,0,0) coordinate (c) -- cycle
(o) -- ++(0,0,-\cubez) coordinate (d) -- ++(0,-\cubey,0) coordinate (e) edge (g) -- (c) -- cycle
(o) -- (a) -- ++(0,0,-\cubez) coordinate (f) edge (g) -- (d) -- cycle;
\path [every edge/.append style={draw=black, |-|}]
(b) +(0,-5pt) coordinate (b1) edge ["$W=8$"'] (b1 -| c)
(b) +(-5pt,0) coordinate (b2) edge ["$C=14h + 7$"] (b2 |- a)
(c) +(3.5pt,-3.5pt) coordinate (c2) edge ["$H=8$"'] ([xshift=3.5pt,yshift=-3.5pt]e);
\node[fit=(b)(d)](group){};
\draw[line width=1pt,orange,decorate,decoration={amplitude=7pt,brace}]
([xshift=1cm]group.north east) -- ([xshift=1cm]group.south east);
\node[right=of group,anchor=west,xshift=0.3cm]{input};
\coordinate (inputend) at (-0.4,0.2);
\draw[arrows=->,line width=1pt,draw=blue](inputend)--(-0.4,1.1);
\node[text=blue,anchor=east] at (-0.4,0.7) {3x3 conv; ReLU};
\node[fit=(b)(d)](group2){};
\draw[line width=1pt,green,decorate,decoration={amplitude=7pt,brace,mirror}]
([xshift=-3cm]group2.north west) -- ([xshift=-3cm]group2.south west);
\node[left=of group2,anchor=east,xshift=-2.3cm]{input $\z^0$};
\end{scope}
\begin{scope}[yshift=2.7cm]
\pgfmathsetmacro{\cubex}{1}
\pgfmathsetmacro{\cubey}{1}
\pgfmathsetmacro{\cubez}{1}
\draw [draw=black, every edge/.append style={draw=gray, densely dashed, opacity=.5}, fill=white]
(0,0,0) coordinate (o) -- ++(-\cubex,0,0) coordinate (a) -- ++(0,-\cubey,0) coordinate (b) edge coordinate [pos=1] (g) ++(0,0,-\cubez)  -- ++(\cubex,0,0) coordinate (c) -- cycle
(o) -- ++(0,0,-\cubez) coordinate (d) -- ++(0,-\cubey,0) coordinate (e) edge (g) -- (c) -- cycle
(o) -- (a) -- ++(0,0,-\cubez) coordinate (f) edge (g) -- (d) -- cycle;
\path [every edge/.append style={draw=black, |-|}]
(b) +(0,-5pt) coordinate (b1) edge ["$W=8$"'] (b1 -| c)
(b) +(-5pt,0) coordinate (b2) edge ["$C=256$"] (b2 |- a)
(c) +(3.5pt,-3.5pt) coordinate (c2) edge ["$H=8$"'] ([xshift=3.5pt,yshift=-3.5pt]e);
\node[fit=(d)(inputend)](group){};
\draw[line width=1pt,orange,decorate,decoration={amplitude=7pt,brace}]
([xshift=1.4cm]group.north east) -- ([xshift=1.4cm]group.south east);
\node[right=of group,anchor=west,xshift=0.7cm]{\emph{layer 1:} convolution to 256 channels};
\coordinate (block0end) at (-0.4,0.2);
\draw[arrows=->,line width=1pt,draw=blue](block0end) to (-0.4,1.1);
\draw[arrows=->,line width=1pt,draw=blue](block0end) ++(-0.1,0) to [out=140,in=240] (-0.5,1.1);
\node[text=blue,anchor=east] at (-0.6,0.7) {3x3 conv; ReLU; 3x3 conv; ReLU};
\node[text=blue,anchor=west] at (-0.3,1.1) {\emph{add}; ReLU};
\node[fit=(b)(d)](group2){};
\draw[line width=1pt,green,decorate,decoration={amplitude=7pt,brace,mirror}]
([xshift=-3cm]group2.north west) -- ([xshift=-3cm]group2.south west);
\node[left=of group2,anchor=east,xshift=-2.3cm]{activations $\z^1$};
\end{scope}
\begin{scope}[yshift=5.0cm]
\pgfmathsetmacro{\cubex}{1}
\pgfmathsetmacro{\cubey}{1}
\pgfmathsetmacro{\cubez}{1}
\draw [draw=black, every edge/.append style={draw=gray, densely dashed, opacity=.5}, fill=white]
(0,0,0) coordinate (o) -- ++(-\cubex,0,0) coordinate (a) -- ++(0,-\cubey,0) coordinate (b) edge coordinate [pos=1] (g) ++(0,0,-\cubez)  -- ++(\cubex,0,0) coordinate (c) -- cycle
(o) -- ++(0,0,-\cubez) coordinate (d) -- ++(0,-\cubey,0) coordinate (e) edge (g) -- (c) -- cycle
(o) -- (a) -- ++(0,0,-\cubez) coordinate (f) edge (g) -- (d) -- cycle;
\path [every edge/.append style={draw=black, |-|}]
(b) +(-5pt,0) coordinate (b2) edge ["$C=256$"] (b2 |- a)
 ([xshift=3.5pt,yshift=-3.5pt]e);
\node[fit=(d)(block0end)](group){};
\draw[line width=1pt,orange,decorate,decoration={amplitude=7pt,brace}]
([xshift=1cm]group.north east) -- ([xshift=1cm]group.south east);
\node[right=of group,anchor=west,xshift=0.3cm]{\emph{layer 2:} ResNet block 1};
\coordinate (block1end) at (-0.4,0.5);
\draw[decorate sep={1mm}{2mm},fill=gray,draw=gray](block1end) -- (-0.4,1.4);
\draw[decorate sep={1mm}{2mm},fill=gray,draw=gray](block1end) ++(3.1,0) -- (2.7,1.4);
\draw[decorate sep={1mm}{2mm},fill=gray,draw=gray](block1end) ++(-5.0,0) -- (-5.4,1.4);
\coordinate (block18end) at (-0.4,1.4);
\draw[arrows=->,line width=1pt,draw=blue](block18end) to (-0.4,2.3);
\draw[arrows=->,line width=1pt,draw=blue](block18end) ++(-0.1,0) to [out=140,in=240] (-0.5,2.3);
\node[text=blue,anchor=east] at (-0.6,1.9) {3x3 conv; ReLU; 3x3 conv; ReLU};
\node[text=blue,anchor=west] at (-0.3,2.3) {\emph{add}; ReLU};
\node[fit=(b)(d)](group2){};
\draw[line width=1pt,green,decorate,decoration={amplitude=7pt,brace,mirror}]
([xshift=-3cm]group2.north west) -- ([xshift=-3cm]group2.south west);
\node[left=of group2,anchor=east,xshift=-2.3cm]{activations $\z^2$};
\end{scope}
\begin{scope}[yshift=8.5cm]
\pgfmathsetmacro{\cubex}{1}
\pgfmathsetmacro{\cubey}{1}
\pgfmathsetmacro{\cubez}{1}
\draw [draw=black, every edge/.append style={draw=gray, densely dashed, opacity=.5}, fill=white]
(0,0,0) coordinate (o) -- ++(-\cubex,0,0) coordinate (a) -- ++(0,-\cubey,0) coordinate (b) edge coordinate [pos=1] (g) ++(0,0,-\cubez)  -- ++(\cubex,0,0) coordinate (c) -- cycle
(o) -- ++(0,0,-\cubez) coordinate (d) -- ++(0,-\cubey,0) coordinate (e) edge (g) -- (c) -- cycle
(o) -- (a) -- ++(0,0,-\cubez) coordinate (f) edge (g) -- (d) -- cycle;
\path [every edge/.append style={draw=black, |-|}]
(b) +(-5pt,0) coordinate (b2) edge ["$C=256$"] (b2 |- a)
 ([xshift=3.5pt,yshift=-3.5pt]e);
\node[fit=(d)(block18end)](group){};
\draw[line width=1pt,orange,decorate,decoration={amplitude=7pt,brace}]
([xshift=1cm]group.north east) -- ([xshift=1cm]group.south east);
\node[right=of group,anchor=west,xshift=0.3cm]{\emph{layer 20:} ResNet block 19};
\node[fit=(b)(d)](group2){};
\draw[line width=1pt,green,decorate,decoration={amplitude=7pt,brace,mirror}]
([xshift=-3cm]group2.north west) -- ([xshift=-3cm]group2.south west);
\node[left=of group2,anchor=east,xshift=-2.3cm]{activations $\z^{20}$};
\coordinate (block19end) at (-0.4,0.2);
\draw[arrows=->,line width=1pt,draw=blue](block19end) to (1.4,1.4);
\node[text=blue,anchor=west] at (1.1,1.0) {1x1 conv; ReLU};
\draw[arrows=->,line width=1pt,draw=blue](block19end) to (-2.4,1.4);
\node[text=blue,anchor=east] at (-2.1,1.0) {1x1 conv; ReLU};
\end{scope}
\begin{scope}[xshift=2.0cm,yshift=11.5cm]
\pgfmathsetmacro{\cubex}{1}
\pgfmathsetmacro{\cubey}{1}
\pgfmathsetmacro{\cubez}{1}
\draw [draw=black, every edge/.append style={draw=gray, densely dashed, opacity=.5}, fill=white]
(0,0,0) coordinate (o) -- ++(-\cubex,0,0) coordinate (a) -- ++(0,-\cubey,0) coordinate (b) edge coordinate [pos=1] (g) ++(0,0,-\cubez)  -- ++(\cubex,0,0) coordinate (c) -- cycle
(o) -- ++(0,0,-\cubez) coordinate (d) -- ++(0,-\cubey,0) coordinate (e) edge (g) -- (c) -- cycle
(o) -- (a) -- ++(0,0,-\cubez) coordinate (f) edge (g) -- (d) -- cycle;
\path [every edge/.append style={draw=black, |-|}]
(b) +(0,-5pt) coordinate (b1) edge ["$W=8$"'] (b1 -| c)
(b) +(-5pt,0) coordinate (b2) edge ["$C=256$"] (b2 |- a)
(c) +(3.5pt,-3.5pt) coordinate (c2) edge ["$H=8$"'] ([xshift=3.5pt,yshift=-3.5pt]e);
\coordinate (policy1end) at (-0.4,0.2);
\draw[arrows=->,line width=1pt,draw=blue](policy1end) to (-0.4,0.9);
\node[text=blue,anchor=west] at (-0.3,0.7) {1x1 conv + biases};
\end{scope}
\begin{scope}[xshift=2.0cm,yshift=13.2cm]
\pgfmathsetmacro{\cubex}{1}
\pgfmathsetmacro{\cubey}{0.7}
\pgfmathsetmacro{\cubez}{1}
\draw [draw=black, every edge/.append style={draw=gray, densely dashed, opacity=.5}, fill=white]
(0,0,0) coordinate (o) -- ++(-\cubex,0,0) coordinate (a) -- ++(0,-\cubey,0) coordinate (b) edge coordinate [pos=1] (g) ++(0,0,-\cubez)  -- ++(\cubex,0,0) coordinate (c) -- cycle
(o) -- ++(0,0,-\cubez) coordinate (d) -- ++(0,-\cubey,0) coordinate (e) edge (g) -- (c) -- cycle
(o) -- (a) -- ++(0,0,-\cubez) coordinate (f) edge (g) -- (d) -- cycle;
\path [every edge/.append style={draw=black, |-|}]
(b) +(-5pt,0) coordinate (b2) edge ["$C=73$"] (b2 |- a)
([xshift=3.5pt,yshift=-3.5pt]e);
\node[text=blue,anchor=center] at (-0.4,0.7) {flatten; softmax};
\coordinate (softmax) at (-0.4,0.7);
\node[fit=(softmax)(block19end)](group){};
\draw[line width=1pt,orange,decorate,decoration={amplitude=7pt,brace}]
([xshift=2.7cm]group.north east) -- ([xshift=2.7cm]group.south east);
\node[right=of group,anchor=west,xshift=2.0cm]{policy head $\p$};
\end{scope}
\begin{scope}[xshift=-2.0cm,yshift=11.0cm]
\pgfmathsetmacro{\cubex}{1}
\pgfmathsetmacro{\cubey}{0.4}
\pgfmathsetmacro{\cubez}{1}
\draw [draw=black, every edge/.append style={draw=gray, densely dashed, opacity=.5}, fill=white]
(0,0,0) coordinate (o) -- ++(-\cubex,0,0) coordinate (a) -- ++(0,-\cubey,0) coordinate (b) edge coordinate [pos=1] (g) ++(0,0,-\cubez)  -- ++(\cubex,0,0) coordinate (c) -- cycle
(o) -- ++(0,0,-\cubez) coordinate (d) -- ++(0,-\cubey,0) coordinate (e) edge (g) -- (c) -- cycle
(o) -- (a) -- ++(0,0,-\cubez) coordinate (f) edge (g) -- (d) -- cycle;
\path [every edge/.append style={draw=black, |-|}]
(b) +(0,-5pt) coordinate (b1) edge ["$W=8$"'] (b1 -| c)
(b) +(-5pt,0) coordinate (b2) edge ["$C=1$"] (b2 |- a)
(c) +(3.5pt,-3.5pt) coordinate (c2) edge ["$H=8$"'] ([xshift=3.5pt,yshift=-3.5pt]e);
\node[text=blue,anchor=center] at (-0.4,0.7) {flatten};
\coordinate (policy1end) at (-0.4,0.9);
\draw[arrows=->,line width=1pt,draw=blue](policy1end) to (-0.4,1.4);
\node[text=blue,anchor=east] at (-0.5,1.1) {linear layer; ReLU};
\end{scope}
\begin{scope}[xshift=-2.0cm,yshift=13.5cm]
\pgfmathsetmacro{\cubex}{1.4}
\pgfmathsetmacro{\cubey}{0.4}
\pgfmathsetmacro{\cubez}{0.4}
\draw [draw=black, every edge/.append style={draw=gray, densely dashed, opacity=.5}, fill=white]
(0,0,0) coordinate (o) -- ++(-\cubex,0,0) coordinate (a) -- ++(0,-\cubey,0) coordinate (b) edge coordinate [pos=1] (g) ++(0,0,-\cubez)  -- ++(\cubex,0,0) coordinate (c) -- cycle
(o) -- ++(0,0,-\cubez) coordinate (d) -- ++(0,-\cubey,0) coordinate (e) edge (g) -- (c) -- cycle
(o) -- (a) -- ++(0,0,-\cubez) coordinate (f) edge (g) -- (d) -- cycle;
\path [every edge/.append style={draw=black, |-|}]
(b) +(0,-5pt) coordinate (b1) edge ["$\mathrm{dim}=256$"'] (b1 -| c) ([xshift=3.5pt,yshift=-3.5pt]e);
\coordinate (value2end) at (-0.4,0.4);
\draw[arrows=->,line width=1pt,draw=blue](value2end) to (-0.4,1.0);
\node[text=blue,anchor=east] at (-0.5,0.7) {linear layer to scalar};
\node[text=blue,anchor=center] at (-0.5,1.3) {tanh};
\coordinate (tanhend) at (-0.5,1.3);
\node[fit=(tanhend)(block19end)](group){};
\draw[line width=1pt,orange,decorate,decoration={amplitude=7pt,brace,mirror}]
([xshift=-3cm]group.north west) -- ([xshift=-3cm]group.south west);
\node[left=of group,anchor=east,xshift=-2.4cm]{value head $v$};
\end{scope}
\end{tikzpicture}
\caption{The AlphaZero network.
Each $3 \times 3$ convolution indicates the application of 
256 filters of kernel size $3 \times 3$ with stride 1.
A ResNet block contains two rectified batch-normalized convolutional layers with a skip connection.
In the input $\z^0$, a history length of $h=8$ plies is used,
encoding the current board position and those of the seven preceding plies. The input is a $8 \times 8 \times 119$-dimensional tensor.}
\label{fig:alphazero-network}
\end{figure}
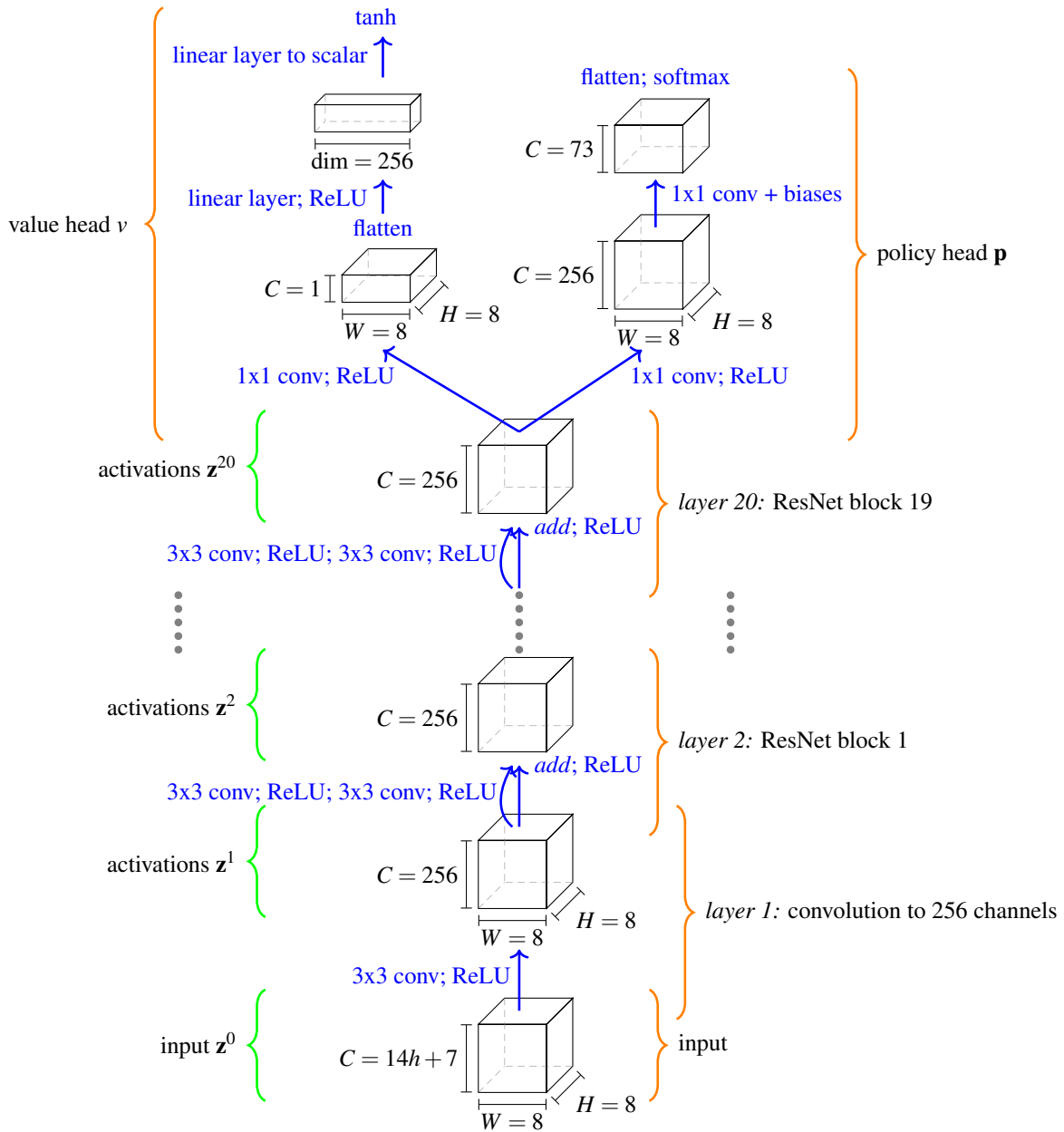

This report is concerned with the evolution of AlphaZero's network; how chess knowledge is progressively acquired and represented.
Figure \ref{fig:alphazero-network} illustrates the AlphaZero network.

The network takes input $\z^0 \in \mathbb{R}^{d_0}$.
In Figure \ref{fig:alphazero-network}, the input is 
$\z^0 \in \mathbb{R}^{8 \times 8 \times (14h+7)}$ for a history length of $h$ plies.
If $h=1$ and only the current position is represented, $\z^0 \in \mathbb{R}^{8 \times 8 \times 21}$.
The first twelve $8 \times 8$ channels in $\z^0$ are binary, encoding the positions of the playing side and opposing side's king, queen(s), rooks, bishops, knights and pawns respectively.
It is followed by $8 \times 8$ binary channels representing the number of repetitions (for three-fold repetition draws), the side to play, and four
binary channels for whether the player and opponent can still castle king and queenside.
Finally, the last two channels are an irreversible move counter (for 50 move rule) and total move counter, both scaled down.
The input representation is always oriented toward the playing side,
so that the board position with black to play is first
flipped horizontally and vertically before being represented in 
the stack of $8 \times 8$ channels $\z^0$.
Even though the state is fully captured with $h=1$ when only the current position is encoded, there is a marginal empirical increase in performance when a few preceding positions are also incorporated into $\z^0$, 
and $\z^0 \in \mathbb{R}^{8 \times 8 \times 119}$ if the board positions of the last eight plies are stacked.
Unless otherwise stated, $h=8$ is used in this report, following \cite{Silver1140}.

If we have a large set of inputs we denote them by
$\left\{ \z^0_n \right\}^N_{n=1}$.
Typically, $N$ is few million, and the inputs would be, for example, all positions from a large collection of grandmaster games.

\subsubsection{Layers}

The network in Figure \ref{fig:alphazero-network} has a
residual neural network 
(ResNet) backbone \cite{he2016deep},
and every ResNet block will form a layer indexed by $l = 1, \ldots, L$.
Each ResNet block contains internal layers, and in this paper we index layers at the points where the skip-connections meet.
We denote the activations at layer $l$ with  $\z^l \in \Rbb^{d_l}$,
with $\z^0$ being the input.
In the AlphaZero network, as illustrated in Figure \ref{fig:alphazero-network}, $\z^l \in \Rbb^{8 \times 8 \times 256}$ for each $l = 1, \ldots, 20$.
There are therefore 16384 activations at the end of each layer,
and we will
use notation $z^l_{i}$ to refer to activation $i$ in layer $l$'s activations $\z^l$.
 
The network progressively transforms input $\z^0$ to $\z^1$, then $\z^2$, and so on through a series of residual blocks and final policy/value heads, as shown in Figure~\ref{fig:alphazero-network}.
The activations of layer $l$ is given by the function
$\z^{l} = f^{l}_{\btheta}(\z^{l-1})$,
and hence $f^{l}_{\btheta}: \mathbb{R}^{d_{l-1}} \to \mathbb{R}^{d_{l}}$, where $d_l$ is dimensionality of layer $l$.
We are going to omit the dependence of the layer on its parameters where it is clear from the context.
For layers $l \ge 2$ the ResNet backbone in Figure \ref{fig:alphazero-network} has the form
\begin{equation} \label{eq:resnet-block}
\z^{l} = f^{l}(\z^{l-1}) = \mathrm{ReLU}(\z^{l-1} + g^l(\z^{l-1})) \ ,    
\end{equation}
which directly copies activations $\z^{l-1}$, adds an additional nonlinear function $g^l(\z^{l-1})$ composed of two more convolution layers to it, and clips the result to be nonnegative through a
rectified nonlinar unit (ReLU).
For stability, activations $\z^l$ are additionally clipped to a maximum value of 15.
The form of Equation \ref{eq:resnet-block} means that
$z^l_i$ is equal to a clipped version of $z^{l-1}_i$ and information from layers of additional local convolutions on $\z^{l-1}$.
We revisit this link in Section \ref{sec:activations} where we show that $z^l_i$ and $z^{l-1}_i$ would typically be activated for similar input patterns.

We introduce additional notation. Considering parts of the network in Figure \ref{fig:alphazero-network},
the neural network function between layers $k$ and $l$ with $k \le l$ takes
$\z^{k-1}$ as input and maps it to outputs $\z^l$.
That is
$f^{k:l}(\z^{k - 1}) = f^{l} \circ \cdots \circ f^{k + 1} \circ f^{k}(\z^{k-1})$, where $\circ$ denotes function composition,
and therefore
\begin{equation} \label{eq:activation-layer-l}
\z^{l} = f^{1:l}(\z^0) = f^{l} \circ \cdots \circ f^{2} \circ f^{1}(\z^{0}) \ .
\end{equation}
Layers 1 to 20 in Figure \ref{fig:alphazero-network} form the `torso' of the network.
We restrict our analysis in this paper to the layers in the torso.
Two `heads' complete the neural network by 
performing a computation on $\z^{20}$, the activations of the last layer in the torso.
The `value head' computes $v$ in \eqref{eq:az-network},
while the `policy head' computes $\p$, a distribution over all moves.
The policy head, before flattening, produces a $8 \times 8 \times 73$
tensor.
For every square, it encodes 73 possible moves to a next square:
7 horizontally left and right; 7 vertically up and down; 7 diagonal moves north west, north east, south west and south east; 8 knight moves; 3 promotion options to \symbishop, \symknight, \symrook \
(a \symqueen \ is default when a pawn reaches the eight rank)
for the three single-square forward moves.

\FloatBarrier

\subsection{AlphaZero training iterations}
\label{sec:training-iterations}

Our experimental setup updates the parameters $\btheta$ of the AlphaZero network over 1,000,000 gradient descent training steps.
A million steps is an arbitrary training time slightly longer than that of AlphaZero in \cite{Silver1140}.
We will use $t$ to index the training step, and
$\btheta_t$ the network parameters after gradient descent step $t$.

The network is trained through positions
with their associated MCTS move probability vectors
that are sampled from self-play buffer containing the previous 1 million positions.
At most 30 positions are sampled from a game on average, as positions on subsequent moves are strongly correlated, and including all of them may lead to increased overfitting.
Stochastic gradient descent steps are taken with a batch size of 4096 in training. A synchronous decaying optimizer is used with initial learning rate 0.2, which is multiplied by 0.1 after 100k, 300k, 500k and 700k iterations.

After every 1000 training steps, the networks that are used to generate self-play games are refreshed, so that MCTS search uses a newer network.
We refer to networks and their parameters $\btheta_t$ that are saved to disk at these points as `checkpoints'.
Self-play moves are executed upon reaching 800 MCTS simulations. Diversity in self-play is increased in two ways: through stochastic move sampling and through adding noise to the prior. The first thirty plies are sampled according to the softmax probability of the visit counts, and only after the thirtieth move are the moves with most visits in the MCTS simulations played deterministically.
To further increase diversity in the self-play games, Dirichlet(0.3) noise is added to 25\% of all priors $\p$ from Equation \eqref{eq:az-network} and renormalized. Of all self-play games, 20\% are played out until the end, whereas in the remaining 80\%, an early termination condition is introduced where a game is resigned if the value gives an expected score of 5\% or less. The maximum game length is capped at 512 plies.

\section{Encoding of human conceptual knowledge}
\label{sec:encoding}

In this section we use a sparse linear probing methodology to determine the extent to which the AlphaZero network represents a wide range of human chess concepts. We describe our notion of concepts in Section~\ref{ss:concepts_defn}, describe the probing methodolgy in Section~\ref{sec:concept-regression} and present our results in Sections~\ref{sec:concept_regression} and~\ref{sec:residuals}. In Section~\ref{ss:probing-challenges} we discuss limitations of our approach and future avenues of research suggested by these limitations. Finally, in Section~\ref{sec:value-regression} we investigate how concepts predict AlphaZero's value function.

\subsection{Concepts}\label{ss:concepts_defn}
We adopt a simple definition of concepts as user-defined functions on network input $\z^0$. Each concept is a mapping of the input space onto the real line,
\begin{equation}
c: \mathbb{R}^{d_0} \to \mathbb{R} \ . 
\end{equation}
A concept could be any function of the position $\z^0$. A simple example concept could detect whether the playing side has a bishop pair,
i.e.~both a dark-squared and a light-squared bishop,
\begin{equation}
c(\z^0) = \begin{cases}
1 & \quad \text{if } \z^0 \text{ contains a \symbishop-pair for the playing side} \\
0 & \quad \text{otherwise}
\end{cases} .
\end{equation}
Most concepts are more intricate than merely looking for bishop pairs, and can take a range of integer values (for instance difference in number of pawns) or continuous values (such as total score as measured by the Stockfish 8 chess engine). An example of a more intricate concept is mobility, where a chess engine designer can write a function
that gives a score for how mobile pieces are in $\z^0$ for the playing side compared to that of the opposing side. What concepts have in common is that they are pre-specified functions which encapsulate a particular piece of domain-specific knowledge.

\paragraph{Concepts from Stockfish's evaluation function}
Stockfish's position evaluation function is comprised of sub-functions that give a score for different features of $\z^0$.
We use the publicly exposed sub-functions from Stockfish 8's evaluation function as concepts.
They are enumerated and explained in Table \ref{tab:concepts}
in Appendix \ref{appendix:concepts}.
The use of Stockfish 8 is intentional, as it allows insights from this report to refer to observations in \cite{game_changer}.
The root concepts in Table \ref{tab:concepts} are for
\verb=material=, \verb=imbalance=, \verb=pawns=, \verb=knights=, \verb=bishops=, \verb=rooks=, \verb=queens=, \verb=mobility=, \verb=king_safety=, \verb=threats=, \verb=passed_pawns=, \verb=space= and \verb=total=.
The root concepts are further enumerated according to whether it is White or Black to play,
and is further enumerated by the phase of the game,
so that a concept like \verb=threats_w_mg= would quantify white's threats during the middle game.
Many Stockfish concepts are computed for the current player as well as the opponent: a given concept $c$ (for instance \verb=threats=) will have values for `white', `black' and `total'.
Assuming we are in the middle game (\verb=mg=) with White (\verb=w=) to play,
the total (\verb=t=) value is simply
\[
c_{\texttt{threats\_t\_mg}}(\z^0) = c_{\texttt{threats\_w\_mg}}(\z^0) - c_{\texttt{threats\_b\_mg}}(\z^0) \ .
\]
We focus on the total values, adjusted for the phase of the game, as these are the highest-level versions of each concept, and the ones that are used in Stockfish's position evaluation.

\paragraph{Additional custom concepts}
In addition to the concepts from Stockfish's public API, we implemented 116 custom concepts, which are enumerated in Tables \ref{tab:concepts-custom} and \ref{tab:concepts-custom-pawns} in Appendix \ref{appendix:concepts}.
These concepts encapsulate more specific lower-level features, such as the existence of forks, pins, or contested files, as well as a range of features regarding pawn structure.

\paragraph{Concepts and activations dataset} We randomly selected $10^5$ games from the full ChessBase archive and computed concept values and AlphaZero activations for every position in this set. This set was then deduplicated by removing any duplicate positions using the position's Forsyth-Edwards Notation (FEN) string. We then randomly sampled training, validation, and test sets from the deduplicated data. Continuous-valued concepts used training sets of $10^5$ unique positions (not games), with validation and test sets consisting of a further $3\times 10^4$ unique positions each. Binary-valued concepts were balanced to give equal numbers of positive and negative examples, which restricted the data available for training as some concepts only occur rarely. The minimum training dataset size for any binary concept was 50,363, and the maximum size was $10^5$.

\subsection{Probing concept learning with sparse linear regression}
\label{sec:concept-regression}
To detect the emergence of the human concepts of Tables \ref{tab:concepts}, \ref{tab:concepts-custom} and \ref{tab:concepts-custom-pawns} within the AlphaZero network, we employ a simple probing methodology related to concept activation vectors~\cite{kim2018interpretability}.
We train a sparse regression model from activations $\z^{l_t}$ at a given layer $l$ and training step $t$ to a human concept $j$,
using a linear predictor for continuous concepts and a logistic predictor for binary ones. More precisely, for each layer $l$, training step $t$ and concept $c_j(\z^0)$ we train a parameterised regression function
\begin{align}
g^j_{lt}(\z^{l_t}) &= \w^T_{jlt} \z^{l_t} + b_{jlt}\quad\text{(continuous concepts)}
\label{eq:concept-regression-continuous}
\\
g^j_{lt}(\z^{l_t}) &= \sigma(\w^T_{jlt} \z^{l_t} + b_{jlt})\quad\text{(binary concepts)}
\label{eq:concept-regression-binary}
\end{align}
from the output of the $l^{\mathrm{th}}$ layer of the $t^{\mathrm{th}}$ network in training, $\z^{l_t} = f^{1:l}_{\btheta_t}(\z^0)$, to approximate the concept value $c_j(\z^0)$.
Function $\sigma$ is the sigmoid function $\sigma(x) = 1 / (1 + e^{-x})$. 
As an example, concept $j$ might be a \verb=mobility_t_mg= score for the playing side in position $\z^0$, and consequently
the trained function $g^j_{lt}$ would indicate how linearly predictable \verb=mobility_t_mg= is from the representation $\z^l$ after layer $l$.
\emph{
The accuracy with $g^j_{lt}(\z^{l_t})$ predicts the concept value $c_j(\z^0)$ on a held-out test set indicates how much information the activations at layer $l$ carry regarding that concept at a given point in training.}

\paragraph{Learning the probes}
We define the data matrix of activations
$\Z^l_t \in \Rbb^{d_l \times N}$ to contain layer $l$'s
activations $\z_n^l = f_{\btheta}^{1:l}(\z_n^0)$ for training set 
inputs $ \z_1^0, \ldots \z_N^0$
at training step $t$.
Furthermore,
let $\c_j \in \Rbb^N$ be a vector that contains the value
of concept $j$ for each of the inputs of the training set,
hence each $c_j(\z_n^0)$.
Vector $\c_j$ has no superscript $l$ or subscript $t$; the concepts are purely functions of the
input and is independent of the neural network layer or training step.

The parameters $\w_{jlt}$ and $b_{jlt}$ for Equation \ref{eq:concept-regression-continuous} are found by
minimizing the empirical mean squared error between $g^j_{lt}(\z^{l_t})$ and $c_j(\z^0)$.
From Figure \ref{fig:alphazero-network}, each
block's activations are $8 \times 8 \times 256$ dimensional, and hence $\z^l \in \Rbb^{16,384}$.
Because of the high dimensionality of $\z^l$, we regularize $\w_{jlt}$ to avoid any overfitting, as well as testing on a held-out test set.
Hence
\begin{align}
\label{eq:sparse-linear}
\w_{jlt}, b_{jlt}
& = \min_{\w, b} \frac{1}{N} \Big\| \w^T \Z^l_t + b \onebf - \c_j \Big\|^2_2 + \lambda \|\w \|_1 + \lambda |b| \ \quad\text{(continuous concepts)}, \\ 
\w_{jlt}, b_{jlt} & = \min_{\w, b} \frac{1}{N} \Big\| \sigma(\w^T \Z^l_t + b \onebf) - \c_j \Big\|^2_2 + \lambda \|\w \|_1 + \lambda |b| \ \quad\text{(binary concepts)},  \label{eq:sparse-linear-binary}
\end{align}
where $\| \cdot \|_p$ denotes the $L_p$ norm, $\sigma(\cdot)$ is applied elementwise,
and $\onebf$ is simply the all-one vector in $\Rbb^N$. The parameter $\lambda$ was determined individually for each regression using cross-validation on a held-out validation set, with $\lambda\in\{0.003, 0.006, 0.01\}$ for continuous concepts and $\lambda\in\{0.01, 0.1\}$ for binary-valued concepts.

\paragraph{Controls: regression from inputs and random concepts} We provide two controls for comparison: regression from network input, and random concept regression. Regression from network input trains the probe from $\mathbf{z}^0$, and random concept regression uses a normally-distributed random target $c_{\rm{random}}(\mathbf{z}^0)\sim\mathcal{N}(0, 1)$. By comparing to the network input we can see what is being added by network computation, as even an untrained network substantially increases the dimensionality of the regression targets compared to the network input. The random concept control ensures that what we are learning relates to specific concepts, rather than simply separability of arbitrary points.

\paragraph{Information-theoretic concerns} From a purely information-theoretic perspective no neural network can be said to be adding any information that is not present in the input; the data processing inequality $I(X, Z_1) \geq I(X; Z_2)$ for random variables $X$, $Z_1$, and $Z_2$ in a Markov chain $X\to Z_1 \to Z_2$ still applies. Even though $f_{\mathbf{\theta}}$ is deterministic, it can still lose information if it is not bijective (consider a function mapping its entire range to zero as a simple example), so it is entirely possible that information relevant to a concept may be lost during network computation, although it can never be created. The data processing inequality therefore presents conceptual challenges to probing, as well as to our intuitive understanding of information. For example, the information content of an image is the same whether it is represented unaltered, with the pixels scrambled, or even encrypted (so long as the encryption function is bijective), but we find it much easier to understand the unaltered image. One way to understand the function of neural network layers is to see them not as creating information in the sense of Shannon information, but making that information available to a computationally-bounded agent (in our case, the probe network), as is suggested by~\cite{xu2019theory}.

\subsection{Visualising concept learning with what-when-where plots}
\label{sec:concept_regression}

We visualise the acquisition of conceptual knowledge by illustrating \emph{what} concept is learned \emph{when} in training time \emph{where} in the network.
This is done by scoring concept regression over the layer ($l$) and temporal ($t$) axes as AlphaZero is trained.

\subsubsection{Scoring concept regression}
The regression models trained in Section \ref{sec:concept-regression} are probing for the existence of information on each individual concept at every layer and checkpoint\footnote{A `checkpoint' refers to a network and its parameters $\btheta_t$ as saved to disk at training step $t$; 
see Section \ref{sec:training-iterations}.}.
This means that the accuracy of these models can be used as a proxy to measure the presence or absence of (linearly-decodable) information.
By comparing accuracy scores across layers and checkpoints we aim to understand what concepts are encoded, when in training they emerge, and where in the network they are strongly encoded.
Because the training process for each sparse regression model is identical, we know that any changes in accuracy between different layers and/or checkpoints must be due to changes in AlphaZero's internal representations. Using a fixed dataset of human play rather than games from AlphaZero self-play means that we also avoid changes in the distribution of positions due to different policies affecting either the distribution of concepts or activations.

To generate what-when-where plots, we require an accuracy measure. For continuous concepts we use the coefficient of determination $r^2$, and for binary concepts we use the increase in accuracy over random label prediction. For concept $j$, checkpoint $t$ and layer $l$, the coefficient of determination $r_{jlt}^2$ using the test set of positions $\z_1^0, \ldots, \z_M^0$ is given by the equation:
\begin{equation} \label{eq:r-squared}
r_{jlt}^2 = 1 - \frac{
\sum_m (c_j(\z^0_m) - g^j_{lt}(\z^{l_t}_m))^2
}{
\sum_m (c_j(\z^0_m) - \overline{c_j})^2
}
\end{equation}
where $\z^{l_t}_m$ is the activation vector of input $\z^0_m$ at layer $l$ at training iteration $t$ for position $m$ (see Equation~\ref{eq:activation-layer-l}), $g^j_{lt}(\z^{l_t}_m)$ is the best (sparse) linear predictor (as found by Equation~\ref{eq:sparse-linear}), and $\overline{c_j}$ is the mean concept value 
\begin{equation}
   \overline{c_j} = \frac{1}{M} \sum_{m} c_j(\z^0_m).
\end{equation}
A score of $r^2 = 1$ means that a concept is perfectly linearly predictable from a layer's activations, whereas a predictor that always returned $\overline{c_j}$ would have $r^2=0$. For binary concepts we report the accuracy increase over random guessing (recall that for all binary concepts we used balanced datasets for training, validation, and testing). For binary concepts the accuracy score $s_{jlt}$ is given by
\begin{equation}
    s_{jlt} = 2\sum_m \delta(c_j(z^0_m) - \tilde{g}^j_{lt}(z^{l_t}_m)) - 1,
\end{equation}
where $\tilde{g}^j_{lt}(z^{l_t}_m)$ is the maximum-probability class output by the logistic regression $g$. This is normalised such that predicting either $0$ or $1$ for all $m$ gives an accuracy $s=0$, whereas predicting the correct class every time gives $s=1$. Normalising in this way allows for direct comparison with continuous-valued concepts.

For every concept $j$, there is a grid of $r_{jlt}^2$ or $s_{jlt}$ values (depending on whether the concept is binary), with one axis representing the neural network layer $l$ and the other axis representing the training step $t$. Plotting this grid gives a what-when-where plot, allowing us to visualise changes in concept regression accuracy with network depth, as well as the evolution of concepts over time. We show a selection of what-when-where plots in Figure~\ref{fig:concept_regression}, and the full set of concepts are visualised in
Appendix~\ref{sec:all_concepts_regression_supplement}.
Results for a second AlphaZero training run are visualised in Appendix~\ref{sec:concept_regression_second_seed}, demonstrating stability of these results under network retraining.

\subsubsection{The evolution of human concepts in AlphaZero}

\begin{figure}
\centering
\begin{subfigure}[t]{0.3\textwidth}
\vskip 0pt
\centering
\includegraphics[width=\textwidth]{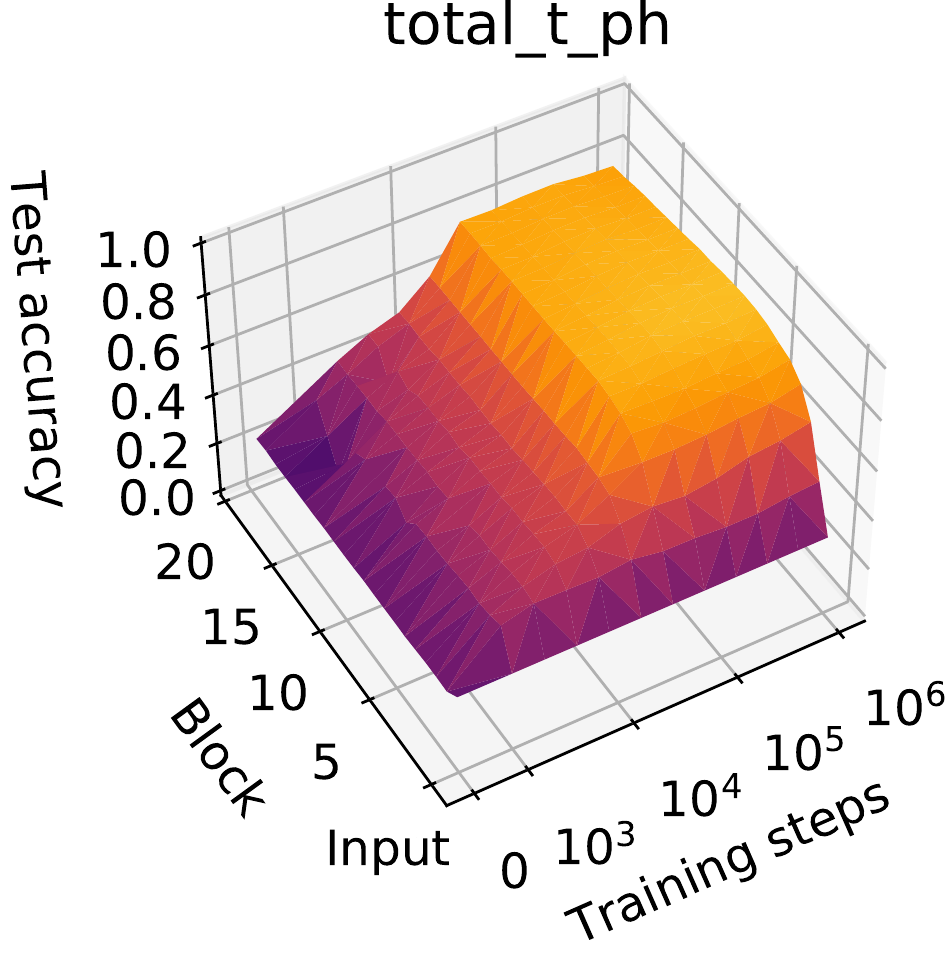}
\caption{Stockfish 8's total score}
\label{fig:concept_regression-total-score}
\end{subfigure}
\hfill
\begin{subfigure}[t]{0.3\textwidth}
\vskip 0pt
\centering
\includegraphics[width=\textwidth]{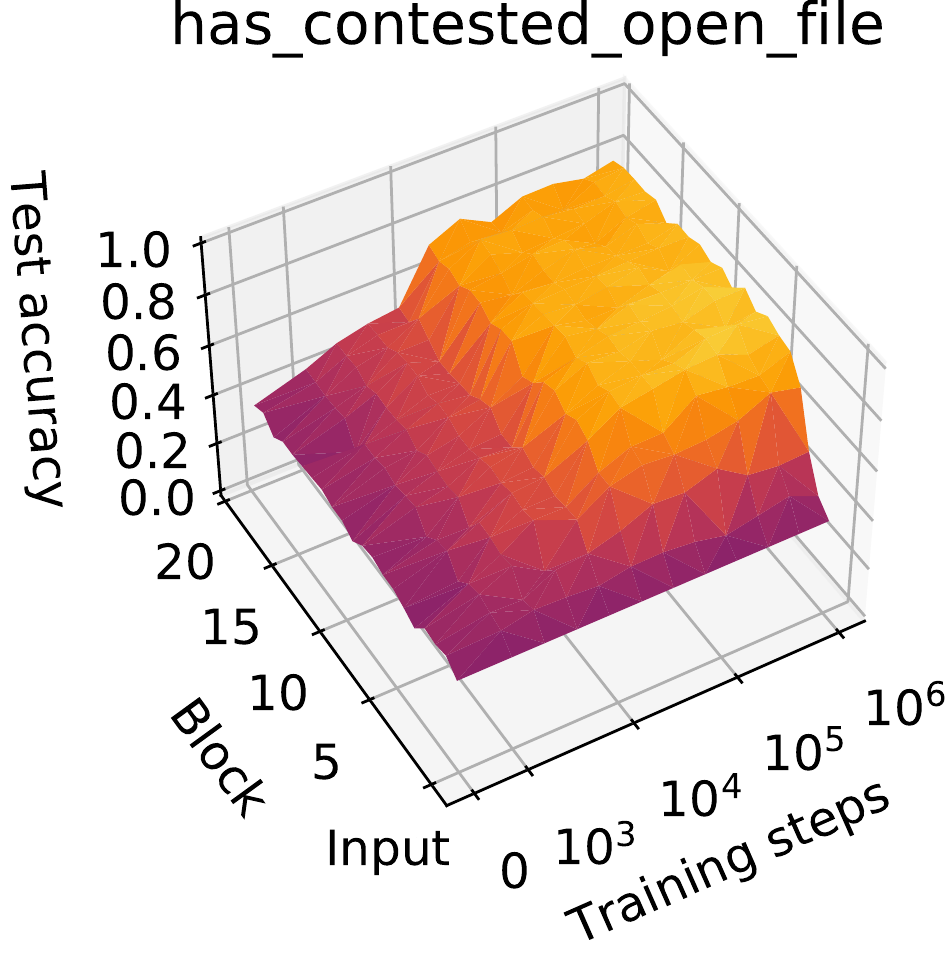}
\caption{A contested open file is occupied by rooks and/or queens of opposite colours}
\label{fig:concept_regression-contested-open-file}
\end{subfigure}
\hfill
\begin{subfigure}[t]{0.3\textwidth}
\vskip 0pt
\centering
\includegraphics[width=\textwidth]{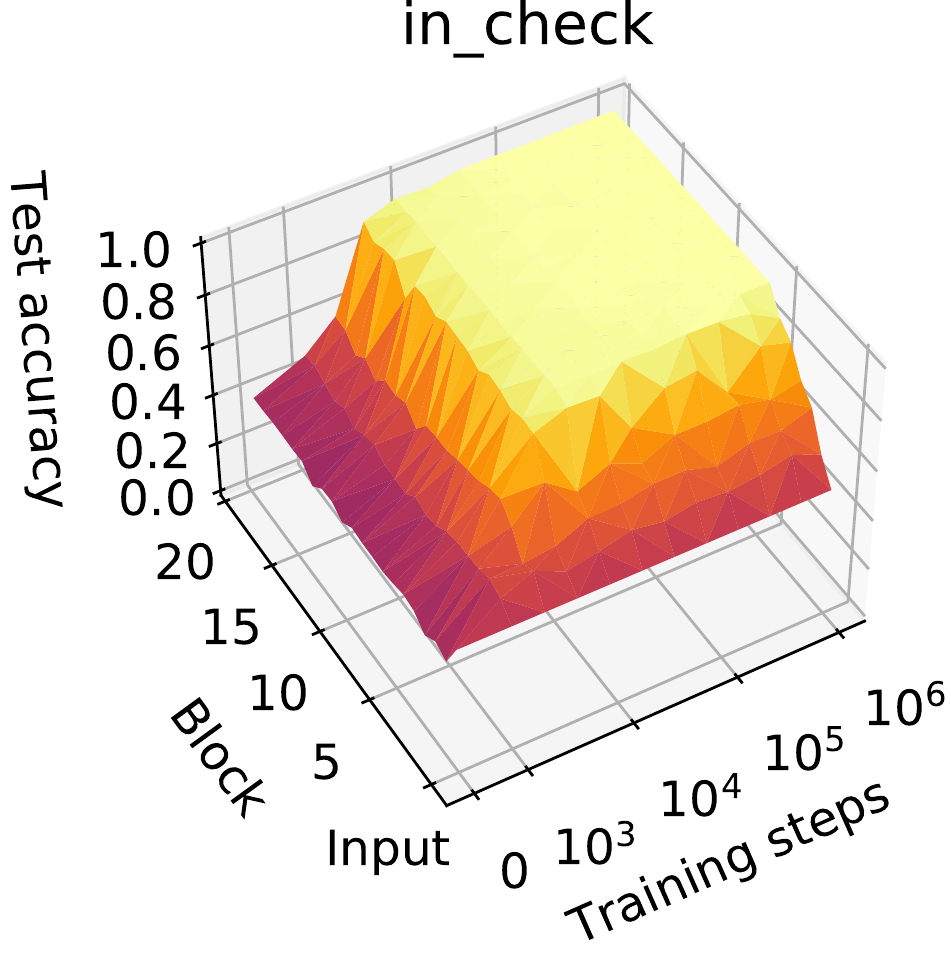}
\caption{Is the playing side in check?}
\label{fig:concept_regression-in-check}
\end{subfigure} \\
\vskip 20pt

\begin{subfigure}[t]{0.3\textwidth}
\vskip 0pt
\centering
\includegraphics[width=\textwidth]{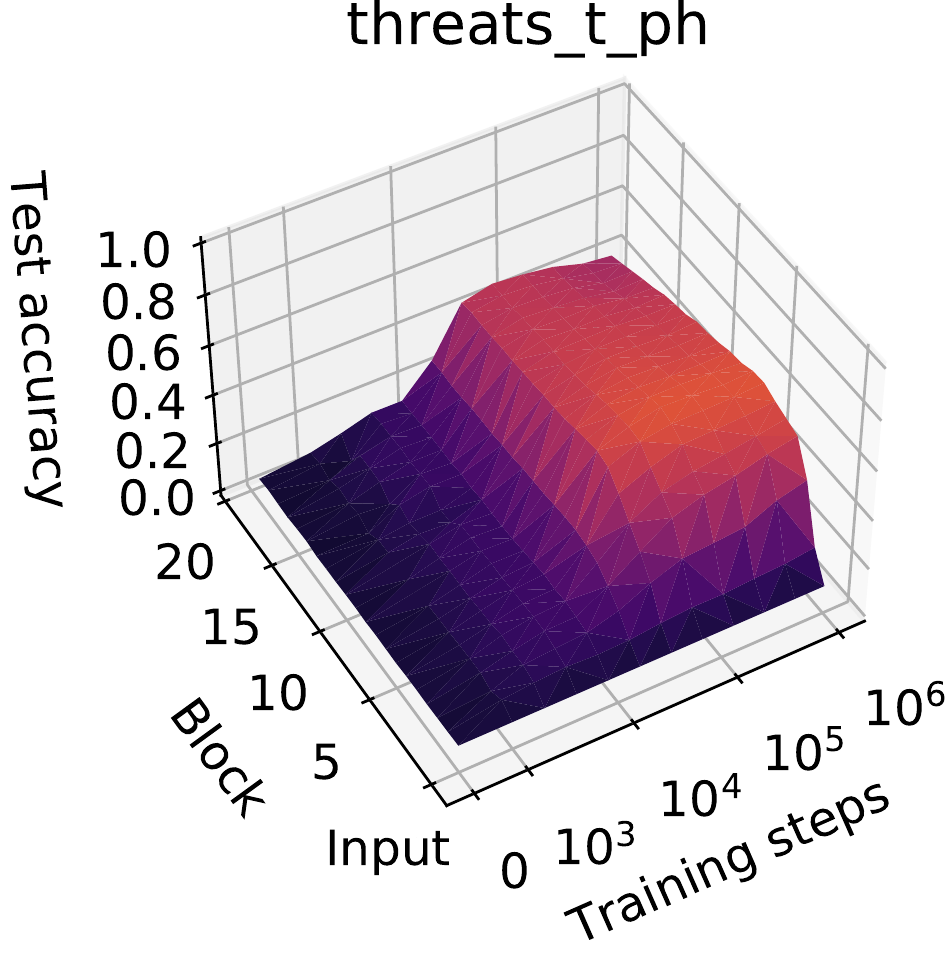}
\caption{Stockfish 8's evaluation of threats.}
\label{fig:concept_regression-threats}
\end{subfigure}
\hfill
\begin{subfigure}[t]{0.3\textwidth}
\vskip 0pt
\centering
\includegraphics[width=\textwidth]{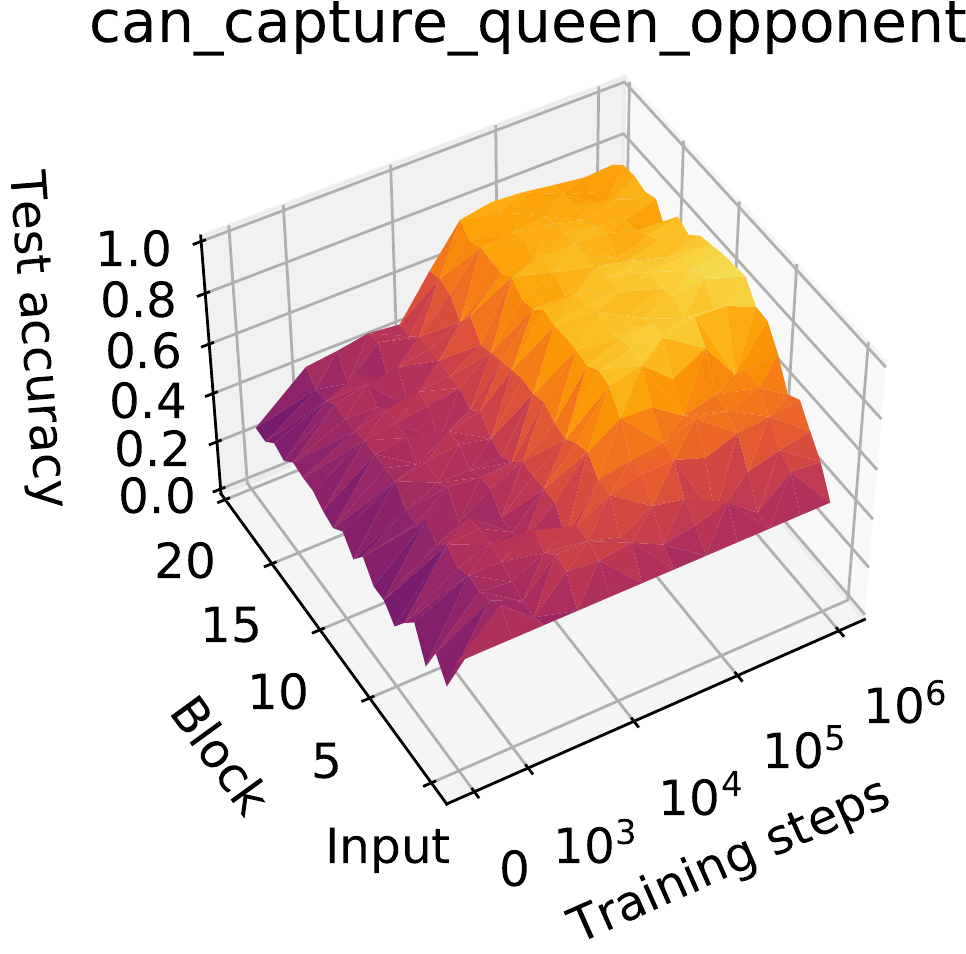}
\caption{Can the playing side capture their opponent's queen?}
\label{fig:concept_regression-capture-queen}
\end{subfigure}
\hfill
\begin{subfigure}[t]{0.3\textwidth}
\vskip 0pt
\centering
\includegraphics[width=\textwidth]{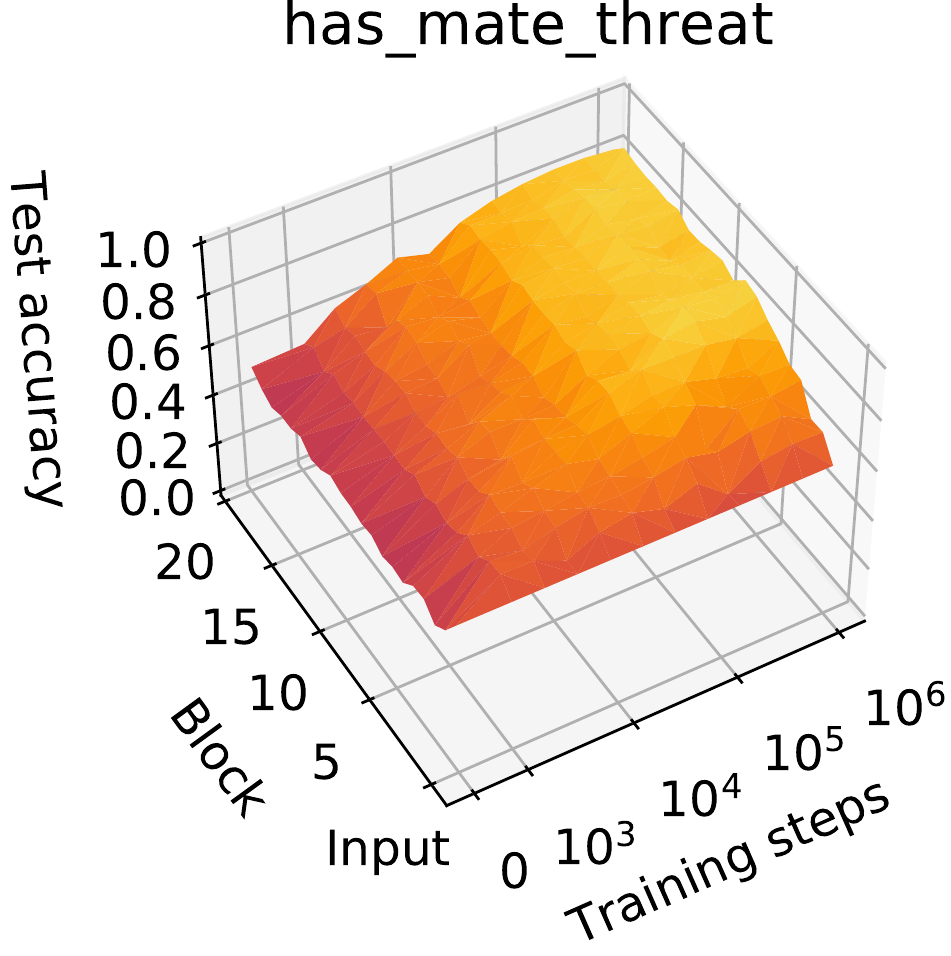}
\caption{Could the opposing side checkmate the playing side in one move?}
\label{fig:concept_regression-checkmate-threat}
\end{subfigure}
\\
\vskip 20pt
\begin{subfigure}[t]{0.3\textwidth}
\vskip 0pt
\centering
\includegraphics[width=\textwidth]{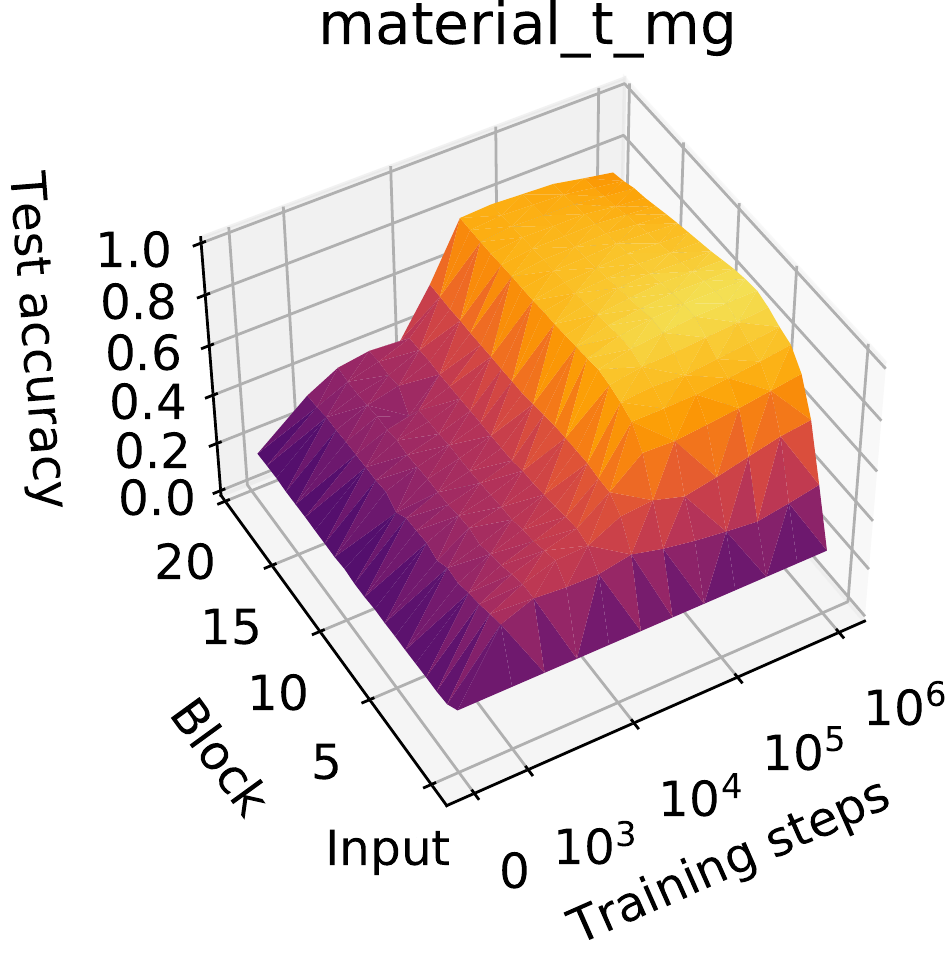}
\caption{Stockfish material score}
\label{fig:concept_regression-material}
\end{subfigure}
\hfill
\begin{subfigure}[t]{0.3\textwidth}
\vskip 0pt
\centering
\includegraphics[width=\textwidth]{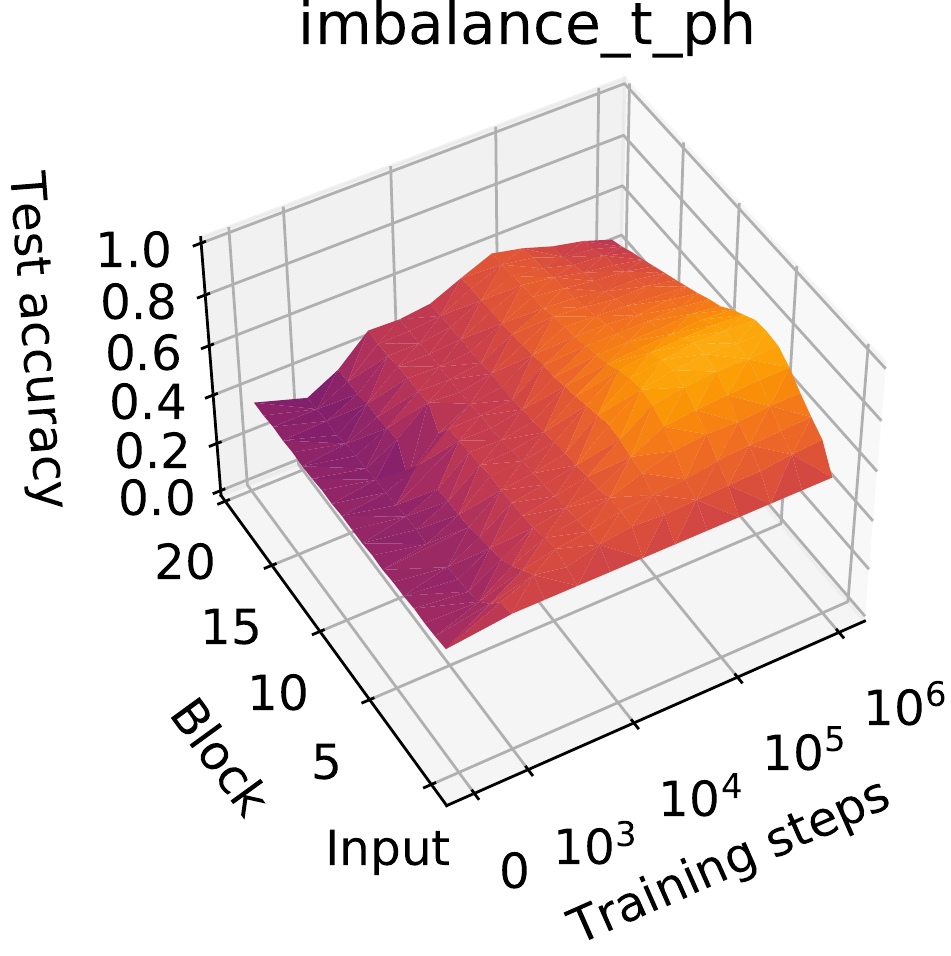}
\caption{Past $10^5$ training steps, StockFish 8's material imbalance score becomes
\emph{less} predictable from AlphaZero's later layers.}
\label{fig:concept_regression-material-imbalance}
\end{subfigure}
\hfill
\begin{subfigure}[t]{0.3\textwidth}
\vskip 0pt
\centering
\includegraphics[width=\textwidth]{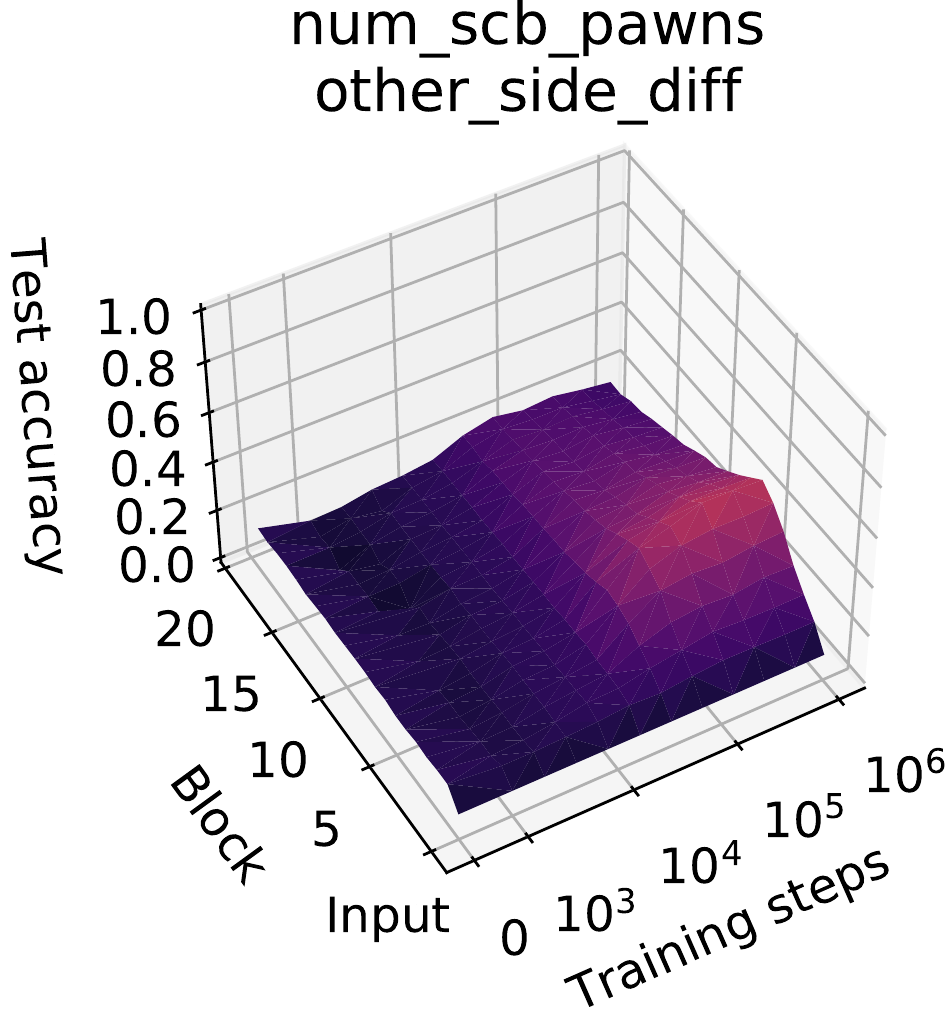}
\caption{When each side has one bishop only,
the difference in number of pawns on squares of the \emph{opposite} colour than the opponent's bishop.}
\label{fig:concept_regression-num-scb-pawns-same-side}
\end{subfigure}
\caption{What-when-where plots for a selection of Stockfish 8 and custom concepts.
Following Figure \ref{fig:alphazero-network}, we count a ResNet `block' as a layer.}
\label{fig:concept_regression}
\end{figure}

At a high level, the data shown in Figure~\ref{fig:concept_regression} and
Appendix~\ref{sec:concept_regression_second_seed} show that information related to human-defined concepts of multiple levels of complexity is being learned over the course of training, and that many of these concepts are computed over the course of multiple blocks. Many surprisingly complex concepts can be regressed with high accuracy, and this accuracy increases substantially over the course of training. Many concepts begin to increase in accuracy around 32,000 steps, which appears to be a period of rapid development in both AlphaZero's representations and opening play (as we demonstrate in later sections).

\paragraph{Stockfish 8 score} We find a surprising level of accuracy on Stockfish 8 total score
\verb=total_t_ph=, with $r^2>0.75$ after 10 layers at 64,000 training steps and beyond;
see Figure \ref{fig:concept_regression-total-score}.
We explore Stockfish 8 score further in Section~\ref{sec:residuals} below. Regression accuracy for Stockfish 8 score is very low early in training, reflecting the complexity of this concept, and only begins to increase substantially after 16,000 steps before plateauing at 128,000 training steps. This pattern occurs repeatedly across a wide range of concepts, and we also observe a rapid change in the policy prior during this window, which we investigate further in Section~\ref{sec:explosion}. 

\paragraph{Threat-related concepts} The progressive increase in regression accuracy for check, queen capture and threats over the initial layers of the trained network indicates that computation of potential moves takes place over the course of early network layers, an observation which is corroborated by studying network activations directly in Section~\ref{sec:intermediate_computations}. Relatively low regression accuracy for threats may be due to a difference in evaluating the value of different threats, or due to highly-distributed representations of threats being penalised by sparsity.
High accuracy for \verb=can_capture_queen_opponent= and \verb=has_mate_threat=
in Figures \ref{fig:concept_regression-capture-queen}
and \ref{fig:concept_regression-checkmate-threat}
suggest that distributed representations may be at least partly responsible, as we can accurately predict more specific threat-related concepts. 

Accuracy for \verb=has_mate_threat= rises throughout the network, suggesting that further processing of threats occurs in later layers, and that AlphaZero is predicting the consequences of its opponent's potential moves. The ability to predict \verb=has_mate_threat= from AlphaZero's activations indicates that AlphaZero is not simply modelling its potential moves, but also its opponent's potential moves and their consequences during position evaluation. This is interesting in light of AlphaZero's training loss $\mathcal{L}$:
\begin{equation}\label{eq:az_policy_loss}
    \mathcal{L} = - \bm{\pi}^T \log \p + \text{value and regularisation terms}
\end{equation}
for policy head output $\p$ and
returned MCTS probability vector $\bm{\pi}$~\cite{Silver1140}. This loss indirectly pushes the policy network to predict the consequences of its actions. Our analysis suggests that training using Equation~\ref{eq:az_policy_loss} has led to at least a single-move lookahead being implemented in the AlphaZero network, rather than this being deferred to MCTS rollouts.

\paragraph{Material} Earlier analysis of AlphaZero's play suggested that \cite{game_changer} that AlphaZero views material imbalance differently from Stockfish 8. The what-when-where plot of Figure \ref{fig:concept_regression-material-imbalance} gives \emph{empirical} evidence that this is the case at the representational level; the test $r^2$ drops for later layers after $10^5$ training steps. Note that Stockfish's material evaluation is not simply a linear combination of piece counts, but also includes position-specific evaluation terms. We also include a simpler material concept measuring only piece count in Appendix~\ref{sec:all_concepts_regression_supplement}.
This simpler material evaluation can be trivially regressed for either player at all network layers (including the input) and at all points during training. This is unsurprising because piece counts are easily available from the input planes. Material difference only becomes predictable after training, and increases with network depth, even for this simpler concept of material. This indicates that our level of sparsity is such that the regression probe must use only a limited part of the input or internal activations, and that material difference is relatively compactly encoded in later layers of the network, i.e. the regression probe is making use of a specific evaluation of material, rather than simply using representations of piece locations.

\paragraph{Drop in linearly-available information} What-when-where plots for
threats (\verb=threats_t_ph=; Figure \ref{fig:concept_regression-threats}),
piece count difference
(\verb=material_t_mg=; Figure \ref{fig:concept_regression-material}) and
imbalance
(\verb=imbalance_t_ph=; Figure \ref{fig:concept_regression-material-imbalance})
reveal a surprising pattern: for well-trained networks, regression accuracy on these concepts
peaks at an early layer before dropping dramatically in later layers. The drop in accuracy indicates that information on these concepts is either being used in further computations and then discarded, or encoded in a more complex nonlinear way, indicating that these concepts are likely to be used as intermediate computations which are useful for later layers but do not feed directly into policy or value calculations. This suggests that the common practice of visualising only last-layer activations (for instance via t-SNE) may miss the presence of important information in complex domains, and underlines the importance of considering intermediate computations for understanding neural networks on complex domains.

\paragraph{Some concepts cannot be regressed, or are trivial to predict} Not all concepts can be regressed from the AlphaZero network, however. Some, such as the number of same colour bishop pawns, have low regression accuracy even at their peak, and a few have constant near-zero accuracy
(see Appendix
\ref{sec:all_concepts_regression_supplement},
Figures \ref{fig:all_sf_concepts_1} to
\ref{fig:custom_pawn_concepts_2} for more examples
).
Concepts with low regression accuracy are disproportionately associated with pawns, and these may be more heavily penalised by sparsity than other concepts, as there are typically more pawns on the board than other pieces. Other concepts (shown in Supplementary Information) have constant high regression accuracy, indicating either a low variety in the number of possible values for the concept, or that they can be trivially regressed from the input.

\paragraph{Implications} Under the assumption that human concepts are represented when they are linearly predictable from a layer,
these results suggest that AlphaZero is developing representations which are closely related to a number of human concepts over the course of training, including high-level evaluation of position, potential moves and consequences, and specific positional features.
The fact that accurate regression can be attained with only a small fraction of neurons at each layer is striking, and indicates that channels learn to specialise in performing distinct functions, and that these functions align to some degree with human concepts. When the network is well-trained many of these these concepts are computed progressively over multiple layers, as shown by the increase in accuracy as network depth increases.
Interestingly, some of these concepts cannot be as accurately predicted from later layers (as indicated by a decrease in test $r^2$), showing that information is either lost in network computations or becomes non-linearly encoded and that not all of the information relevant to network computation is linearly available in the last layer.

\subsection{Systematic semantic differences in Stockfish score outliers}
\label{sec:residuals}

There are instances when the AlphaZero network's value head and Stockfish 8's evaluation function take on fundamentally different roles, purely because of the way \emph{search} differs between the engines.
AlphaZero's MCTS runs simulations up to a fixed ply depth. The AlphaZero network therefore 
has to encode some minimal form of \emph{look-ahead};
we've already seen evidence of this in
Figure \ref{fig:concept_regression-checkmate-threat}
which indicates that the network encodes whether the opposing side can checkmate the playing side in one move.
This minimal form of look-ahead would also be required when an exchange sequence is simulated only halfway before the
maximum MCTS simulation depth is reached. 
Stockfish, on the other hand, dynamically increases evaluation depth during exchange sequences.
There is information that comes from looking ahead that Stockfish's evaluation function
\emph{doesn't have to encode},
simply because there is other code that ensures such information is incorporated in the final evaluation.

Remarkably, differences between the two styles of chess engines
surface in the positions that are outliers in Stockfish 8 score
\verb=total_t_ph= concept regression.
The prediction error outliers have semantic meaning, which we explore in this section.

\begin{figure}
    \centering
    \includegraphics[width=0.9\textwidth]{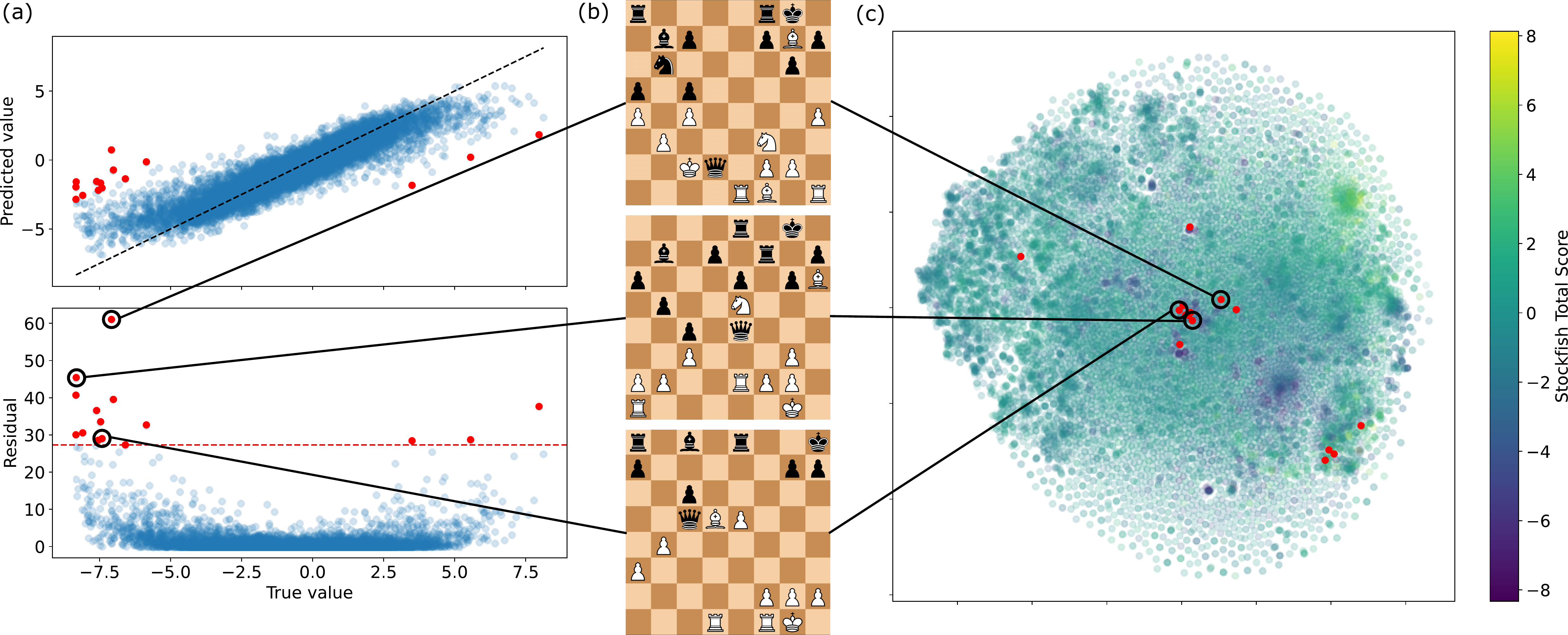}
    \caption{Evidence of patterns in regression residuals. \textbf{(a)} Upper panel: true and predicted values for Stockfish 8 score \texttt{total\_t\_ph}, regressed from block 10 at 1,000,000 training steps, evaluated on the test set.
    Dashed line indicates perfect fit. Red markers indicate predictions in the 99.95th percentile of residuals. Lower panel: True value and residual for Stockfish 8 score, as in top panel. Red dashed line indicates 99.95th percentile cutoff. \textbf{(b)} High-residual positions, corresponding to the data points marked in (a) and (c). Note that Black's queen can be taken in all positions. This is true for all 12 high-residual pieces where the regressed score is more favourable to White than the Stockfish score. \textbf{(c)} t-SNE of block activations at 1,000,000 steps. Red markers indicate high-residual positions shown in (a), showing substantial clustering.}
    \label{fig:residuals}
\end{figure}

\paragraph{Why can we learn from prediction errors?}
In the previous section (\ref{sec:concept_regression}),
we showed that many complex concepts can be predicted with surprisingly high accuracy by probing the AlphaZero network. However, in almost all cases concept regression was not perfect.
In this section we investigate whether we can learn anything from these prediction errors.
It may seem surprising to expect this to be possible, as prediction errors could simply be noise with no interesting structure, the relevant data may not be present in AlphaZero's activations, or sparsity constraints may artificially limit regression accuracy (as we discuss in Section~\ref{ss:probing-challenges}). In these cases we would expect to learn little or nothing from prediction errors.
For concepts which are simple and objective (for example \verb=has_contested_open_file=) one or more of these possibilities is very likely to be true. Where concepts are more complex or have an element of subjectivity, however (for example \verb=total_t_ph=), there is another possibility:
regression errors may point to a `difference of opinion'.
We alluded to the example of evaluating the Stockfish 8 score during an exchange sequences,
where Stockfish can afford to differ from AlphaZero, simply because their search algorithms differ in fundamental ways.
Prediction errors due to `differences of opinion' should appear as consistent structure in the regression errors: inputs on which the network fails to predict well should have something in common.

\paragraph{Interpretable structure in Stockfish score regression residuals}
We investigated this possibility using the Stockfish score concept \verb=total_t_ph=.
For each position $n$ in the test set we computed the Stockfish 8 score concept
which we denote as $c_n = c_{\texttt{total\_t\_ph}}(\z^0_n)$ in this section.
Following Equation \ref{eq:concept-regression-continuous},
we additionally compute the predicted score
$\hat{c}_n = g_{lt}^{\texttt{total\_t\_ph}}(\z^{l_t}_n)$ from the linear regression model trained on block $l = 10$ of the $t = 1,000,000$ steps checkpoint.
The block and checkpoint were 
chosen because regression achieves high accuracy at this point.
We then calculated the residuals
\begin{equation}
    \varepsilon_n = (c_n - \hat{c}_n)^2.
\end{equation}
These residuals are shown in the bottom panel of Figure~\ref{fig:residuals}a. We then visualised the positions corresponding to the residuals in the 99.95th percentile, representing the positions with the most extreme prediction errors.
These outliers are marked in red in Figure~\ref{fig:residuals}a, and a selection are shown in Figure~\ref{fig:residuals}b.
In all 12 outlier positions where the regressed score is more favourable to White than the Stockfish score, Black's queen can be taken, often without requiring an exchange: the `difference of opinion' is maximal halfway through a sequence of moves exchanging queens.
This suggests that the regression model is picking up on White's ability to take Black's queen, rendering these positions more favourable than the Stockfish score would suggest.
Although this case is likely to be exceptional, and in many cases prediction errors are likely to indicate a simple failure of the regression model, this example shows that we can sometimes learn more from the regression probes trained when generating what-when-where plots.
Furthermore, the outlier positions also cluster in activation space: Figure~\ref{fig:residuals}c shows a t-SNE projection~\cite{van2008visualizing} of AlphaZero activations from the same layer and checkpoint, coloured by Stockfish score.
Outliers are again shown by red markers, and show a surprising level of clustering. Extreme positive and negative values of Stockfish score also cluster, and many of these clusters do not contain any outlier values, suggesting that the similarity in the outliers is due to representational similarity, rather than simply a common feature of low Stockfish score positions.

Although the degree of structure shown here is surprising, this example is not cherry-picked;
it was the first concept/layer/checkpoint combination we tried, and the positions presented are simply those whose residuals are past a cutoff that we did not tune.

\subsection{Challenges and limitations for concept probing}\label{ss:probing-challenges}
\paragraph{What's the right probing architecture?} In this work we have used sparse linear probes because they are a simple to understand and low-capacity regression model. Keeping the capacity of the regression model low is important in order to ensure that the model captures structure in AlphaZero's representations, rather than learning complex relationships of its own, but we believe that better probing architectures are possible. Although sparsity is essential due to the high dimensionality of neural network activations, it can cause difficulties with spatially-represented concepts, which AlphaZero's architecture is naturally biased towards. For example, consider a channel representing the possibility of capturing the opponent's queen by a positive activation at the position of the queen and zero activations otherwise. There are locations on the board in which the queen is more likely to be captured than other locations, and therefore there are levels of sparsity that lead to positive regression weights only at the positions where the queen is most likely to be captured. In this (hypothetical) example, all of the information is there in the network, but a sparse linear regression model is unable to capture all of it and thus will have $r^2<1$. More sophisticated regression methods such as data-dependent sparsity using Gated Linear Networks~\cite{veness2019gated}, information-theoretic regularisation using the information bottleneck~\cite{schulz2020restricting}, minimum description length probing~\cite{voita2020information} or Bayesian probing~\cite{pimentel2021bayesian} may provide ways to mitigate this problem.

\paragraph{How should we interpret complex or subjective concepts?} Many concepts in our dataset have a subjective element as well as a more objective one. Some concepts, for instance \verb=has_contested_open_file=, simply indicate the existence of a specific feature, whereas others such as \verb=threats= combine both the presence of various threats as well as a subjective assessment of their value. Complex concepts like this can pose a challenge to interpreting regression results: when regression accuracy is low, is this because the components of the concept aren't present (i.e. the relevant threats aren't being computed by the model) or because our assessment of the value of those threats differs from that of the network? Our evidence in Sections~\ref{sec:concept-regression} and~\ref{ss:nmf} indicate that threats are being computed, suggesting that either our assessment of the value of different threats differs from AlphaZero's or much of the relevant information in the network remains distributed or nonlinearly encoded throughout the residual stack.

Complex concepts often incorporate a degree of judgement or subjectivity. In our experiments, we rely on the components of the public API of Stockfish 8, along with a number of independently implemented low-level features, to provide us with a notion of \emph{ground truth}, which we then subsequently try to identify within the AlphaZero network
This choice is in itself arbitrary, as there would be differences between different chess engines and their implementations, as well as across different versions of Stockfish.
It should also be clear that the positional assessment of any individual chess grandmaster isn't equivalent to that of Stockfish's evaluation function.

\paragraph{When can we definitively say a concept is represented?} There is no clear line between $r^2=0$ and $r^2=1$ at which we can say a concept is definitely present or absent. More sophisticated probing methodologies (as discussed above) could go some way to alleviating this problem by allowing better use of available data, which would either increase regression accuracy to a level where we are confident the concept is being represented or increase our confidence that the relevant concept is not present. The comparative approach introduced by what-when-where plots means that we are able to tell when linearly-decodable information is being generated by the network, and sharp jumps to high regression values intuitively seem (at least to us) to provide stronger evidence of a concept being computed by the network. Understanding prediction failures can also be useful, particularly in the case of complex composite concepts: if a concept is accurately predicted half of the time, and completely wrongly the other half, then we may be able to refine our concepts to better reflect network activations by understanding what the successes or failures have in common. We give an example of this in Section~\ref{sec:residuals}.

Spatial representations can also lead to $r^2<1$ for reasons unrelated to sparsity: consider the previous example of determining whether the opponent's queen can be captured. As we demonstrate in Section~\ref{ss:nmf} the set of possible moves is developed over the course of multiple network layers. This is necessarily the case because convolutional blocks can capture only local information: a 3x3 convolutional kernel must take several layers to propagate a bishop's diagonal moves across a chessboard, for instance. Because of this gradual development, only potential captures where the capturing piece is near the queen can be regressed from the earliest layers, with longer-distance captures becoming predictable later in the network. This will lead to a steady increase in regression accuracy with network depth. We notice this pattern of gradual increase in a broad range of threat-related concepts.

\paragraph{When is a network `really' representing or using a concept?} High regression score is only a proxy of concept encoding. It is entirely possible that the network contains a concept that confounds with the human concept, rather than a representation of the concept itself. When we train a probe we cannot tell if we are getting a confounder or the concept itself. Our use of held-out validation and test sets ensure that any such confounders carry over accurately to novel positions, but cannot determine whether the network is `really' computing a given concept (a philosophically tricky question in itself). Dealing with confounders is typically achieved by intervention, but the correct way to intervene is not obvious. We could intervene on the input to the network (as is done in~\cite{goyal2019explaining}) or use instrumental variables (as is done in~\cite{bahadori2020debiasing}), but even with such measure, many concepts are still entangled with one another. For example, consider a situation where a queen is in an absolute pin (it cannot be moved without putting the king in check). In this case, we cannot intervene to set the concept \verb=can_capture_queen= to 0 without setting \verb=in_check= to 1, in addition to altering a whole range of other concepts. Alternatively, we could intervene on the network's internals directly, altering the value of the activations from which we can regress \verb=can_capture_queen= and seeing what other concepts change in later layers. This is not, strictly speaking, an intervention on the concept itself, but is rather an intervention on the parts of the network that we believe encode the relevant concept. As such it would whether the network is using the concept, rather than the effect of changing the external situation to alter the value of the concept. The challenge here is that such an intervention may push the network far off-distribution and result in incoherent downstream computation.

\subsection{Relating human concepts to AlphaZero's value function}
\label{sec:value-regression}
We have found that many human-defined concepts can be predicted from intermediate representations in AlphaZero when training has progressed sufficiently (Section~\ref{sec:concept-regression}), but this does not determine how these concepts relate to the outputs of the network. In this section we investigate the relation of concepts (piece count differences and high-level Stockfish concepts) to AlphaZero's value predictions. We use concept values to predict the output of AlphaZero's value function using a linear predictor. By analysing the regression coefficients we can investigate how AlphaZero's value function relates to human concepts over the course of training. A schematic of our approach is shown in Figure~\ref{fig:value_regression-schema}, and full details are given below.

Equation \ref{eq:az-network} stated the AlphaZero neural network as
$\p, v = f_{\btheta_t}(\z^0)$ at training step $t$.
In this section we consider only the value head output, which we denote by $v_{\btheta_t}(\z^0)$.
Given a vector of concept values
$\c(\z^0) = [ c_1(\z^0), c_2(\z^0), \ldots, c_J(\z^0) ]$ as input vector,
we train a generalized linear model to predict
$v_{\btheta_t}(\z^0)$ from these concepts.
The generalized linear model has weights $\w$ and bias $b$,
\begin{equation}
\label{eq:value_prediction}
\hat{v}_{\w, b}(\mathbf{z}^0) = \tanh\big(\w^T \c(\mathbf{z}^0) + b\big) \ .
\end{equation}
The use of $\tanh$ nonlinearity is because $v_{\btheta_t}(\z^0) \in (-1, 1)$, and it also corresponds to the final non-linear function in the value head; see
Figure \ref{fig:alphazero-network}.
The weights $\w$ and biases $b$ are trained using the $L_1$ loss
\begin{equation}
\w_t, b_t = \min_{\w, b} \frac{1}{N} \sum_n \Big|\hat{v}_{\w, b}(\mathbf{z}_n^0) - v_{\btheta_t}(\z_n^0) \Big| \ .
\end{equation}
The weights $\w_t$ and biases $b_t$ will differ between checkpoints $t$ because AlphaZero's value network changes over the course of self-play, leading to different $v_{\btheta_t}$ outputs. We train on $N = 10^5$ training positions, deduplicated as in Section~\ref{sec:concept-regression},
before testing on $3\times 10^4$ held-out test positions.
We use the $L_1$ loss rather than $L_2$ in training as we found that the $L_2$ loss systematically underestimated piece weights.
\begin{figure}
\centering
\begin{subfigure}[t]{0.3\textwidth}
\centering
\includegraphics[width=0.7\textwidth]{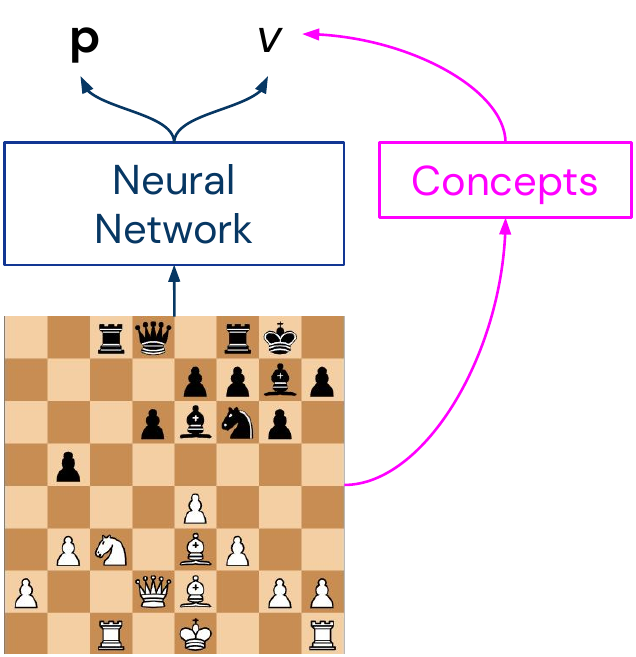}
\caption{Value regression methodology: we train a generalized linear model on concepts to predict AlphaZero's value head for each neural network checkpoint.}
\label{fig:value_regression-schema}
\end{subfigure}
\begin{subfigure}[t]{0.34\textwidth}
\includegraphics[width=\textwidth]{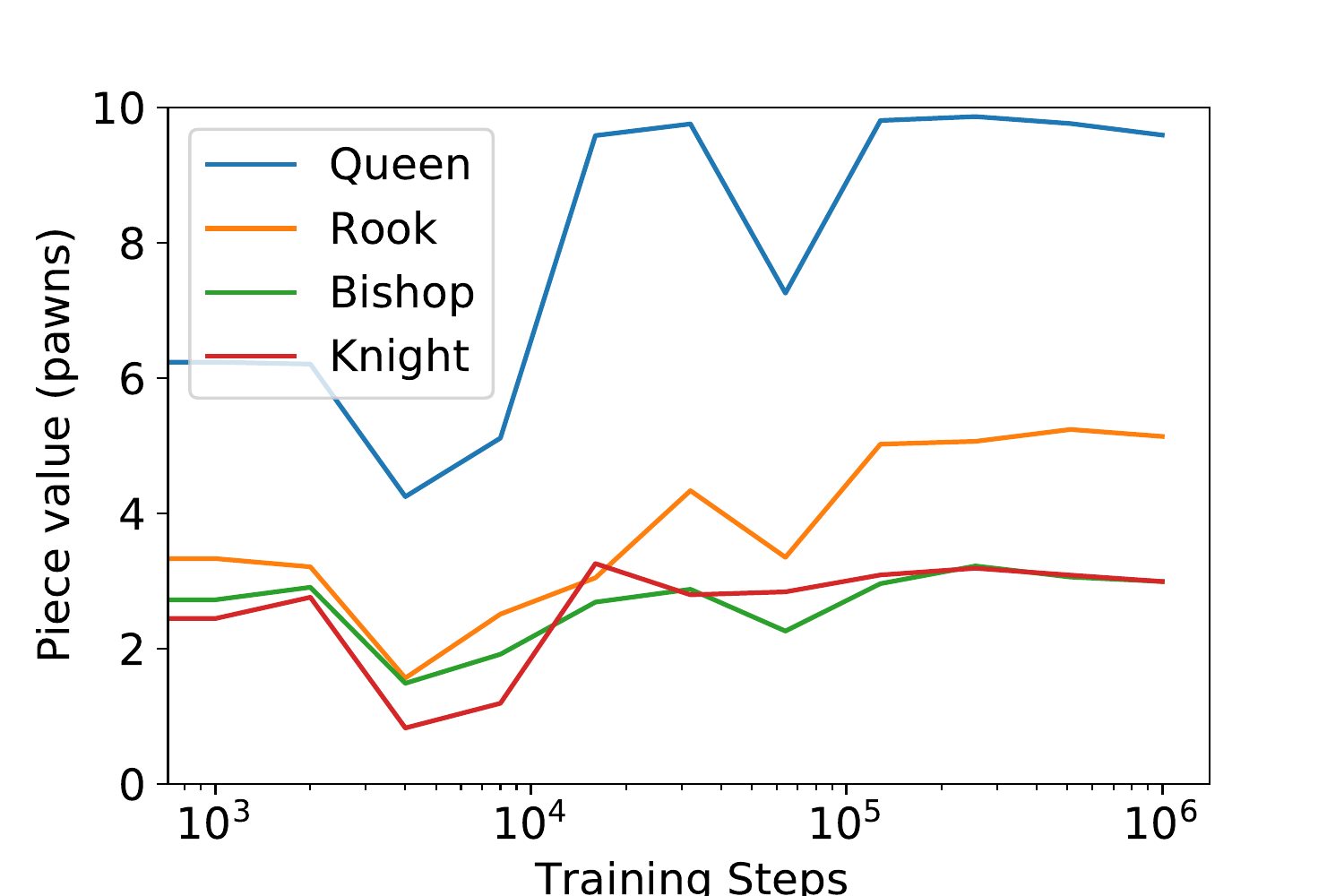}
\caption{Piece value weights converge to values close to those predicted by conventional \\ theory.}
\label{fig:value_regression-pieces}
\end{subfigure}
\begin{subfigure}[t]{0.34\textwidth}
\includegraphics[width=\textwidth]{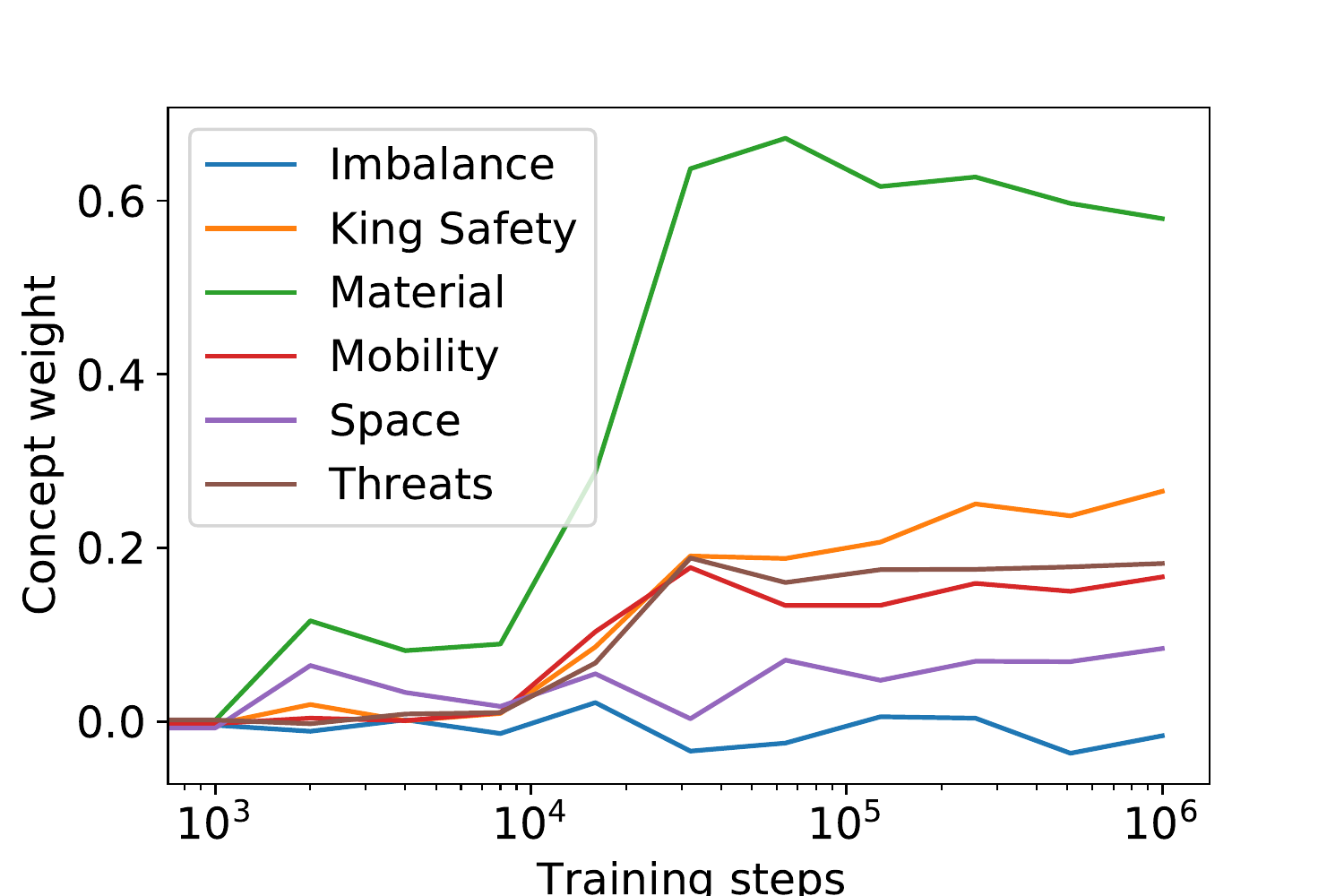}
\caption{Material predicts value early in training, with more subtle concepts such as mobility and king safety emerging later.}
\label{fig:value_regression-stockfish}
\end{subfigure}
\caption{Value regression from human-defined concepts over time.}
\label{fig:value_regression}
\end{figure}

\paragraph{Piece value}
Simple values for material are one of the first things a beginner chess player learns, and allow for basic assessment of a position.
We began our investigation of the value function by using only piece count difference as the `concept vector'.
In this case, similar to \cite{az_variants_preprint}, the vector
$\c(\z^0) = [ d_{\textrm{\sympawn}}(\z^0),
d_{\textrm{\symknight}}(\z^0), d_{\textrm{\symbishop}}(\z^0),
d_{\textrm{\symrook}}(\z^0), d_{\textrm{\symqueen}}(\z^0) ]    
$
contains the difference $d$ in numbers of pawns, knights, bishops, rooks and queens between the current player and the opponent.
For piece value regression we use only positions where at least one piece count differs between White and Black.
The evolution of piece weights are shown in Figure~\ref{fig:value_regression-pieces}, showing that piece values converge towards commonly-accepted values after 128,000 training steps.

\paragraph{Higher-level concepts}
We furthermore explored the relative contributions of high-level Stockfish concepts
imbalance, king safety, material, mobility, space and threats
(see Table~\ref{tab:concepts} for details)
to predicting $v_{\btheta_t}(\z^0)$.
We normalized the high-level Stockfish concepts by dividing by their
respective standard deviations $c' = c / \textrm{std dev}$,
and use the vector
$
\c(\z^0) = [
c_{\texttt{imbalance}}'(\z^0), \,
c_{\texttt{king\_safety}}'(\z^0), \,
c_{\texttt{material}}'(\z^0), \,
c_{\texttt{mobility}}'(\z^0), \,
c_{\texttt{space}}'(\z^0), \,
c_{\texttt{threats}}'(\z^0) ]
$
with the concepts' `\texttt{t\_ph}' function calls\footnote{See Table \ref{tab:concepts} for reference. `\texttt{t}' stands for the `total' side (White-Black difference), while `\texttt{ph}'
indicates a phased value, a composition that is made up of a
middle game `\texttt{mg}', end game `\texttt{eg}' and the phase of the game evaluations.}
in the generalized linear model of \eqref{eq:value_prediction}.
Figure \ref{fig:value_regression-stockfish} illustrates the progression of the weights $\w_t$ over training steps $t$.
Material is the first concept to be learned, consistent with initial human learning, and begins to be used at 16,000 steps.
More sophisticated concepts, such as king safety, threats, and mobility, are used at 32,000 steps and beyond.
This coincides with the point at which many of these concepts begin to be accurately regressed from some layers of the neural network.
The space concept has a comparatively low weight, and emerges late in training.
The imbalance concept receives a small negative weight,
and `imbalance' makes a negligible contribution to linearly predicting $v_{\btheta_t}(\z^0)$.
Interestingly, Figure \ref{fig:concept_regression-material-imbalance} shows that regression accuracy for the imbalance concept declines
in later layers at all points in training,
suggesting that this information is not preserved in a linear form in the later layers of the network.

\section{Progression through AlphaZero and human history}
\label{sec:human-history}

In this section we depart from the progression of human concepts in the AlphaZero network as the system trains, and compare AlphaZero training to the progression of human knowledge.

There is a marked difference between AlphaZero's progression of move preferences through its history of training steps,
and what is known of the progression of human understanding of chess since the fifteenth century.
AlphaZero starts with a uniform opening book, allowing it to explore all options equally, and largely narrows down plausible options over time.
Recorded human games over the last five centuries point to an opposite pattern:
an initial overwhelming preference for 1.~e4, with an expansion of plausible options over time.

\subsection{Five centuries of data}

We base our analysis on a subsample of games from the ChessBase Mega Database \cite{chessbase}.
Much of the early history of chess has not been recorded.
The earliest recorded game in the data was played in Valencia in 1475,
and we use all 135 available recorded games played between
the years
1400 and 1800.
It is followed by 1175 games played between 1800 and 1850,
and a further 9615 games played between 1850 and 1900.
From the twentieth century onward we used roughly
10,000 games per decade,
either by randomly sampling games before 1970 subject to a minimum length of 20 moves,
or taking the games played by players with the highest ELO ratings after 1970.
It is true that there is a difference in the average quality of games recorded across different periods,
especially given the emergence of professional chess players
and the benefits of computer chess engines analysis in the recent years.
However, these are all part of the human progression of the understanding of chess.

\subsection{First move progression}
\label{sec:first-move-progression}

\begin{figure}[t]
\centering
\begin{subfigure}{\textwidth}
\centering
\includegraphics[width=0.7\linewidth,valign=t]{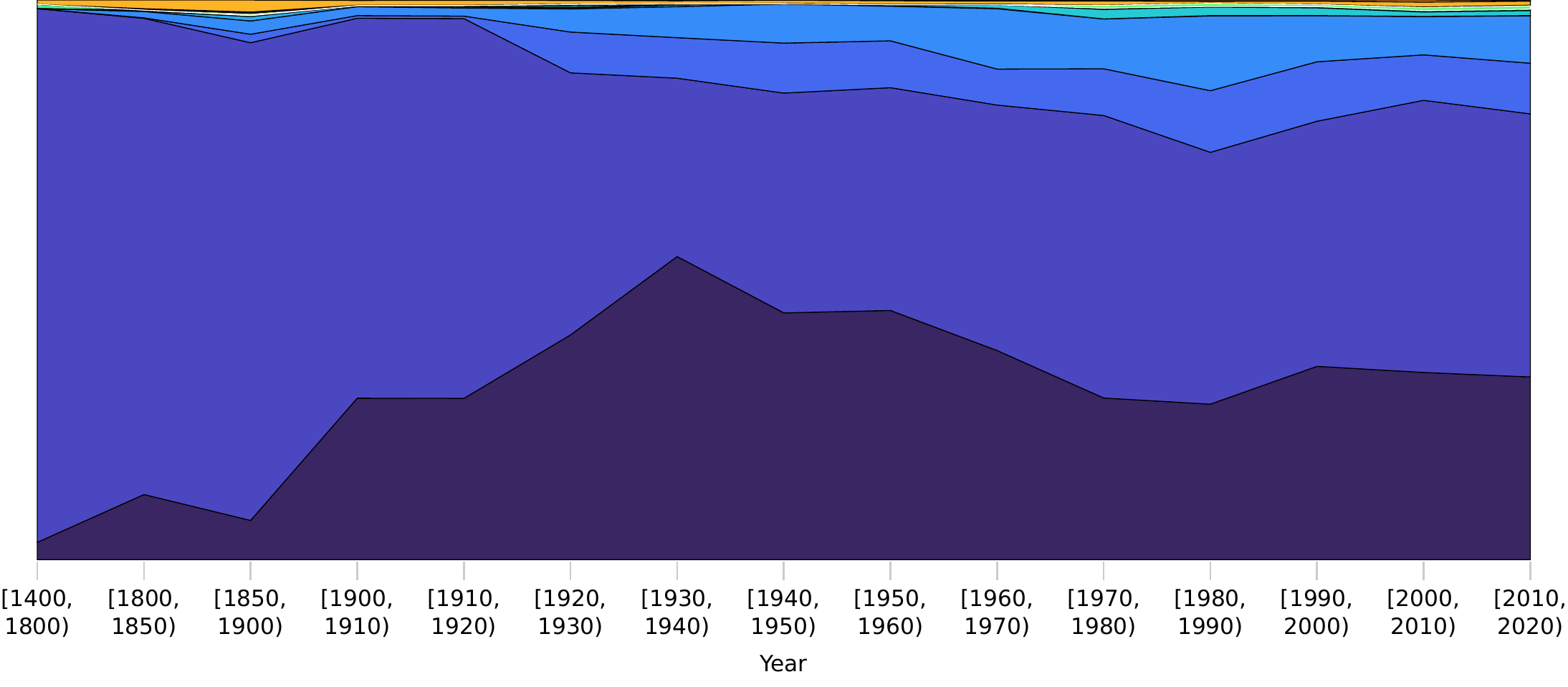}
\includegraphics[width=0.04\linewidth,valign=t,trim={0 0 7.5cm 0}]{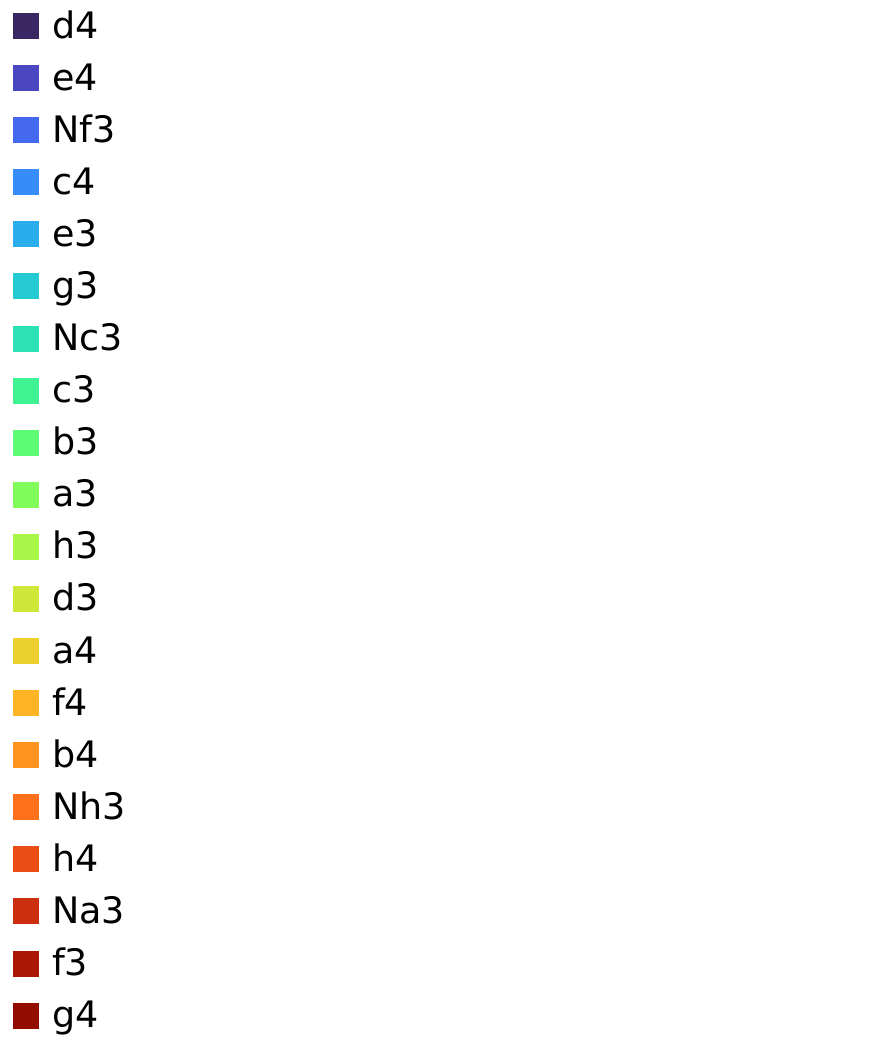}
\caption{The evolution of the first move preference for White over the course of human history, spanning back to the earliest recorded games of modern chess in the Chessbase database.
The early popularity of 1.~e4 gives way to a more balanced exploration of different opening systems and an increasing adoption of more flexible systems in modern times.}
\label{fig:human-history-starting}
\end{subfigure} \\
\vspace{10pt}
\begin{subfigure}{\textwidth}
\centering
\includegraphics[width=0.3\linewidth,valign=t]{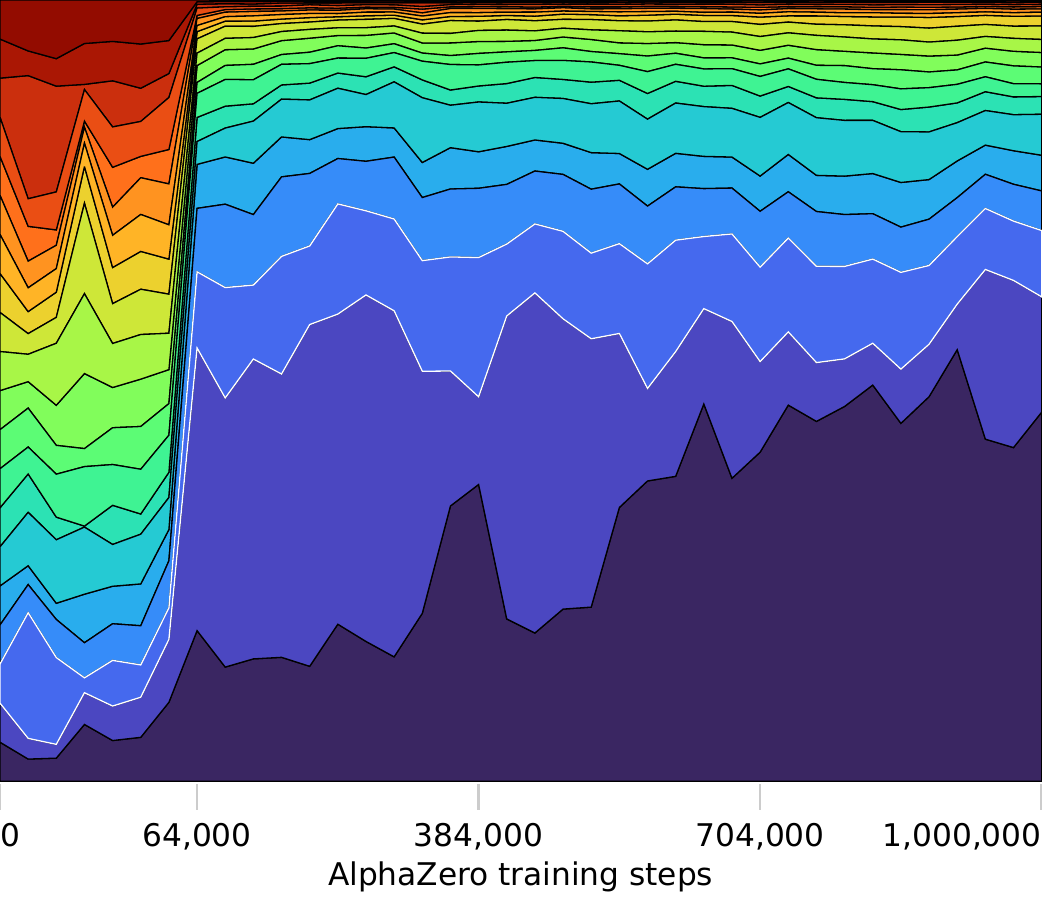} \hfill
\includegraphics[width=0.3\linewidth,valign=t]{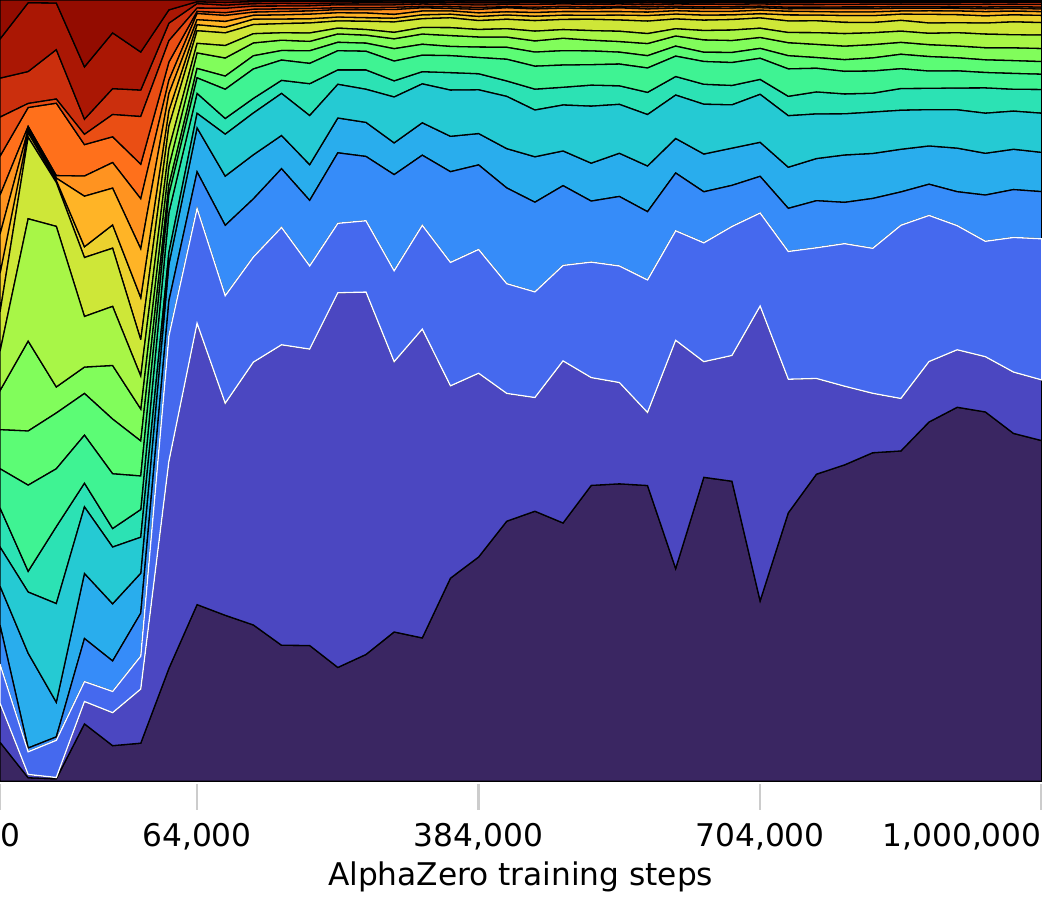} \hfill
\includegraphics[width=0.3\linewidth,valign=t]{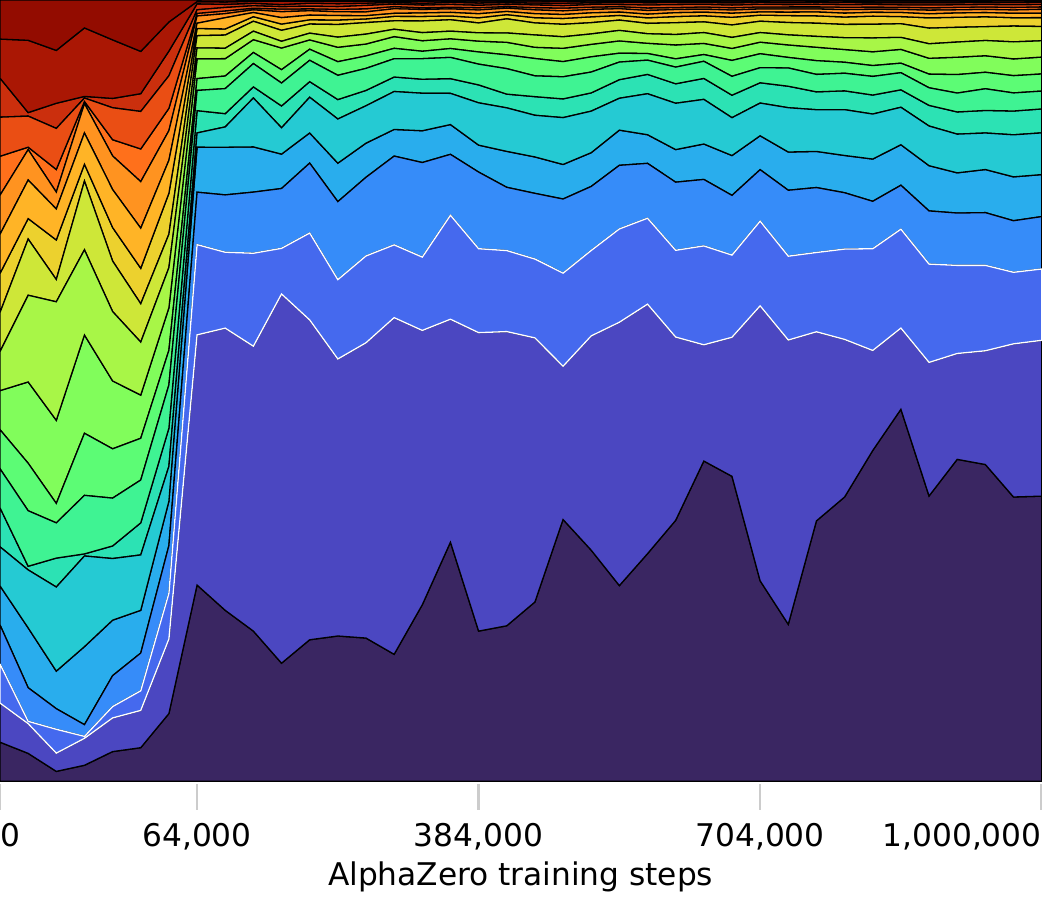} \hfill
\includegraphics[width=0.04\linewidth,valign=t,trim={0 0 7.3cm 0}]{figures/history/first_move_legend.pdf}
\caption{The AlphaZero policy head's preferences of opening move, as a function of training steps.
Here AlphaZero was trained three times from three different random seeds.
AlphaZero's opening evolution starts by weighing all moves equally, no matter how bad, and then narrows down options.
It stands in contrast with the progression of human knowledge, which gradually expanded from 1.~e4.
}
\label{fig:alphazero-prior-starting}
\end{subfigure}
\caption{A comparison between AlphaZero's and human first-move preferences over training steps and time.} 
\label{fig:starting-comparison}
\end{figure}

Figure~\ref{fig:human-history-starting} shows the opening move preference for White at move 1.
Most of the earliest recorded games seem to feature 1.~e4 as the opening move.
As this is one of the best theoretical tries, it remains popular throughout human history, but its dominance in the early years gives way to a more balanced distribution of chess openings, with 1.~d4 being slightly more popular in the early 20th century, and an increasing popularity of more flexible systems like 1.~c4 and 1.~\symknight f3.
This is different to AlphaZero's exploration,
shown in Figure \ref{fig:alphazero-prior-starting}.
It is more simultaneous in comparison, rather than relying mainly on a single initial move before branching into alternatives, which seems to be the case in early human play.

By and large, AlphaZero decreases first-move entropy through training.
What we know from data, human first-move entropy increases over the last five centuries.
With a uniform prior,
the AlphaZero neural network's initial move preference encodes $- \sum_{i=1}^{20} \frac{1}{20} \log_2 \frac{1}{20} = 4.32$ bits per move.
This entropy reduces to 
2.76, 2.81 and 3.02
bits for the first move prior after 1 million training steps for each of the seeds in Figure \ref{fig:alphazero-prior-starting} respectively.
The prior reflects the data in training games,
which were played with 800 MCTS simulations, injected noise and some stochastic move sampling (see Section \ref{sec:training-iterations}).
As a result, the first-move priors in Figure \ref{fig:alphazero-prior-starting} places \emph{some} nonzero mass on each move.

From recorded data, the first move preferences between years 1400 and 1800 empirically encode a mere 0.33 bits of information.
At the end of the twentieth century, the first move in top level play empirically encodes 1.87 bits of information.

\FloatBarrier

\subsection{The Ruy Lopez}

\begin{table}[t]
\centering
\begin{tabular}{l|llll}
 & seed \emph{(left)} & seed \emph{(centre)} & seed \emph{(right)} & seed \emph{(additional)} \\
\hline
3\ldots~\symknight f6   & 5.5\%  & 92.8\% & 88.9\% & 7.7\%  \\
3\ldots~a6              & 89.2\% & 2.0\%  & 4.6\%  & 85.8\% \\
3\ldots~\symbishop c5   & 0.7\%  & 0.8\%  & 1.3\%  & 1.3\%
\end{tabular}
\caption{The AlphaZero prior network preferences after
1.~e4 e5 2.~\symknight f3 \symknight c6 3.~\symbishop b5, for 
four different training runs of the system.
The prior is given after 1 million training steps, and the seeds \emph{(left, centre, right)} correspond to those in Figure \ref{fig:alphazero-prior-starting}.}
\label{table:ruy-lopez}
\end{table}

We consider a specific example, the Ruy Lopez 1.~e4 e5 2.~\symknight f3 \symknight c6 3.~\symbishop b5, where AlphaZero's preferences and human knowledge can take different paths.
Table \ref{table:ruy-lopez} shows the response of four different versions of AlphaZero, trained from four different seeds.
Training either converges to the Berlin defense 3\ldots~\symknight f6 as its main reply to the Ruy Lopez,
or to the more traditional 3\ldots a6.
From a theoretical standpoint, there isn't really a major difference between the two, so the partial interchangeability is ultimately not really surprising.

\begin{figure}[t]
\centering
\includegraphics[width=\textwidth]{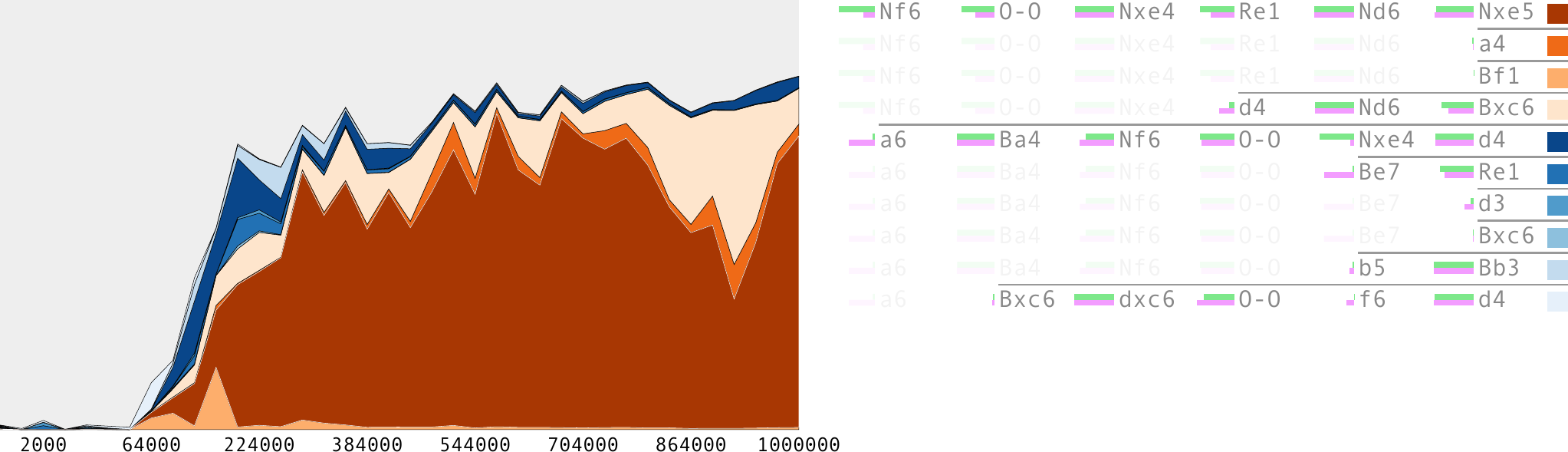}
\small
AlphaZero training steps\verb=                                         =
\normalsize
\begin{flushleft}
\includegraphics[width=0.528\textwidth]{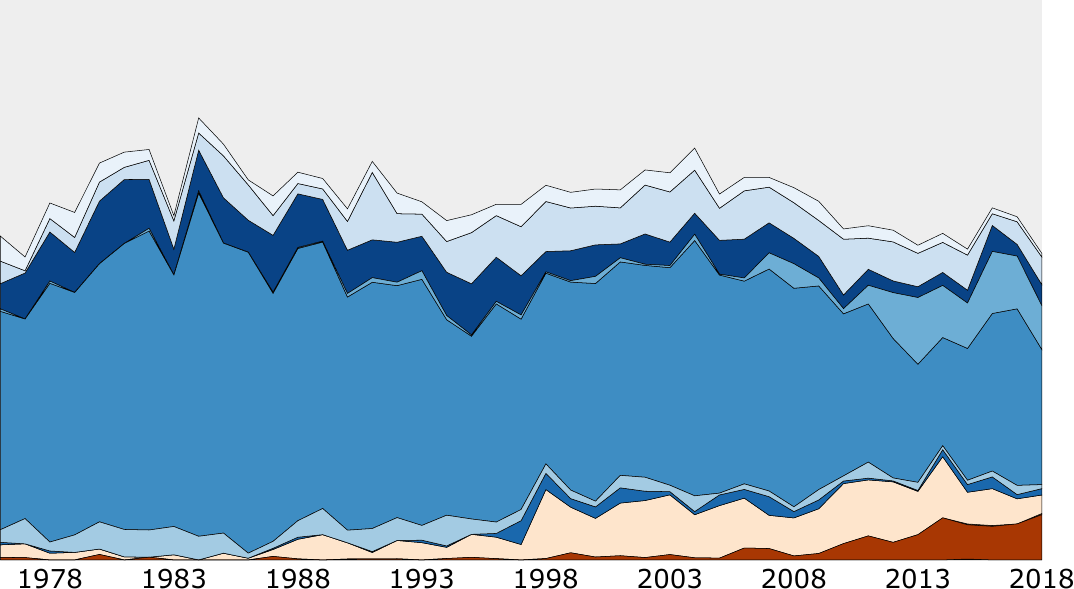} \\
\small
\verb=                    =Human years
\normalsize
\end{flushleft}
\caption{The prior continuation after 1.~e4 e5 2.~\symknight f3 \symknight c6 3.~\symbishop b5.
The figures show the proportional weight out of one of the top 11 continuations that are six plies deep.
The 10 continuations are AlphaZero's top 4 and human grandmasters' top 6 over the last 30 years.
Assuming 25 moves per ply, there should be another quarter of a billion ($25^6$) lines in the figure.
The light green bars next to the moves show AlphaZero's prior's preference compared to grandmaster human frequency over the last 30 years in pink.
}
\label{fig:openings-explosion}
\end{figure}

Figure~\ref{fig:openings-explosion} illustrates a case when AlphaZero's preferences take a different path from human history, and where the Berlin defense system (3\ldots~\symknight f6) 
is preferred.
This system was championed by GM Vladimir Kramnik at top level, and has since become a highly fashionable modern equalizing attempt in the Ruy Lopez, leading to a deeply theoretical Berlin endgame.
Yet, for a long period of time, it was thought of as being slightly worse for Black, compared to the more typical 3\ldots~a6.
Looking back in time, it took a while for human chess opening theory to fully appreciate the benefits of Berlin defense and to establish effective ways of playing with Black in this position.
On the other hand, AlphaZero develops a preference for this line of play quite rapidly, upon mastering the basic concepts of the game. This already highlights a notable difference in opening play evolution between the humans and the machine.

\subsection{Remarks}

The lens of comparison through which we view AlphaZero training and human history is narrow; we've only sketched a main difference
by looking at the narrowing (AlphaZero) or expansion (human) of options at the start of the game.
For initial opening moves, an empirical density could easily be estimated from human positions.
For the same reason we have not taken account of middle game themes or a greater understanding of endgames in human play,
which require a different approach of estimating and predicting human preferences from data.

\section{Rapid increase of basic knowledge}
\label{sec:explosion}

\begin{figure}
\centering
\begin{subfigure}{0.32\textwidth}
\centering
\includegraphics[width=\textwidth]{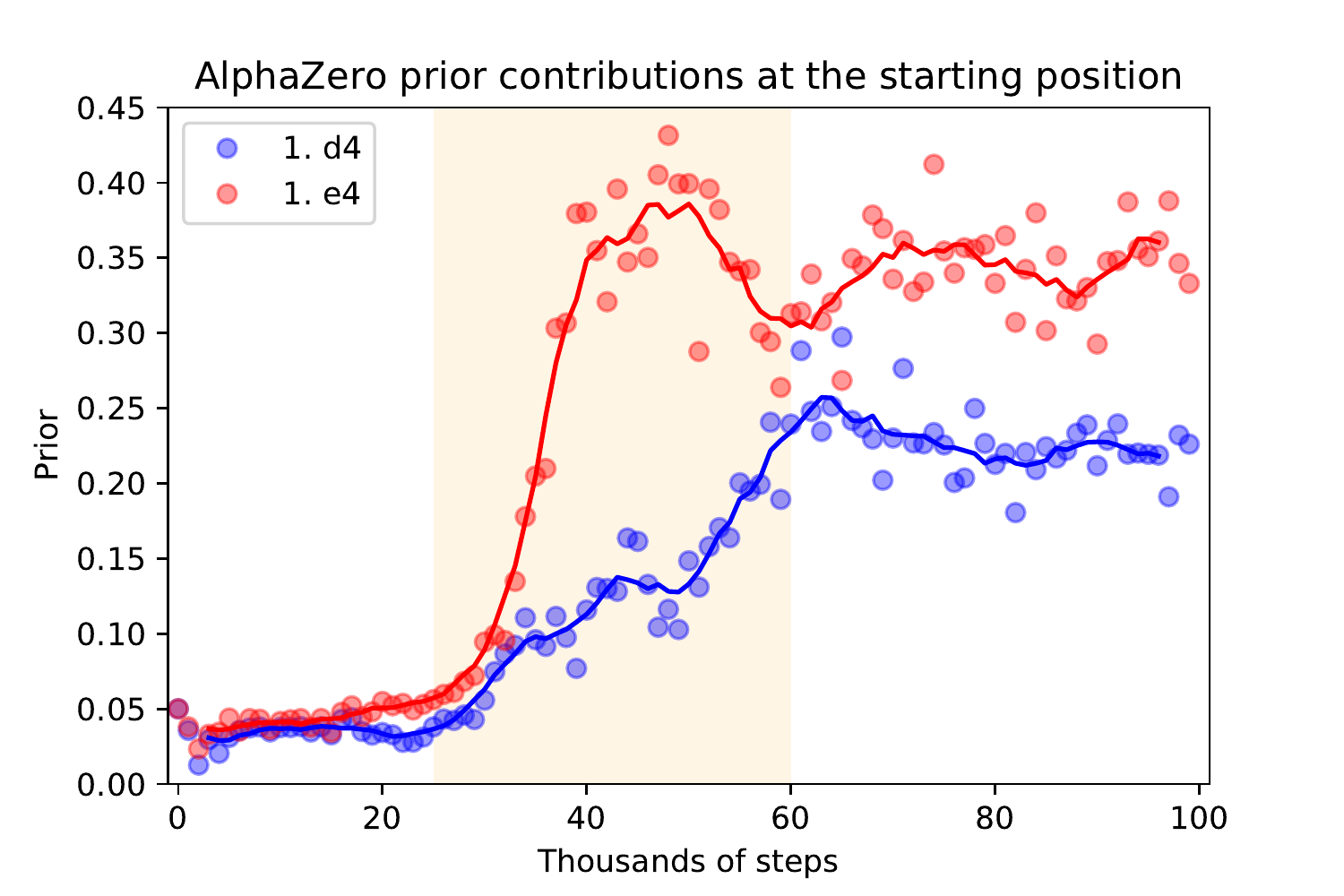}
\caption{After 25k training iterations, e4 and d4 are discovered to be good opening moves, and \textbf{rapidly adopted} within a short period of around 30k training steps.}
\label{fig:detailed-openings-explosion-starting}
\end{subfigure}
\hfill
\begin{subfigure}{0.32\textwidth}
\centering
\includegraphics[width=\textwidth]{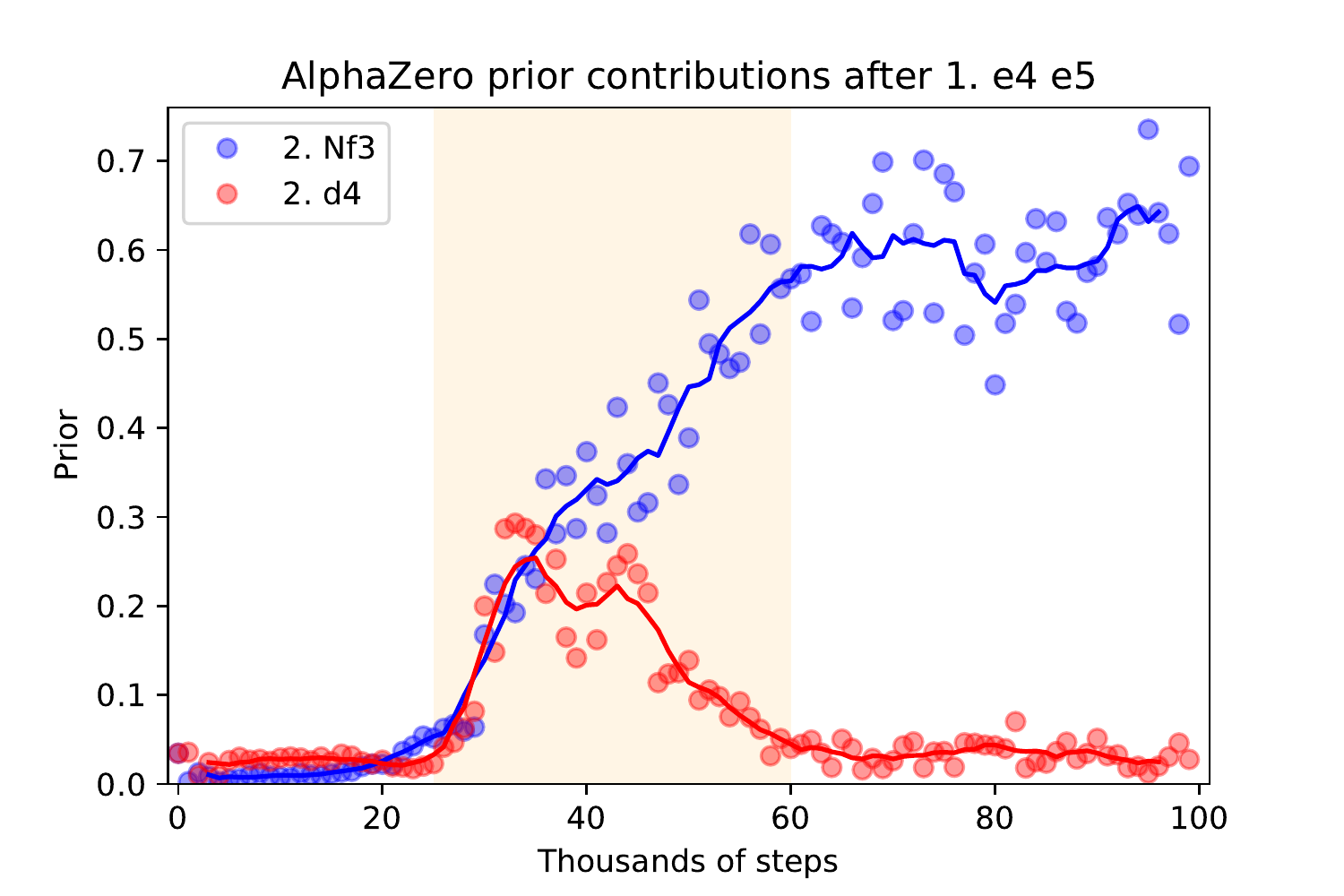}
\caption{Rapid discovery of options given 1.~e4 e5.
Within a short space of time, \symknight f3 is settled on as a standard reply, whereas d4 is considered and discarded.}
\label{fig:detailed-openings-explosion-e4-e5}
\end{subfigure}
\hfill
\begin{subfigure}{0.32\textwidth}
\centering
\includegraphics[width=\textwidth]{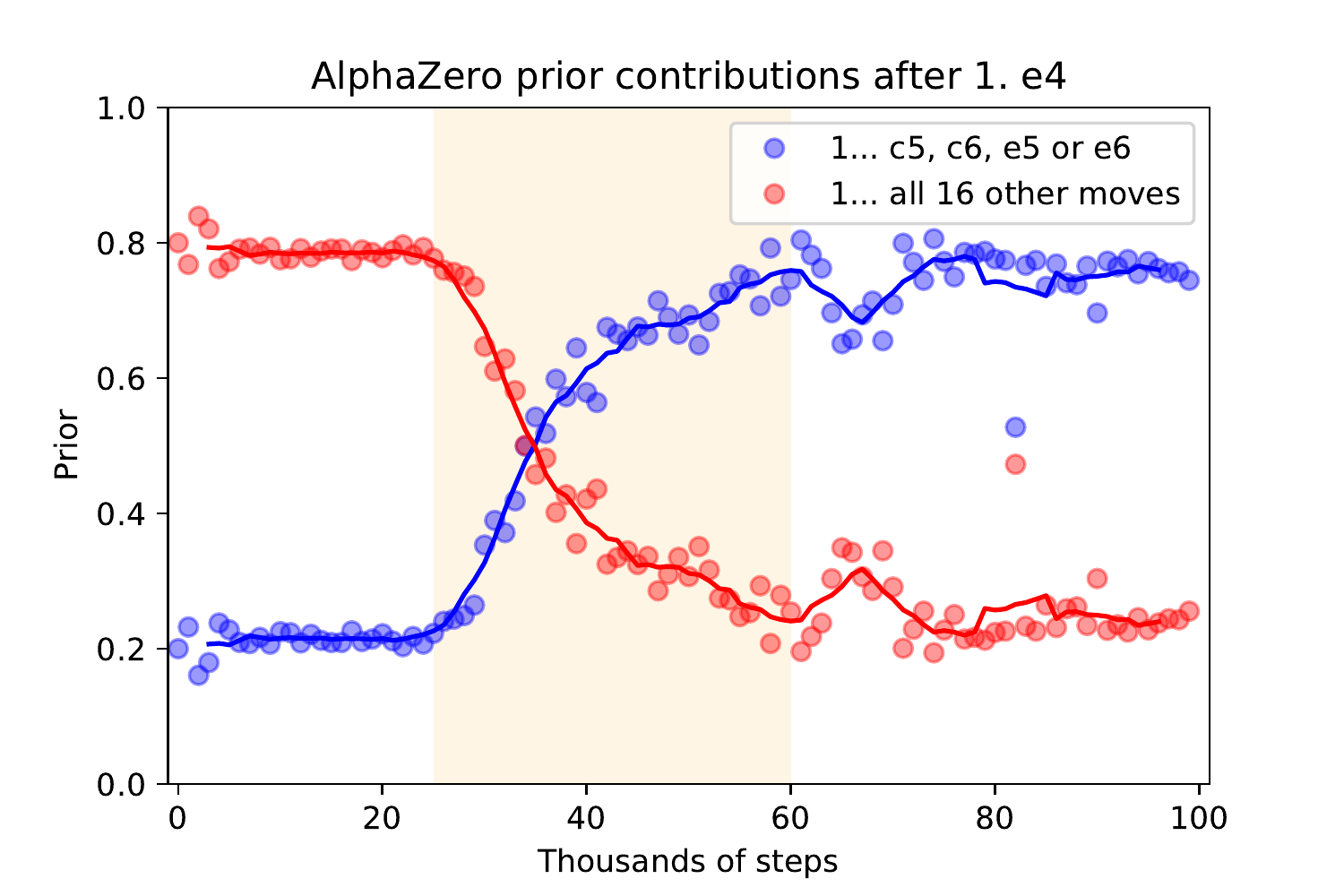}
\caption{In the window between 25k and 60k training steps, AlphaZero learns to put 80\% of its mass on four replies to e4,
and 20\% of its mass on all other 16 moves.}
\label{fig:detailed-openings-explosion-prior-e4}
\end{subfigure}
\caption{Rapid discovery of basic openings.
The randomly initialized AlphaZero network gives a roughly uniform prior over all moves.
The distribution stays roughly uniform for the first 25k training iterations, after which popular opening moves quickly gain prominence.
In particular, 1.~e4 is fully adopted as a sensible move in a window of 10k training steps,
or in a window of 1\% of AlphaZero's training time.
}
\label{fig:detailed-openings-explosion}
\end{figure}

During the course of AlphaZero's million training steps, there is an inflection point where the network understands ``enough'' about piece value that
playing basic opening sequences (like playing 1.~e4 and not 1.~f3) lead to tangible advantages.

From the evidence we have in Section \ref{sec:value-regression}, notably Figure \ref{fig:value_regression}, 
Stockfish's evaluation sub-function of \verb=material= is strongly indicative of the AlphaZero network's value assessment.
Figure \ref{fig:value_regression} suggests that 
the concept of \verb=material= and its importance in the evaluation of a position
is largely learned between training steps 10k and 30k, and then refined.
Furthermore, the concept of piece \verb=mobility= is progressively incorporated in AlphaZero's value head
in the same period.
It is plausible that a basic understanding of the material value of pieces should precede a rudimentary understanding that
greater piece mobility is advantageous. 

\subsection{Discovery of standard openings}

In this section we examine the discovery of standard openings.
Evidence suggests recognizable opening theory develops between 25k and 60k training steps, after the period between 10k and 30k steps when knowledge of basic material value is acquired.
Section \ref{sec:first-move-progression} and notably Figure \ref{fig:alphazero-prior-starting} illustrates a rapid transition from a largely uniform prior to commonly played moves.
Figure \ref{fig:openings-explosion} shows how the Ruy Lopez opening is refined over a million training steps, with
the inflection point where it starts gaining significant pass being clearly visible.
We investigate the transition further here.

To examine the transition, we consider
the contribution of familiar opening moves as a fraction of the AlphaZero prior over all moves; see 
Figure \ref{fig:detailed-openings-explosion}.
Figure \ref{fig:detailed-openings-explosion-starting}
shows that 
after 25k training iterations, 1.~d4 and 1.~e4 are discovered to be good opening moves, and are \textbf{rapidly adopted} within a short period of around 30k training steps.
Similarly, AlphaZero's preferred continuation after 1.~e4 e5 is determined in the same short temporal window.
Figure \ref{fig:detailed-openings-explosion-e4-e5} illustrates
how both 2.~d4 and 2.~\symknight f3 are quickly learned as reasonable White moves, but 2.~d4 is then dropped almost as quickly in favour
of 2.~\symknight f3 as a standard reply.
Looking beyond individual moves,
we group AlphaZero's responses to 1.~e4 into two sets,
1\ldots c5, c6, e5 or e6 and then all 16 other moves.
Figure \ref{fig:detailed-openings-explosion-prior-e4} shows how
1\ldots c5, c6, e5 or e6 together account for 20\% of the prior mass when the network is initialized, as they should, as they are four out of twenty possible Black responses.
However, 
after 25k training steps their contribution rapidly grows to account for 80\% of the prior mass.

After the rapid adoption of basic opening moves in the window from 25k to 60k training steps, opening theory is progressively refined through repeatedly updating the training buffer of games with fresh self-play games.
While these examples of the rapid discovery of basic openings are not exhaustive, further examples across multiple training seeds are
given in Appendix \ref{appendix:policy-progression}.

\paragraph{Training configuration dependence}
Our conclusions are dependent on the particular hyperparameter settings that empirically yielded the strongest version of the system over a hyperparameter search. We refer the reader to Section \ref{sec:training-iterations} for training details for a specific configuration:
At the start of training, the self-play training buffer (queue) is filled with 1 million training positions from completely random games. At 30 positions per game, at least 30k games are there played using the randomly initialized network in MCTS. With a batch size of 4096, the queue is empty before 250 training steps are reached.
The policy head $\p$ in Figure \ref{fig:alphazero-network} is a $8 \times 8 \times 73 = 4672$ tensor of possible moves, and from these random games, the network starts assigning most weight to legal moves.
After every 1000 training steps, the networks responsible for generating self-play games are refreshed with the latest copy of the training network.
After 25k training steps, when familiar openings start gaining traction,
the neural networks responsible for generating the self-play games would have been reloaded with fresh copies from the training job 25 times.
The onset is therefore not a result of the initial random training buffer being depleted of examples.
The onset is dependent on the training configuration settings; 
it would shift if the self-play networks were refreshed at a faster or slower rate, for example.
We do not examine suboptimal training configuration settings leading to significantly weaker versions of AlphaZero in this work.

\paragraph{Sequential knowledge acquisition}
Figures \ref{fig:value_regression}, \ref{fig:detailed-openings-explosion} and \ref{fig:openings-explosion} suggest a sequence: that piece value is learned before basic opening knowledge; that once discovered, there is an explosion of basic opening knowledge in a short temporal window; that the network's opening theory is slowly refined over hundreds of thousands of training steps.

\subsection{Vladimir Kramnik's qualitative assessment}

To further shed light on the early period in the evolution of chess knowledge within the AlphaZero network,
we complemented our quantitative experiments by a qualitative assessment by GM Vladimir Kramnik, a former world chess champion.
This assessment is an attempt to identify themes and differences in style of play between AlphaZero checkpoints
in the early phase of training.
By generating games played between versions of AlphaZero
at different training steps,
their strengths and weaknesses would potentially be exposed in the games.
We generated games played between AlphaZero at 16k and 32k, 32k and 64k, and 64k and 128k training steps.
For each, Vladimir Kramnik was presented with
a random sample of 100 games from each pairing.
These qualitative assessments were provided prior to seeing our other quantitative results in order to avoid introducing biases into the analysis.

Perhaps surprisingly, GM Kramnik was able to easily spot recurring themes in the shared game samples and to thereby formulate opinions as hypotheses for the differences between model performance at these points in time:
\begin{description}
\item[16k to 32k]
The comparison between 16k and 32k was the simplest.
AlphaZero at 16k has a crude understanding of material value and fails to accurately assess material in complex positions.
This leads to potentially undesirable exchange sequences, and ultimately losing games on material.
On the other hand, AlphaZero at 32k seemed to have a solid grasp on material value, thereby being able to capitalize on the material assessment weakness of 16k.
\item[32k to 64k]
The main difference between 32k and 64k, in GM Kramnik's view,
lies in the understanding of king safety in imbalanced positions.
This manifests in the 32k version potentially underestimating the attacks and long-term material sacrifices of the 64k version,
as well as the 32k version overestimating its own attacks, resulting in losing positions.
Despite the emphasis on king safety, GM Kramnik also comments that 64k appears to be somewhat stronger in all aspects than 32k, which is well-aligned with our quantitative observations for the changes in the period.
\item[64k to 128k]
The theme of king safety in tactical positions resurfaces in the comparisons between 64k and 128k, where it seems that 128k has a much deeper understanding of which attacks will succeed and which would fail, allowing it to sometimes accept material sacrifices made by 64k for purposes of attack, yet then proceed to defend well, keep the material advantage, and ultimately convert to a win. There seems to be less of a difference in terms of positional and endgame play.
\end{description}

\subsection{Are tactical skills acquired before positional skills?}

From observing AlphaZero's self-play games over increasing training steps,
Vladimir Kramnik remarked that tactical skills appear to precede positional skills as AlphaZero learns.

The GM’s observation can be explained by the difference in the abilities that are required to master tactical and positional lines. Playing tactical lines requires finding short term moves that force the opponent to lose material, being checkmated, etc.; while playing the positional lines requires foreseeing the development of the game many plies ahead, i.e. being able to perform deeper search. The depth of the search, that AlphaZero can perform given a fixed amount of compute, is directly related to the amount of entropy in the AlphaZero policy priors. As demonstrated in Section \ref{sec:human-history}, the entropy is maximal in the beginning of the training process when the priors are distributed almost uniformly. Over time, AlphaZero learns to discard least promising moves; that makes deeper search possible. Since AlphaZero cannot properly evaluate positional lines before deeper search is possible, the emergence of the positional skills is delayed.

To collect more supporting evidence, GM Kramnik selected two sets of 10 opening 
lines based on their known theoretical properties.
The first set contained starting positions corresponding to 
quiet and positional opening systems like the Berlin defense,
Queen's gambit declined,
Queen's gambit accepted and the English opening.
The second set contained starting positions corresponding to
aggressive and tactical opening systems
like the Najdorf Sicilian, Nimzowitsch defense, the French Winawer and the King's Indian defense
Both the tactical and the positional lines were selected such that they are considered equal, not conferring an advantage to either player.
In the self-play games, each version of AlphaZero had a chance to play as White and Black from the same position.

We generated self-play games played between versions of AlphaZero after 32k, 64k and 128k training steps.
For each of the 10 opening lines in each of the two sets,
we generated 100 games played at approximately 1 second per move, resulting in 1000 games per set.
The outcome of these games were:
\begin{description}
\item[32k to 64k; tactical openings]
In the tactical lines, AlphaZero at 64k won 82.3\% of the games, lost 4.1\% of the games, and drew 13.6\% of the games, resulting in an estimated ELO difference of 365 points.
\item[32k to 64k; positional openings]
In the positional lines, AlphaZero at 64k won 76.6\% of the games,
lost 1.2\% of the games, and drew 22.2\% of the games, resulting in the estimated ELO difference of 343 points.
\item[64k to 128k; tactical openings]
In the tactical lines, AlphaZero at 128k won 53.7\% of the games, lost 12.5\% of the games, and drew 33.8\% of the games, resulting in an estimated ELO difference of 152 points.
\item[64k to 128k; positional openings]
In the positional lines, AlphaZero at 128k won 37.8\% of the games,
lost 3.8\% of the games, and drew 58.4\% of the games, resulting in the estimated ELO difference of 123 points.
\end{description}
These experiments show that
there is a marginal difference between game outcomes from tactical positions
and game outcomes from more positional openings.
The slight difference in favour of tactical positions persists both in the comparison between the 128k and 64k model, as well as the 64k and 32k model.
The observed difference in ELO between the checkpoints playing positional and tactical lines is
circumstantial evidence that tactical skill acquisition
might happen before or faster than positional skill acquisition.
However, this is only one potential explanation of the differences.
GM Kramnik's remark points to further work to 
understand the order in which skills are acquired.

\section{Exploring activations with unsupervised methods}
\label{sec:intermediate_computations}

In Section \ref{sec:concept-regression} we investigated when and where the AlphaZero network encodes human conceptual knowledge.
The approach was confined to labeled data of different human concepts, and
was made possible through supervised methods.
The public sub-functions of Stockfish's evaluation function and a large collection of custom chess concept functions allow the automation of large supervised data sets.
Our approach so far has relied on supervised learning, either to regress predefined concepts from the network or to predict the value function from those concepts. This runs the risk of overlooking anything we have not included in our concept set, as well as biases our understanding of the AlphaZero network towards those concepts.
In this section we take an alternative approach, using simple unsupervised approaches involving matrix decomposition (Section~\ref{ss:nmf}) and correlation analysis (Section~\ref{sec:activations}).
These methods generate a wealth of data to analyse at each layer of the network.
We present some highlights here and make the full dataset available in the Supplementary Materials.

\subsection{Non-negative matrix factorisation}
\label{ss:nmf}

Non-negative matrix factorisation (NMF)~\cite{lee1999learning} is an
approach to discover features in an unsupervised way.
It has previously been used to interpret vision~\cite{olah2018building} and simple RL models~\cite{hilton2020understanding}.
In this section we explore NMF as a complementary unsupervised approach to the probes trained in Section~\ref{sec:encoding}: rather than probe activations for specific concepts, we instead simplify the activations in a concept-agnostic way. This allows the structure of the activations to reveal itself, instead of imposing our assumptions on it.

\subsubsection{Methodology}

Layer $l$'s activations $\z^l \in \Rbb^{H \times W \times C} = \Rbb^{8 \times 8 \times 256}$
are non-negative.
Each of the 256 channels is an $8 \times 8$ plane.
We treat each plane as a vector and reshape $\z^l$ into a matrix $\hat{\z}^l \in \Rbb^{HW \times C}$.
This is a square-by-channel matrix, where every row corresponds to
a square and is a non-negative $C$-dimensional vector.
We compress the activations $\z^l$ by
representing each square as a non-negative $K$-dimensional weight vector, with $K < C$.
It means the compression would be into a
matrix $\bOmega \in \Rbb^{HW \times K}$.
In the reduced representation,
the rows of $\bOmega$ correspond to squares, and its entries are
$K$ weights that sum together 
non-negative global factors $\f_k \in \Rbb^C$ for $k = 1, \ldots, K$
so that the original row in $\hat{\z}^l$ is closely approximated.
If the factor matrix $\F \in \Rbb^{K \times C}$ contains the global factors as columns,
the activations are approximated with
$\hat{\z}^l \approx \bOmega \F$.
For brevity we omit the dependence of the factor and weight matrices on the layer index $l$.

To determine the global factors $\F$, we stack 
the network activations of $N$ randomly selected inputs
$\hat{\z}_1^l, \ldots, \hat{\z}_N^l$ into a matrix
$\hat{\Z}^l \in \Rbb^{NHW \times C}$ (we use $N=10^4$, and randomly select 50 positions to visualise).
The factors and their weights
$\bOmega_{\mathrm{all}} \in \Rbb^{NHW \times K}$ 
are found by minimizing
\begin{align}
\F^{*}, \bOmega_{\mathrm{all}}^{*} &= \min_{\F, \bOmega_{\mathrm{all}}} \big\| \hat{\Z}^l -
\bOmega_{\mathrm{all}} \F \big\|_2^2 \nonumber \\
\F, \bOmega_{\mathrm{all}} & \geq \zerobf \ .
\end{align}
The stacked submatrices of $\bOmega_{\mathrm{all}}^{*}$ correspond to the NMF weights for each input's activations. Alternatively,
given $\F^*$, the NMF weights for any activations $\z^l$ could be retrieved by
\begin{align}
\bOmega^{*} &= \min_{\bOmega} \big\| \hat{\z}^l -
\bOmega \F^{*} \big\|_2^2 \nonumber \\
\bOmega & \geq \zerobf \ .
\end{align}
To visualize the NMF factors for activations $\z^l$, we overlay the $K$ columns of $\bOmega^{*}$
onto the input $\z^0$.
The visualization of factor $k$'s contributions to $\z^l$ 
is done by reshaping the column $k$ of $\bOmega^{*}$ into a $H \times W$ or $8 \times 8$ matrix.
The visualization shows how much NMF factor $k$ contributes to each neuron's representation. This visualisation makes a strong assumption that representations in the residual block are spatially-aligned, i.e. an activation at a given spatial position can be interpreted in light of the pieces around that position. Although the network architecture biases towards this correspondence (all hidden layers are identical in shape to a chess board, and the residual network structure biases activations towards spatial correspondence) this is not strictly enforced by the architecture.

Using 36 factors per block, the full NMF dataset consists of 720 block/factor pairs. We report selected factors below, and the full dataset is \href{https://storage.googleapis.com/uncertainty-over-space/alphachess/index.html}{available online}. Most factors in later layers remain unexplained, however - we view explaining these factors (and developing the methods necessary to do so) as an important avenue for future work.%

\subsubsection{Results}

\begin{figure}[t]
     \centering
     \begin{subfigure}[b]{0.24\textwidth}
         \centering
         \includegraphics[width=\textwidth]{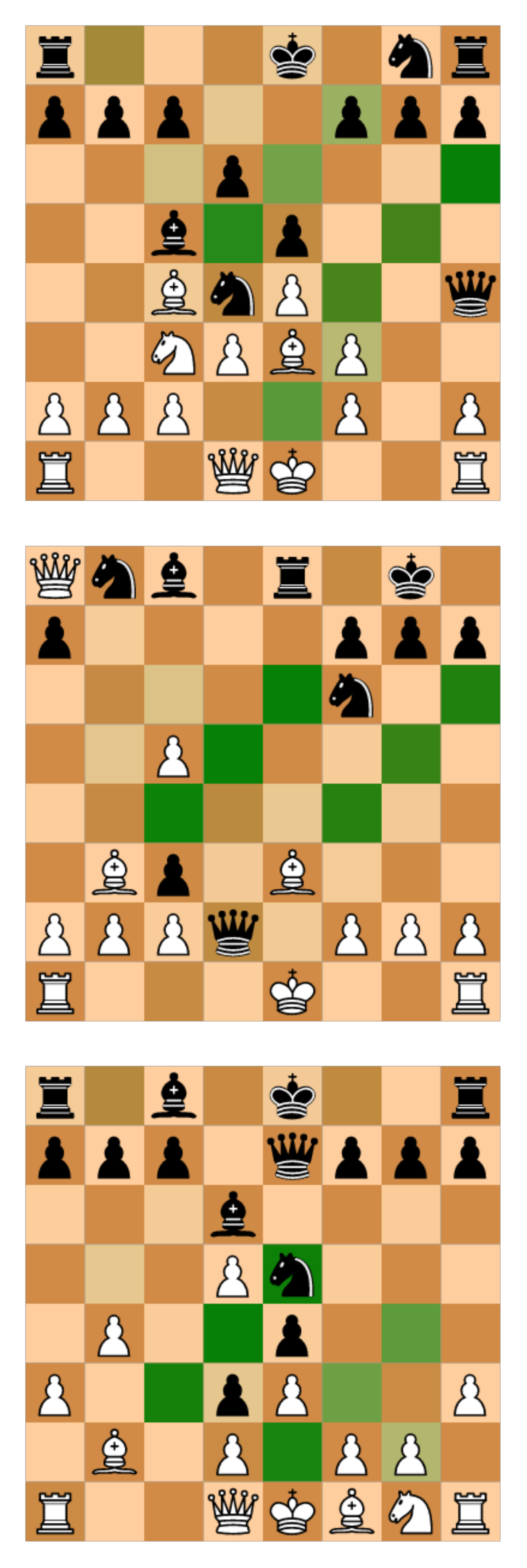}
         \caption{Development of diagonal moves for player (block 1, factor 26 of 36).}
         \label{fig:factor_diag1}
     \end{subfigure}
     \hfill
     \begin{subfigure}[b]{0.24\textwidth}
         \centering
         \includegraphics[width=\textwidth]{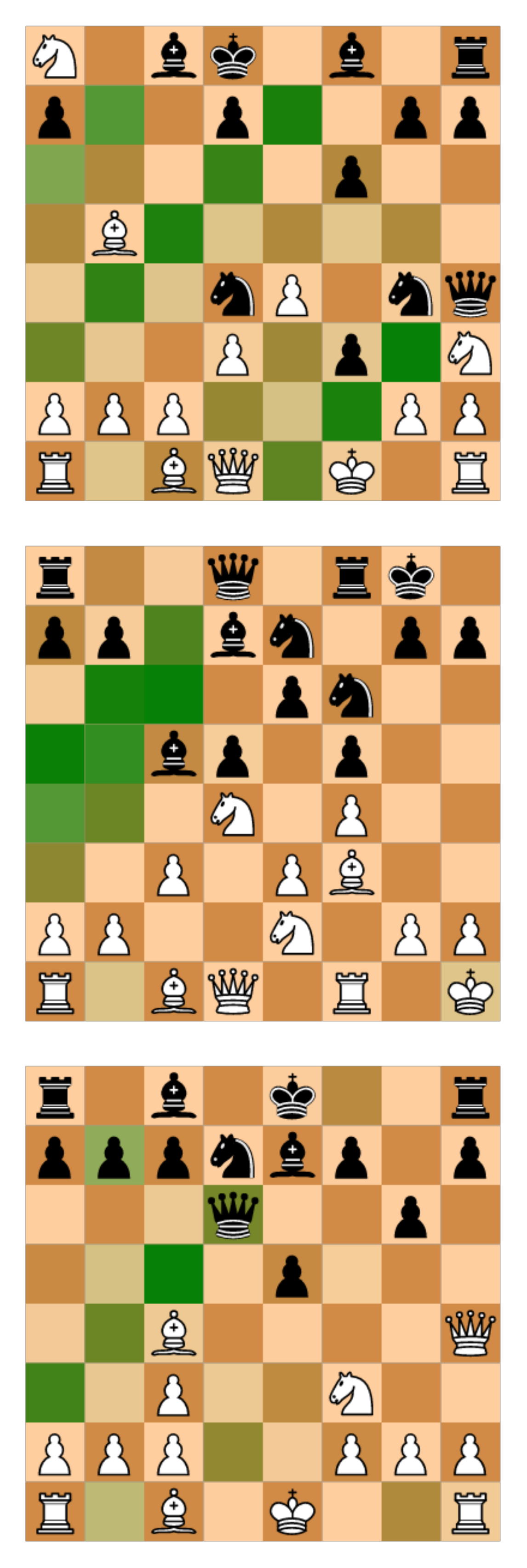}
         \caption{Fully developed diagonal moves for opponent (block 3, factor 22 of 36).}
         \label{fig:factor_diag2}
     \end{subfigure}
     \hfill
     \begin{subfigure}[b]{0.24\textwidth}
         \centering
         \includegraphics[width=\textwidth]{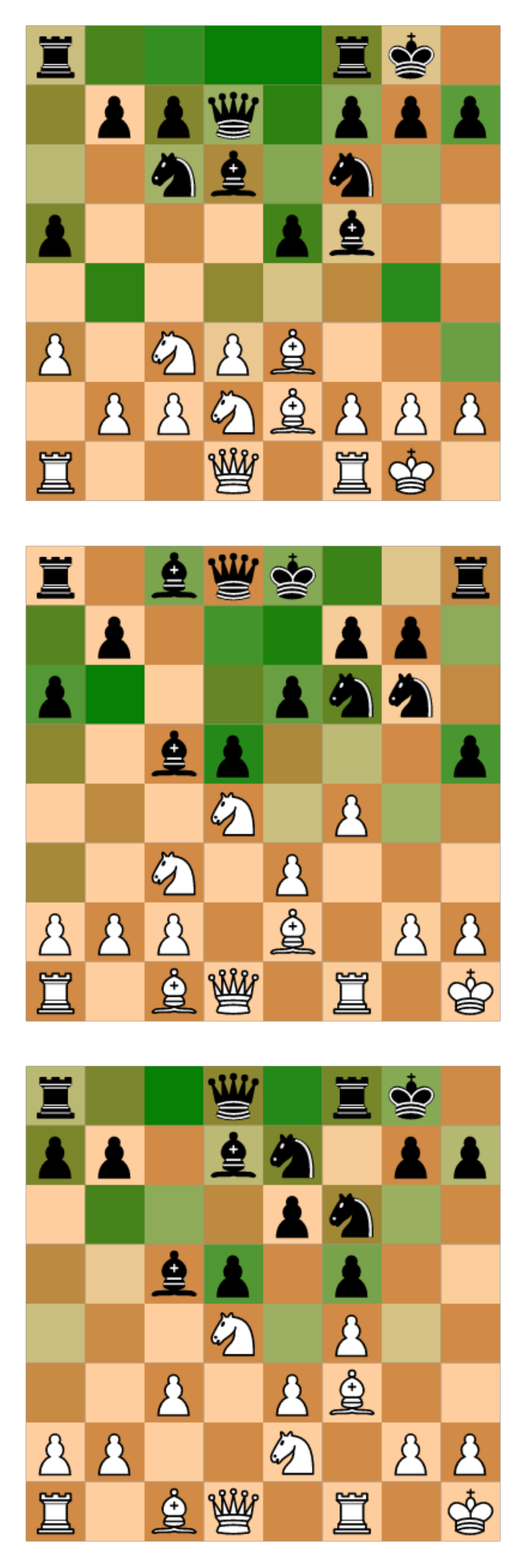}
         \caption{Count of opponent's potential piece moves (block 3, factor 11 of 36).}
         \label{fig:factor_threatsquares}
     \end{subfigure}
      \hfill
     \begin{subfigure}[b]{0.24\textwidth}
         \centering
         \includegraphics[width=\textwidth]{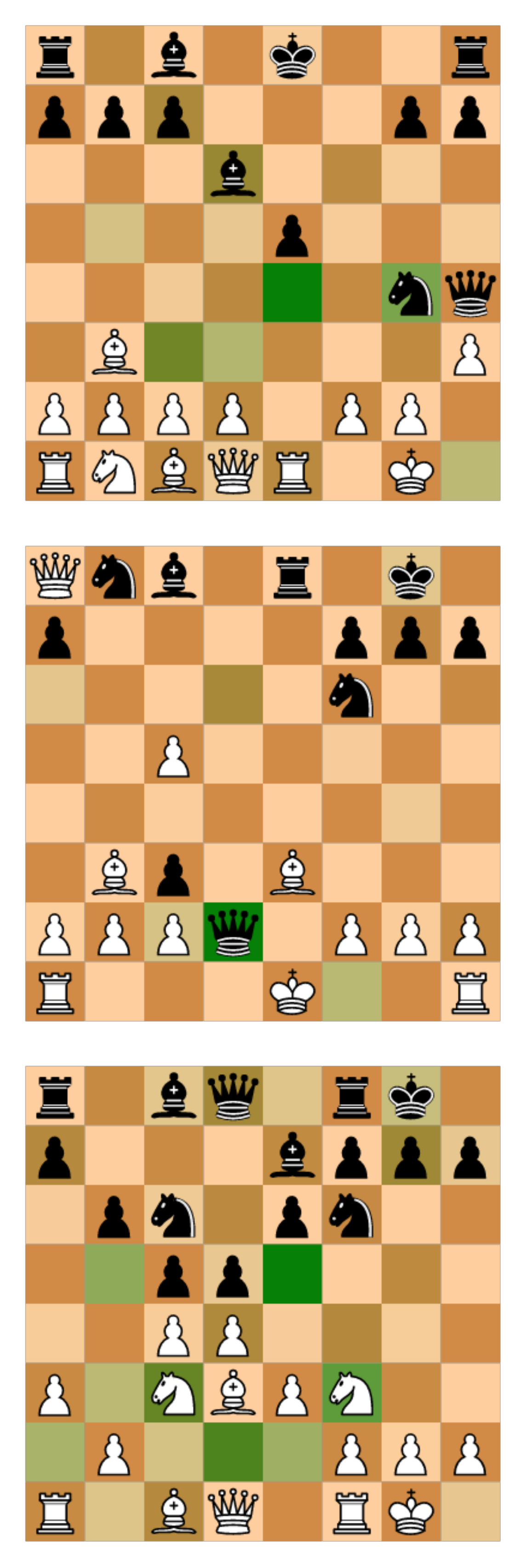}
         \caption{Potential good squares to move to? (block 18, factor 22 of 36).}
         \label{fig:factor_piecestotake}
     \end{subfigure}
\caption{Visualisation of NMF factors in a fully-trained AlphaZero network, showing development of threats and anticipation of possible moves by the opponent, as well as a factor that may be involved in move selection.
Following Figure \ref{fig:alphazero-network}, we count a ResNet `block' as a layer.}
\label{fig:factors}
\end{figure}

This section highlights some illustrative examples of interpretable factors in AlphaZero's activations. The factors shown in Figure~\ref{fig:factors}a and Figure~\ref{fig:factors}b show the development of potential move computations for the player's and opponent's diagonal moves respectively. In the first layer moves of only three squares or fewer are shown (and only those towards the upper right of the board), demonstrating that move calculations take multiple blocks to complete. The convolutional structure of the AlphaZero network means that all computations from one layer to the next involve only spatially adjacent neurons. Because of this, move computations must occur over the course of multiple layers. This partially explains the gradual increase in predictive power over the initial layers for threat-related concepts shown in Figure~\ref{fig:concept_regression}b. %

Figure~\ref{fig:factors}c shows a more complex factor in layer 3: a count of the number of the opponent's pieces that can move to a given square (darker weights indicates more pieces can move to that square). This factor is likely to be useful in computing potential exchanges, and indicates that AlphaZero is also considering potential opponent moves even early in the network. Figure~\ref{fig:factors}d appears to show potential moves for the current player - darker squares indicate better moves (for instance the opponent's hanging queen on d2).

The \href{https://storage.googleapis.com/uncertainty-over-space/alphachess/index.html}{NMF factor data} contains many factors we have not yet interpreted, especially in later layers of the network. Where factors are uninterpretable, this could be because the number of factors used is incorrect, spatial correspondence is broken, or the factor represents something too complex to understand without reference to earlier layers. Development of methods for relating complex later-layer factors to well-understood early layer factors is an important priority for further interpretability work in complex domains. Finally, we note that although these interpretations of the above factors bear out in the majority of the randomly selected positions shown in the online database of factors, an interpretation can only be considered definitive once it has been quantitatively validated~\cite{leavitt2020towards}, ideally by intervening on the input.

\subsection{Covariance between inputs and activations}
\label{sec:activations}

Which input features in $\z^0$ covary most with a particular activation $z_i^l$?
If the activation $z_i^l$ increases, which input features in
$\z^0$ are most correlated with the increase, and which input features have no bearing on a change in $z_i^l$? 
This section departs from earlier sections to look at individual activations.
While our goal is to use complex concepts to understand AlphaZero,
the concepts in layer $l$ are composed of a distributed representation through the individual activations in $\z^l$.
A simpler unsupervised test is to determine how specific positions in a board $\z^0$ are ``mapped'' in each layer of the neural network,
and here we measure how related inputs are to individual activations $z_i^l$.

\paragraph{Activation-input covariance}
There are many gradient-based methods to investigate how changes in $\z^0$ affect $z_i^l$ by using some form of
$\partial z_i^l / \partial \z^0$ \cite{integrated-gradients}.
However, with the exception of a move-counter plane,
$\z^0$ is binary
and $f_{\btheta}(\z^0)$ has never seen examples that are not on the $\{ 0, 1\}^{8 \times 8 \times (14h + 7)}$ hypercube; see Figure \ref{fig:alphazero-network}.
We take a different approach. As $\z^0$ is much smaller than
a typical image, we compute the covariance between $\z^0$ and each activation in the network. 
We use only the fully trained network,
and determine which input features $\z^0$ are most correlated with each activation $z_i^l$ in each layer $l$,
$\z^l = f^{1:l}_{\btheta}(\z^0)$.
The \textbf{activation-input covariance}
\begin{equation} \label{eq:activation-input-covariance}
\mathrm{cov}(z^l_{i} , \z^0) =  \Ebb\big[z^l_{i} \, \z^0 \big] - \Ebb \big[ z^l_{i} \big] \, \Ebb \big[ \z^0 \big]
\end{equation}
gives an indication of how correlated each component of $\z^0$ is with an activation output at layer $l$.
Because $z^l_{i} \in [0, 15]$ due to gradient clipping and $\z^0 \in [0, 1]^{8 \times 8 \times (14h+7)}$ (due to input scaling, with piece location inputs being either 1 for a piece or 0 for an absence), the covariance is between -15 and 15.
We estimate $\mathrm{cov}(z^l_{i} , \z^0)$ in \eqref{eq:activation-input-covariance} empirically over 1.4 million randomly selected positions from grandmaster games.

\paragraph{Visualizing the activation-input covariance}
We visualize the first twelve $8 \times 8$ planes of
$\mathrm{cov}(z^l_{i} , \z^0)$ in this analysis -- 
in particular in Figures 
\ref{fig:activation-input-covariance-layer-1},
\ref{fig:activation-input-covariance-layer-1-special},
\ref{fig:activation-input-covariance-column-e4-9} and 
\ref{fig:activation-input-covariance-column-e4-7}
-- as these
correspond to the planes in $\z^0$ that specify the board position.
The $8 \times 8 \times 12$ submatrix corresponding to the piece planes is clipped at zero and scaled so that its maximum value is one, so that only positive correlations -- the presence and not the absence of pieces -- are visualized.
Recall that each of the first twelve planes of 
$\mathrm{cov}(z^l_{i} , \z^0)$ correspond to different piece types.
All twelve planes are rendered on the same board.
The \emph{opacity} of each piece in a plane corresponds to the covariance of activation with that piece on a given square.
The input encoding of the pieces in $\z^0$ is from the point of view of the playing side.
The input encoding for Black to play would simply flip the standard chess board horizontally and vertically so that it is aligned from the point of view of the playing side.
The visualizations are from the point of view of the playing side, and colour the playing side as White.
In our illustrations of the activation-input covariances, we used a lighter board colour than other visualizations in this paper.
This is solely to make the semi-translucent pieces more visible.

\begin{figure}[t]
\centering
\includegraphics[width=0.4\textwidth]{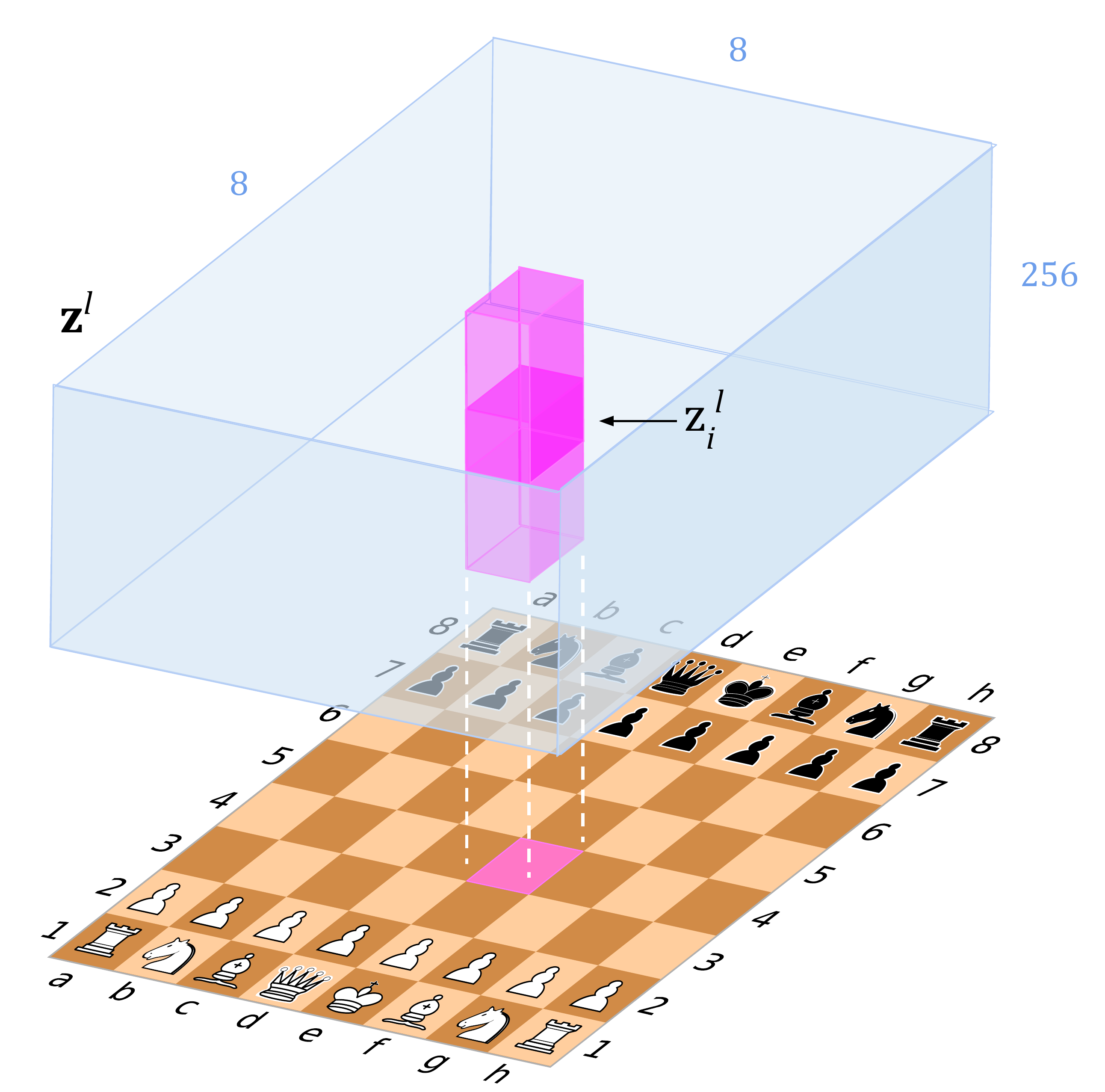}
\caption{Because of the ResNet backbone of the AlphaZero network in Figure \ref{fig:alphazero-network}, activations $z_i^l$ for $i = (5, 4, \cdot)$ \emph{align spatially} with the e4 square.
There are activations $z_i^l$ in the same position in each $\z^l$ for $l = 1, \ldots 20$ that simply propagate local features like piece placement forward from the first layers to the last in the backbone (Section 
\ref{sec:persisted-feature-detectors}; Figure \ref{fig:activation-input-covariance-column-e4-9}).
Other activations $z_i^l$ in the same position in each $\z^l$ for $l = 1, \ldots 20$ are active on an evolving pattern of global features (Section \ref{sec:evolving-feature-detectors}; Figure \ref{fig:activation-input-covariance-column-e4-7}).
}
\label{fig:spatial-structure}
\end{figure}

The sequence of convolutions in the ResNet backbone
of $f_{\btheta}(\z^0)$ has a spatial structure and
each $\z^l$'s $8 \times 8 \times 256$ tensor aligns with the input board encoding.
This is illustrated in Figure \ref{fig:spatial-structure}.
We let $i$ index the width, height and channel in $\z^l$.
As there are 16384 activations at the end of each layer,
we use indexes $i = (5, 4, \cdot)$ in
Figures 
\ref{fig:activation-input-covariance-layer-1},
\ref{fig:activation-input-covariance-layer-1-special},
\ref{fig:activation-input-covariance-column-e4-9} and 
\ref{fig:activation-input-covariance-column-e4-7}.
These indexes align with the central e4 square on a chess board.

We illustrate different roles of individual activations as
feature detectors below, including where they detect basic patterns on the input board position,
seem to copy the presence of features over from one layer to the next,
or seem to combine lower level features.

\subsubsection{Feature detectors}

\begin{figure}
\centering
\includegraphics[width=0.18\textwidth]{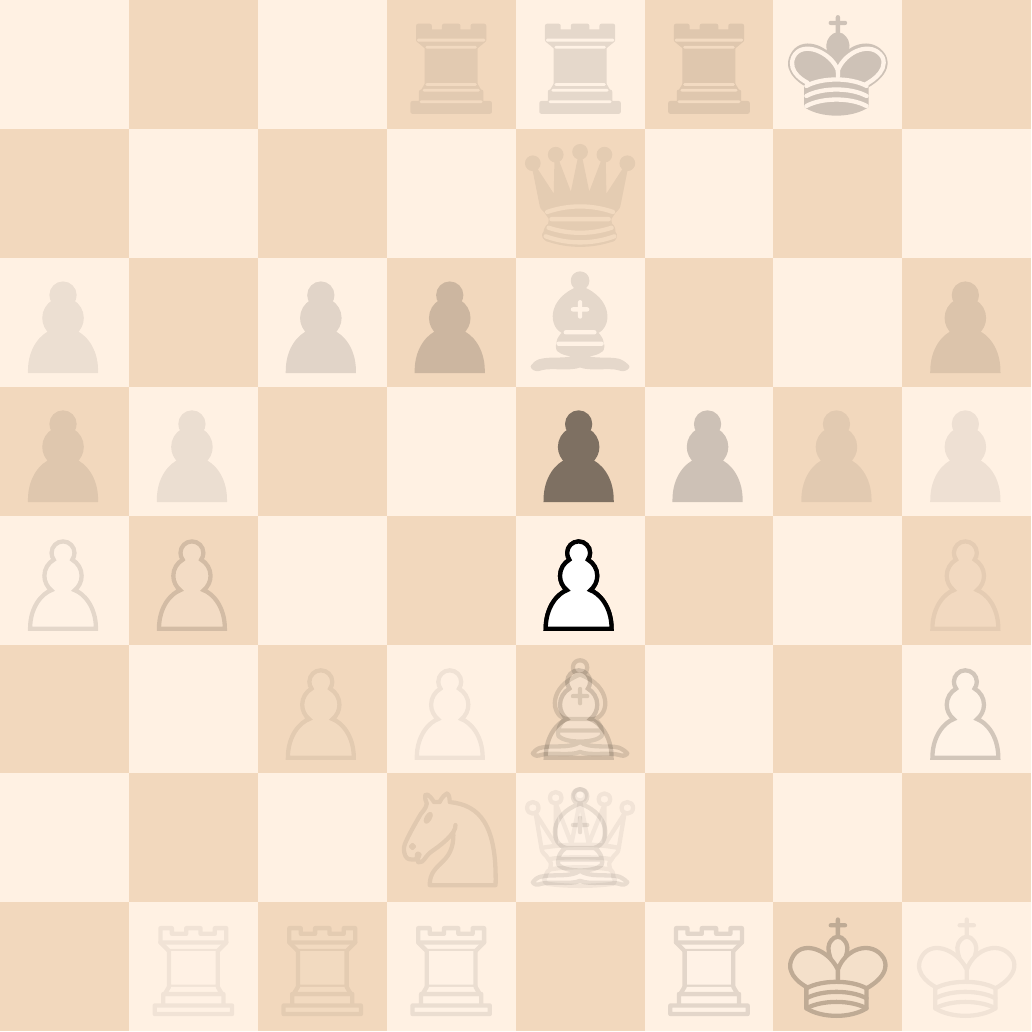}
\includegraphics[width=0.18\textwidth]{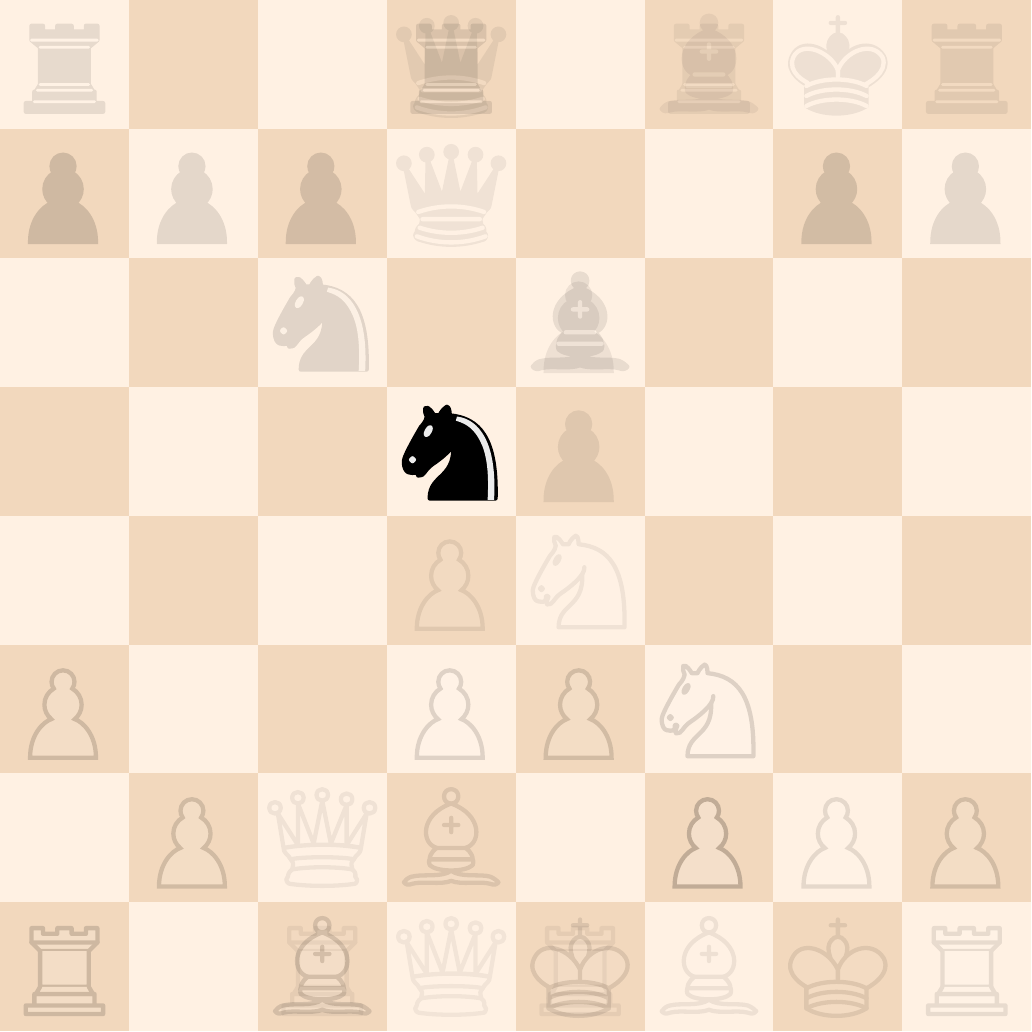}
\includegraphics[width=0.18\textwidth]{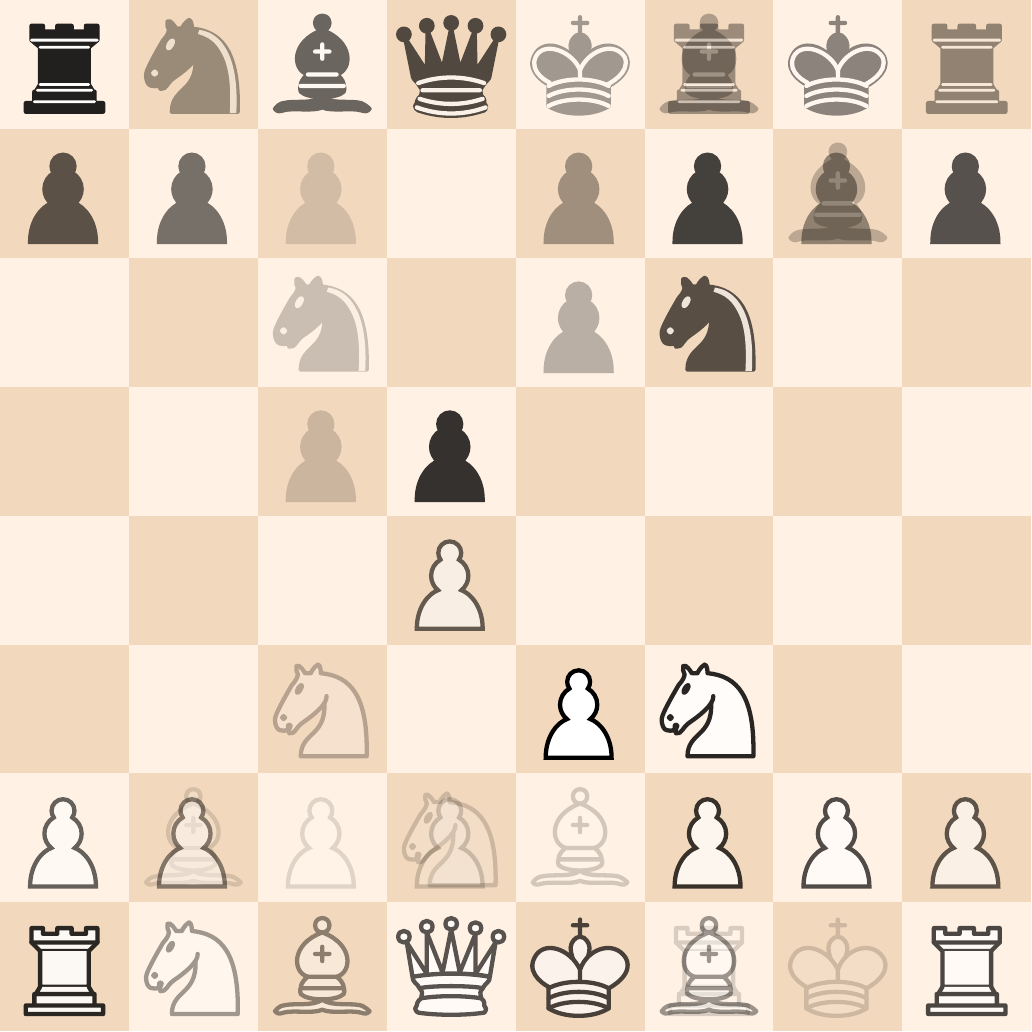}
\includegraphics[width=0.18\textwidth]{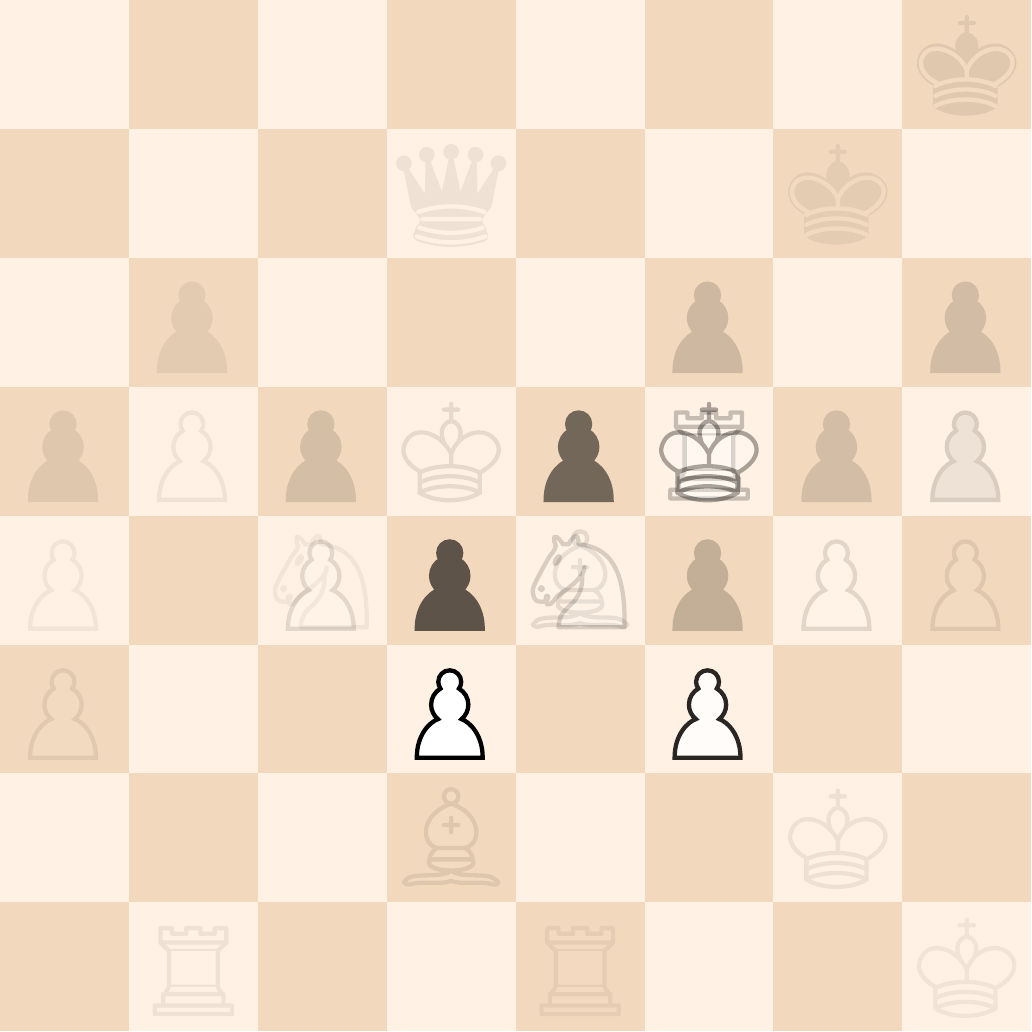}
\includegraphics[width=0.18\textwidth]{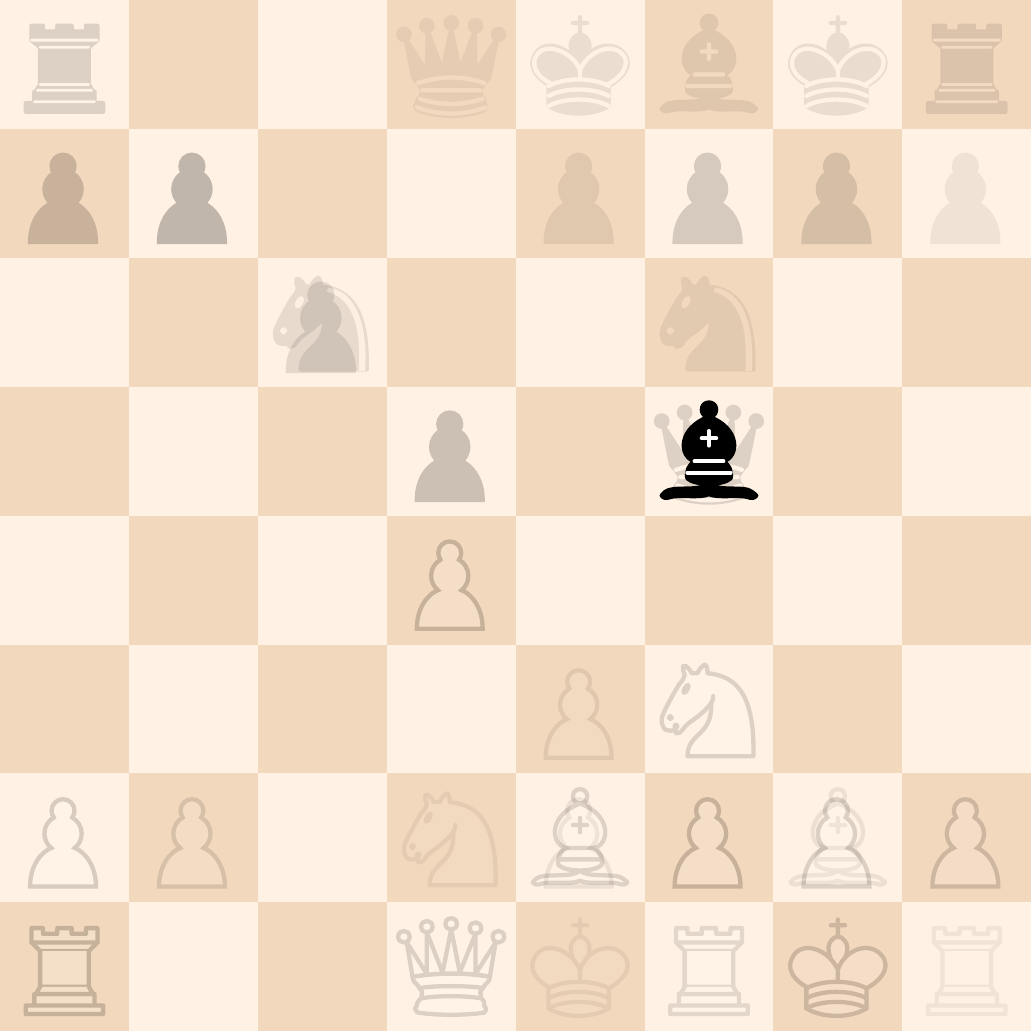} \\
\vspace{2pt}
\includegraphics[width=0.18\textwidth]{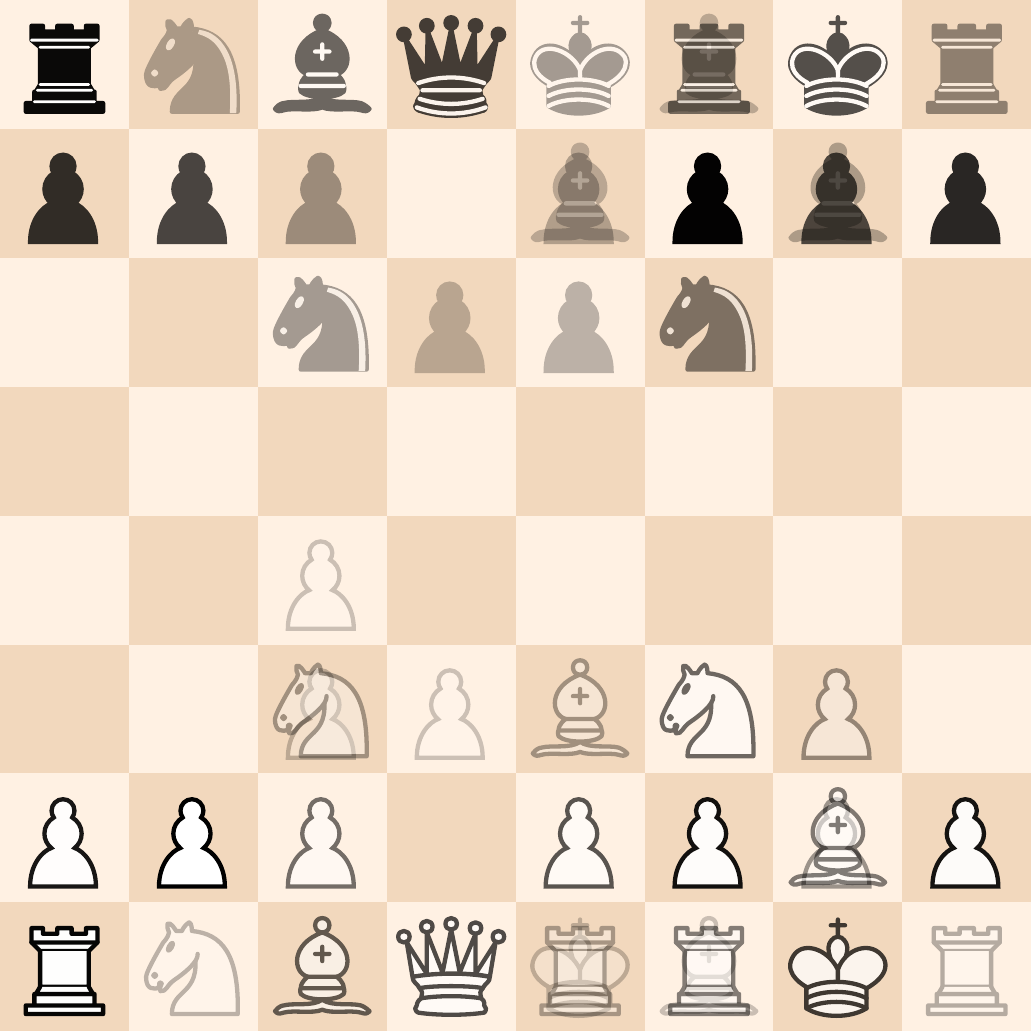}
\includegraphics[width=0.18\textwidth]{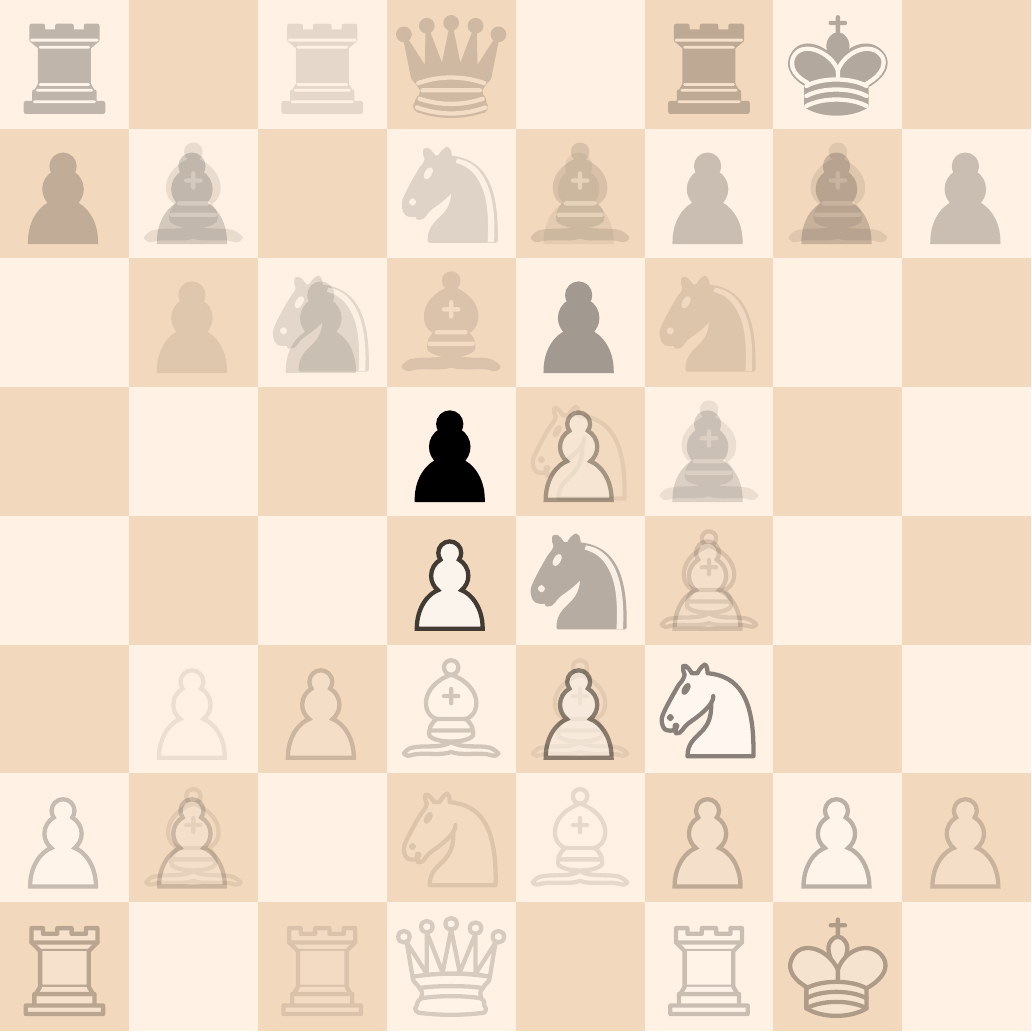}
\includegraphics[width=0.18\textwidth]{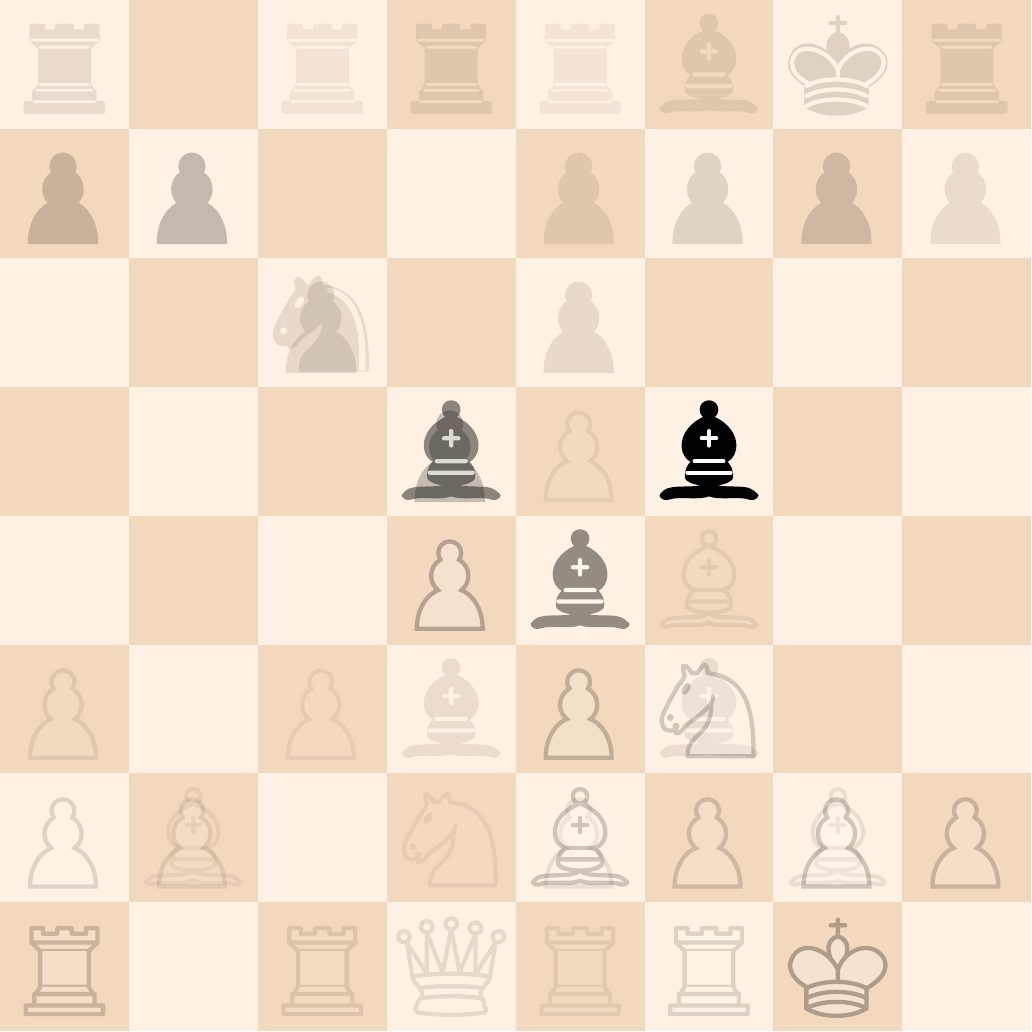}
\includegraphics[width=0.18\textwidth]{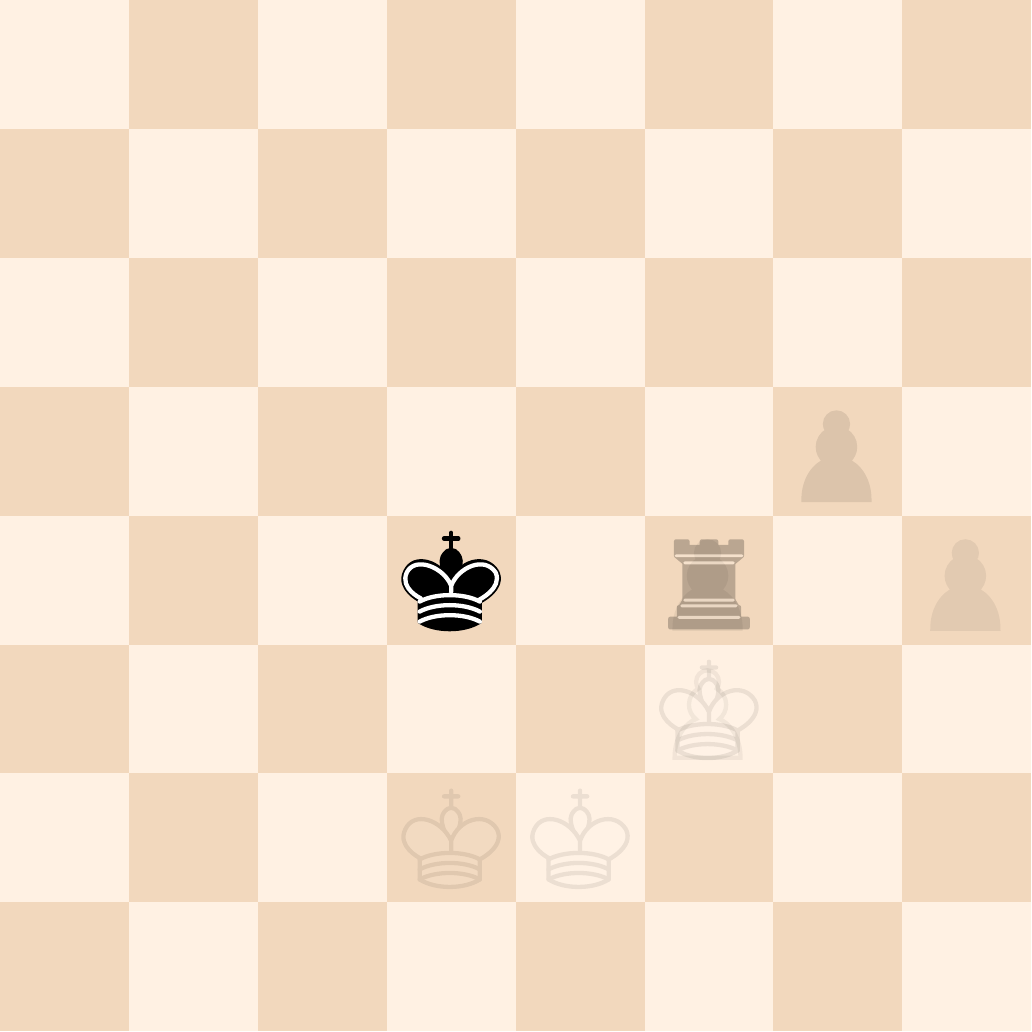}
\includegraphics[width=0.18\textwidth]{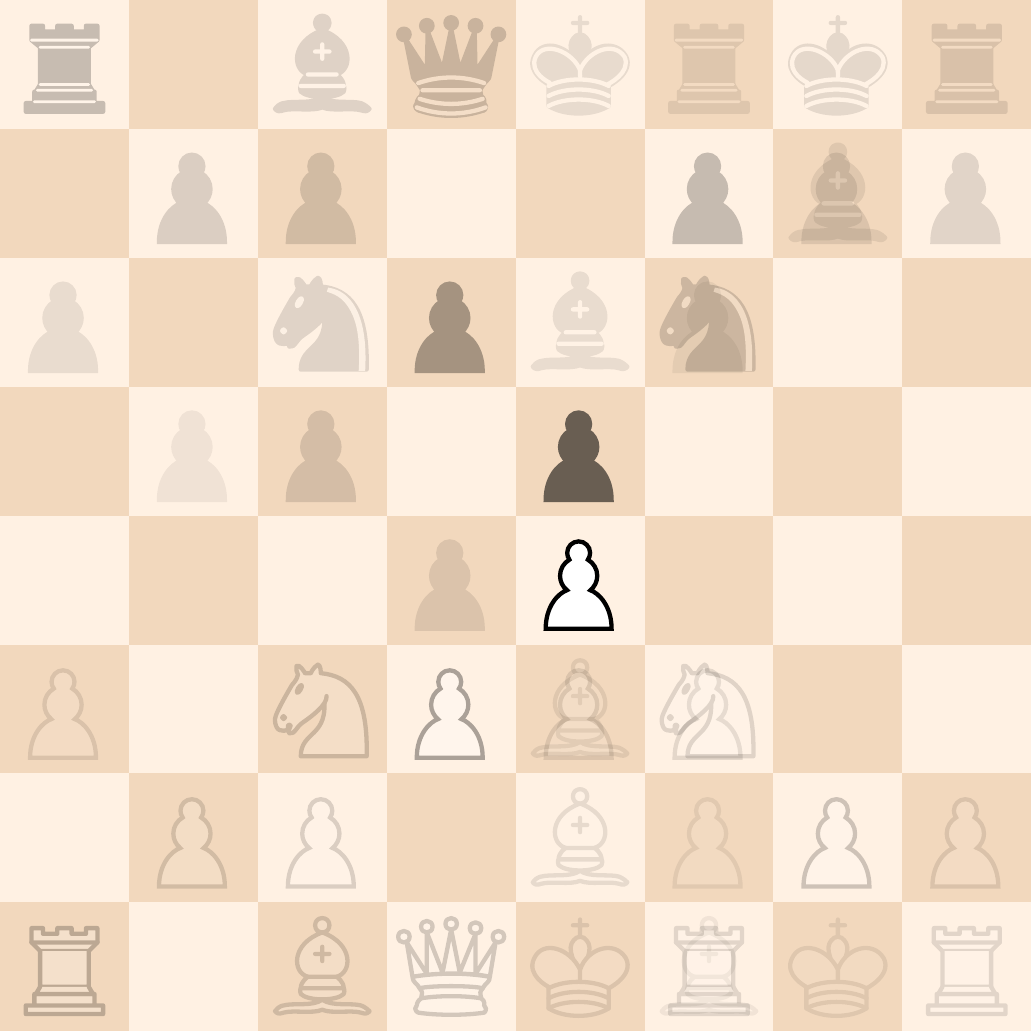} \\
\vspace{2pt}
\includegraphics[width=0.18\textwidth]{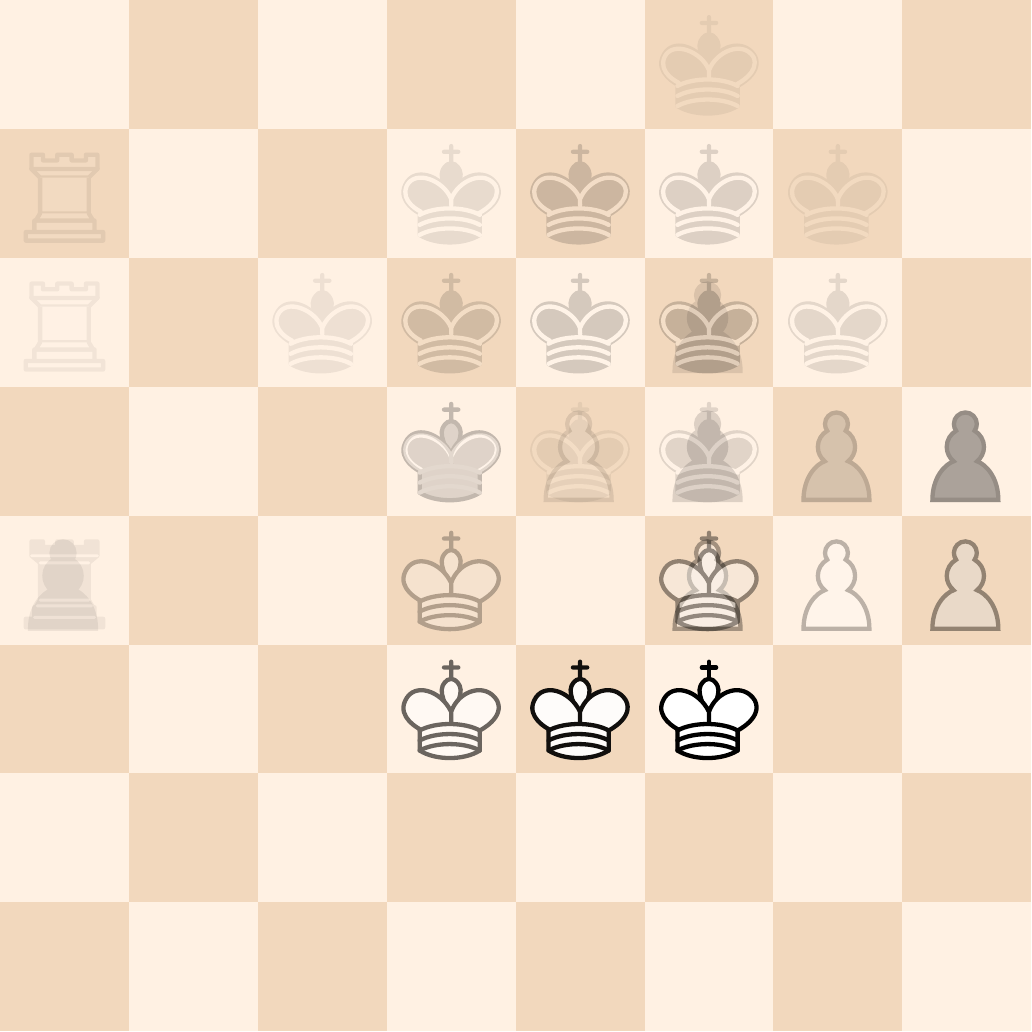}
\includegraphics[width=0.18\textwidth]{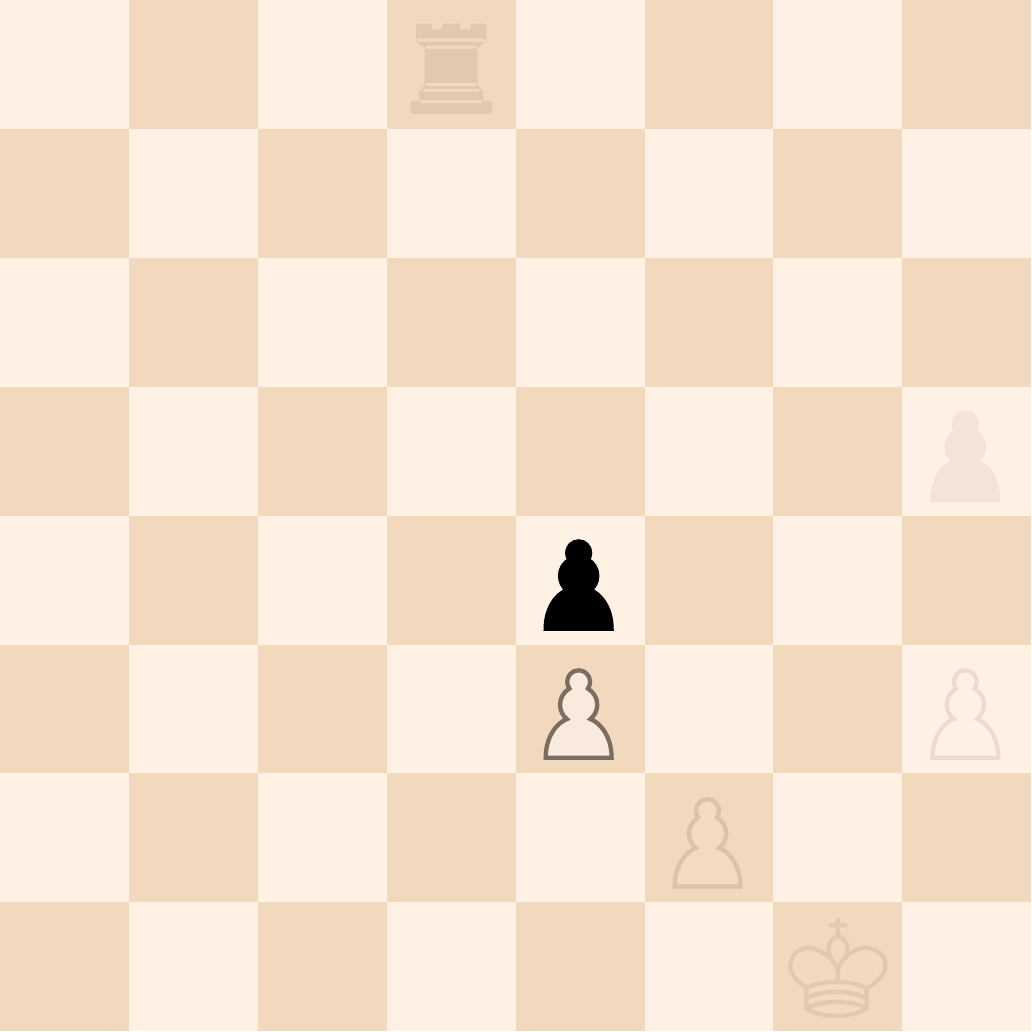}
\includegraphics[width=0.18\textwidth]{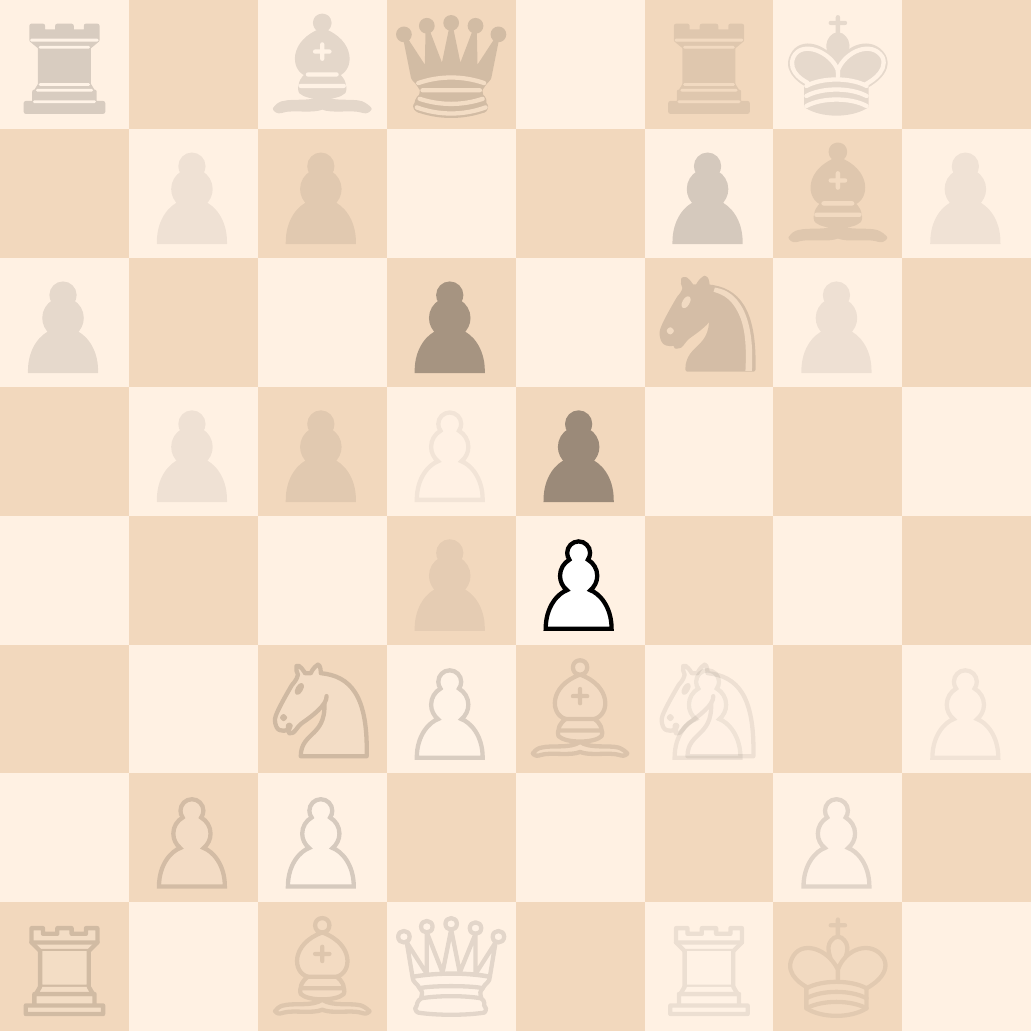}
\includegraphics[width=0.18\textwidth]{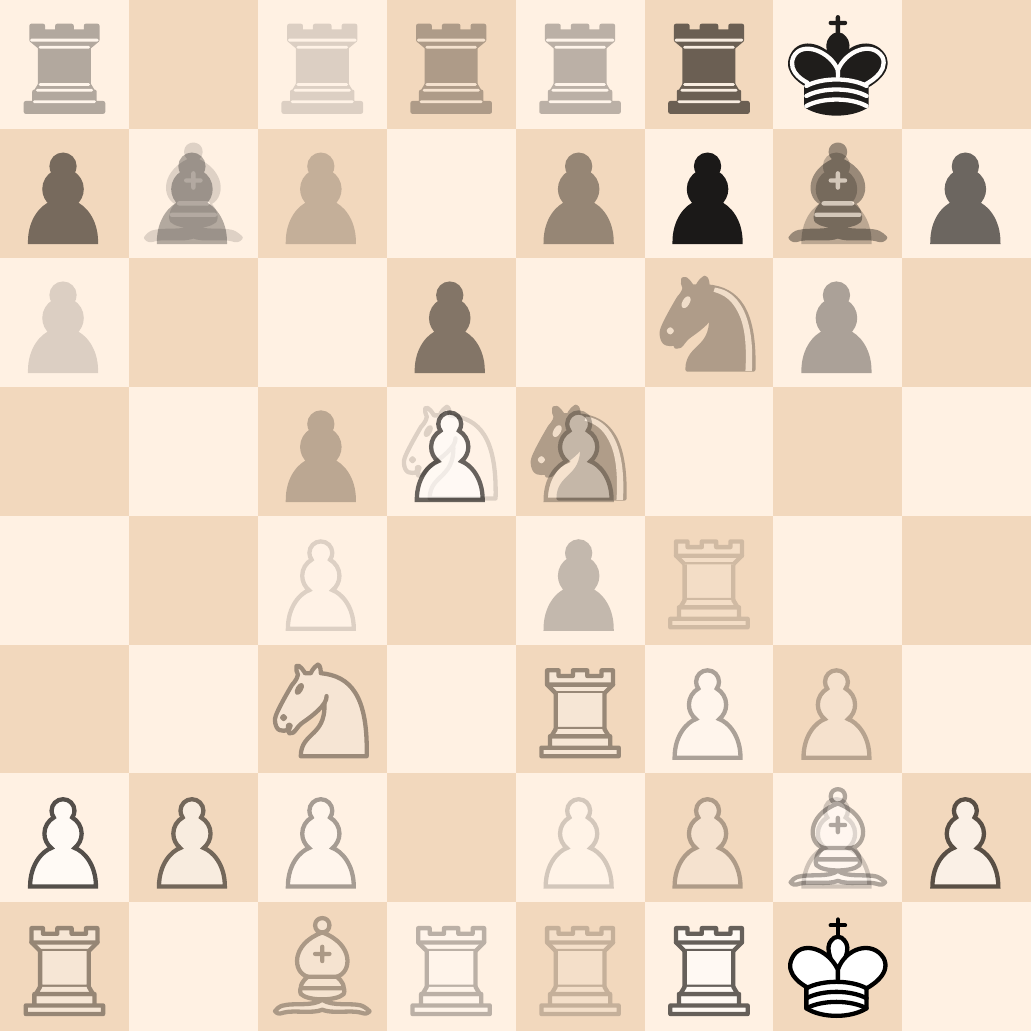}
\includegraphics[width=0.18\textwidth]{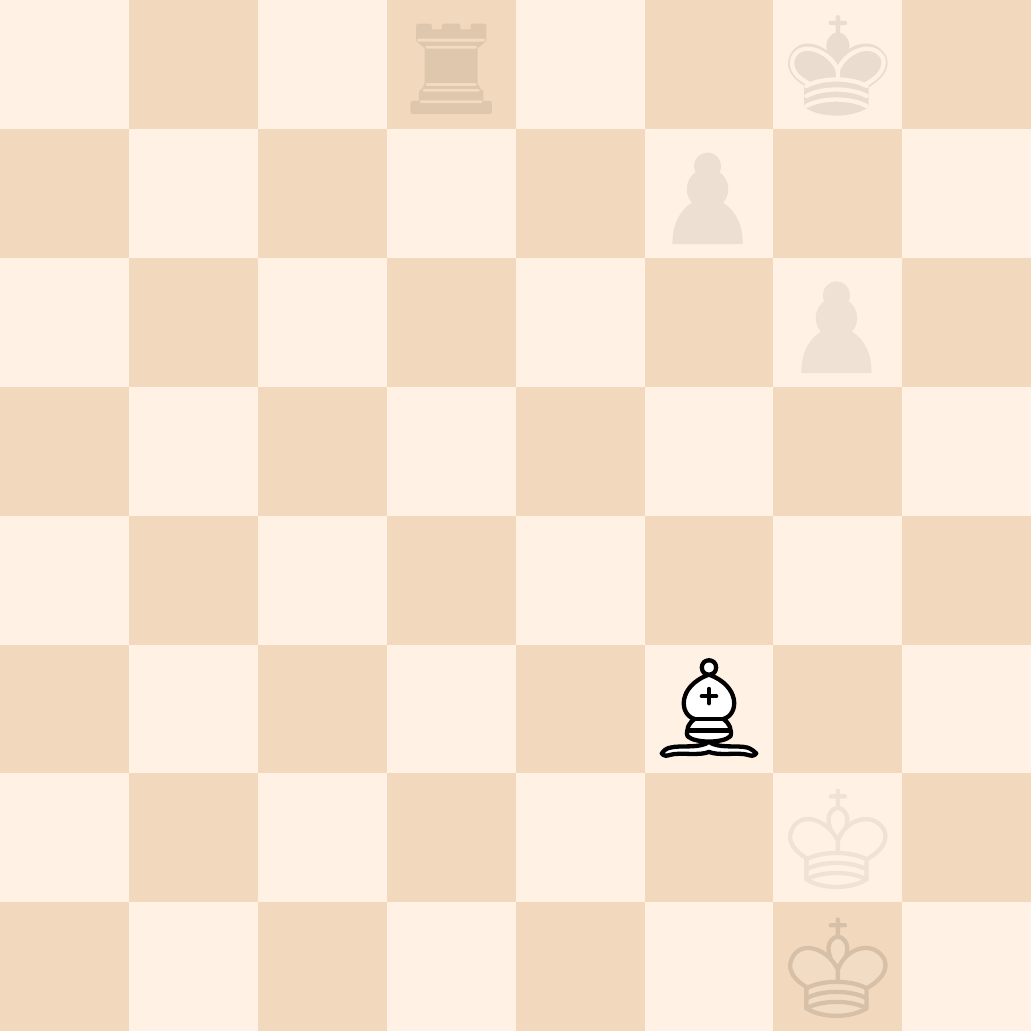}
\caption{The activation input covariance $\mathrm{cov}(z^1_{i} , \z^0)$ for the
\textbf{first layer}, with activations $i = (5, 4, c)$, iterating from channels $c = 1, \ldots, 15$ for layer $l=1$.
The top row shows channels 1 to 5,
the middle row channels 6 to 10,
and the bottom row channels 11 to 15.
Because $(5, 4, c)$ is centered at square e4,
this is a feature detector for groupings of pieces around e4.}
\label{fig:activation-input-covariance-layer-1}
\end{figure}

There is evidence that lower layers are likely to encode less abstract and therefore more location specific concepts \cite{bau2017network}.
To illustrate the role of lower layers as feature detectors, we pick square e4 and visualize the activations from layer 1 that align with it spatially.
In Figure \ref{fig:activation-input-covariance-layer-1}
we show the activation-input covariances of channels $c = 1, \ldots, 15$ of $i = (5, 4, c)$ in $\z^1$.
The choice of the first fifteen channels over any of the other 256 channels is arbitrary.
Layer 1 in Figure \ref{fig:alphazero-network} follows after applying a bank of $3 \times 3$ convolution filters, and we expect only local $3 \times 3$ patterns around e4 to be correlated with the activations in all channels.
The correlations are indeed localized; index $(5,4)$ corresponds to the e4 square on the board, and after a $3 \times 3$ convolution (see Figure \ref{eq:az-network}) piece patterns around the e4 square are detected.

There are also a number of global positions that appear in the covariance visualizations in Figure \ref{fig:activation-input-covariance-layer-1}.
It is because of \emph{correlations in the data}: the features they detect are globally correlated to other piece positions on the board in the sample of grandmaster chess positions.
As an example,
activation $i = (5,4,6)$ detects the pattern ({\sympawn d3}, {\symbishop e3}, {\symknight f3}) with both rows (d4, e4, f4) and (d5, e5, f5) in front of it being empty. This piece pattern predominantly appears 
in certain opening positions and almost never elsewhere in middle- and endgames, and hence rooks on a1 and a8 are also visible.

\begin{figure}
\centering
\includegraphics[width=0.18\textwidth]{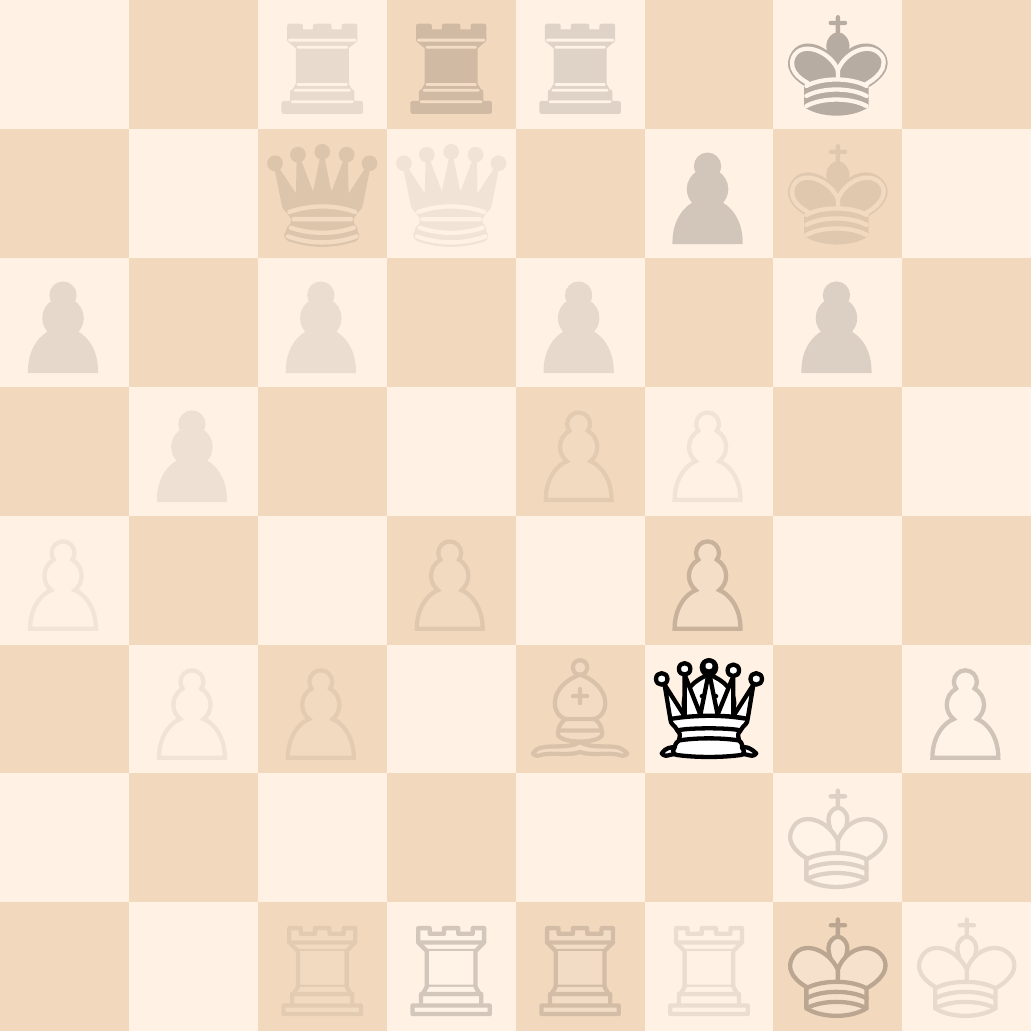}
\includegraphics[width=0.18\textwidth]{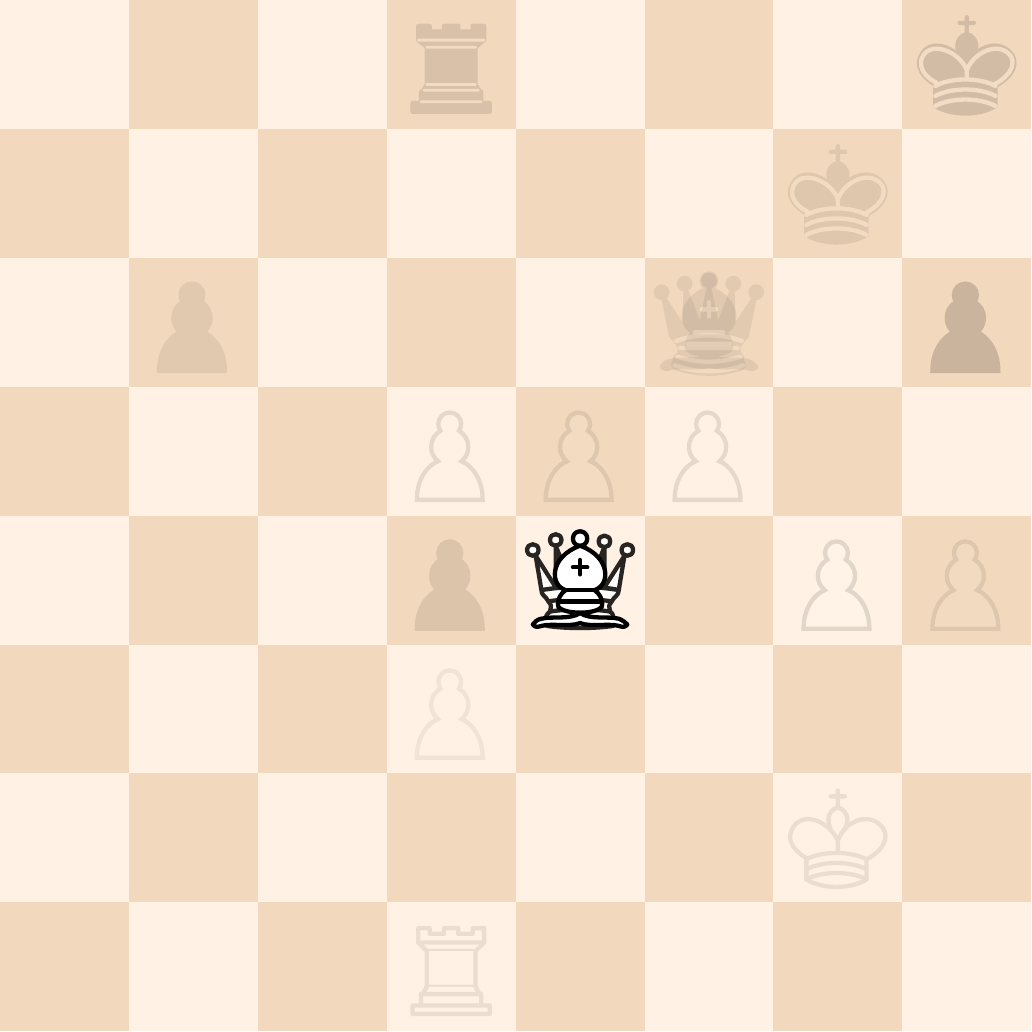}
\includegraphics[width=0.18\textwidth]{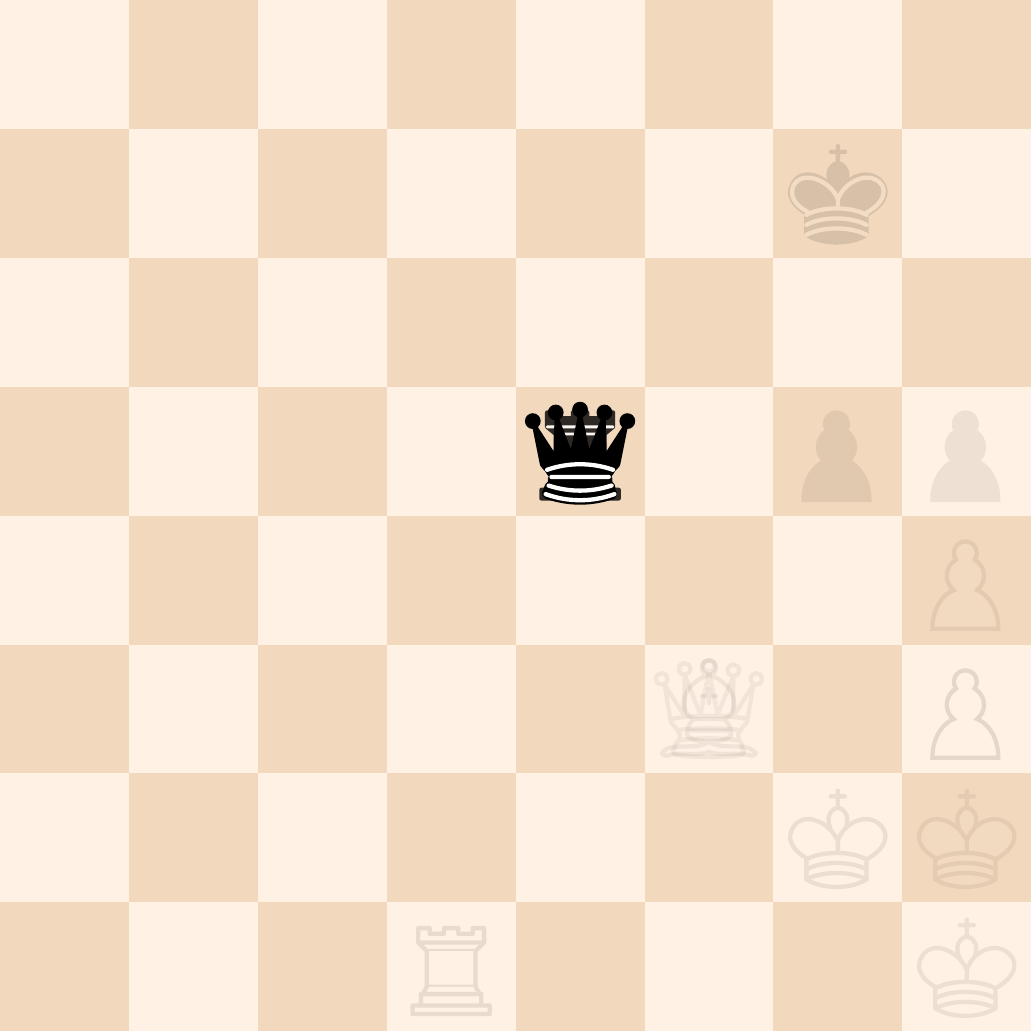}
\includegraphics[width=0.18\textwidth]{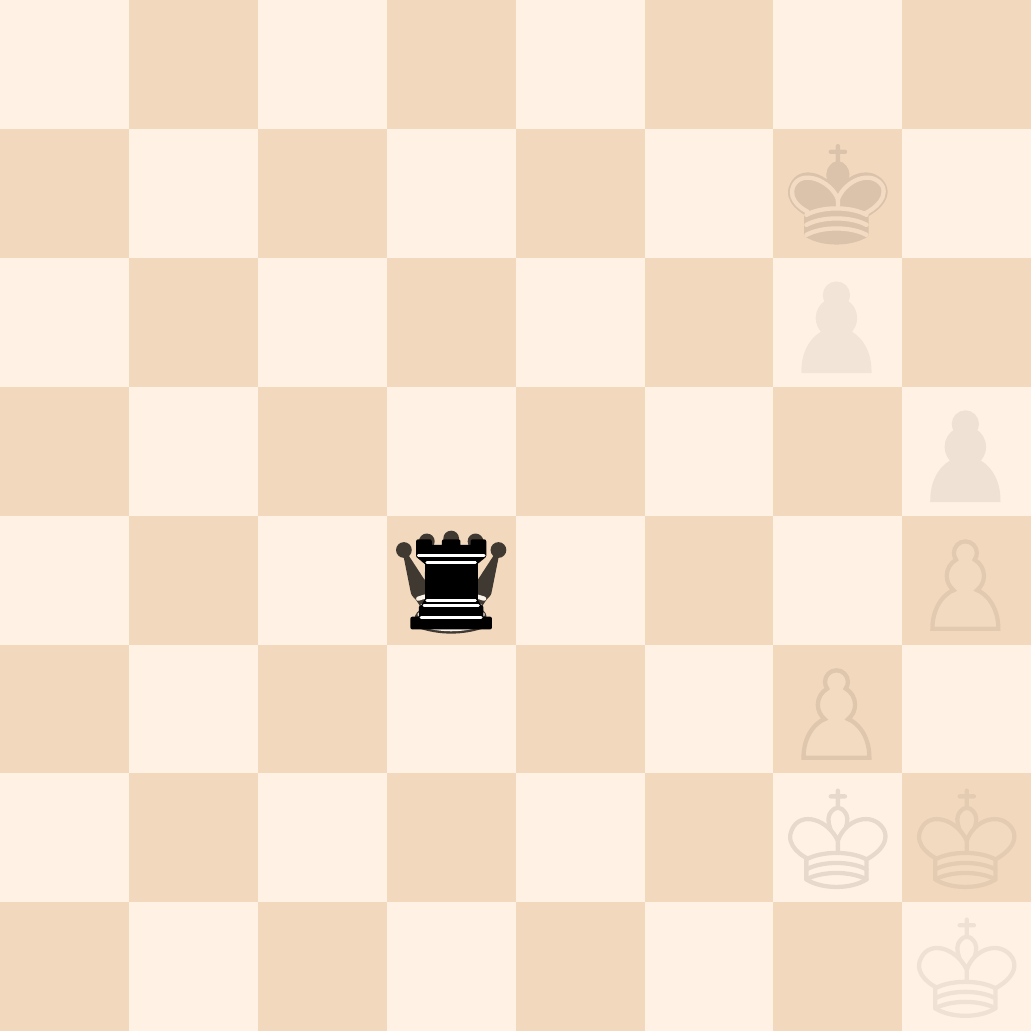}
\includegraphics[width=0.18\textwidth]{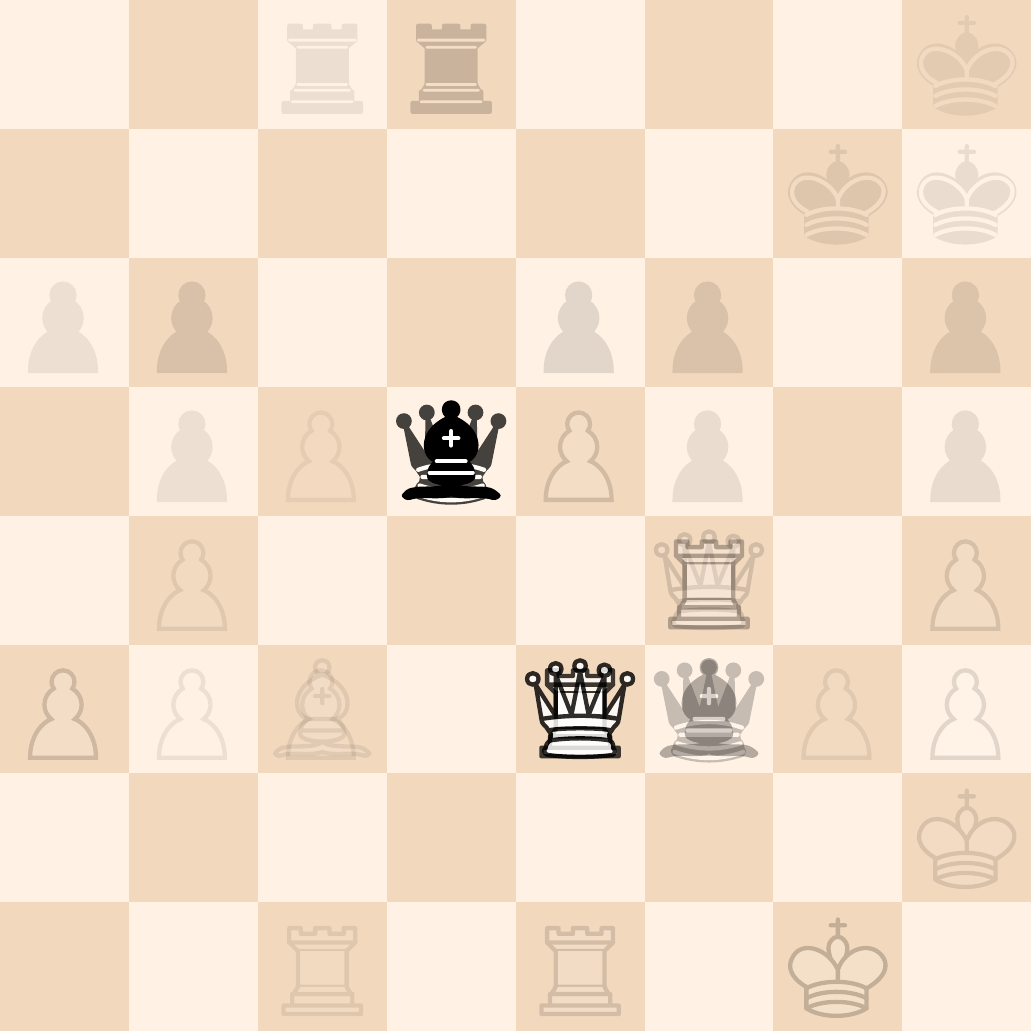}
\caption{Activations detecting \textbf{move types from a square}, illustrated with the activation input covariance $\mathrm{cov}(z^1_{i} , \z^0)$ for the
\textbf{first layer}.
From left to right:
Activations $i = (5, 4, 23)$ and $(5, 4, 90)$
show a \symbishop \ and \symqueen \ overlaid, and
detect possible long-range diagonal piece movement from f3 and e4 (compare $i = (5, 4, 15)$ in the bottom right of Figure \ref{fig:activation-input-covariance-layer-1}, which only detects the bishop).
Activations $i = (5, 4, 33)$ and $(5, 4, 92)$
show a \symrook \ and \symqueen \ overlaid, and detect when the (opponent's) piece on e5 can move horizontally or vertically across the board.
Activation $i = (5, 4, 91)$
looks for two patterns: a diagonal moving piece (\symbishop \ or \symqueen) on d5 and a horizontal-vertical moving piece (\symrook \ or \symqueen) on e3.}
\label{fig:activation-input-covariance-layer-1-special}
\end{figure}

The features detected by the activations in $\z^1$ go beyond detecting specific localized patterns of pieces. There are a small number of activations that clearly detect \emph{move types} from a square.
In Figure \ref{fig:activation-input-covariance-layer-1-special} we identified the first five such activations for the channels corresponding to the e4 square.
There, the activation-input correlations are large when there is a long-range diagonal-moving piece on a square (either a Bishop or a Queen) or a long-range horizontal or vertical-moving piece (either a Rook or a Queen).

\subsubsection{Persisted feature detectors}
\label{sec:persisted-feature-detectors}

The structure of the ResNet links activations $z^l_i$ and $z^{l-1}_i$ with
\begin{equation} \label{eq:resnet-block-i}
z^l_i = \mathrm{ReLU}(z^{l-1}_i + g^l(\z^{l-1})_i) \ ,    
\end{equation}
as given in \eqref{eq:resnet-block}.
Depending on the purpose of the activation $z^l_i$ in the distributed
encoding $\z^l$, it combines
\begin{enumerate}
\item propagating information forward through $z^{l-1}_i$ and
\item evolving by incorporating additional local context through $g^l(\z^{l-1})_i$.
\end{enumerate}
In the first case, information is persisted.
Some feature might be useful in a distributed representation at any layer in the network,
and the feature propagated forward in each layer of the network.
As an example, it is 
important that information about the position of pieces are
propagated through the network, as the prior $\p$ in \eqref{eq:az-network} has to put mass on legal moves.
Equivalently, other local features might be universally useful
in each layer of representation, and the structure of
\eqref{eq:resnet-block-i}
allows activations to be propagated forward.

\begin{figure}[t]
\centering
\includegraphics[width=0.18\textwidth]{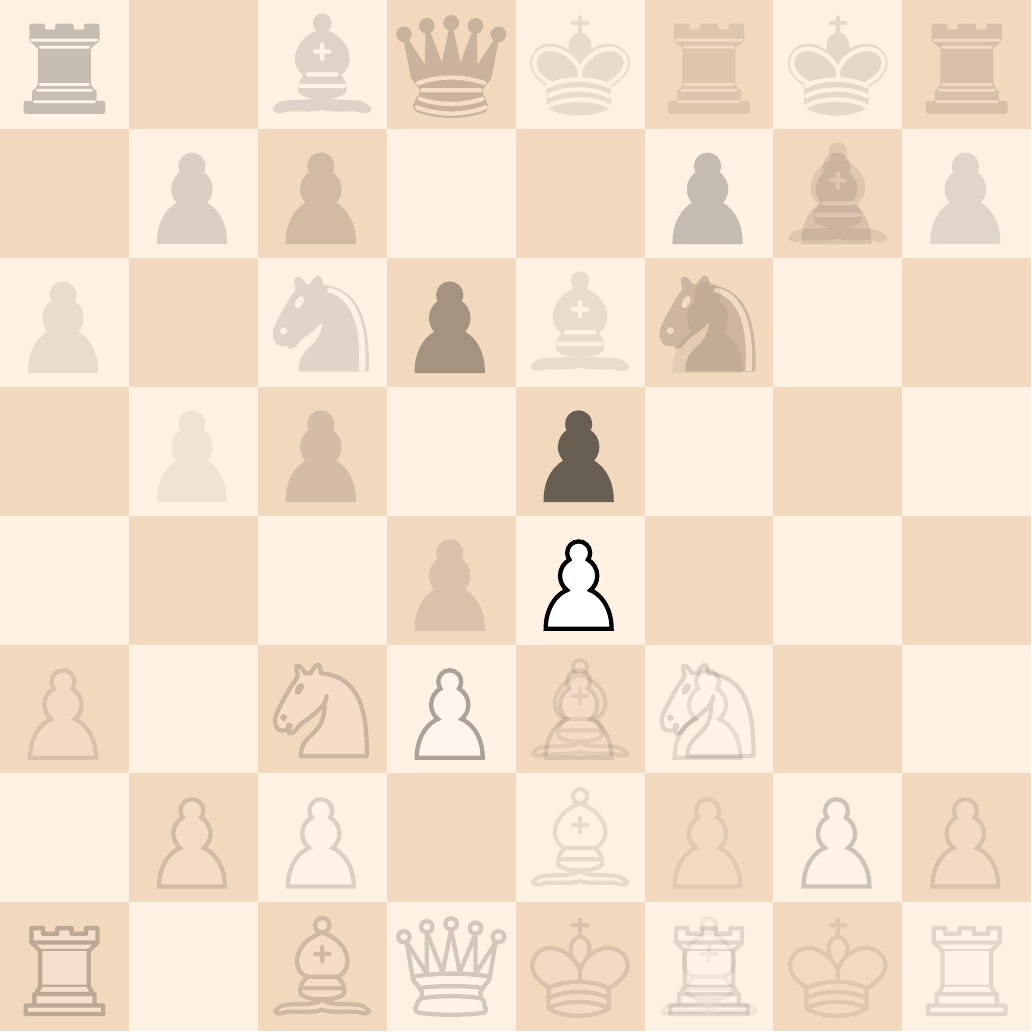}
\includegraphics[width=0.18\textwidth]{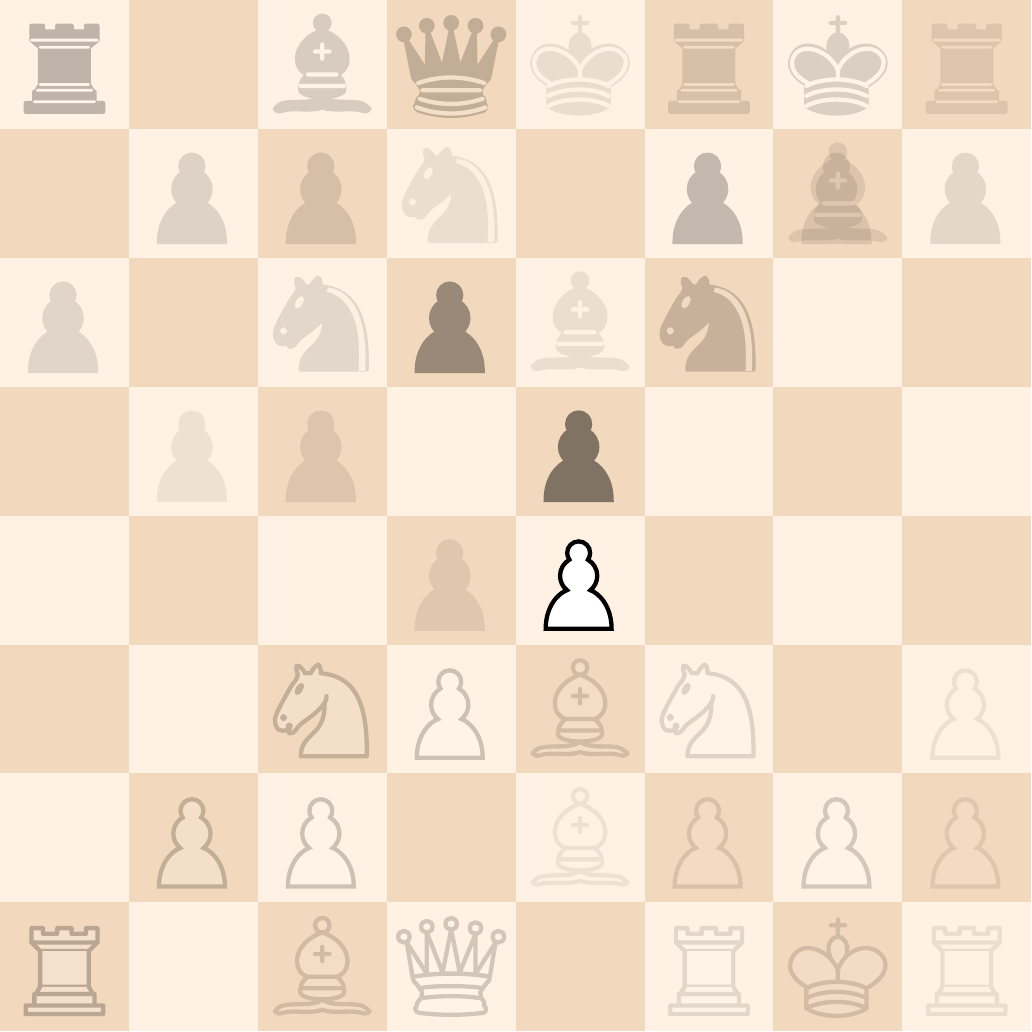}
\includegraphics[width=0.18\textwidth]{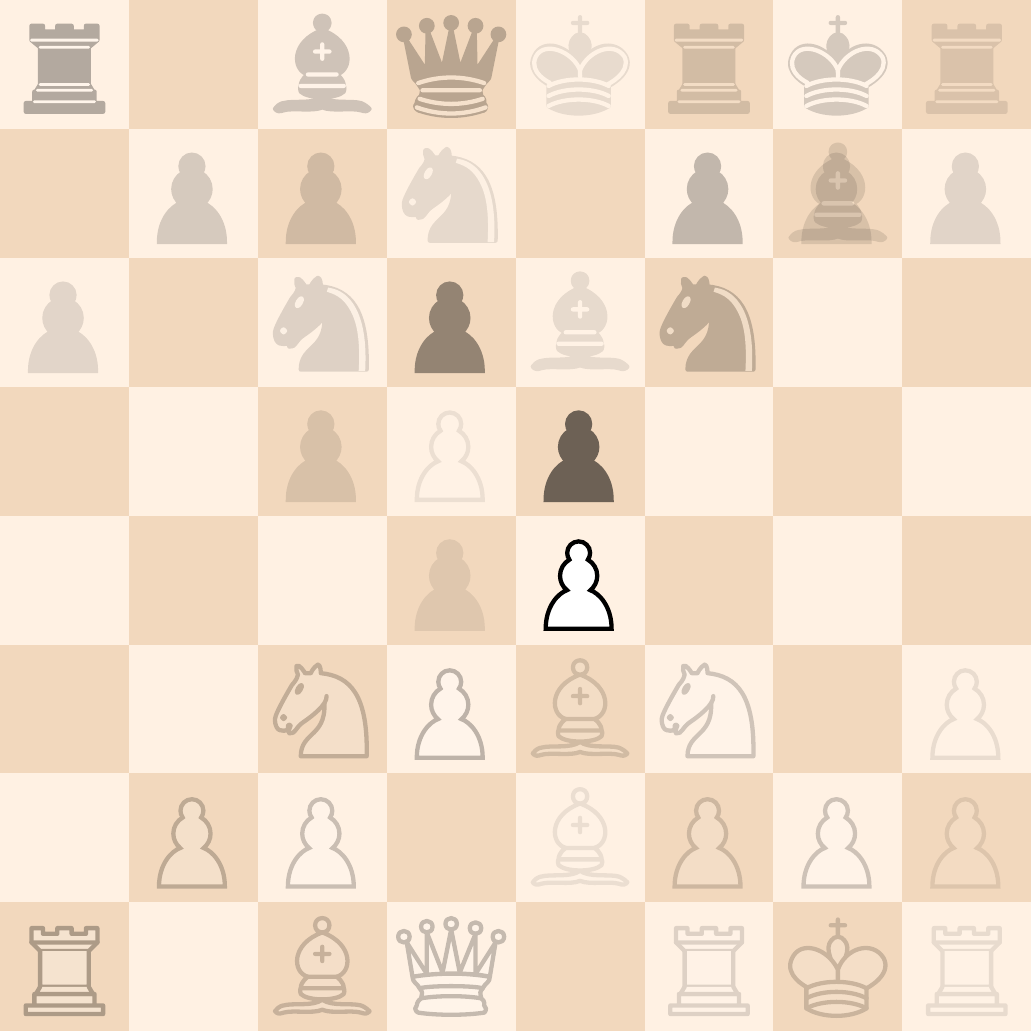}
\includegraphics[width=0.18\textwidth]{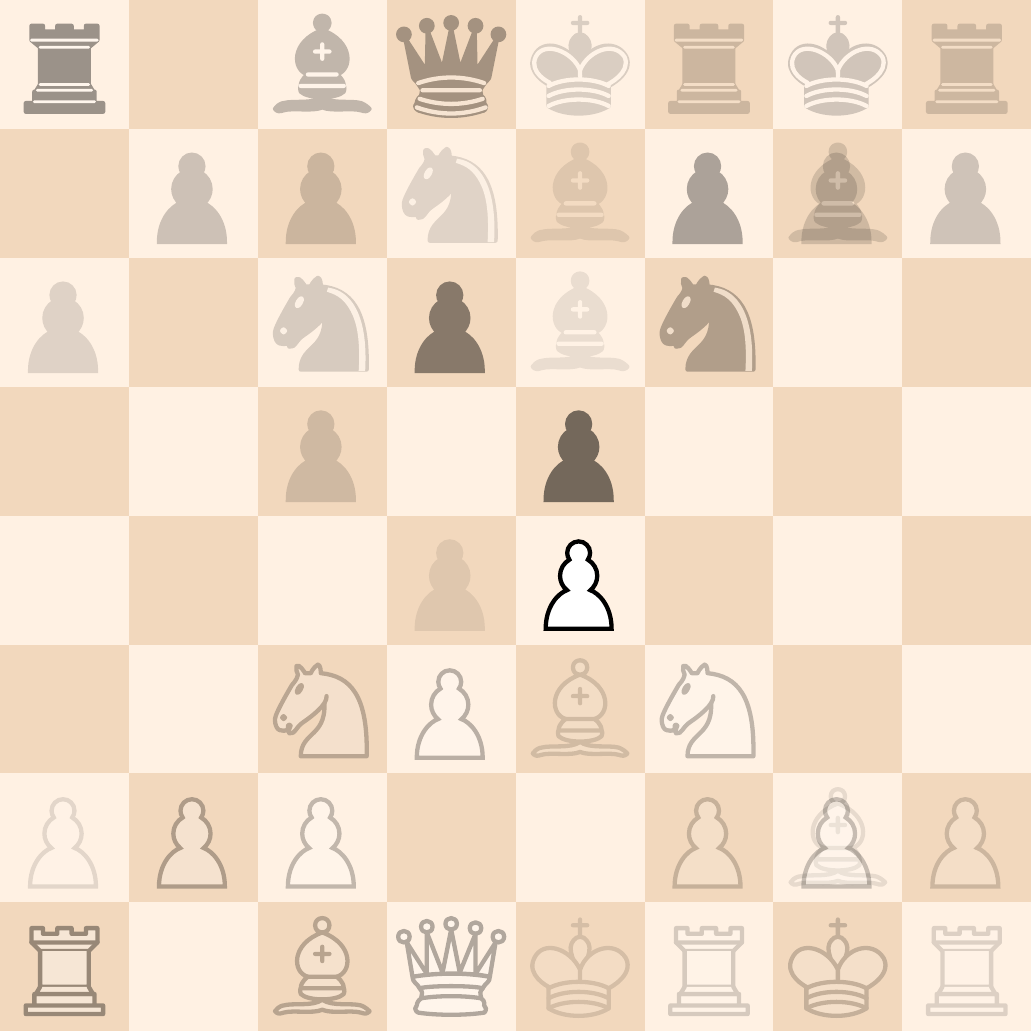}
\includegraphics[width=0.18\textwidth]{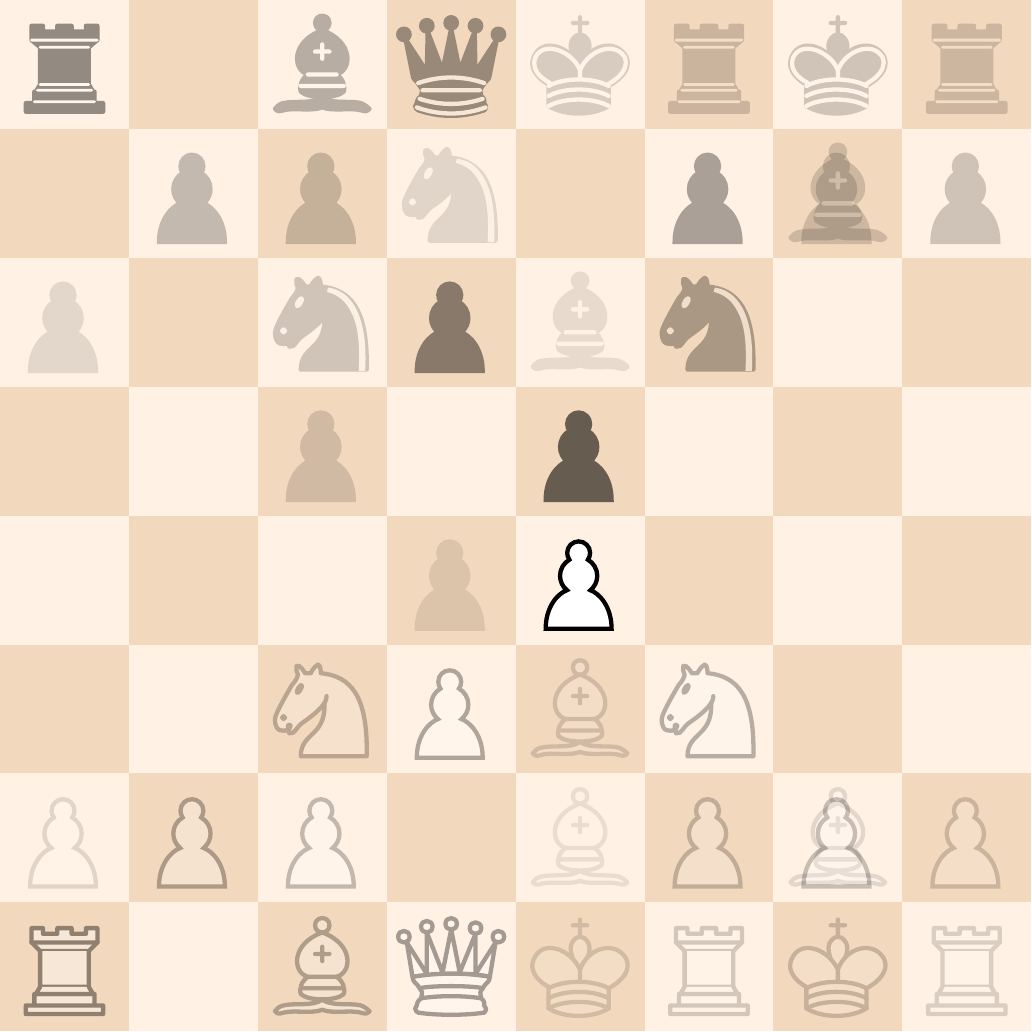} \\
activation-input covariance for activation $i = (5, 4, 10)$; layers 1 to 5 \\
\vspace{2pt}
\includegraphics[width=0.18\textwidth]{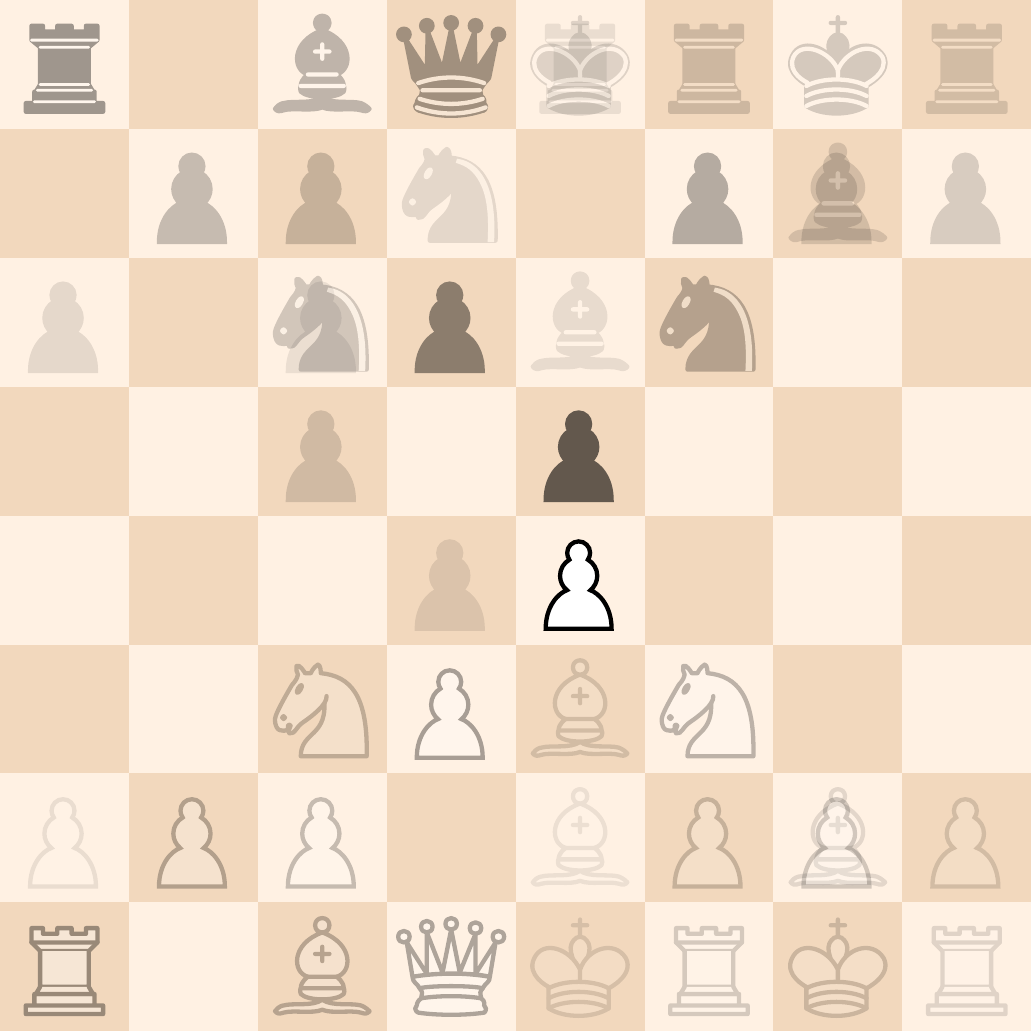}
\includegraphics[width=0.18\textwidth]{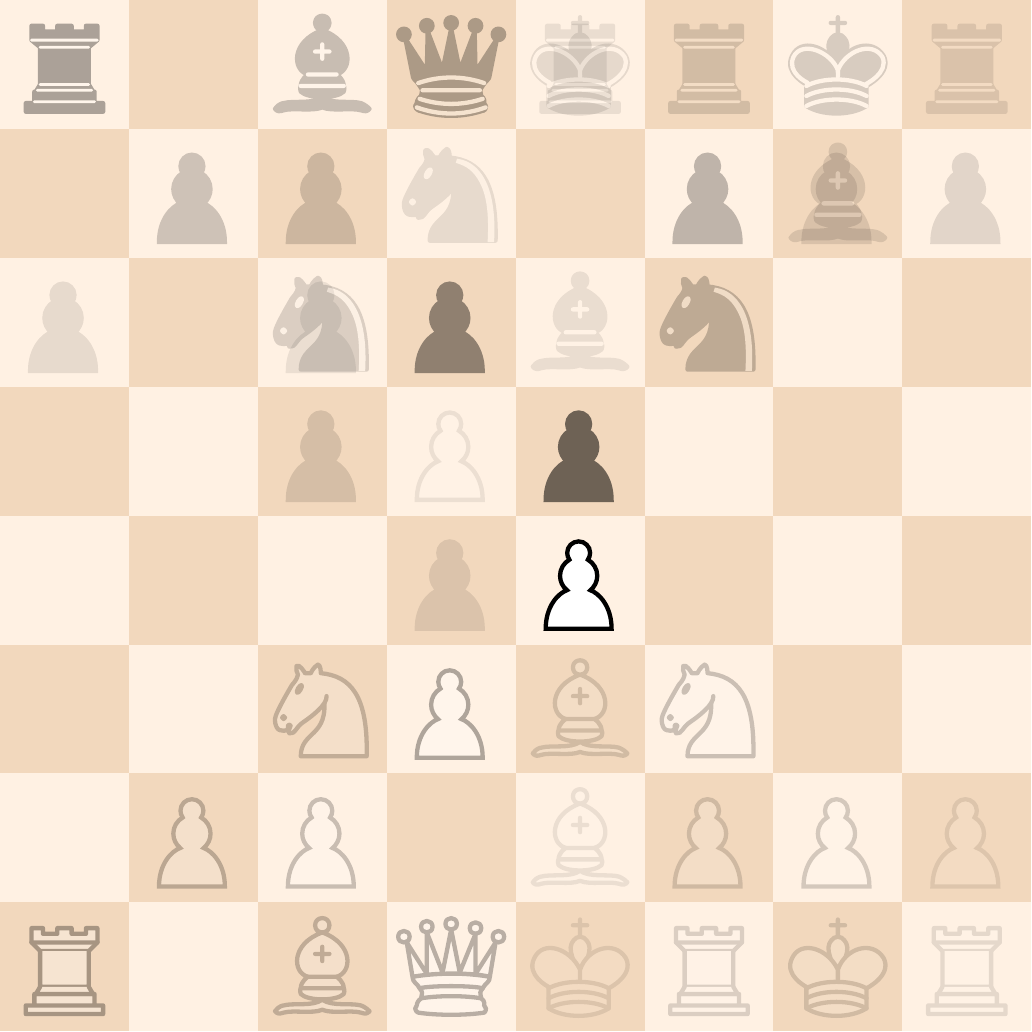}
\includegraphics[width=0.18\textwidth]{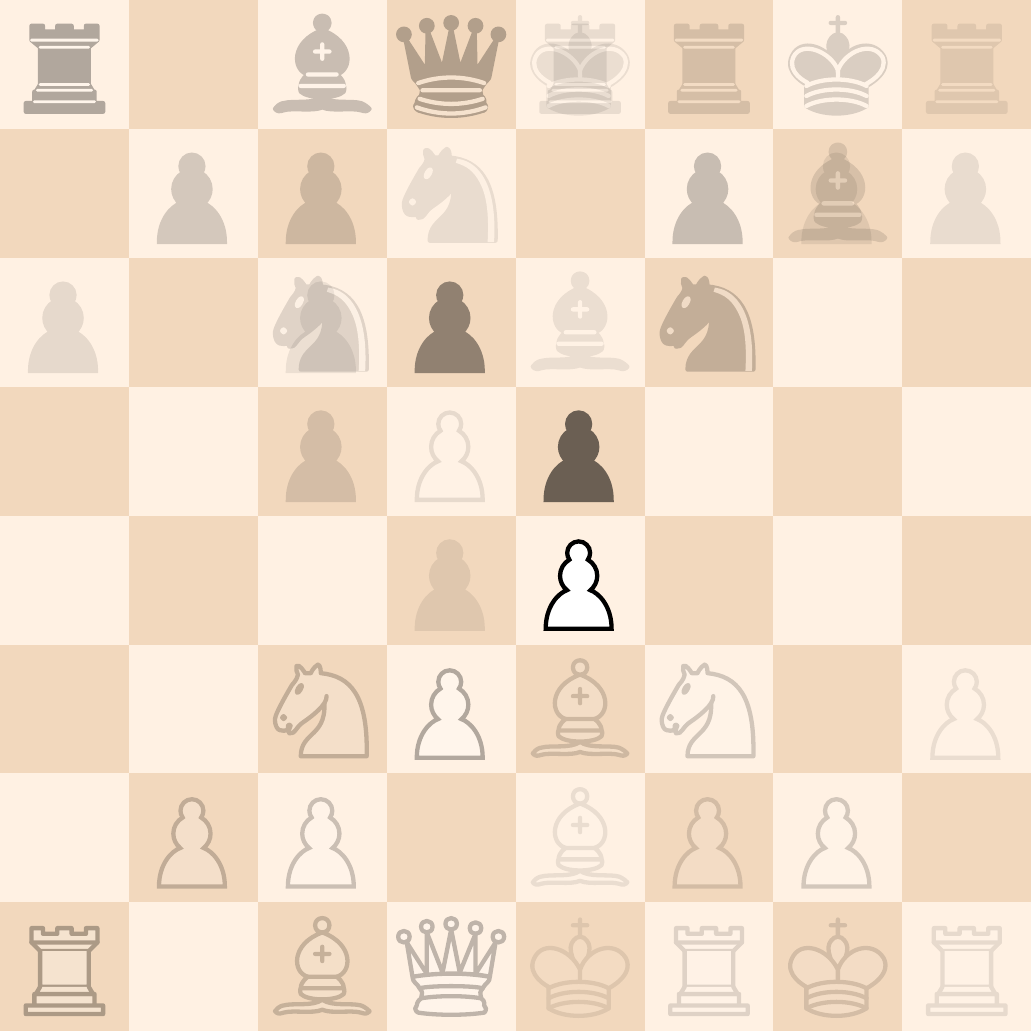}
\includegraphics[width=0.18\textwidth]{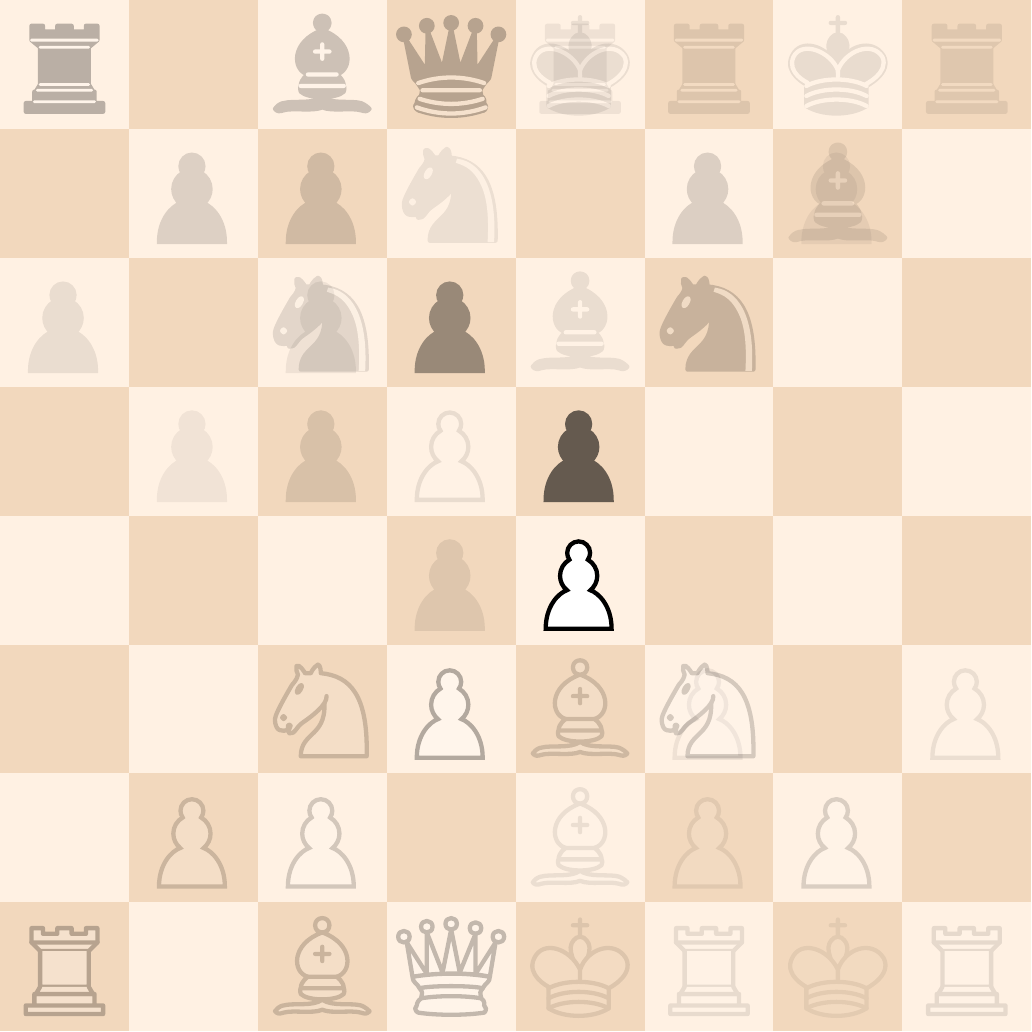}
\includegraphics[width=0.18\textwidth]{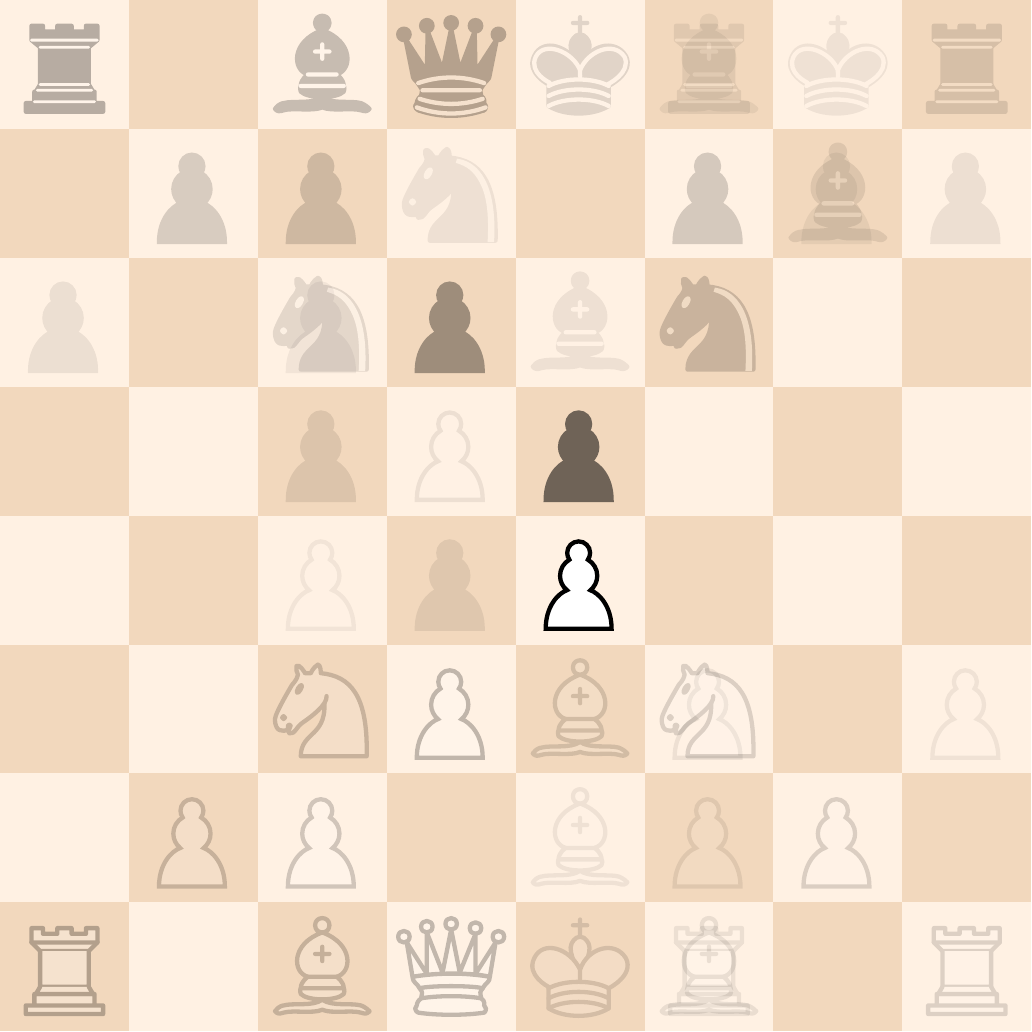} \\
activation-input covariance for activation $i = (5, 4, 10)$; layers 6 to 10 \\
\vspace{2pt}
\includegraphics[width=0.18\textwidth]{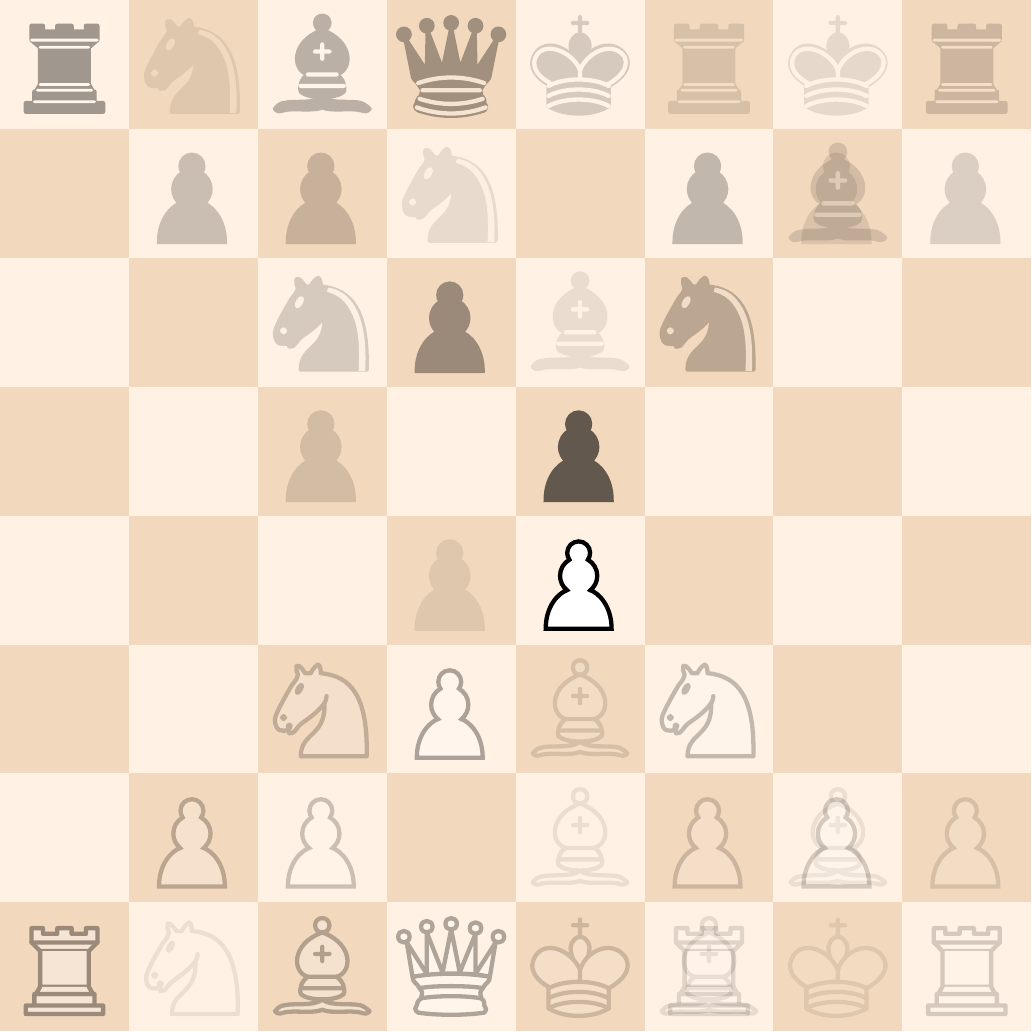}
\includegraphics[width=0.18\textwidth]{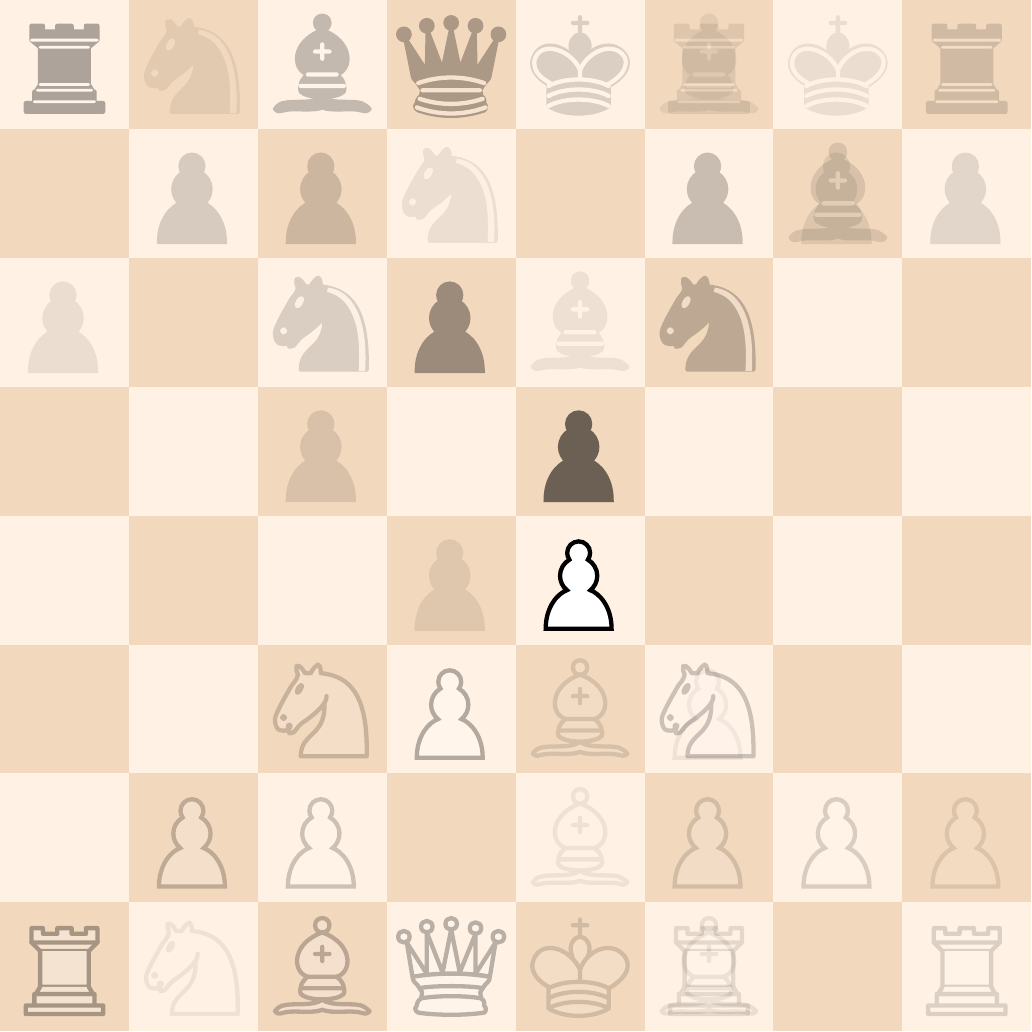}
\includegraphics[width=0.18\textwidth]{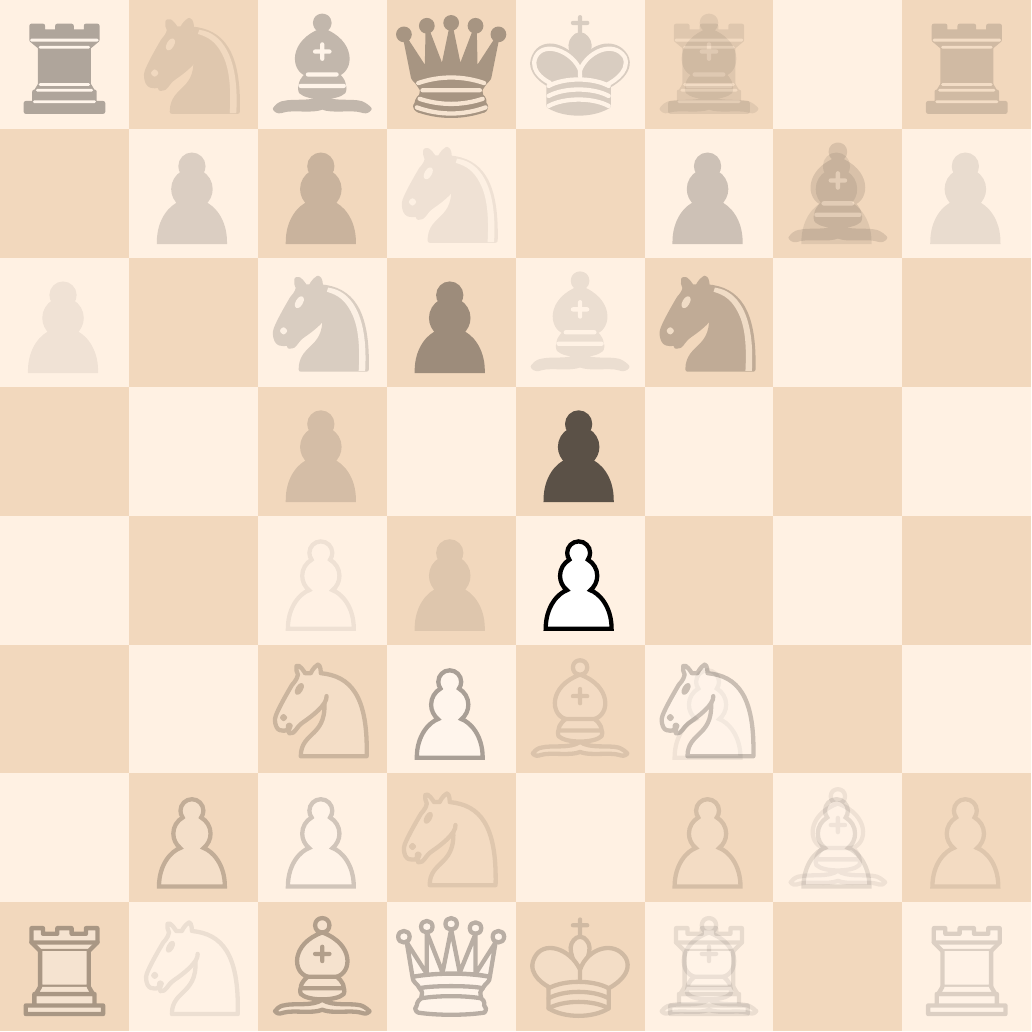}
\includegraphics[width=0.18\textwidth]{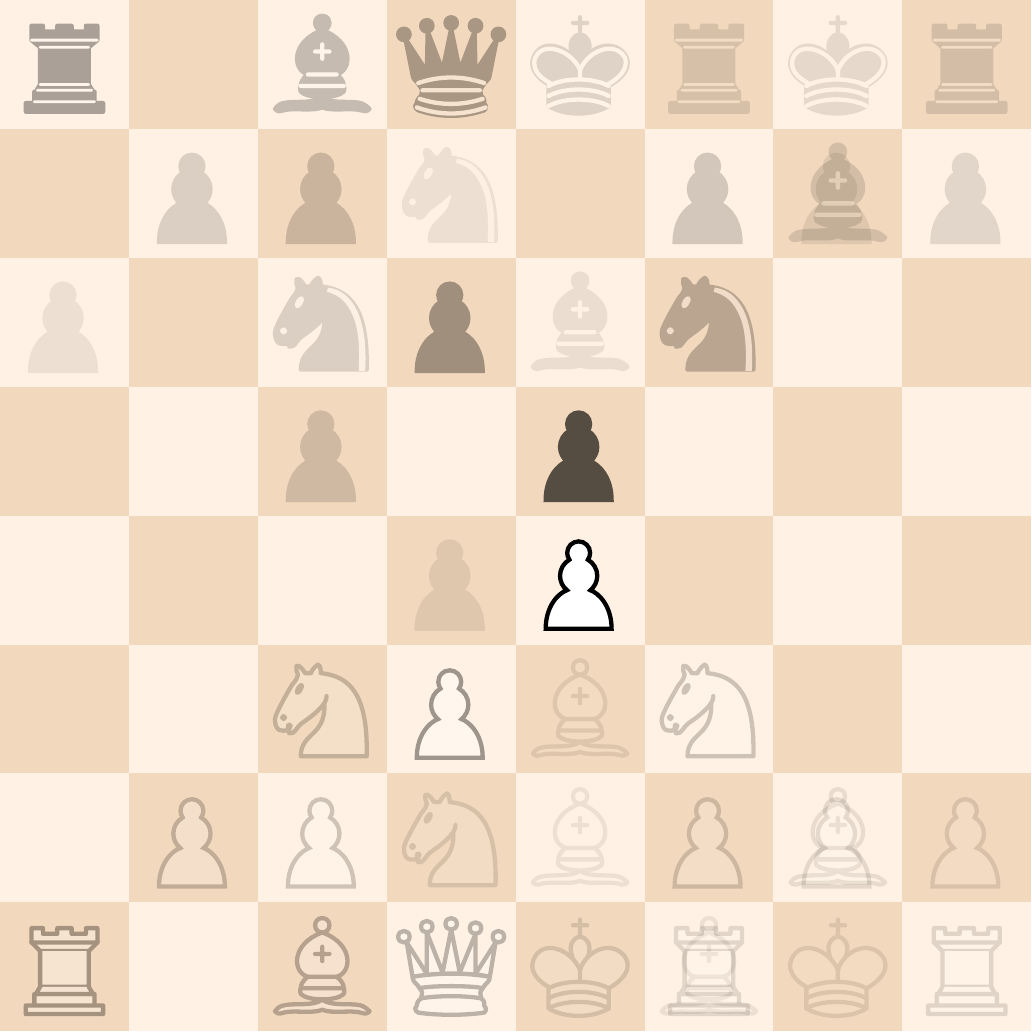}
\includegraphics[width=0.18\textwidth]{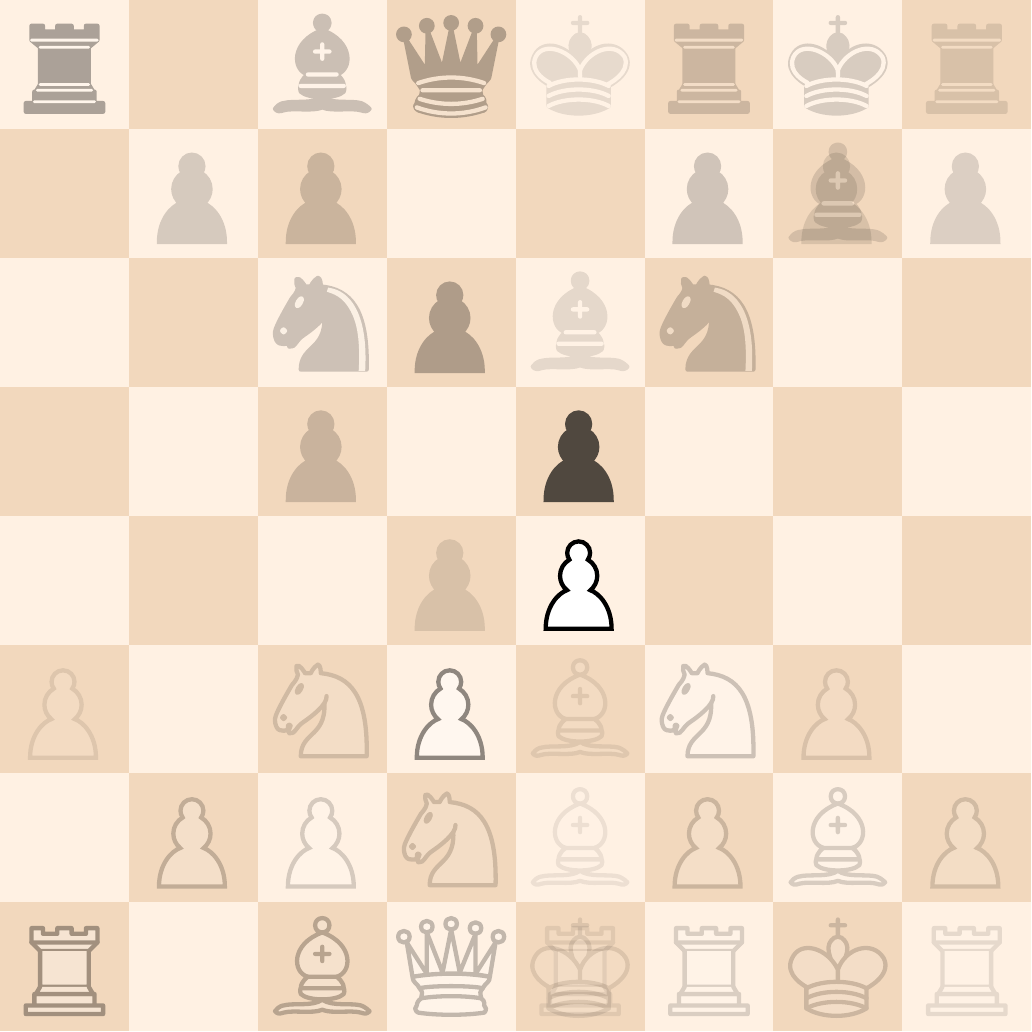} \\
activation-input covariance for activation $i = (5, 4, 10)$; layers 11 to 15 \\
\vspace{2pt}
\includegraphics[width=0.18\textwidth]{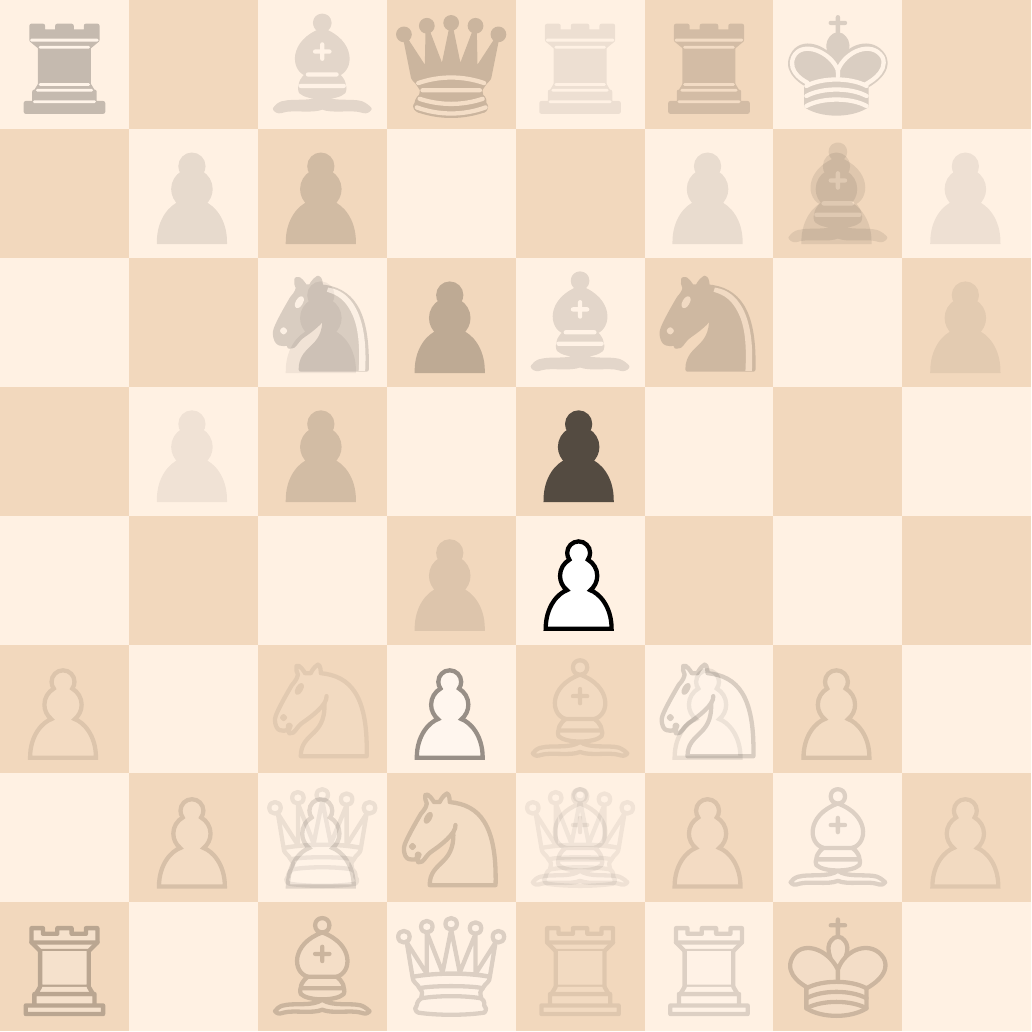}
\includegraphics[width=0.18\textwidth]{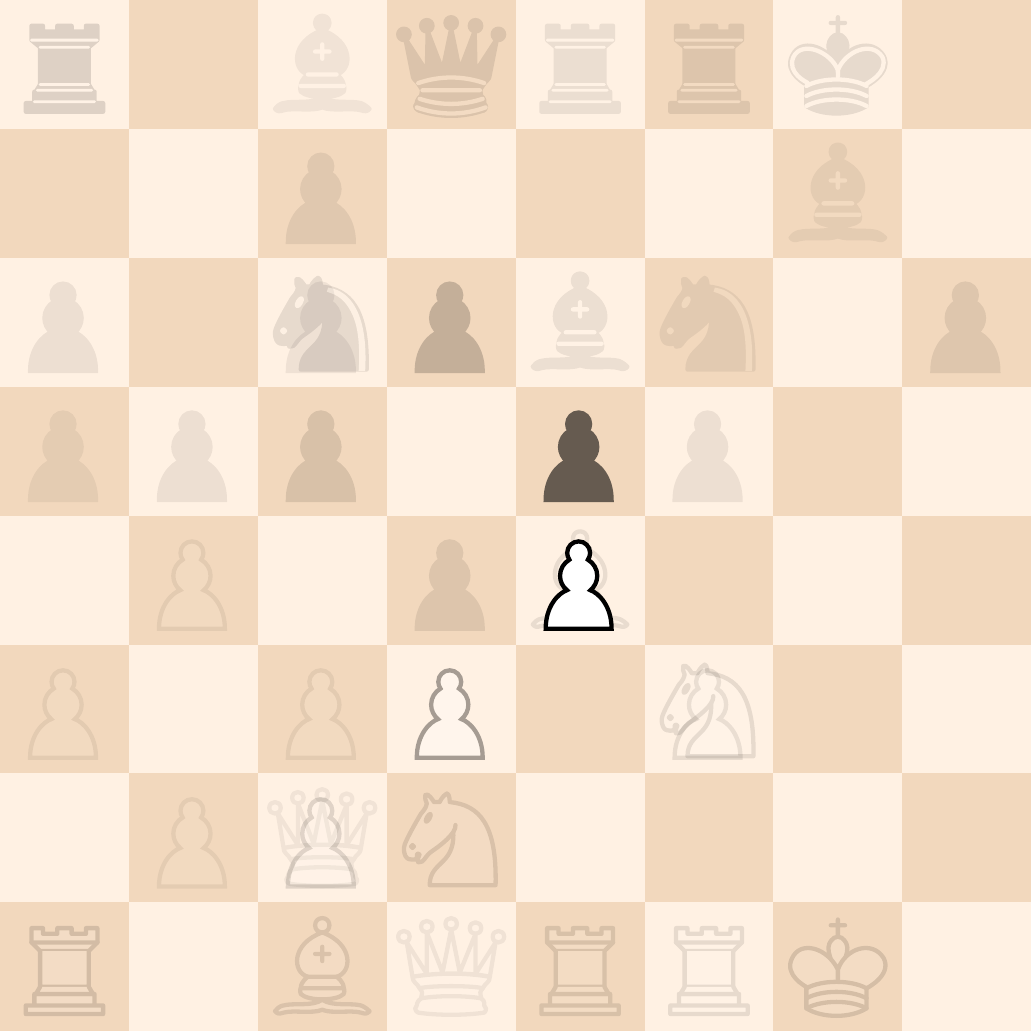}
\includegraphics[width=0.18\textwidth]{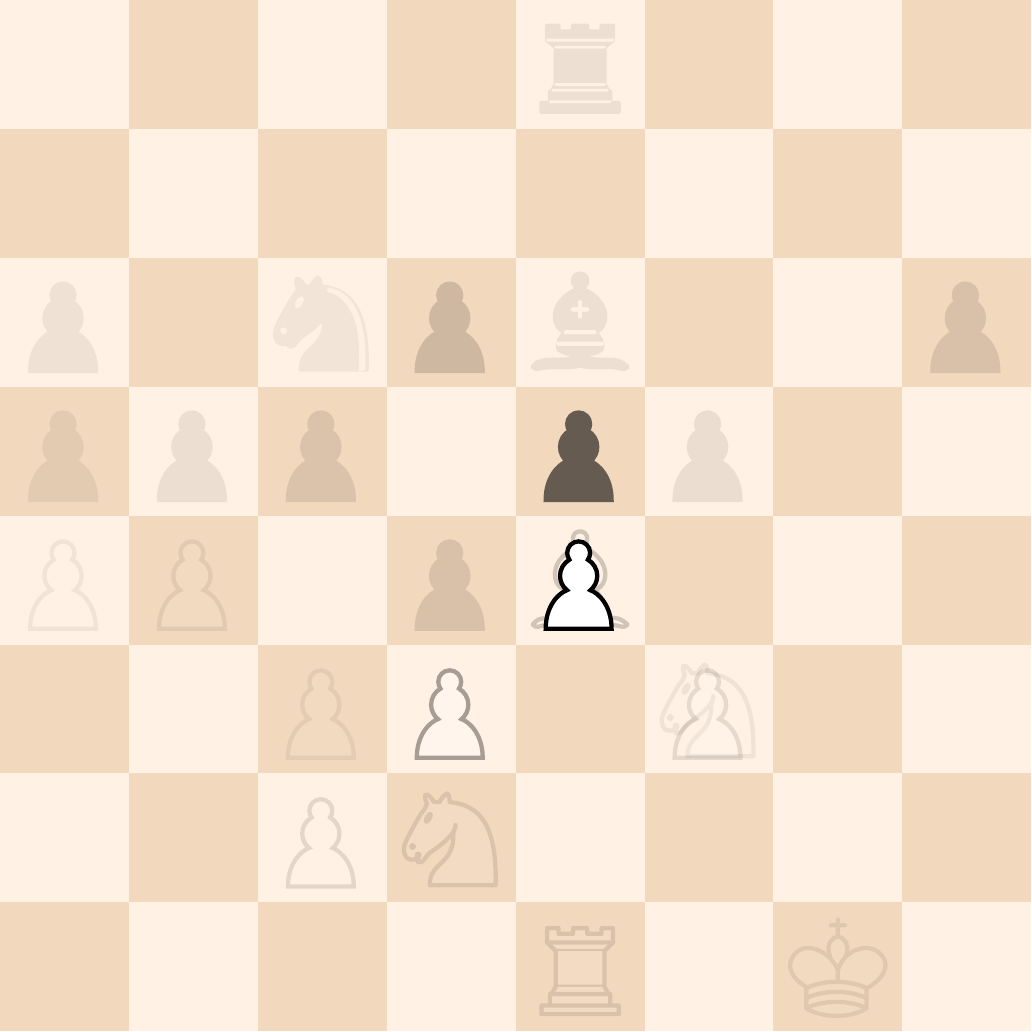}
\includegraphics[width=0.18\textwidth]{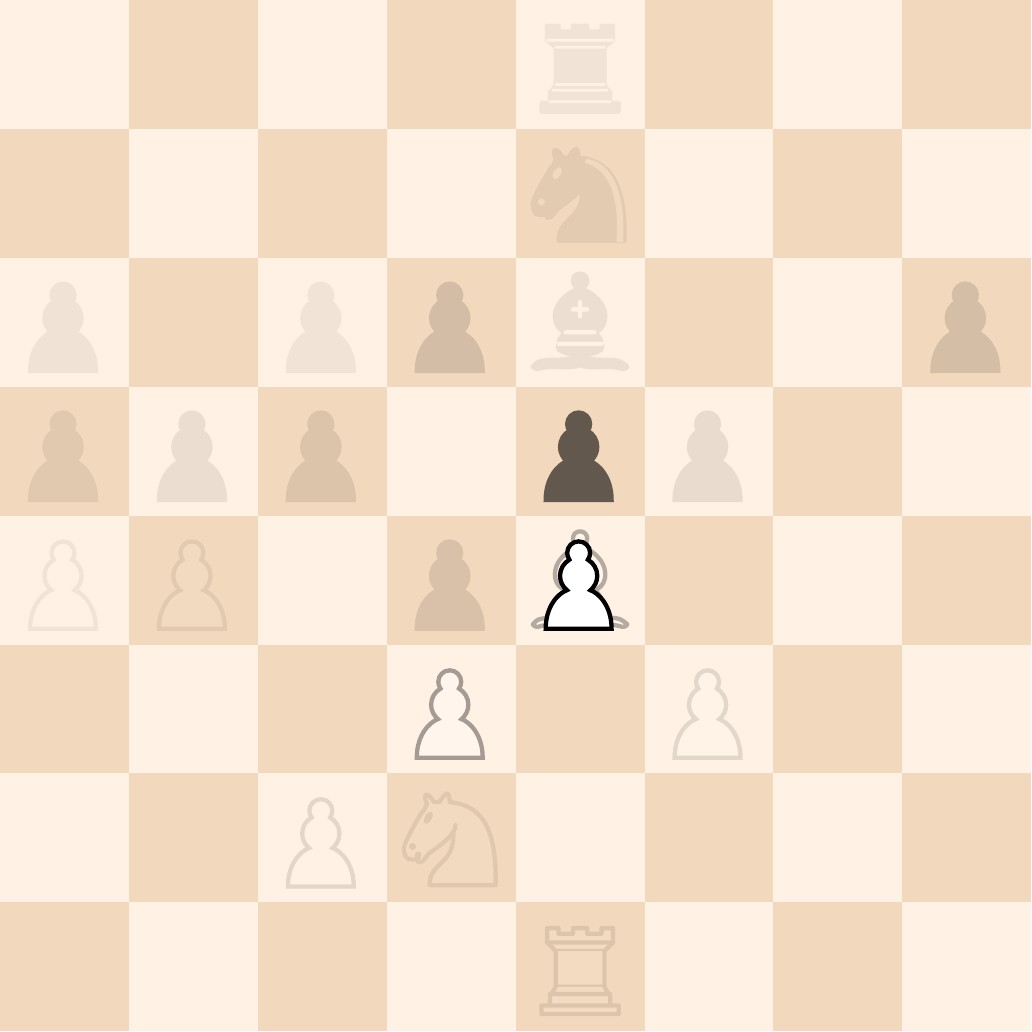}
\includegraphics[width=0.18\textwidth]{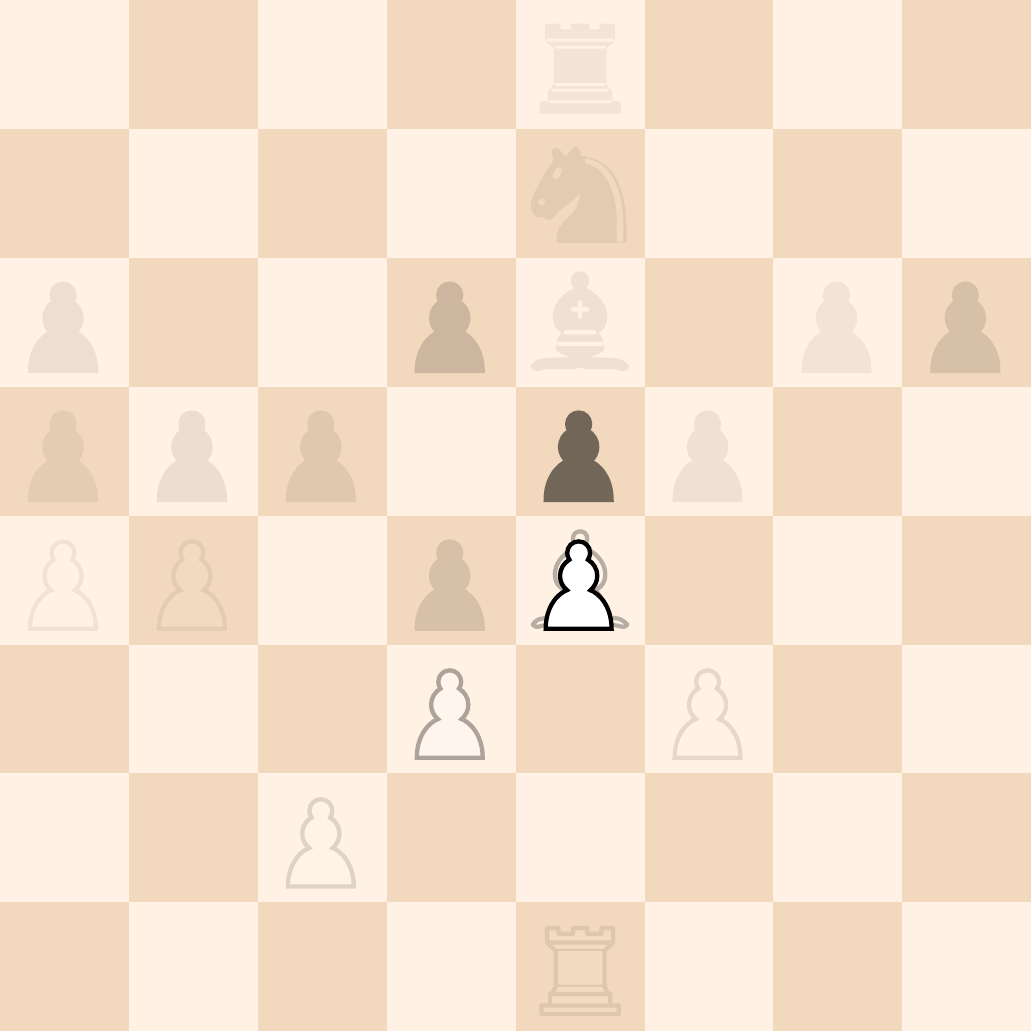} \\
activation-input covariance for activation $i = (5, 4, 10)$; layers 16 to 20
\caption{The activation-input covariance $\mathrm{cov}(z^l_{i} , \z^0)$ for
\textbf{layers 1 to 20} for activations $i = (5, 4, 10)$ in each layer $l$.
Unlike Figure \ref{fig:activation-input-covariance-column-e4-7}, the unit is persistently activated when there are opposing central pawn; the detection of two opposing central pawns are propagated through all layers of the network.
}
\label{fig:activation-input-covariance-column-e4-9}
\end{figure}

We illustrate how a feature detector $z^l_i$ is persistent from layer 1 to 20.
Activation $i=(5,4,10)$ in layer 1 in Figure \ref{fig:activation-input-covariance-layer-1} detects opposing central pawns, and we follow that activation through all layers of the network in 
Figure \ref{fig:activation-input-covariance-column-e4-9}. 
The activation-input covariance is largely similar, 
suggesting that
the feature is persisted and reused through layers of the network, being propagated up from the first layer to the last.

\subsubsection{Evolving feature detectors}
\label{sec:evolving-feature-detectors}

\begin{figure}
\centering
\includegraphics[width=0.18\textwidth]{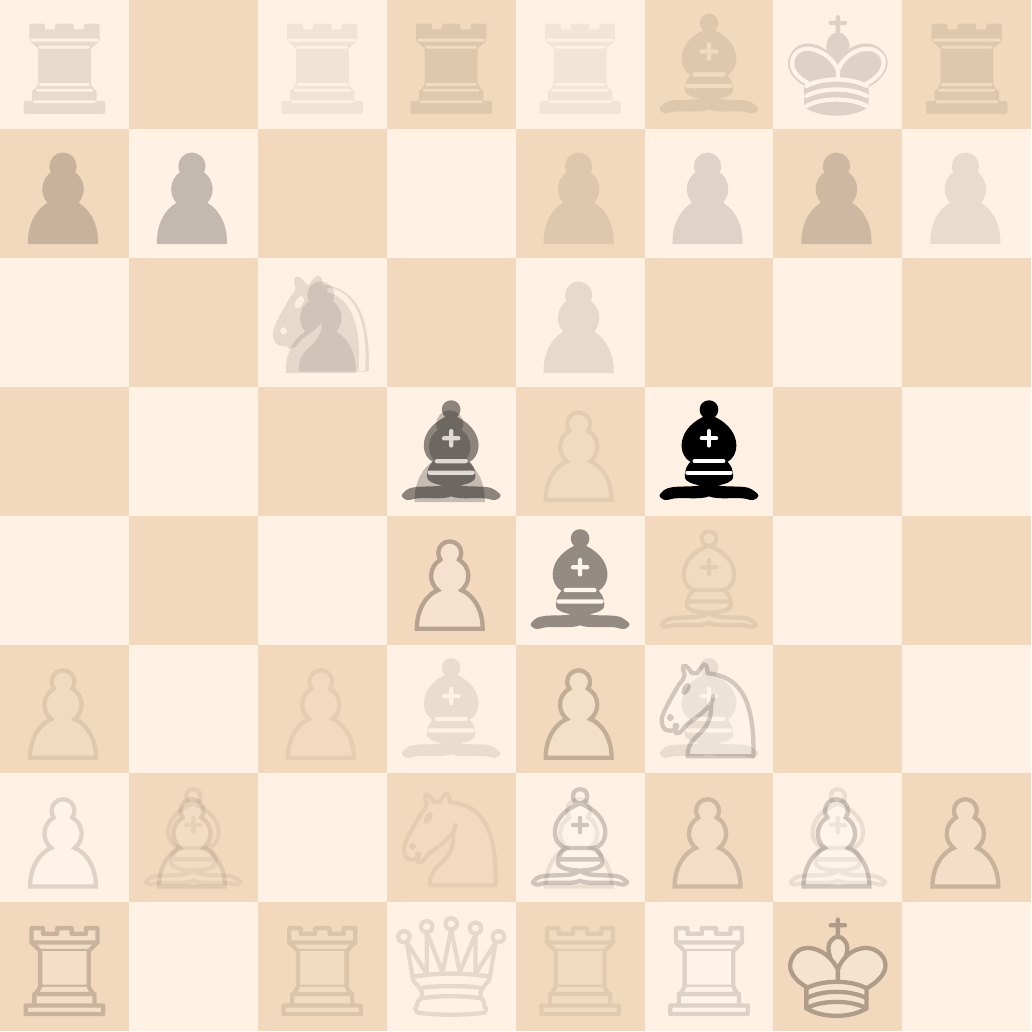}
\includegraphics[width=0.18\textwidth]{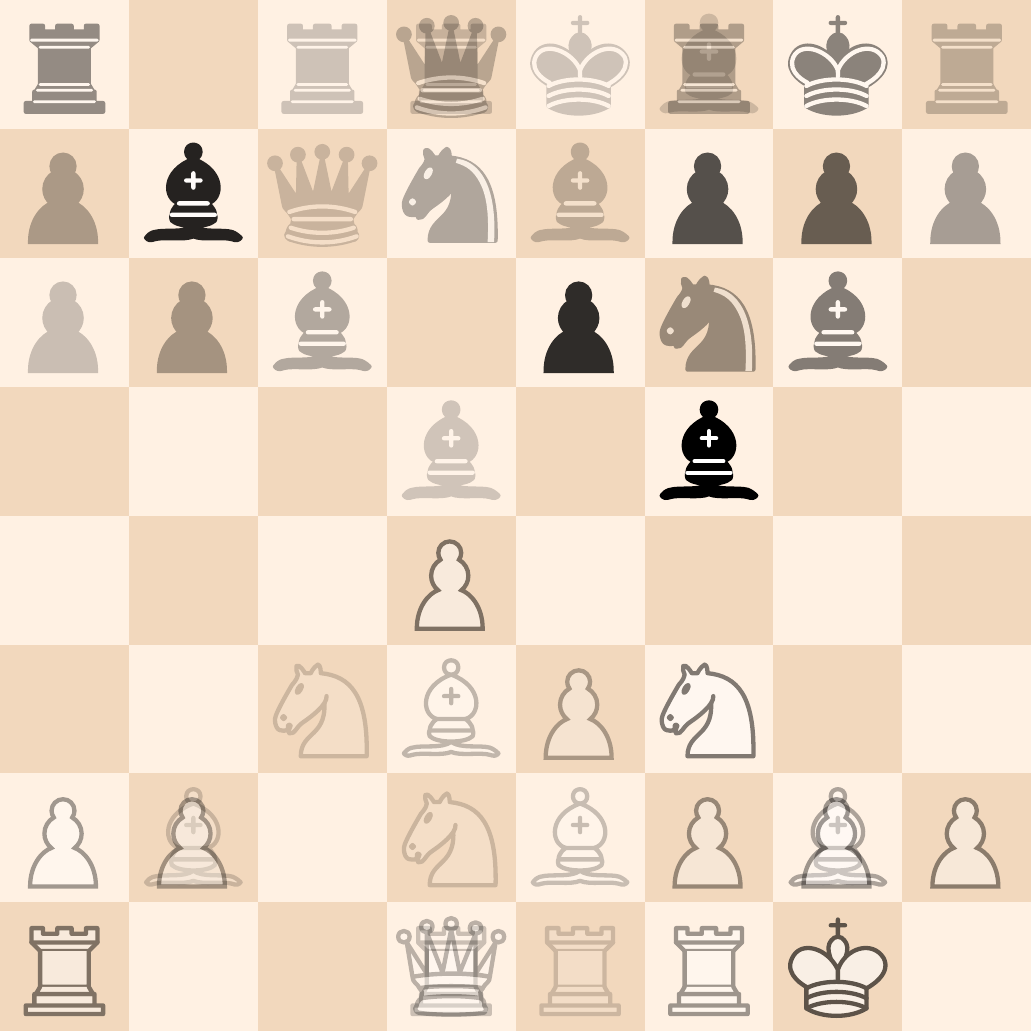}
\includegraphics[width=0.18\textwidth]{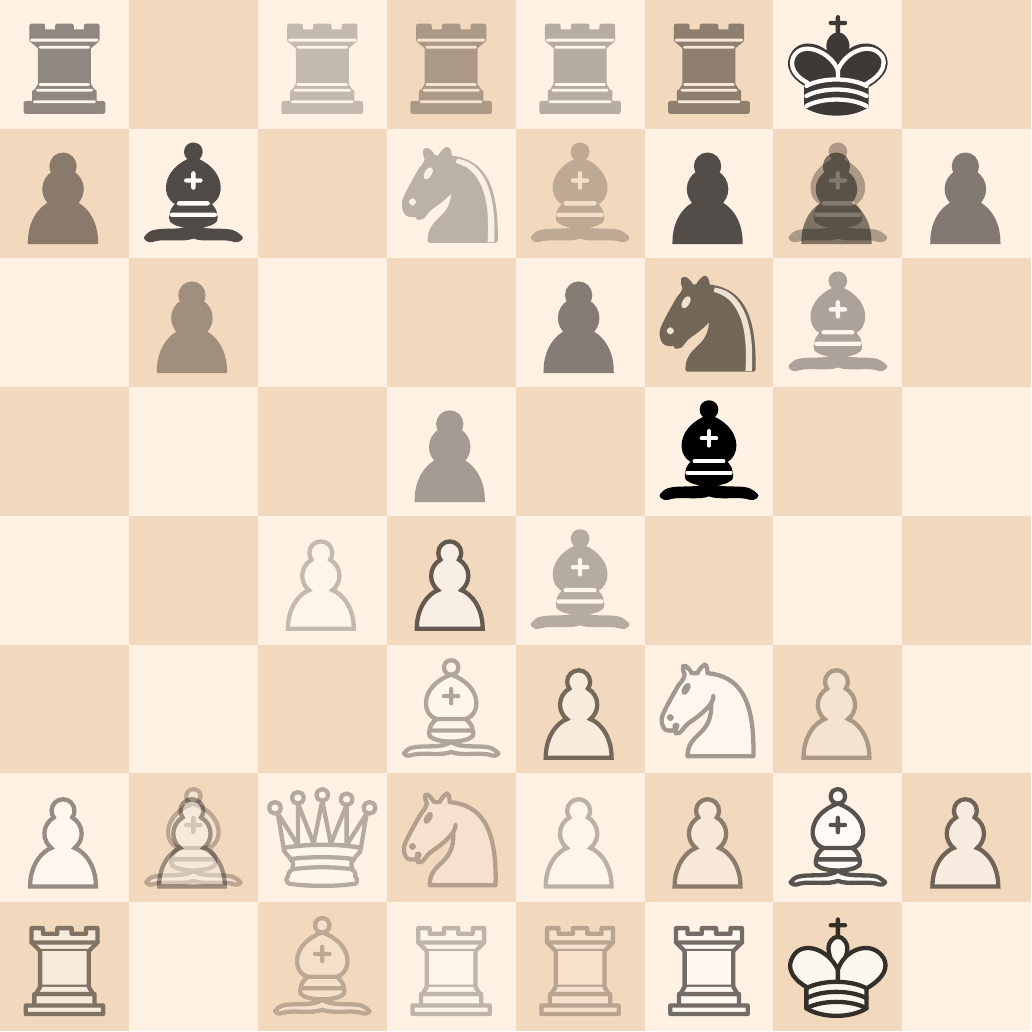}
\includegraphics[width=0.18\textwidth]{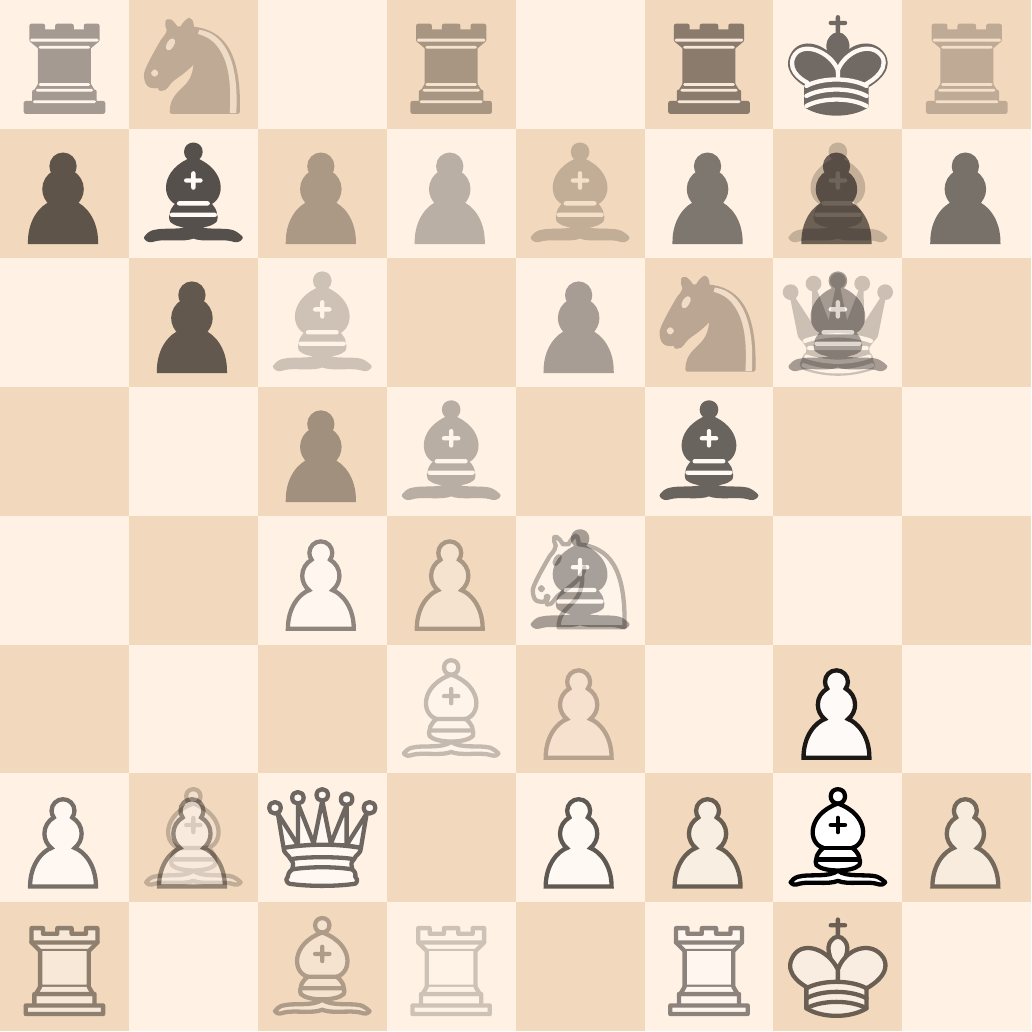}
\includegraphics[width=0.18\textwidth]{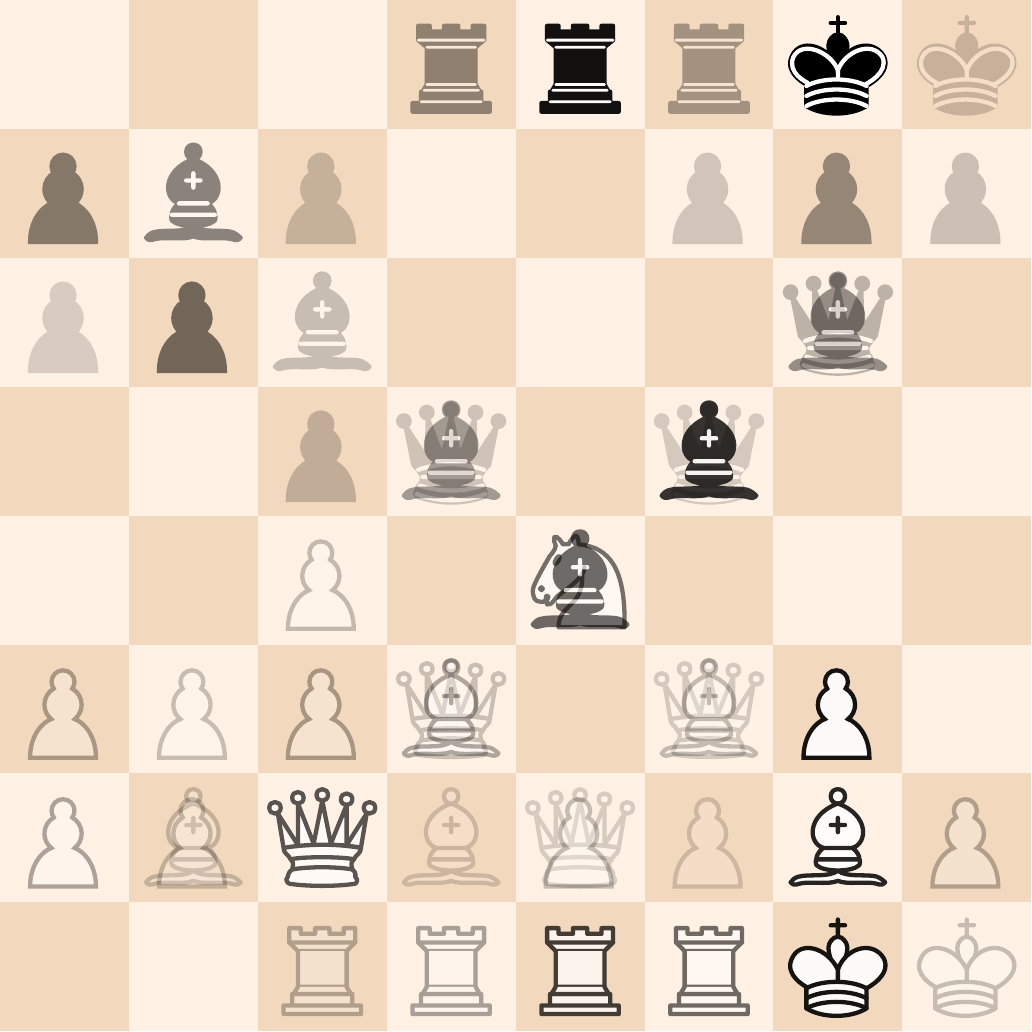} \\
activation-input covariance for activation $i = (5, 4, 8)$; layers 1 to 5 \\
\vspace{2pt}
\includegraphics[width=0.18\textwidth]{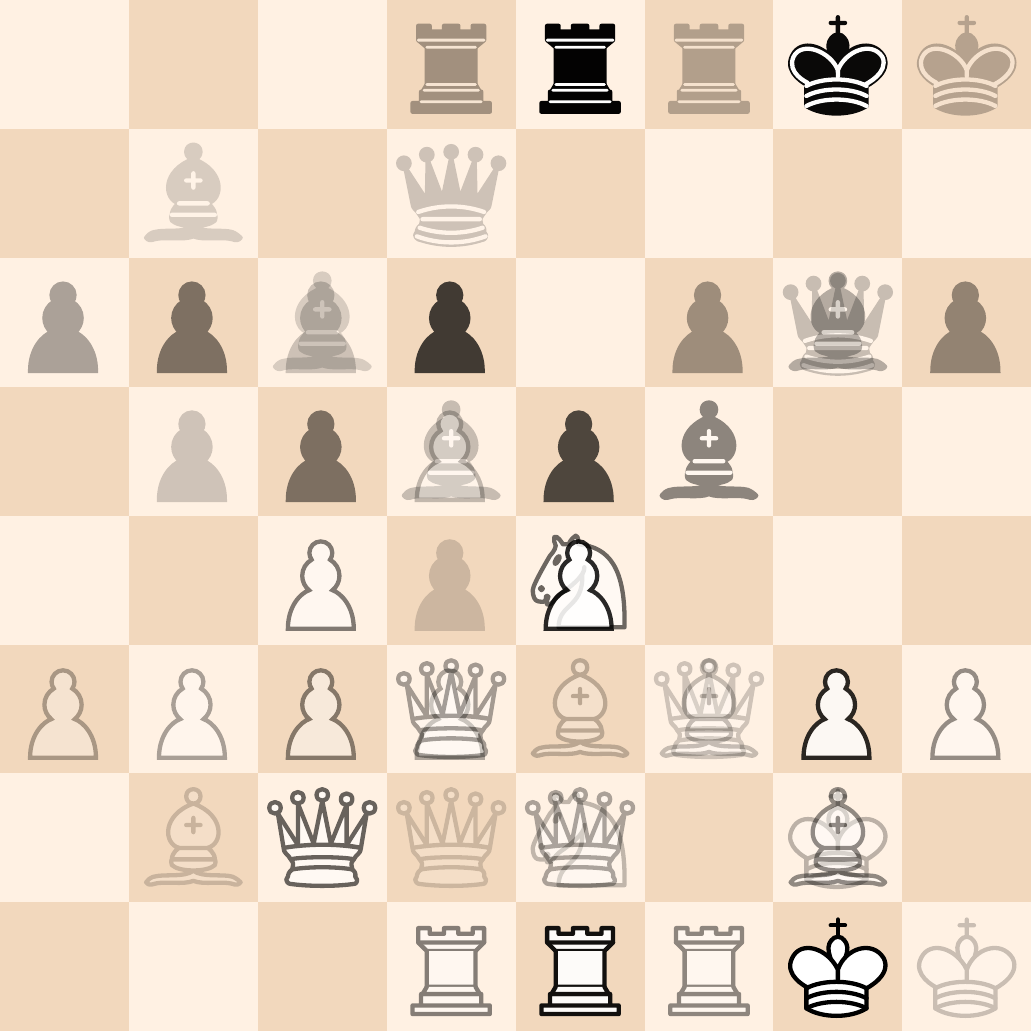}
\includegraphics[width=0.18\textwidth]{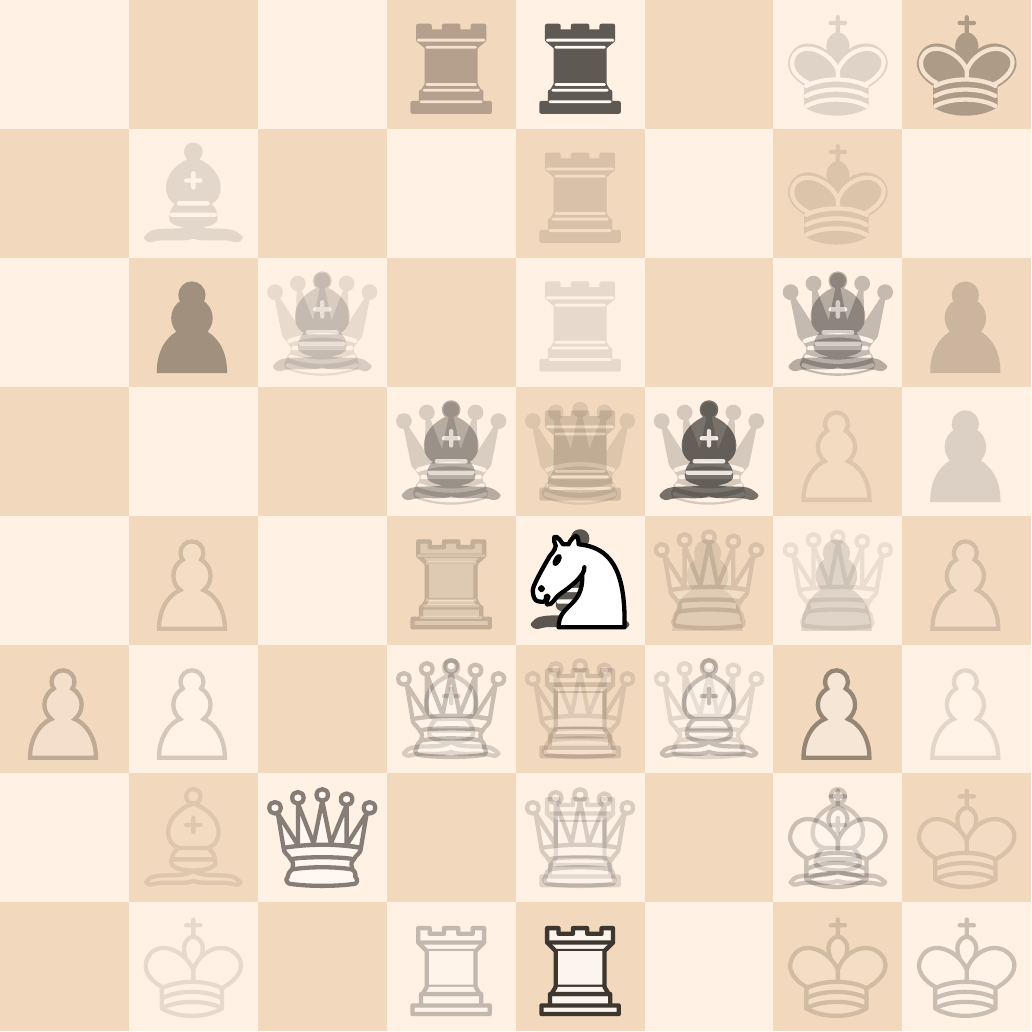}
\includegraphics[width=0.18\textwidth]{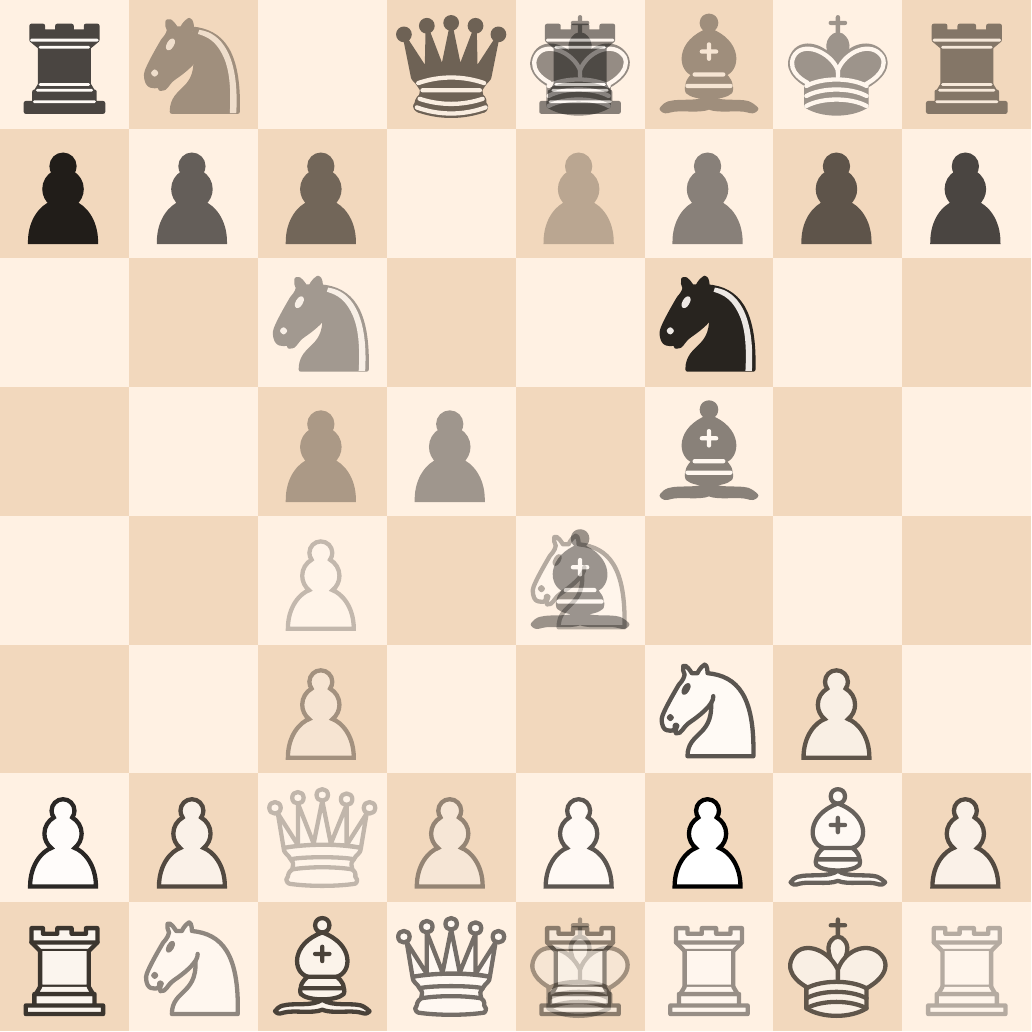}
\includegraphics[width=0.18\textwidth]{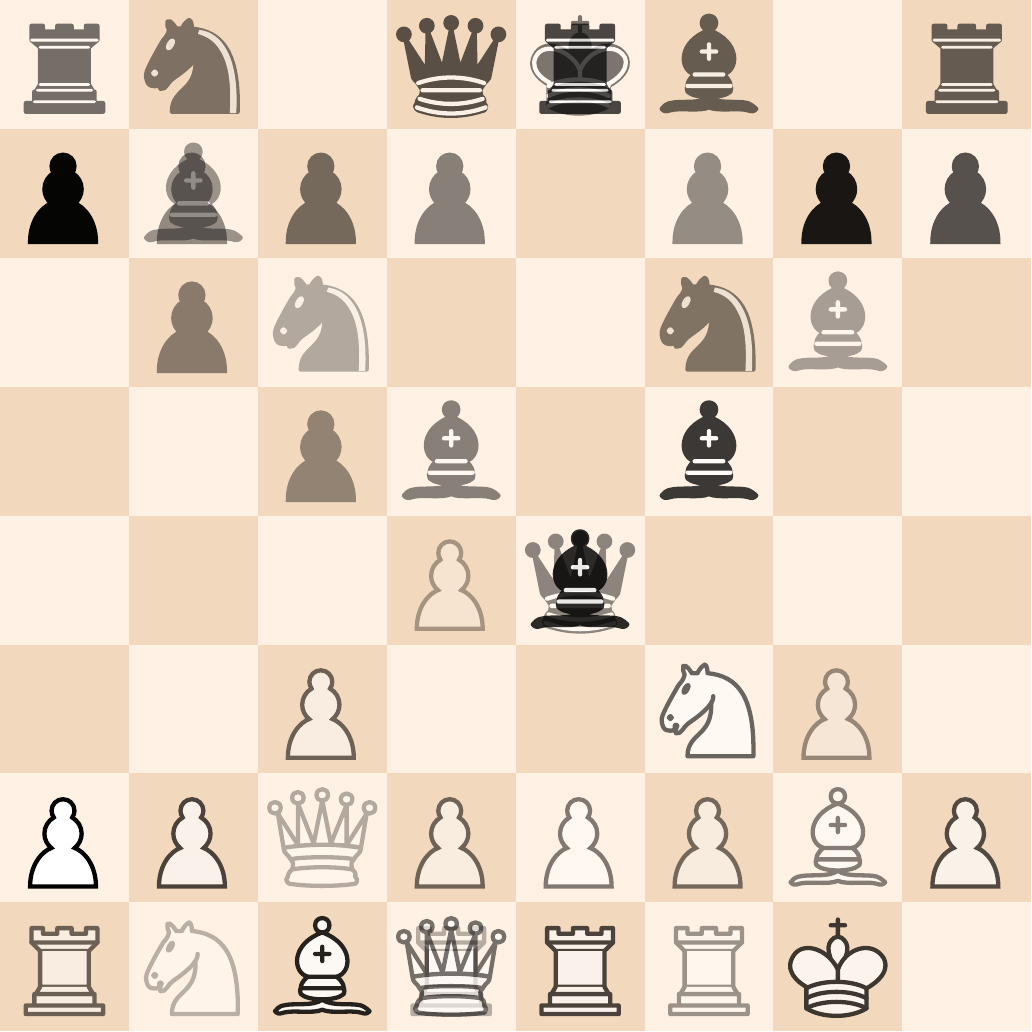}
\includegraphics[width=0.18\textwidth]{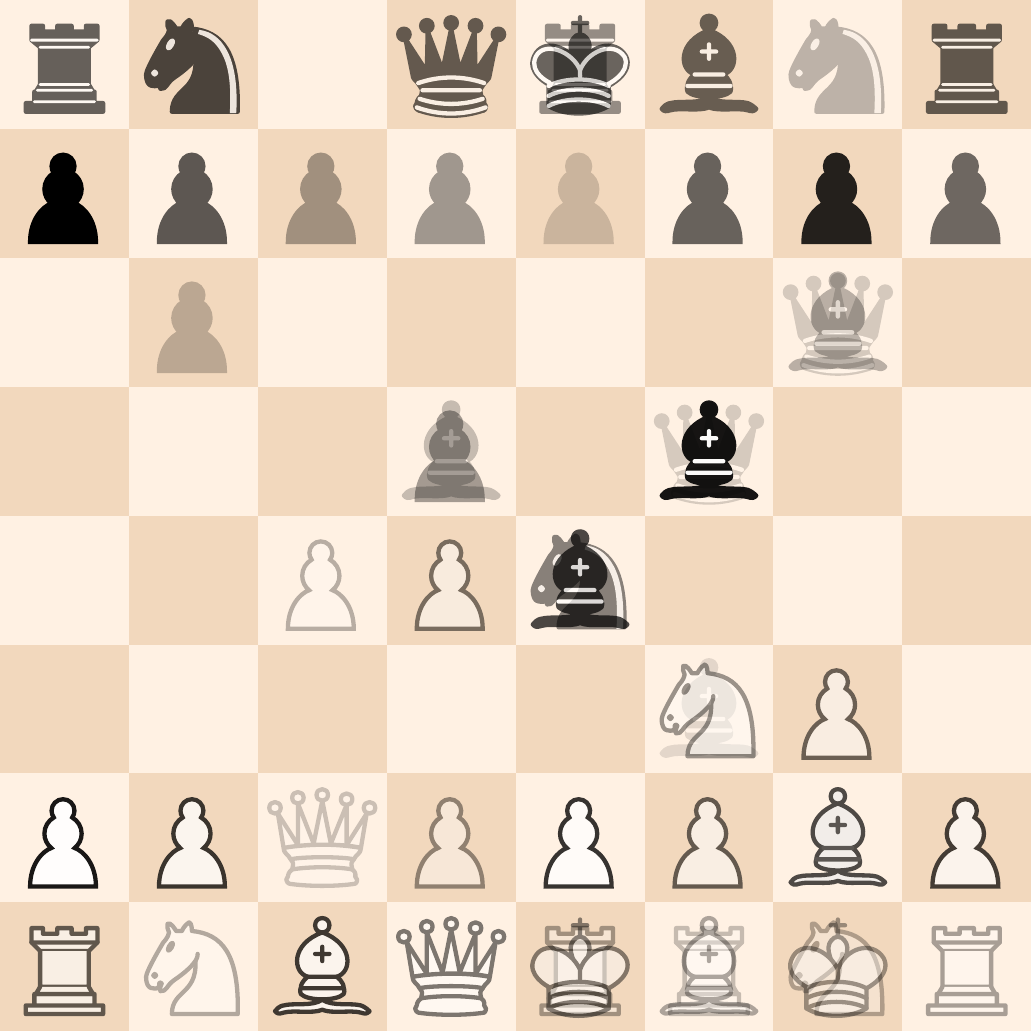} \\
activation-input covariance for activation $i = (5, 4, 8)$; layers 6 to 10 \\
\vspace{2pt}
\includegraphics[width=0.18\textwidth]{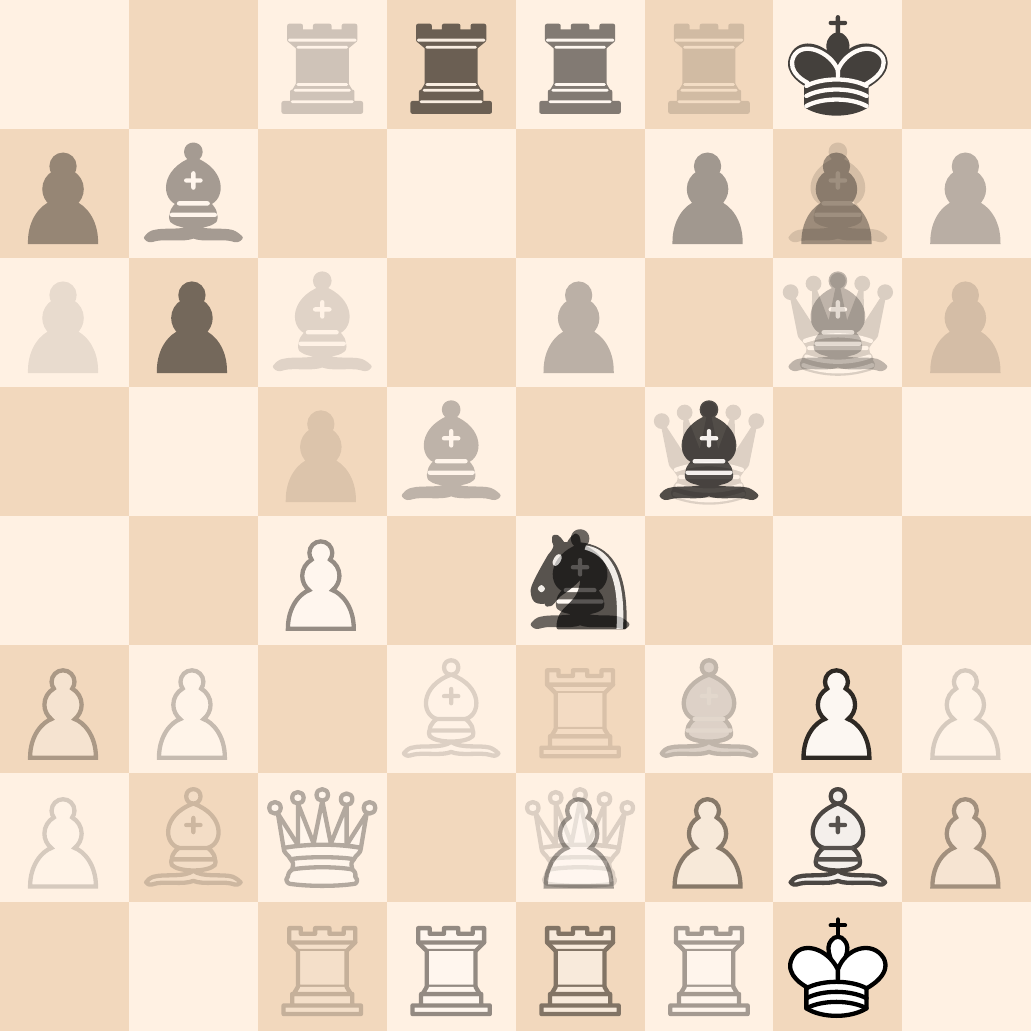}
\includegraphics[width=0.18\textwidth]{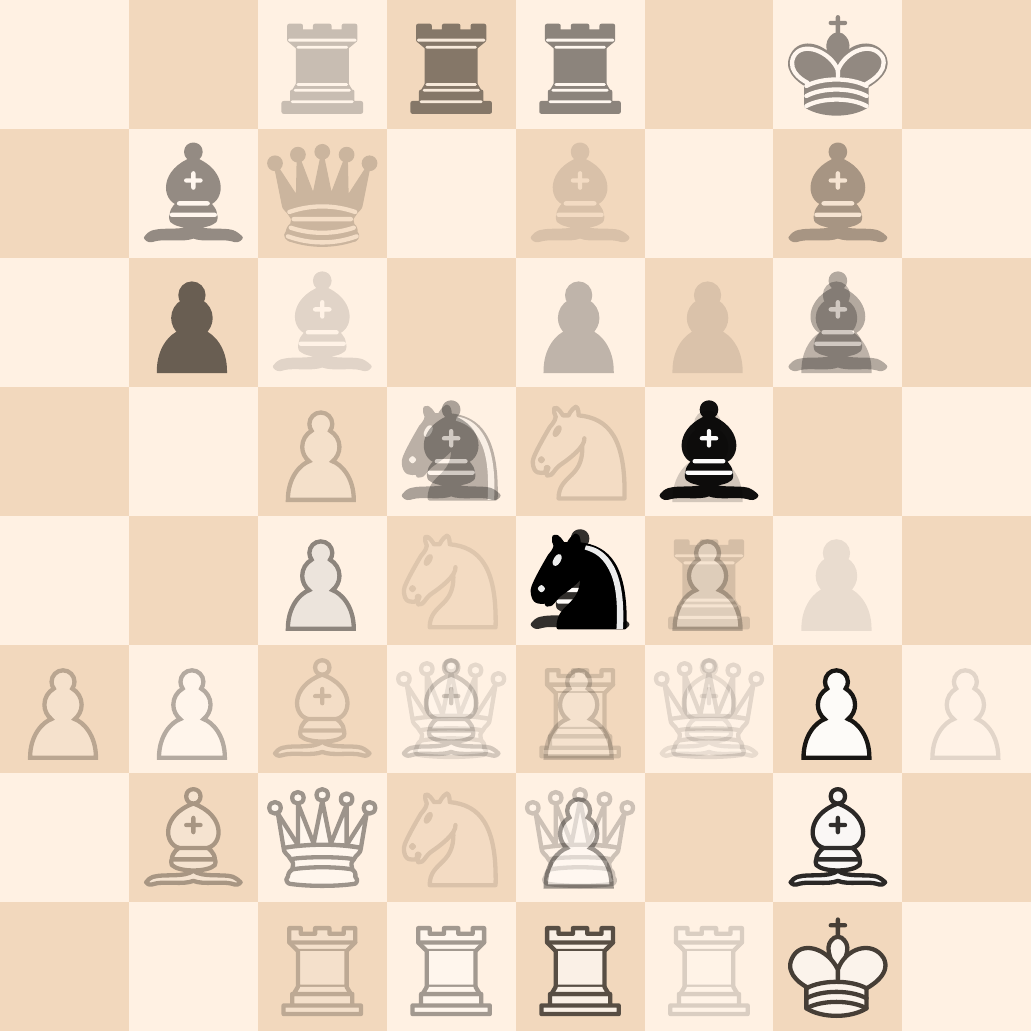}
\includegraphics[width=0.18\textwidth]{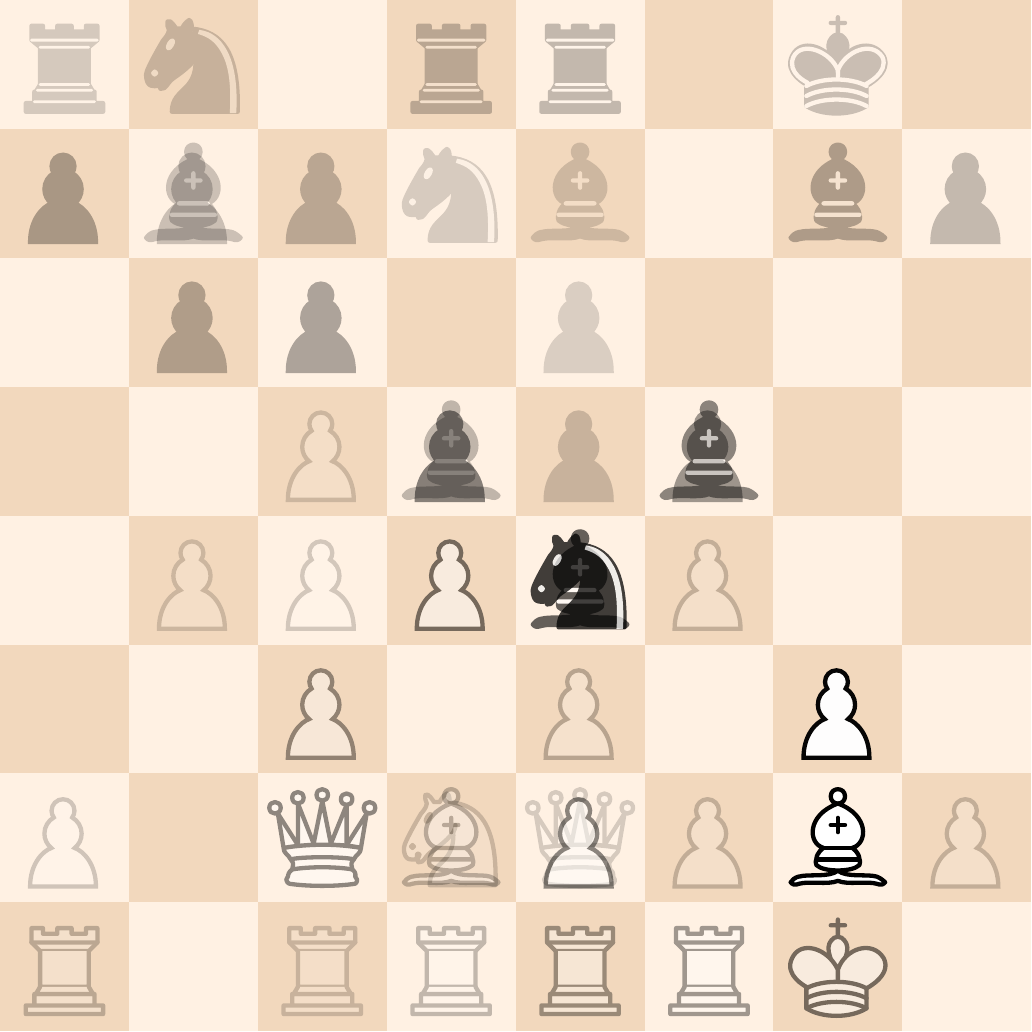}
\includegraphics[width=0.18\textwidth]{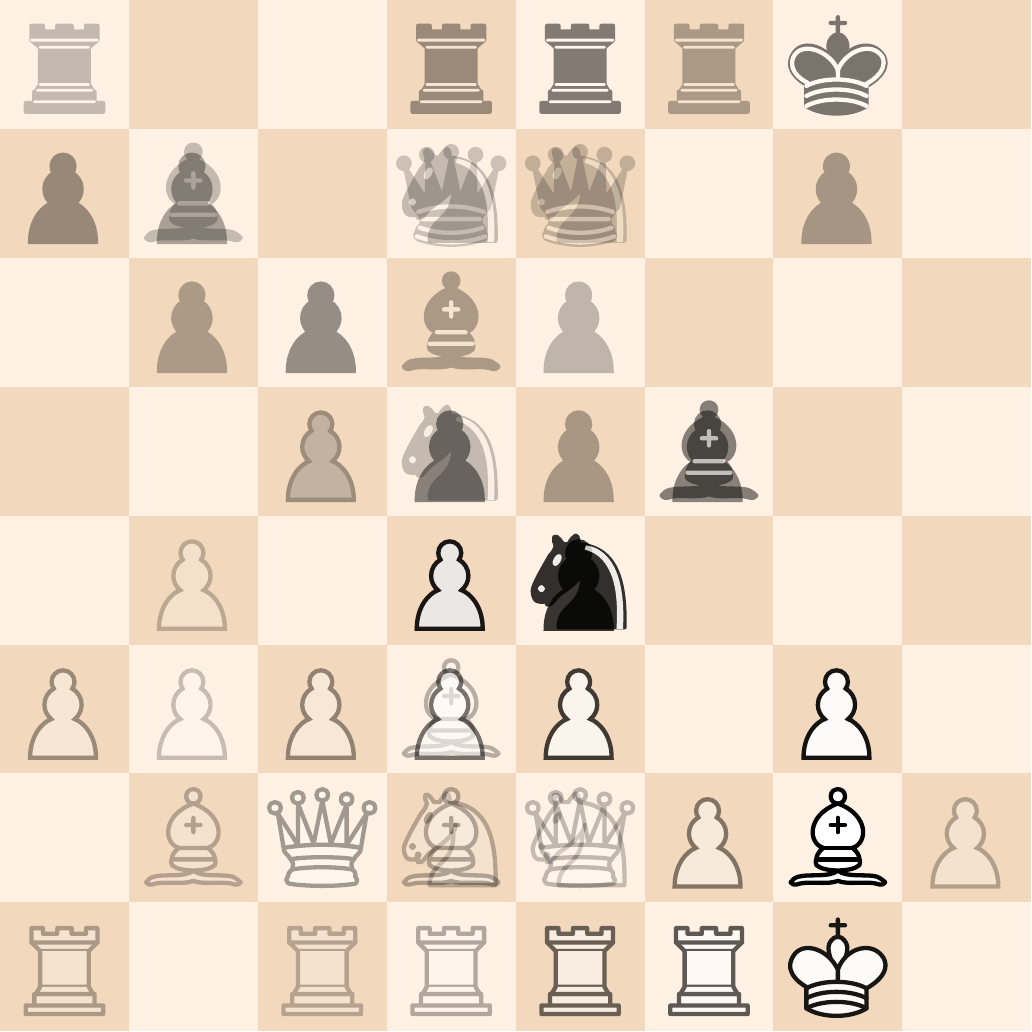}
\includegraphics[width=0.18\textwidth]{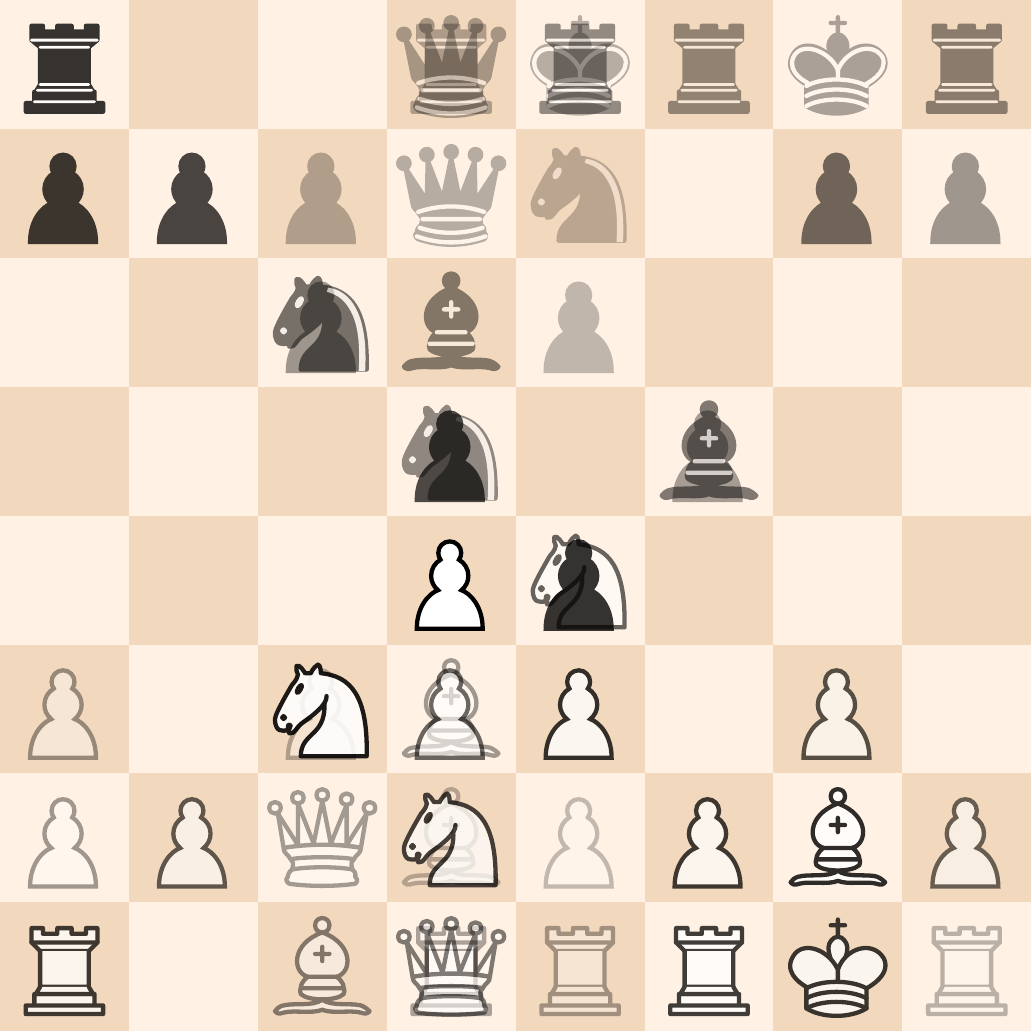} \\
activation-input covariance for activation $i = (5, 4, 8)$; layers 11 to 15 \\
\vspace{2pt}
\includegraphics[width=0.18\textwidth]{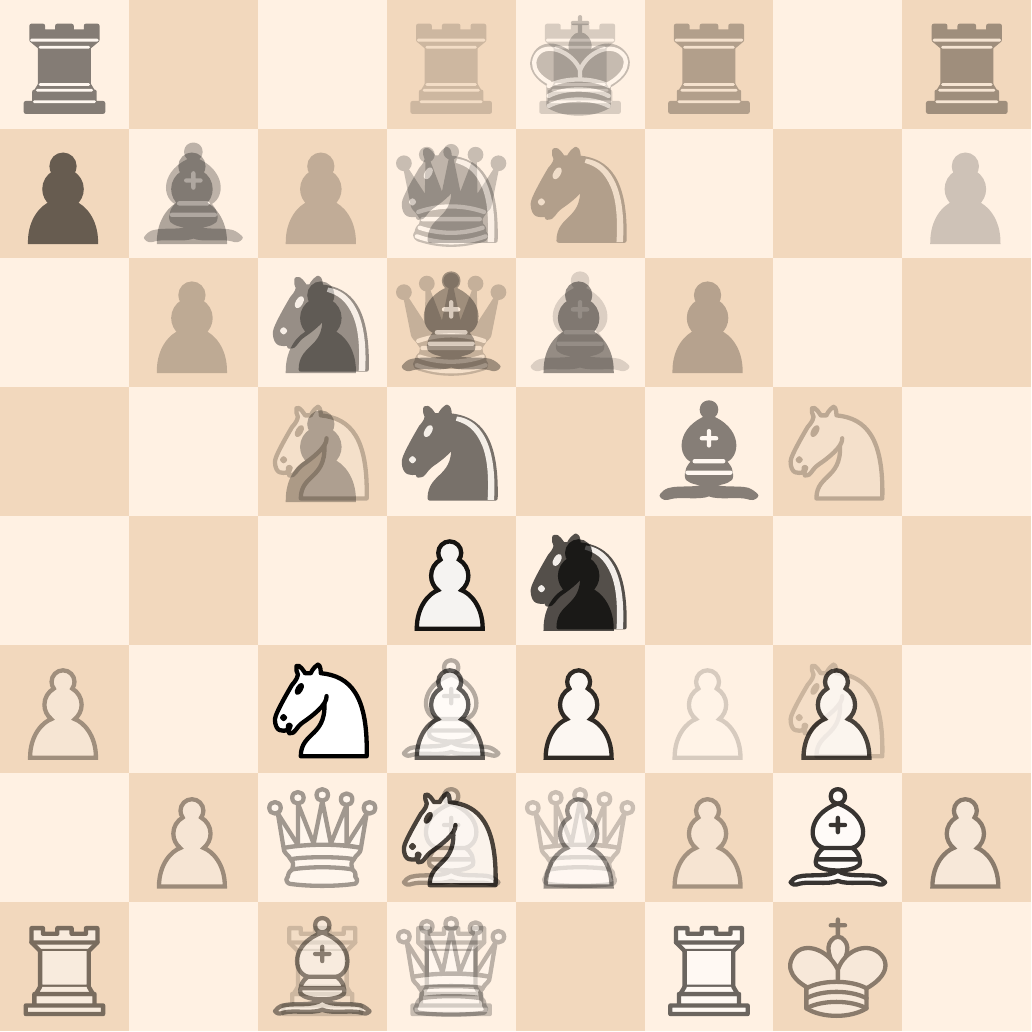}
\includegraphics[width=0.18\textwidth]{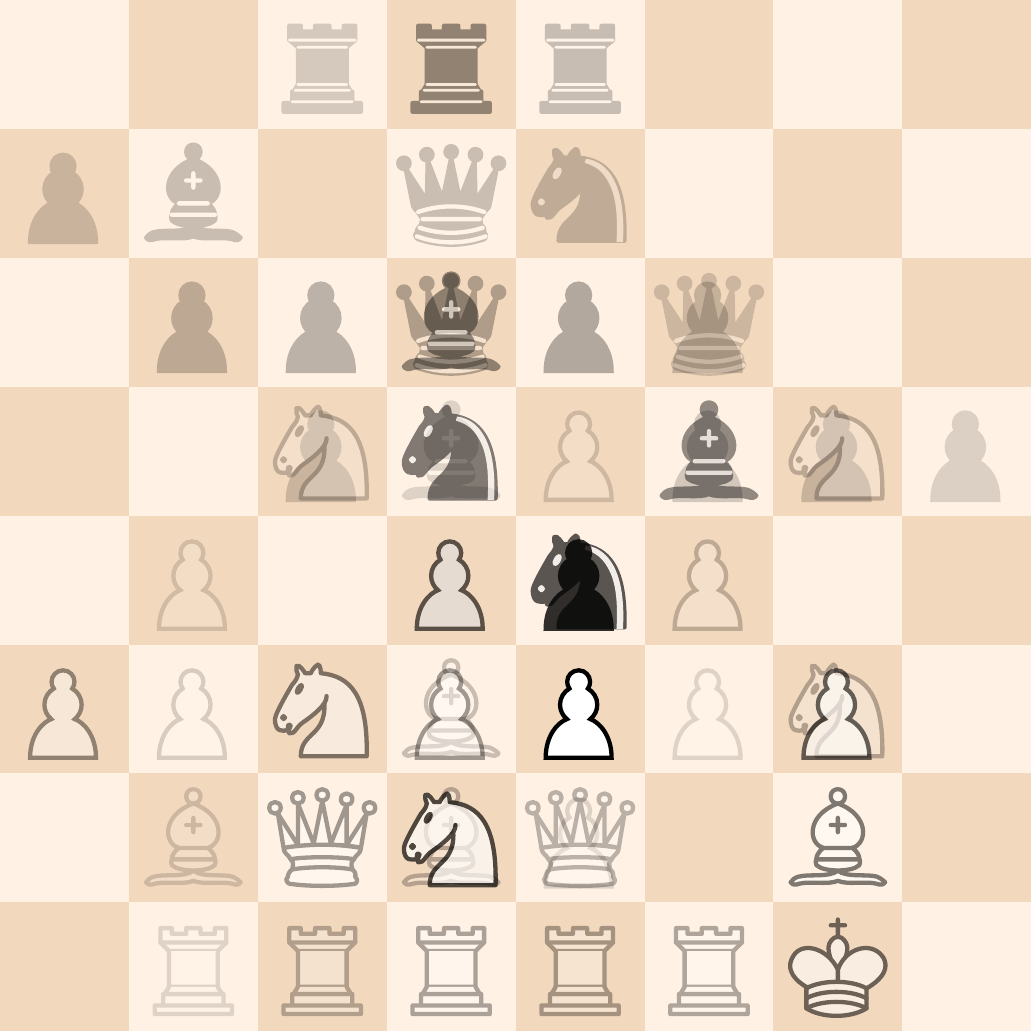}
\includegraphics[width=0.18\textwidth]{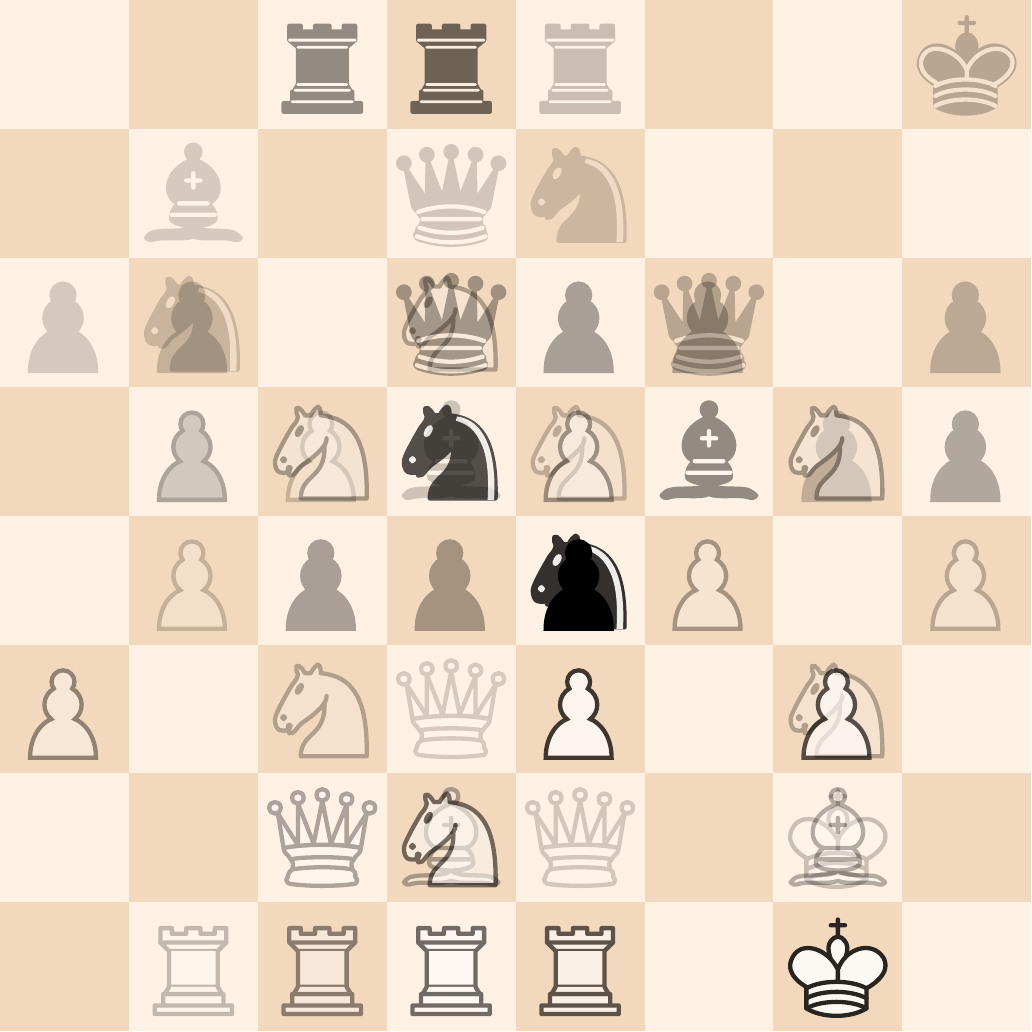}
\includegraphics[width=0.18\textwidth]{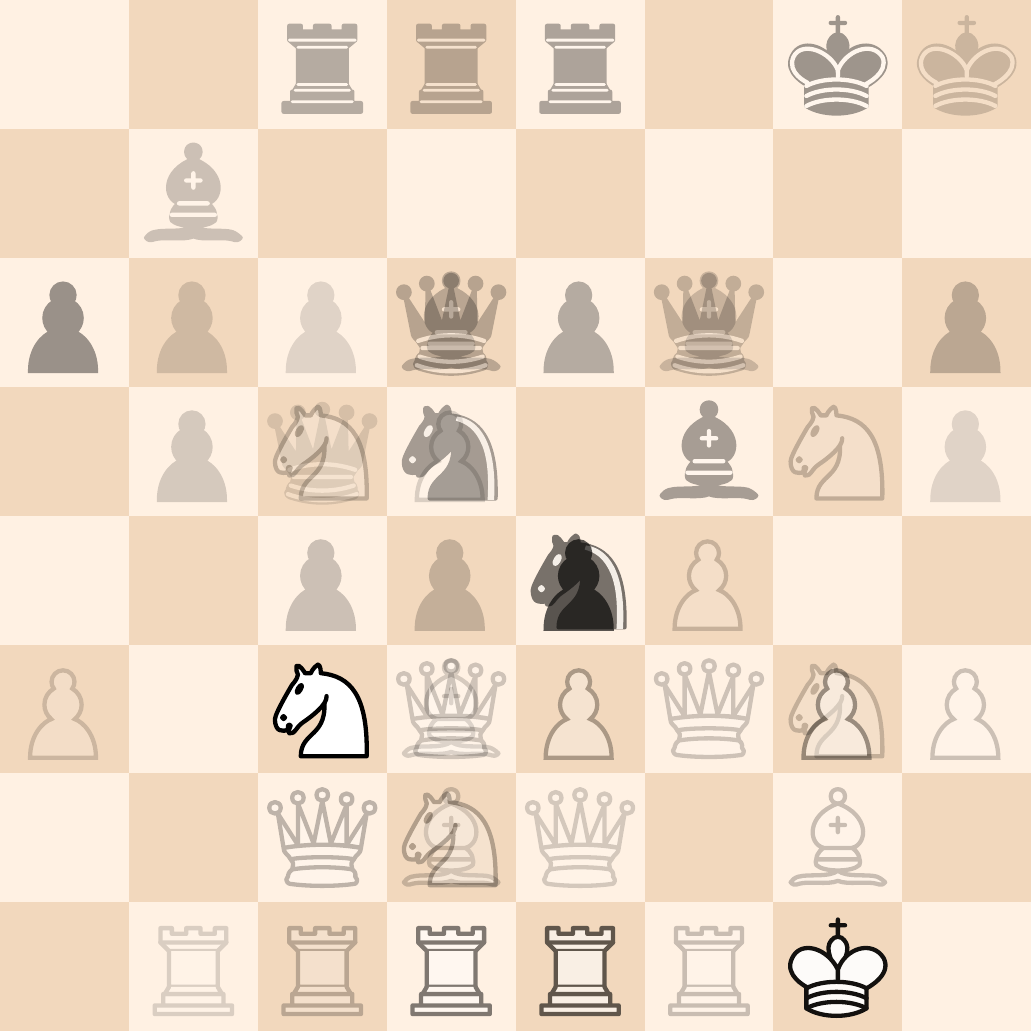}
\includegraphics[width=0.18\textwidth]{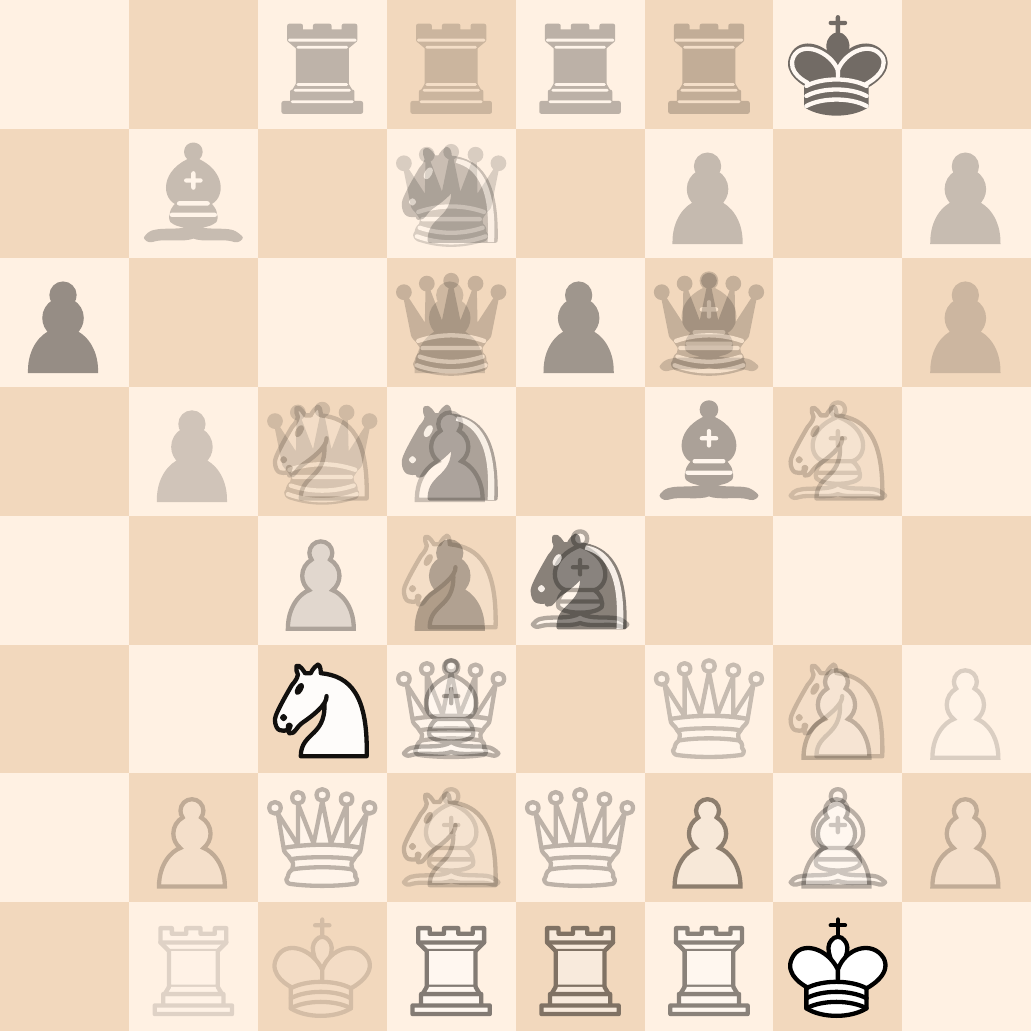} \\
activation-input covariance for activation $i = (5, 4, 8)$; layers 16 to 20
\caption{The activation-input covariance $\mathrm{cov}(z^l_{i} , \z^0)$ for
\textbf{layers 1 to 20} for activations $i = (5, 4, 8)$ in each layer $l$.
Moving up the network, with White to play, features revolve around a central light-squared bishop. These evolve to a central opposing pawn on e4, to finally a \symknight c3 (attacking e4).}
\label{fig:activation-input-covariance-column-e4-7}
\end{figure}

Some information is useful for the distributed representation of $\z^0$ in layer $l$ as $\z^l$, but possibly not at higher layers in the network architecture.
The activations in layer 1 detect localized patterns, but these could become increasingly global as activations are propagated through the network.
We consider activation $i = (5,4,8)$ from Figure \ref{fig:activation-input-covariance-layer-1},
which detects a central light-squared bishop for the opposing side.
Figure \ref{fig:activation-input-covariance-column-e4-7} shows how that activation
evolves through the layers of the trained AlphaZero network.
Moving up the network, with White to play, features revolve around a central light-squared bishop. These evolve to a central opposing pawn on e4, to finally a knight on c3 attacking e4.

\subsubsection{Open questions: distributed encoding}

On a micro scale, the activation-input covariances in this section show how
each of the individual activations $z_i^l$ in $\z^l$
is correlated with input positions.
On a macro scale, Section \ref{sec:encoding} showed how common human recognizable concepts could be regressed (linearly) from the distributed encoding. 
On a macro scale, the 16384 activations in $\z^l$ together form a distributed
encoding of many concepts.
An interesting open direction is the relation between the micro and the macro scale: for a given chess position,
decomposing a concept (approximated as a linear function of the activations; see Section \ref{sec:encoding})
into principal features in the position.

\FloatBarrier

\section{Conclusions}
In this work we studied the evolution of AlphaZero's representations and play through a combination of concept probing, behavioural analysis, and examination of AlphaZero's activations. Examining the evolution of human concepts using probing showed that many human concepts can be accurately regressed from the AZ network after training, even though AlphaZero has never seen a human game of chess, and there is no objective function promoting human-like play or activations. Following on from our concept probing we reflected on the challenges associated with this methodology and suggested future directions of research. Our concept regression findings are supported by finding multiple human-interpretable factors in AlphaZero's activations. We have released these online and are excited to see what other researchers can find in this data. We have also analysed the evolution of AlphaZero's play style through a study of opening theory and an expert qualitative analysis of AlphaZero's self-play during training. Our analysis suggests that opening knowledge undergoes a period of rapid development around the same time that many human concepts become predictable from network activations, suggesting a critical period of rapid knowledge acquisition. The fact that human concepts can be located even in a superhuman system trained by self-play broadens the range of systems in which we should expect to find human-understandable concepts. We believe that the ability to find human-understandable concepts in the AZ network indicates that a closer examination will reveal more. The next question is: can we go beyond finding human knowledge and learn something new?

\section*{Acknowledgements}

We would like to thank ChessBase for providing us with an extensive database of historical human games 
\cite{chessbase}.
This was used in our experiments to evaluate the explainability of AlphaZero, as well as contrast it to the historical developments in human chess theory.
We are grateful to Frederic Friedel at ChessBase for his enduring support, and to Shane Legg, Neil Rabinowitz, Aleksandra Faust, Lucas Gill Dixon, Martin Wattenberg, and Nithum Thain for insightful comments and discussions.

\bibliography{references}

\appendix

\newpage

\section{Full concept list}
\label{appendix:concepts}

Table \ref{tab:concepts} summarizes 93 concepts from Stockfish 8's public API. The concepts are taken from
Stockfish 8 and not a later version, as there is a wealth of observations on the difference between AlphaZero and Stockfish 8 \cite{game_changer}.
We implemented a long list of additional concepts programatically,
and these are presented in Tables
\ref{tab:concepts-custom} and
\ref{tab:concepts-custom-pawns}.

\begin{table}[]
\centering
\begin{tabular}{l|l}
\textbf{Concept names} & \textbf{Description} \\
\hline
\verb=material [t] [mg|eg|ph]= & Material score, where each piece on the board has a predefined value \\
& that changes depending on the phase of the game.\\
\hline
\verb=imbalance [t]= & Imbalance score that compares the piece count of each piece type for both \\
\verb=          [mg|eg|ph]= &  colours. E.g., it awards having a pair of bishops vs a bishop and a knight. \\
\hline
\texttt{pawns [t] [mg|eg|ph]} & Evaluation of the pawn structure. E.g., the evaluation considers isolated, \\
& double, connected, backward, blocked, weak, etc. pawns.\\
\hline
\verb=knights [t]= & Evaluation of knights. E.g., extra points are given to knights \\
\verb=        [mg|eg|ph]= & that occupy outposts protected by pawns.\\
\hline
\verb=bishops [t]= & Evaluation of bishops. E.g., bishops that occupy the same color squares\\
\verb=        [mg|eg|ph]= & as pawns are penalised.\\
\hline
\verb=rooks [t]= & Evaluation of rooks. E.g., rooks that occupy open or semi-open files\\
\verb=      [mg|eg|ph]= & have higher valuation.\\
\hline
\verb=queens [t]= & Evaluation of queens. E.g., queens that have relative pin or discovered\\
\verb=       [mg|eg|ph]= & attack against them are penalized.\\
\hline
\verb=mobility [t]= & Evaluation of piece mobility score. It depends on \\
\verb=         [mg|eg|ph]= & the number of squares attacked by the pieces. \\
\hline
\verb=king_safety [t]= & A complex concept related to king safety. It depends on the number \\
\verb=            [mg|eg|ph]= & and type of pieces that attack squares around the king, shelter strength, \\
& number of pawns around the king, penalties for being on pawnless flank, etc.\\
\hline
\verb=threats [t]= & Evaluation of threats to pieces, such as whether a pawn can safely advance \\
\verb=        [mg|eg|ph]= & and attack an opponent's higher value piece, hanging pieces,\\
& possible xray attacks by rooks, etc. \\
\hline
\verb=passed_pawns [t]= & Evaluates bonuses for passed pawns. The closer a pawn is to the promotion \\
\verb=             [mg|eg|ph]= & rank, the higher is the bonus.\\
\hline
\verb=space [t]= & Evaluation of the space. It depends on the number of safe squares available \\
\verb=      [mg|eg|ph]= & for minor pieces on the central four files on ranks 2 to 4.\\
\hline
\verb=total [t] [mg|eg|ph]= & The total evaluation of a given position. It encapsulates all the above concepts. \\
\hline
\end{tabular}
\caption{A summary of 93 concepts from Stockfish 8's public API.
The concepts are enumerated as
\texttt{<concept\_name>\_<side>\_<game\_phase>},
where the side is \texttt{[w|b|t]} for White, Black, total (difference) respectively.
The game phase abbreviations \texttt{[mg|eg|ph]} stands for middle game, end game and phased value respectively. The phased value is a weighted sum of the middle and end game values based on the actual phase of a given position.
As AlphaZero represents positions from the playing side's view,
``side'' is orientated and represented as \texttt{mine} and \texttt{opponent} instead of \texttt{w} and \texttt{b}.
}
\label{tab:concepts}
\end{table}

\begin{table}[]
\centering
\begin{tabular}{l|l}
\textbf{Concept names} & \textbf{Description} \\
\hline
\verb=pawn_fork [m|o]=
& True if a pawn is attacking two pieces of higher value (knight, bishop, \\
& rook, queen, or king) and is not pinned. \\
\hline
\texttt{knight\_fork [m|o]}
& True if a knight is attacking two pieces of higher value (rook, queen, \\ & or king) and is not pinned. \\
\hline
\texttt{bishop\_fork [m|o]}
& True if a bishop is attacking two pieces of higher value (rook, queen, \\ & or king) and is not pinned. \\
\hline
\texttt{rook\_fork [m|o]}
& True if a rook is attacking two pieces of higher value (queen, or king) \\
& and is not pinned. \\
\hline
\hline
\texttt{has\_pinned\_pawn [m|o]}
& True if the side has a pawn that is pinned to the king of the same colour. \\
\hline
\texttt{has\_pinned\_knight [m|o]}
& True if the side has a knight that is pinned to the king of the same colour. \\
\hline
\texttt{has\_pinned\_bishop [m|o]}
& True if the side has a bishop that is pinned to the king of the same colour. \\
\hline
\texttt{has\_pinned\_rook [m|o]}
& True if the side has a rook that is pinned to the king of the same colour. \\
\hline
\texttt{has\_pinned\_queen [m|o]}
& True if the side has a queen that is pinned to the king of the same colour. \\
\hline
\hline
\texttt{material [m|o|diff]} &
Material calculated as (\verb|#|\sympawn) + 3 * (\verb|#|\symknight) + 3 * (\verb|#|\symbishop) + 5 * (\verb|#|\symrook) + 9 * (\verb|#|\symqueen) \\
\hline
\texttt{num\_pieces [m|o|diff]} 
& Number of pieces that a side has. \\
\hline
\texttt{in\_check} &
True if the side that makes a turn is in check. \\
\hline
\texttt{has\_bishop\_pair [m|o]}
& True if the side has a pair of bishops. \\
\hline
\texttt{has\_connected\_rooks [m|o]}
& True if the side has connected rooks. \\
\hline
\texttt{has\_control\_of\_open\_file [m|o]}
& True if the side controls an open file (with the rooks, queen) \\
\hline
\texttt{has\_mate\_threat}
& True if the opponent could mate the current side in a single move\\
& if the turn was passed to the opponent.\\
\hline
\texttt{has\_check\_move [m|o]}
& True if the side can check the opponent's King.\\
\hline
\texttt{can\_capture\_queen [m|o]}
& True if the side can capture the opponent’s queen.\\
\hline
\verb=num_king_attacked_squares=
& The number of squares around the opponent's king that the playing \\
\verb=[m|o|diff]= & side attacks. Can include occupied squares.\\
\hline
\texttt{has\_contested\_open\_file}
& True if an open file is occupied simultaneously by a rook and/or queen\\
& of both colours.\\
\hline
\verb=has_right_bc_ha_promotion=
& True if 1) the side has a passed pawn on a or h files and 2) the side has\\
\verb=[m|o]= &  a bishop that is of the colour of the promotion square of that pawn.\\
\hline
\hline
\verb=num_scb_pawns_same_side=
& The number of own pawns that occupy squares of the same colour as the\\
\verb=[m|o|diff]= & colour of own bishop.
Applicable only when the side has a single bishop. \\
\hline
\verb=num_ocb_pawns_same_side=
& The number of own pawns that occupy squares of the opposite colour to \\
\verb=[m|o|diff]= & that of own bishop.
Applicable only when the side has a single bishop. \\
\hline
\verb=num_scb_pawns_other_side=
& The number of opponent's pawns that occupy the squares of the same  \\
\verb=[m|o|diff]= & colour as the colour of own bishop.
Applicable only when the side \\
& has a single bishop.\\
\hline
\verb=num_ocb_pawns_other_side=
& The number of opponent's pawns that occupy the squares of the opposite  \\
\verb=[m|o|diff]= & colour to the colour of own bishop. Applicable only when the side \\
& has a single bishop.\\
\hline
\hline
\verb=capture_possible_on_{sq} [m|o]=
& True is the side can capture a piece on the given square.\\
\texttt{sq=[d1|d2|d3|e1|e2|e3|g5|b5]}
& The squares are named as if the side were playing White.\\
\hline
\verb=capture_happens_next_move_= \ldots
& True if the capture of a piece on the given square had happened \\
\ldots \verb=on_{sq}=
& according to the game data. The squares are named as if the side were \\
\texttt{sq=[d1|d2|d3|e1|e2|e3|g5|b5]} & playing White. \\
\hline
\end{tabular}
\caption{Custom chess concepts (self implemented; i.e.~not from Stockfish 8's API)  used in this paper.
We use \texttt{m} as shorthand for \texttt{mine} and
\texttt{o} as shorthand for \texttt{opponent}. \texttt{diff} stands for the \texttt{difference} between the mine and opponent values of the same concept.
}
\label{tab:concepts-custom}
\end{table}

\begin{table}[]
\centering
\begin{tabular}{l|l}
\textbf{Concept names} & \textbf{Description} \\
\hline
\verb=num_double_pawn_files [t]=
& Number of files that contain one or more \\
\verb=[m|o|diff]= & pawns of a given colour. \\
\hline
\texttt{has\_double\_pawn [m|o]}
& True if a file contains one or more pawns of a given colour. \\
\hline
\texttt{num\_isolated\_pawns} & Number of pawns that have no friendly pawns \\
\texttt{[m|o|diff]} & in the files to their left and right. \\
\hline
\texttt{has\_isolated\_pawn [m|o]}
& True if there is a pawn that has no pawns in the file \\
& to their left or right. \\
\hline
\texttt{has\_pawn\_on\_7th\_rank} 
& True if the side has a pawn that reached the 7th rank. \\
\texttt{[m|o]} & \\
\hline
\texttt{pawns\_on\_7th\_rank}
& Number of pawns that reached the 7th rank. \\
\texttt{[m|o|diff]} & \\
\hline
\texttt{has\_passed\_pawn [m|o]}
& True if the side has a pawn with no opposing pawns to prevent it from \\
& advancing to the eighth rank. \\
\hline
\texttt{num\_passed\_pawns [m|o|diff]} & The number of passed pawns.\\
\hline
\texttt{has\_protected\_passed\_pawn [m|o]}
& True if the side has a passed pawn that is protected by its own pawn.\\
\hline
\texttt{num\_protected\_passed\_pawns} & The number of protected passed pawns.\\
\texttt{[m|o|diff]} & \\
\hline
\texttt{num\_pawn\_islands [m|o|diff]}
& The number of pawn islands.\\
\hline
\texttt{has\_iqp [m|o]} &
True if the side has an isolated queen's pawn (d file). \\
\hline
\texttt{has\_connected\_passed\_pawns [m|o]}
& True if the side has two or more passed pawns on adjacent files. \\
\hline
\verb=num_connected_passed_pawns=
& The number of connected passed pawns that the side has.\\
\verb=[m|o|diff]= & \\
\hline
\end{tabular}
\caption{Custom chess concepts related to pawns (self implemented; i.e.~not from Stockfish 8's API)  used in this paper.
We use \texttt{m} as shorthand for \texttt{mine} and
\texttt{o} as shorthand for \texttt{opponent}. \texttt{diff} stands for the \texttt{difference} between the mine and opponent values of the same concept.
}
\label{tab:concepts-custom-pawns}
\end{table}

\FloatBarrier

\section{Regression results for all concepts}
\label{sec:all_concepts_regression_supplement}

In our experiments, we have considered a large number of potential human chess concepts within a set of concepts we have tried to identify, localize, and explore the acquisition of within the AlphaZero chess model.
Even this extended list is far from being able to explicitly capture the vast chess knowledge that has accumulated over centuries and the multitude of patterns that can appear on the board. It is merely a starting point for further exploration.

\begin{figure}
    \centering
    \includegraphics[width=\textwidth]{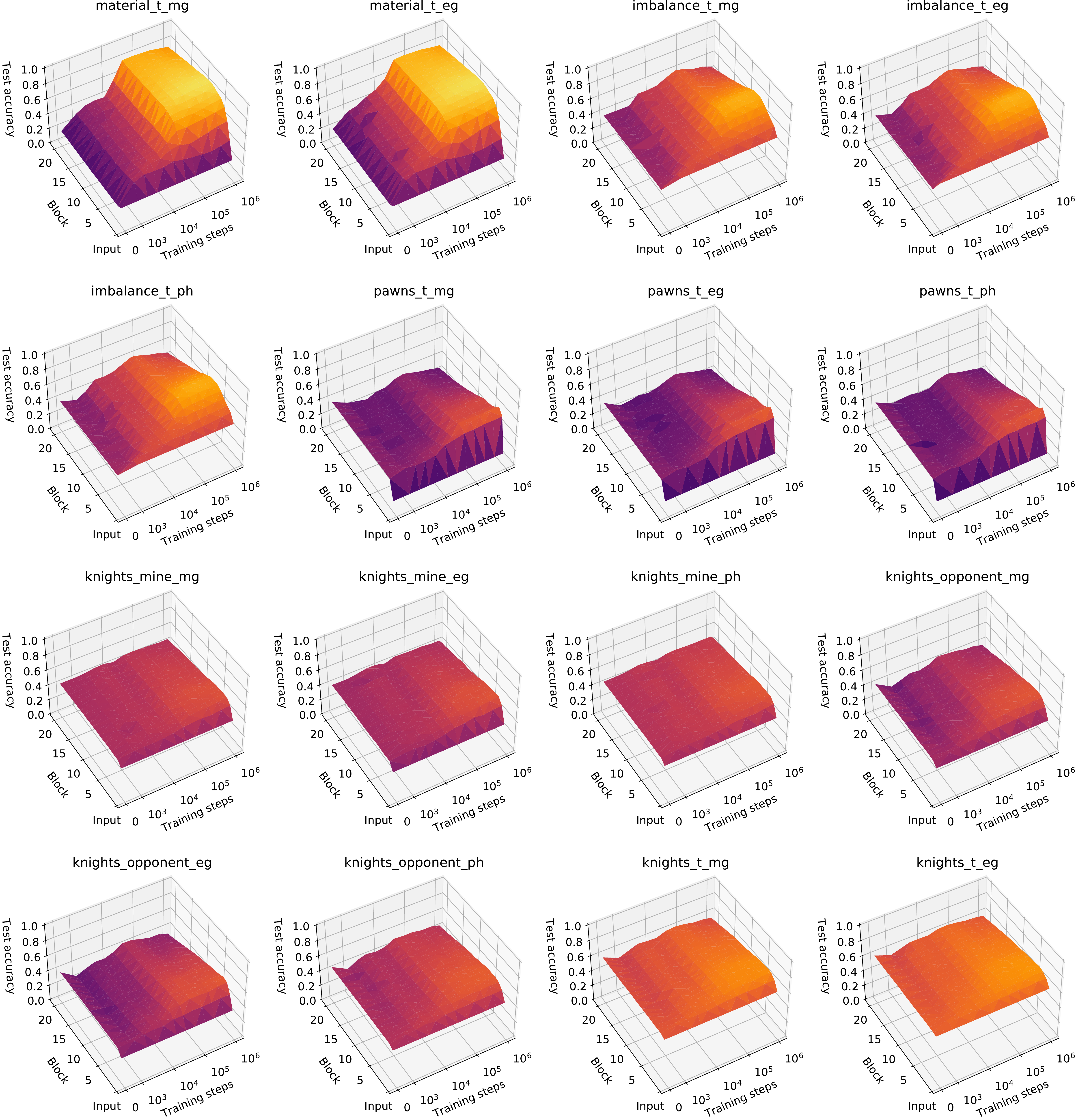}
    \caption{Regression results for Stockfish concepts from Table~\ref{tab:concepts}.}
    \label{fig:all_sf_concepts_1}
\end{figure}

\begin{figure}
    \centering
    \includegraphics[width=\textwidth]{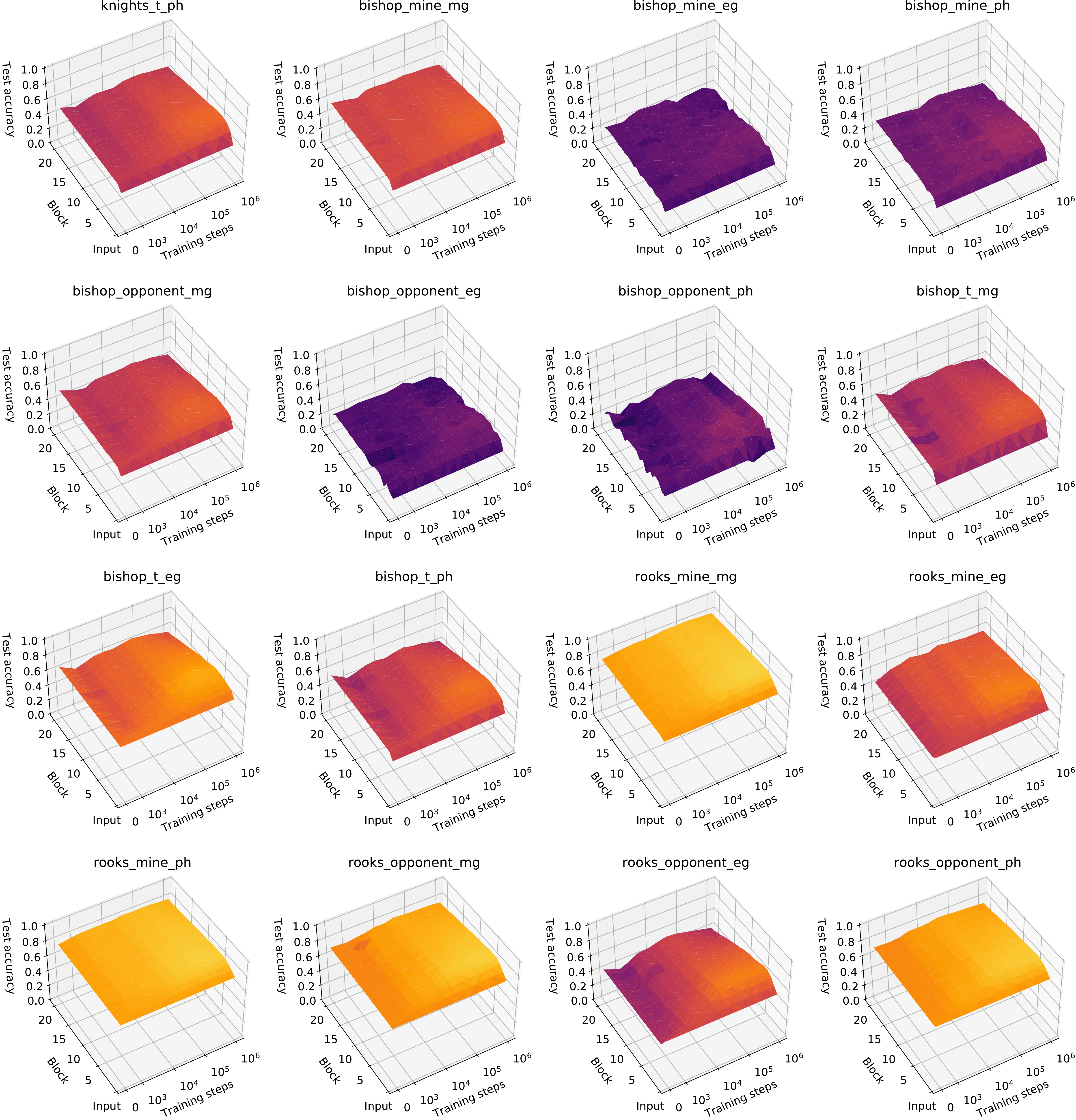}
    \caption{Regression results for Stockfish concepts from Table~\ref{tab:concepts}, continued.}
    \label{fig:all_sf_concepts_2}
\end{figure}

\begin{figure}
    \centering
    \includegraphics[width=\textwidth]{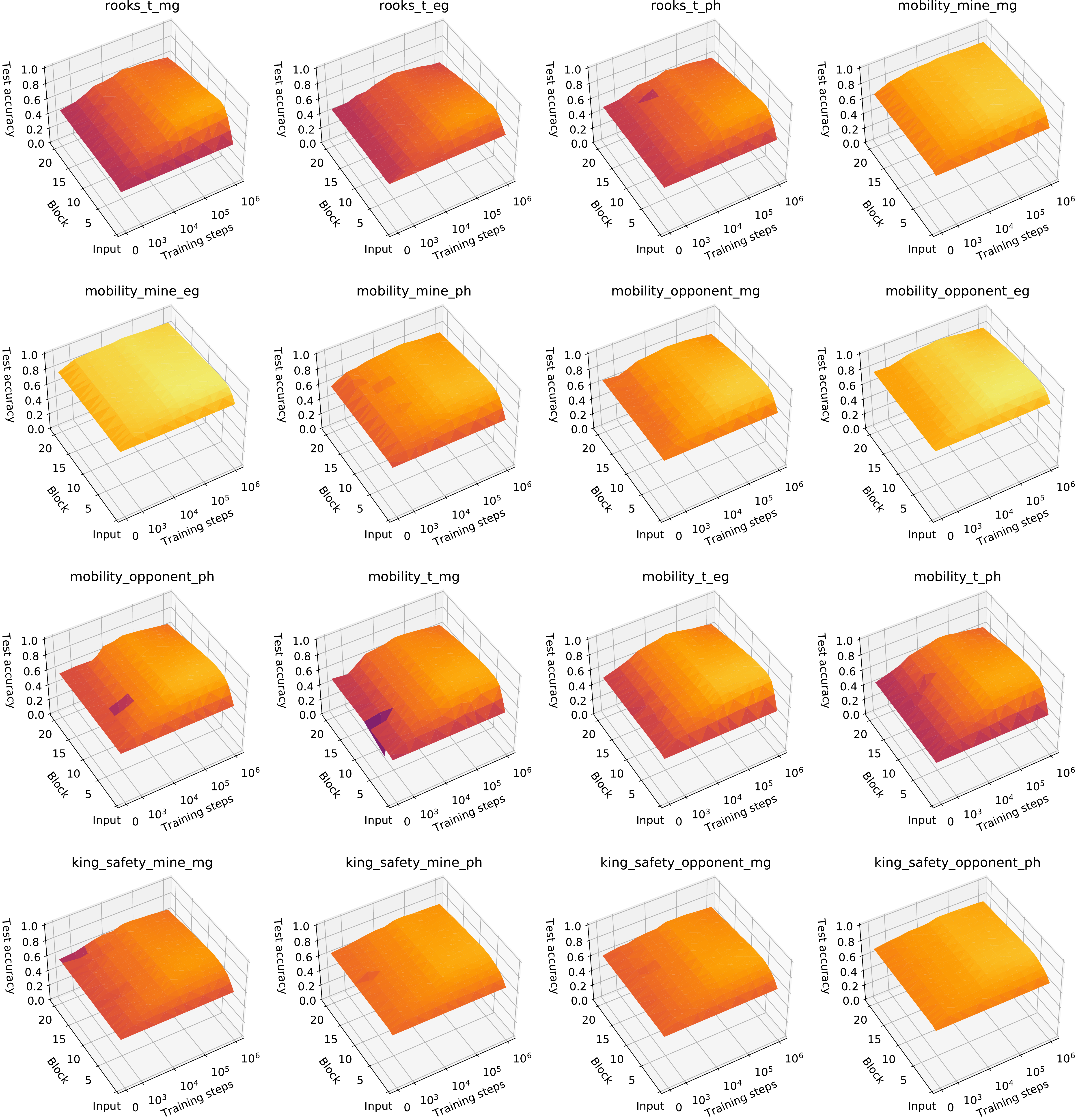}
    \caption{Regression results for Stockfish concepts from Table~\ref{tab:concepts}, continued.}
    \label{fig:all_sf_concepts_3}
\end{figure}

\begin{figure}
    \centering
    \includegraphics[width=\textwidth]{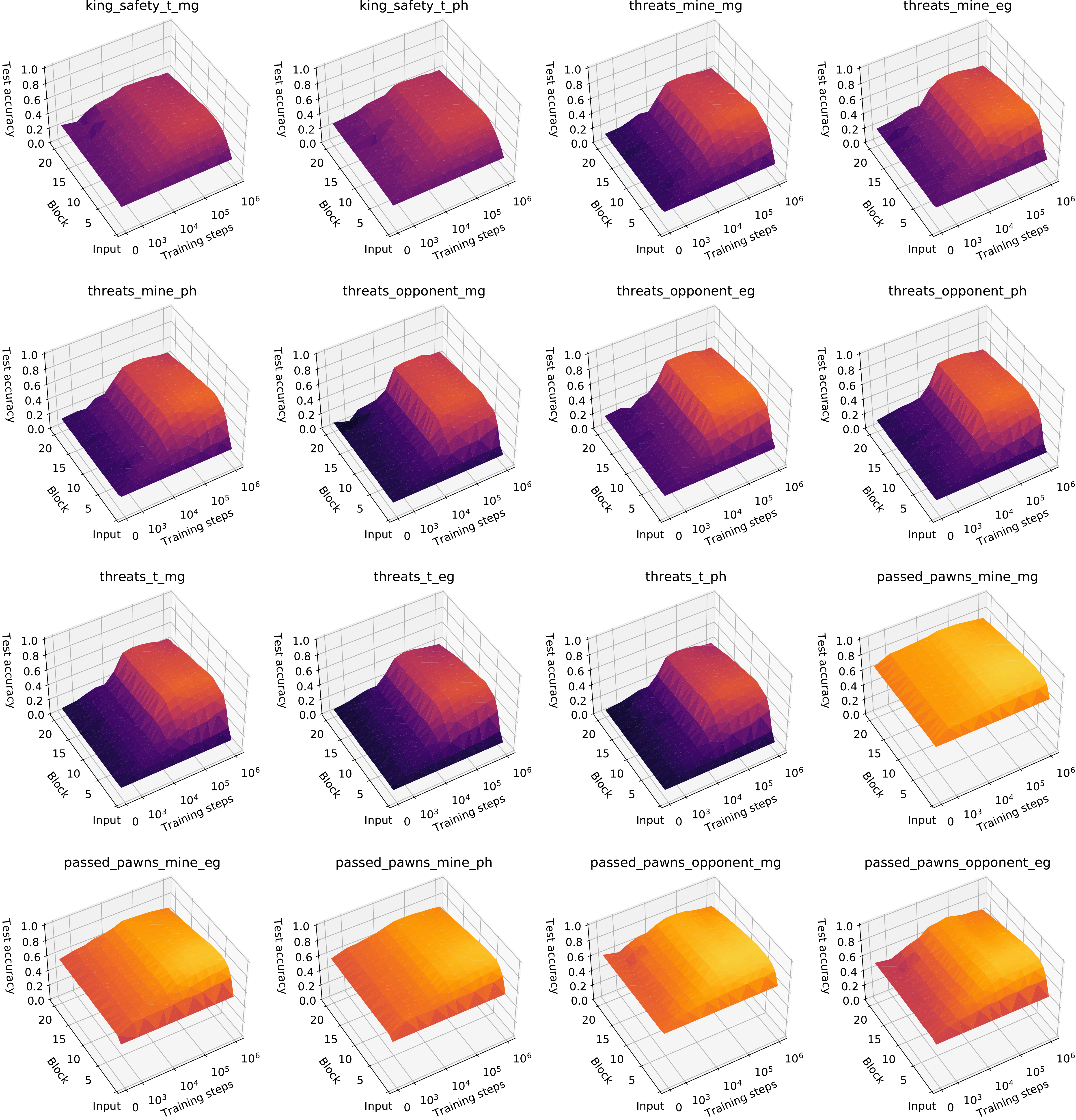}
    \caption{Regression results for Stockfish concepts from Table~\ref{tab:concepts}, continued.}
    \label{fig:all_sf_concepts_4}
\end{figure}

\begin{figure}
    \centering
    \includegraphics[width=\textwidth]{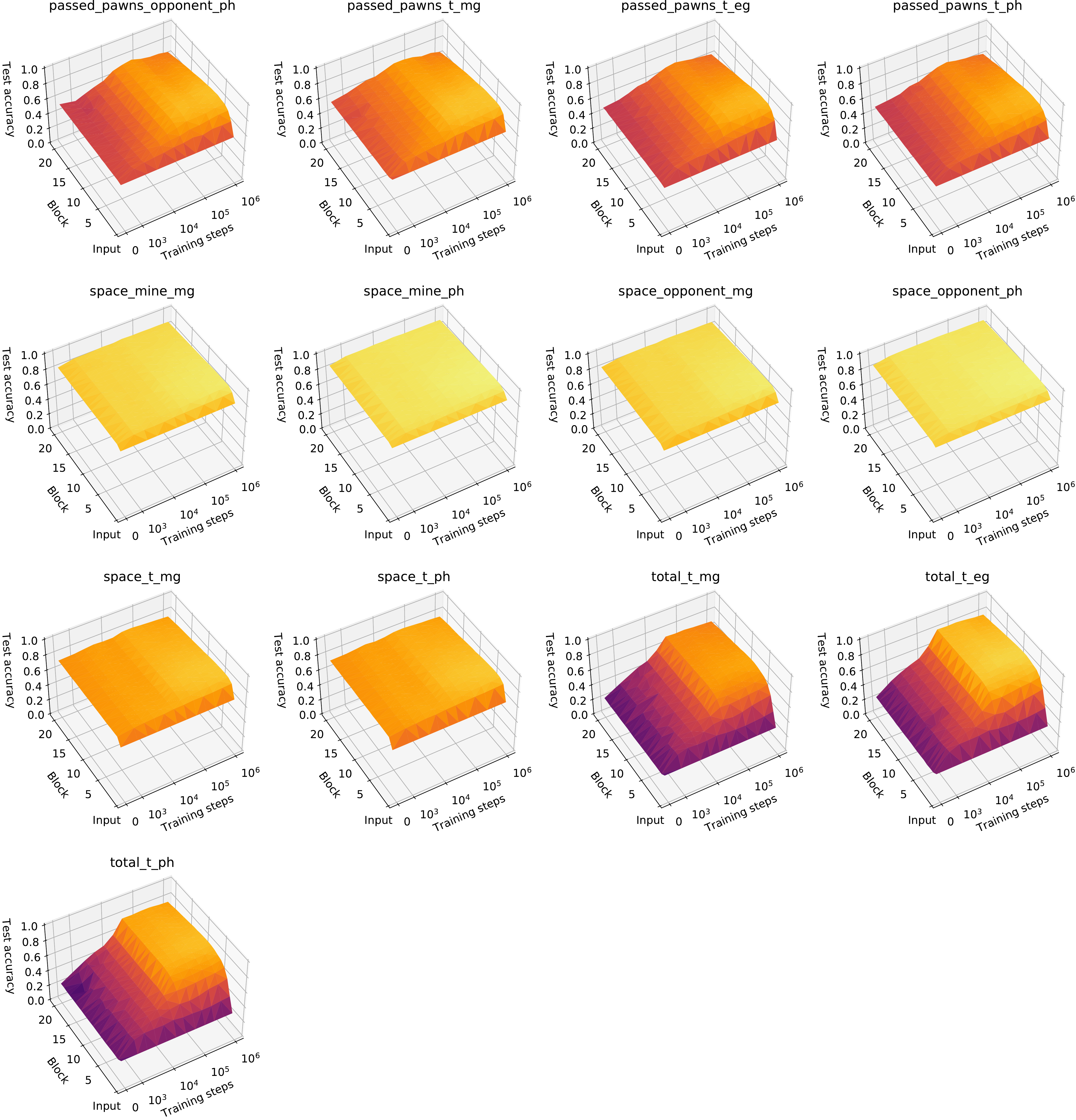}
    \caption{Regression results for Stockfish concepts from Table~\ref{tab:concepts}, continued.}
    \label{fig:all_sf_concepts_5}
\end{figure}

\begin{figure}
    \centering
    \includegraphics[width=\textwidth]{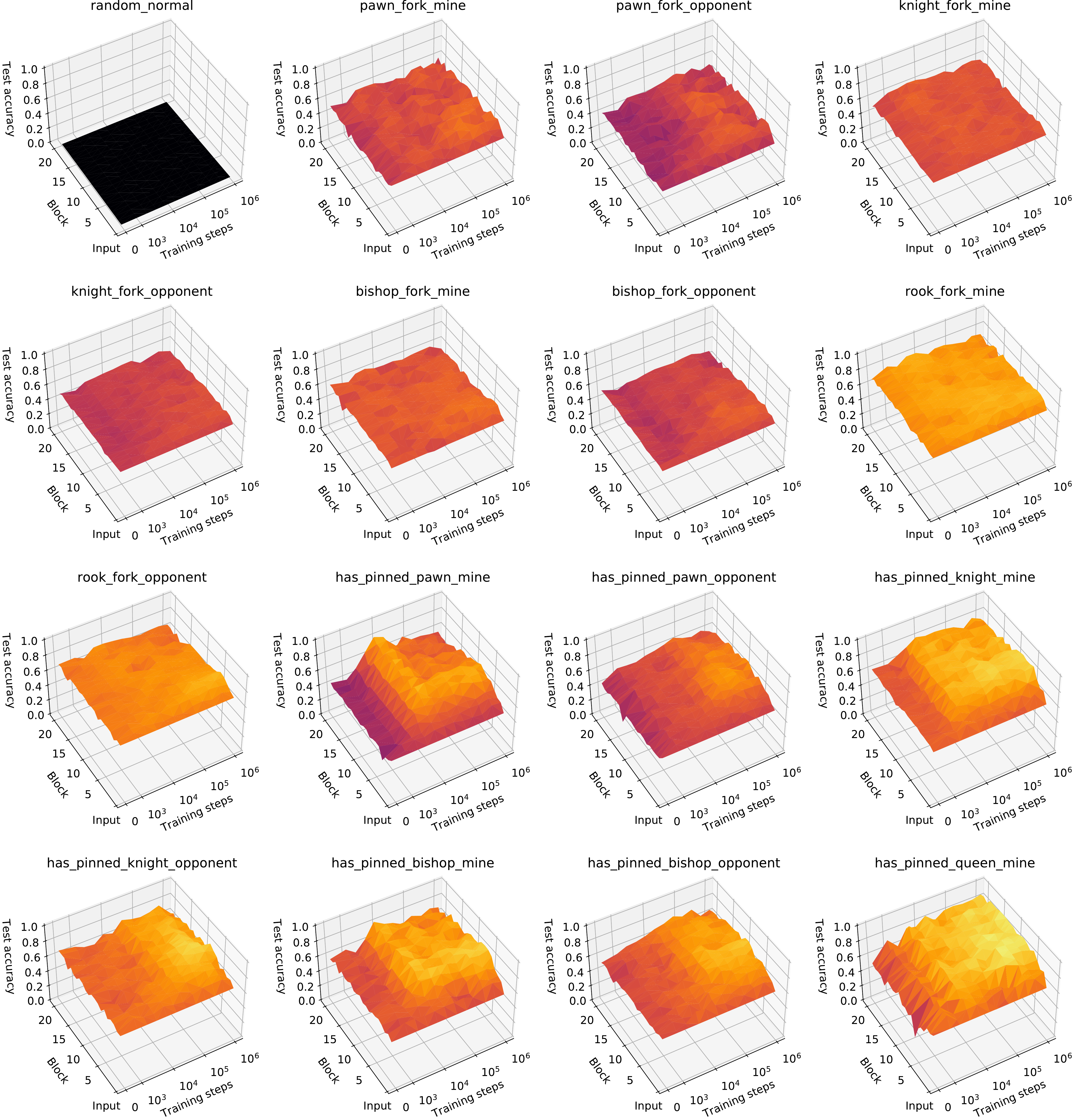}
    \caption{Regression results for custom concepts from Table~\ref{tab:concepts-custom}, excluding capture-related concepts.}
    \label{fig:custom_concepts_nocapture_1}
\end{figure}

\begin{figure}
    \centering
    \includegraphics[width=\textwidth]{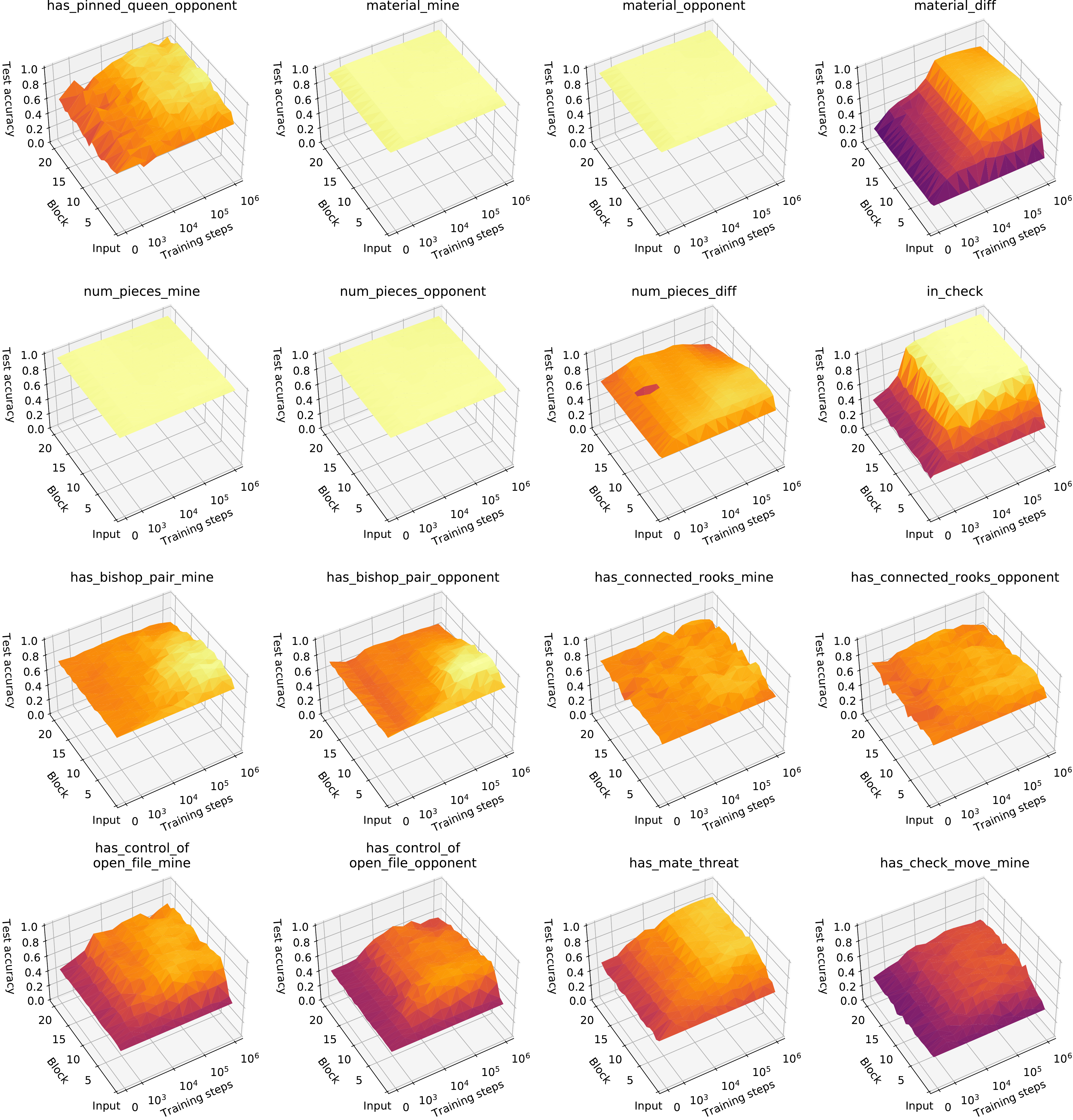}
    \caption{Regression results for custom concepts from Table~\ref{tab:concepts-custom}, excluding capture-related concepts, continued.}
    \label{fig:custom_concepts_nocapture_2}
\end{figure}

\begin{figure}
    \centering
    \includegraphics[width=\textwidth]{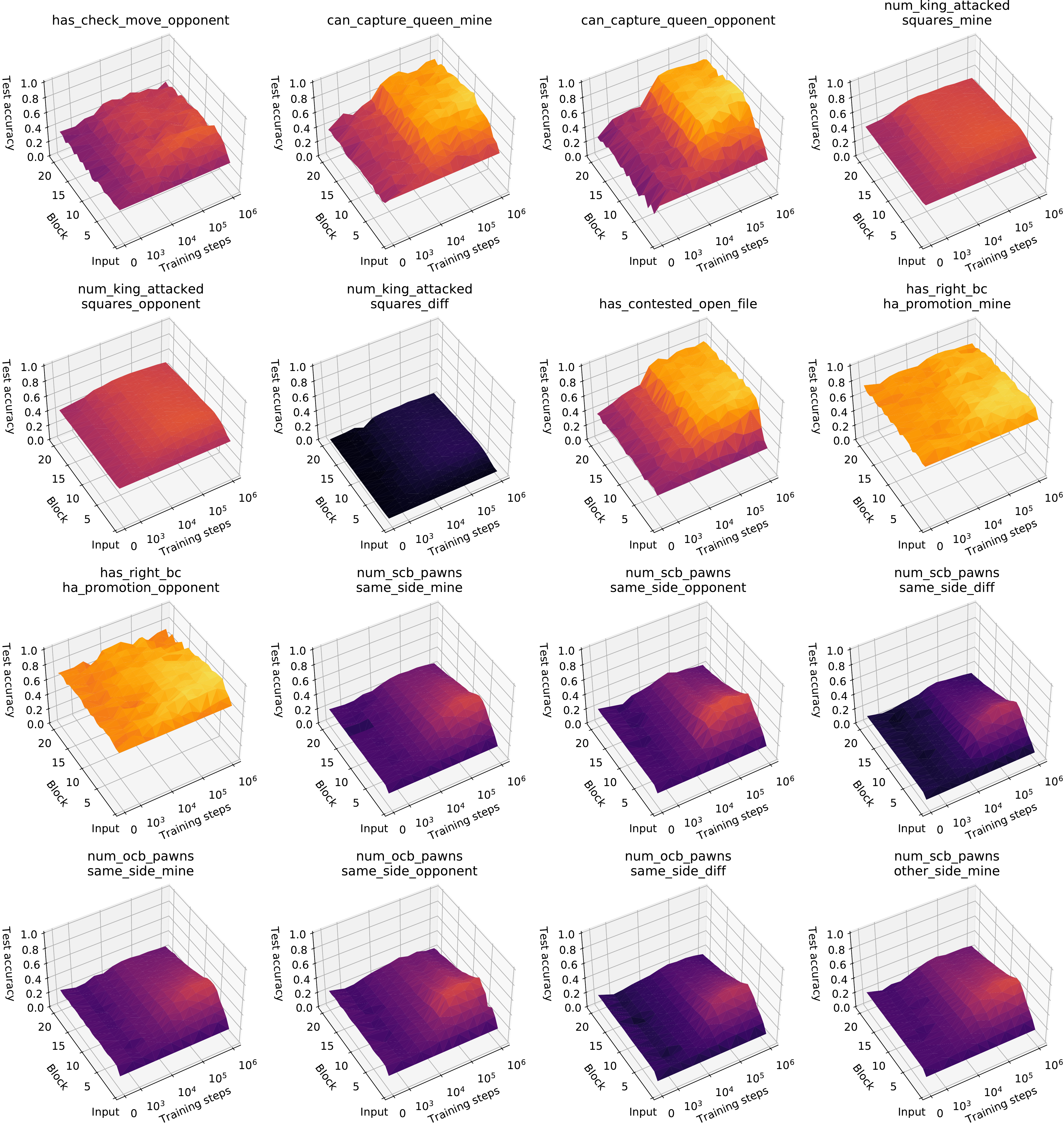}
    \caption{Regression results for custom concepts from Table~\ref{tab:concepts-custom}, excluding capture-related concepts, continued.}
    \label{fig:custom_concepts_nocapture_3}
\end{figure}

\begin{figure}
    \centering
    \includegraphics[width=\textwidth]{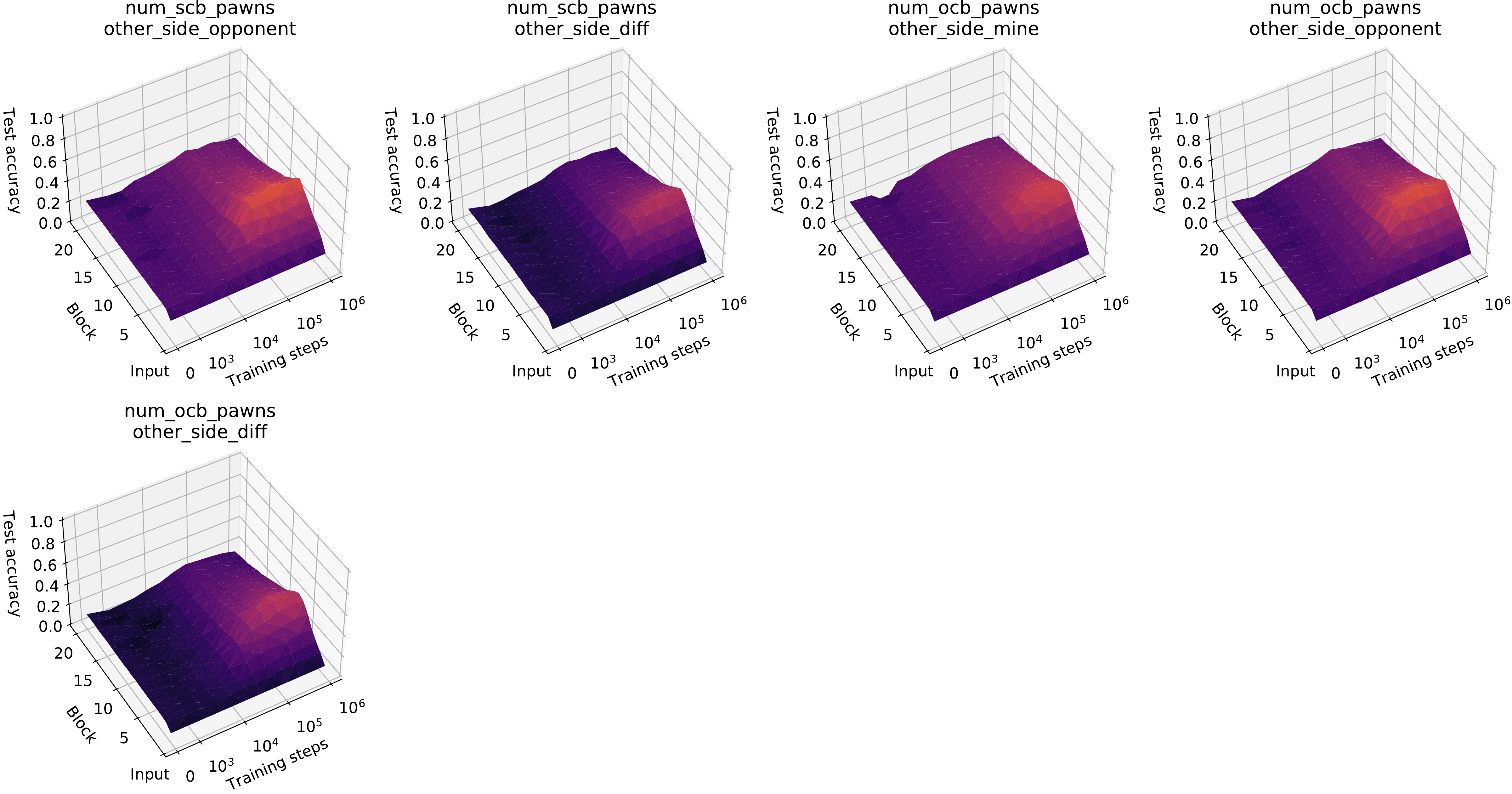}
    \caption{Regression results for custom concepts from Table~\ref{tab:concepts-custom}, excluding capture-related concepts, continued.}
    \label{fig:custom_concepts_nocapture_4}
\end{figure}

\begin{figure}
    \centering
    \includegraphics[width=\textwidth]{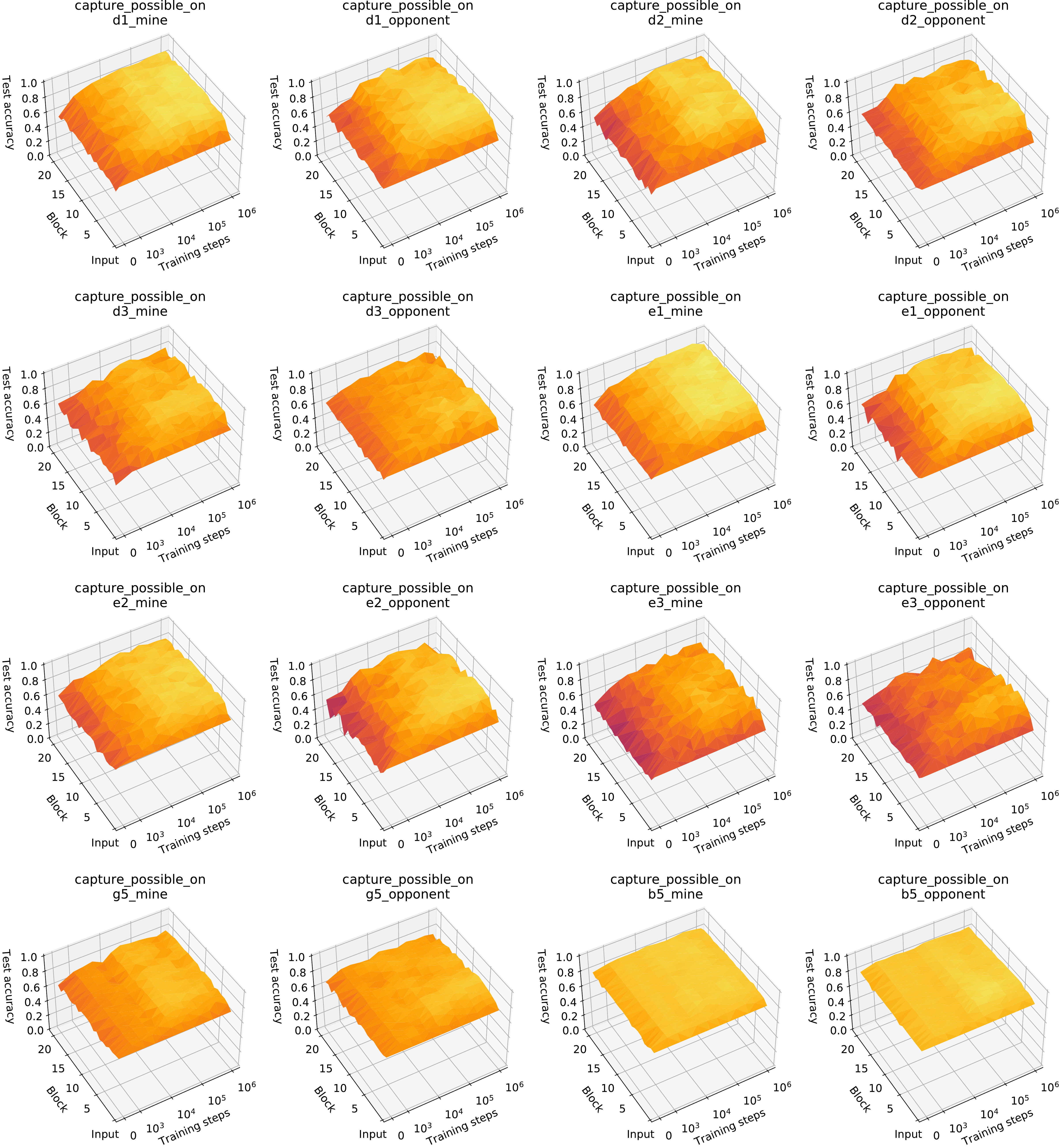}
    \caption{Regression results for custom concepts from Table~\ref{tab:concepts-custom} related to captures.}
    \label{fig:custom_concepts_capture}
\end{figure}

\begin{figure}
    \centering
    \includegraphics[width=\textwidth]{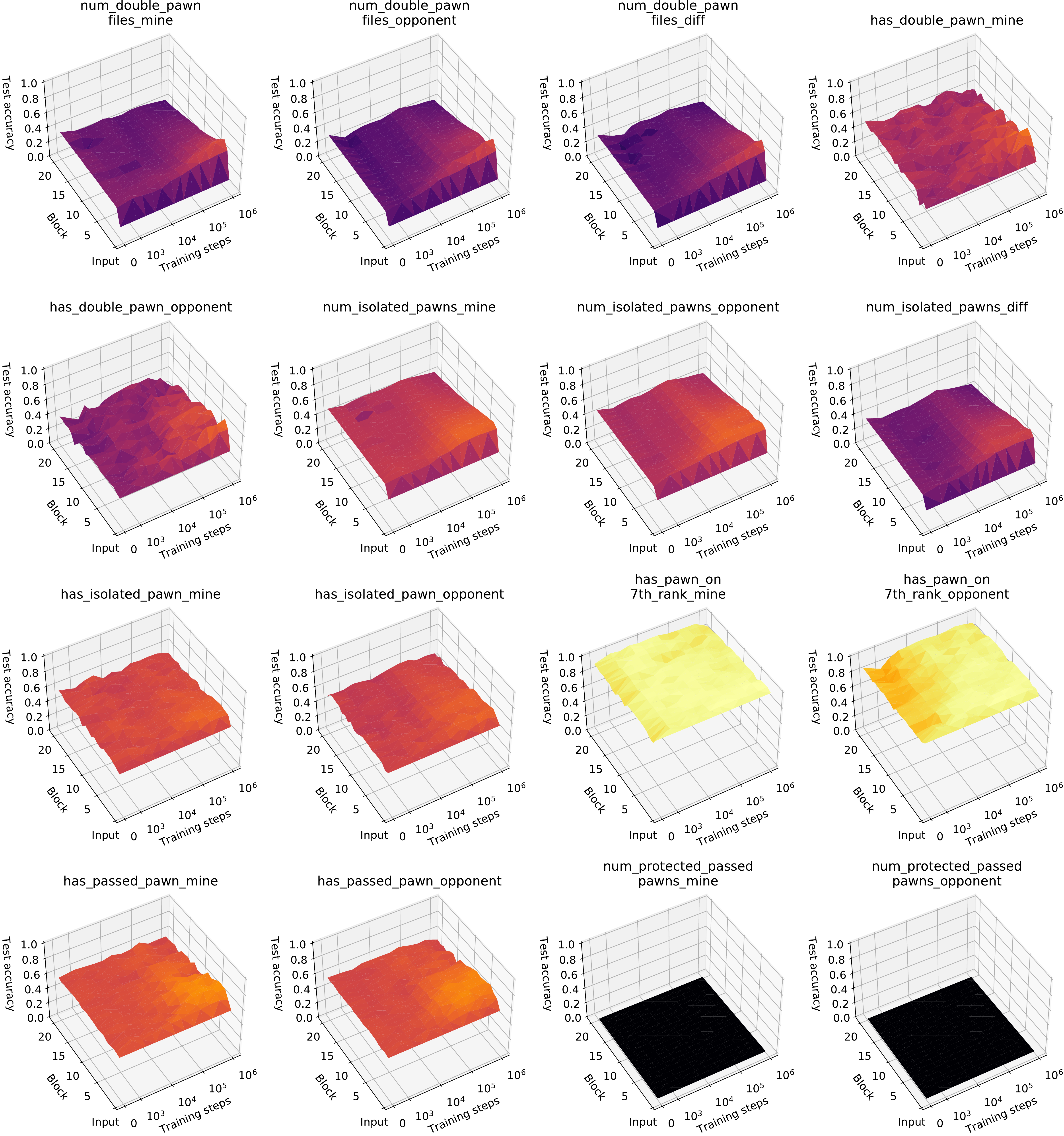}
    \caption{Regression results for custom pawn-related concepts from Table~\ref{tab:concepts-custom-pawns}.}
    \label{fig:custom_pawn_concepts_1}
\end{figure}

\begin{figure}
    \centering
    \includegraphics[width=\textwidth]{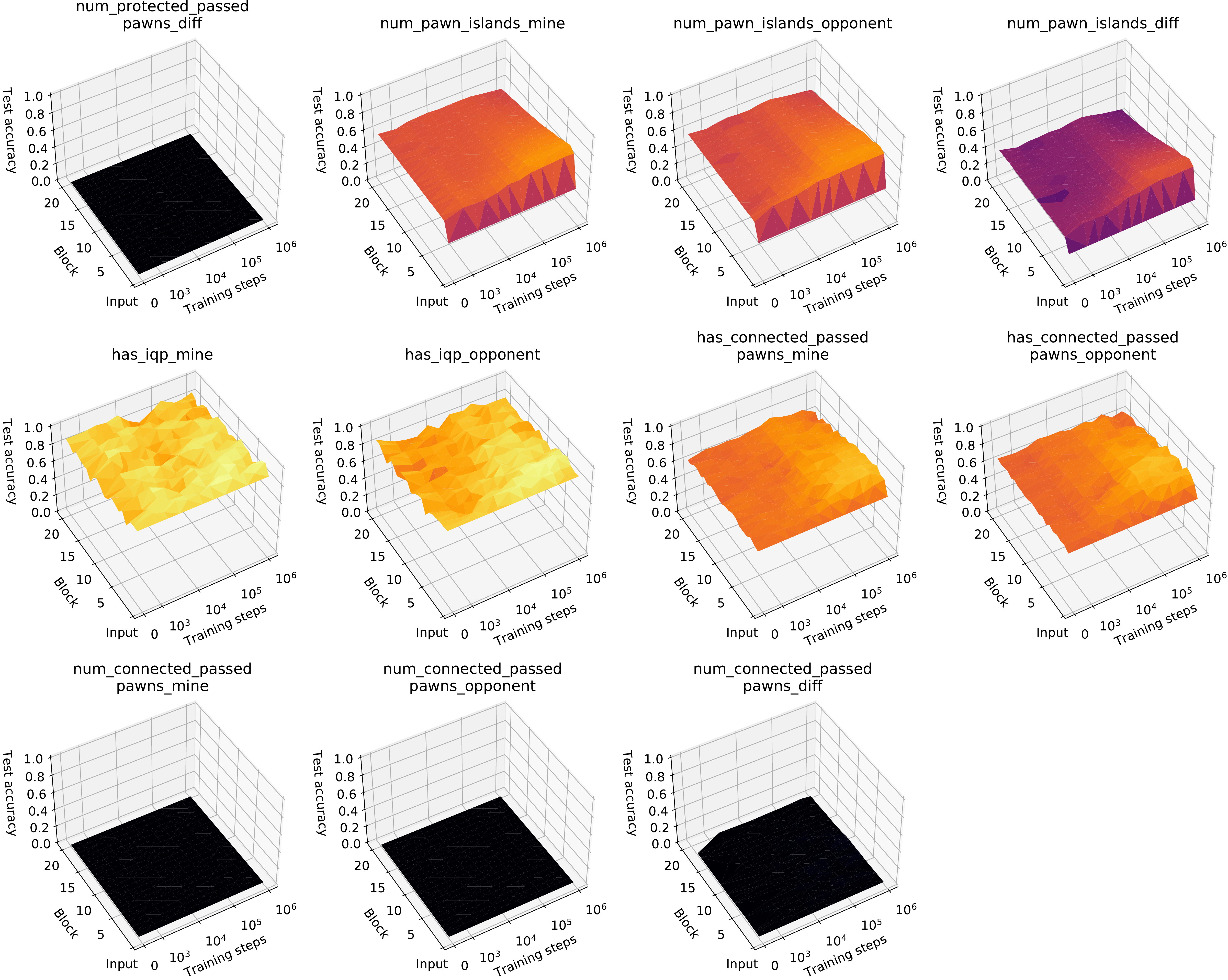}
    \caption{Regression results for custom pawn-related concepts from Table~\ref{tab:concepts-custom-pawns}, continued.}
    \label{fig:custom_pawn_concepts_2}
\end{figure}

While we focused our discussion in the main text on a smaller number of relevant human concepts and their regression from different layers in the AlphaZero network over time, we present the results for the extended list of concepts we've used in our experiments here in Figures \ref{fig:all_sf_concepts_1}
to \ref{fig:custom_pawn_concepts_2}.

\FloatBarrier

\section{AlphaZero policy progression through training for different training seeds}
\label{appendix:policy-progression}

\begin{figure}[h!]
\vspace{-10mm}
\centering
\begin{subfigure}[t]{.4\textwidth}
  \centering
  \includegraphics[width=1.0\linewidth]{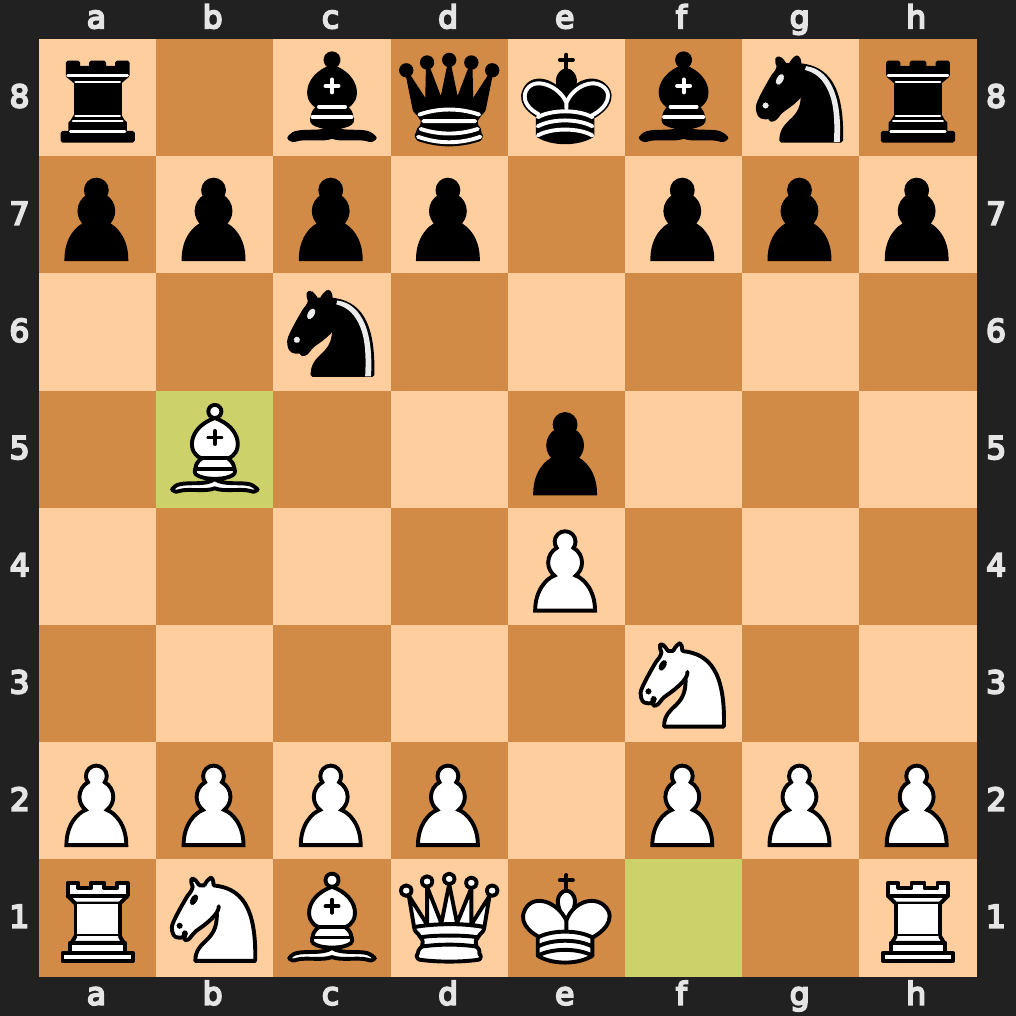}  
  \caption{Sequence start position.}
\end{subfigure}
\hspace{0.14\linewidth}
\begin{subfigure}[t]{.34\textwidth}
  \centering
  \includegraphics[width=1.0\linewidth]{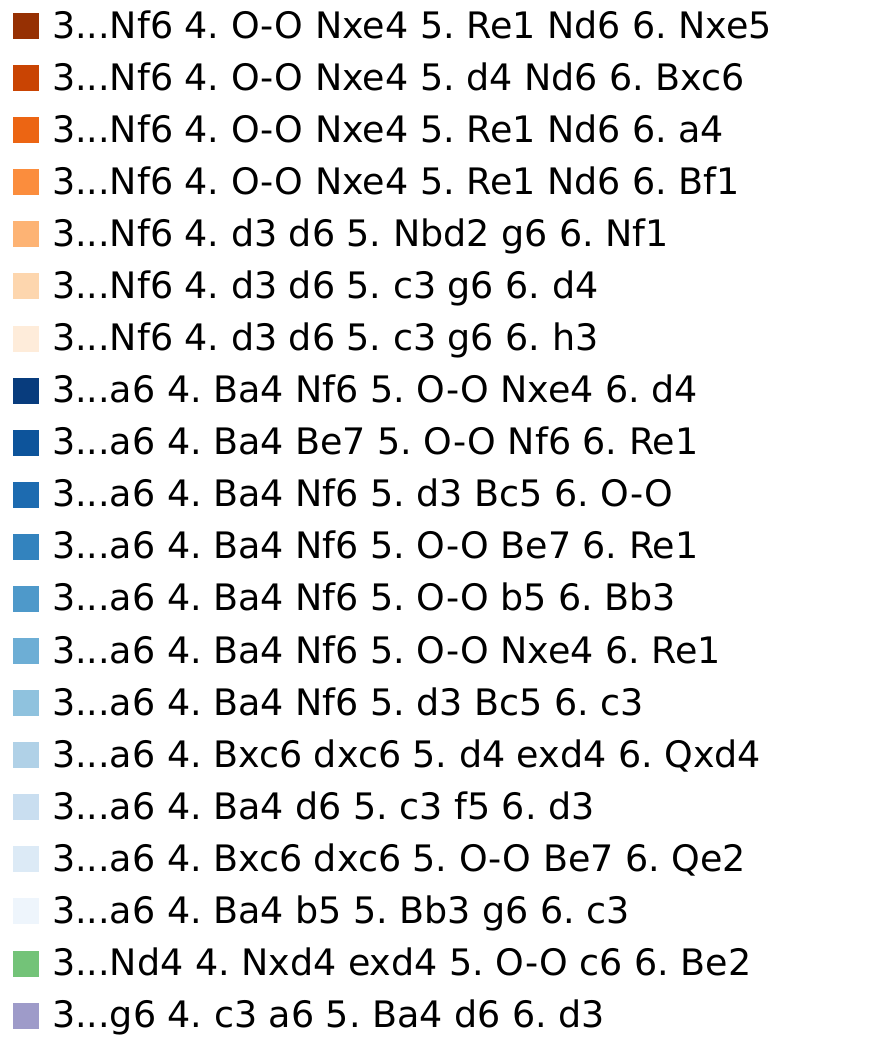}  
  \caption{Legend.}
\end{subfigure}
\vspace{2mm}

\begin{subfigure}{.49\textwidth}
  \centering
  \includegraphics[clip,trim=0 15 0 0,width=\linewidth]{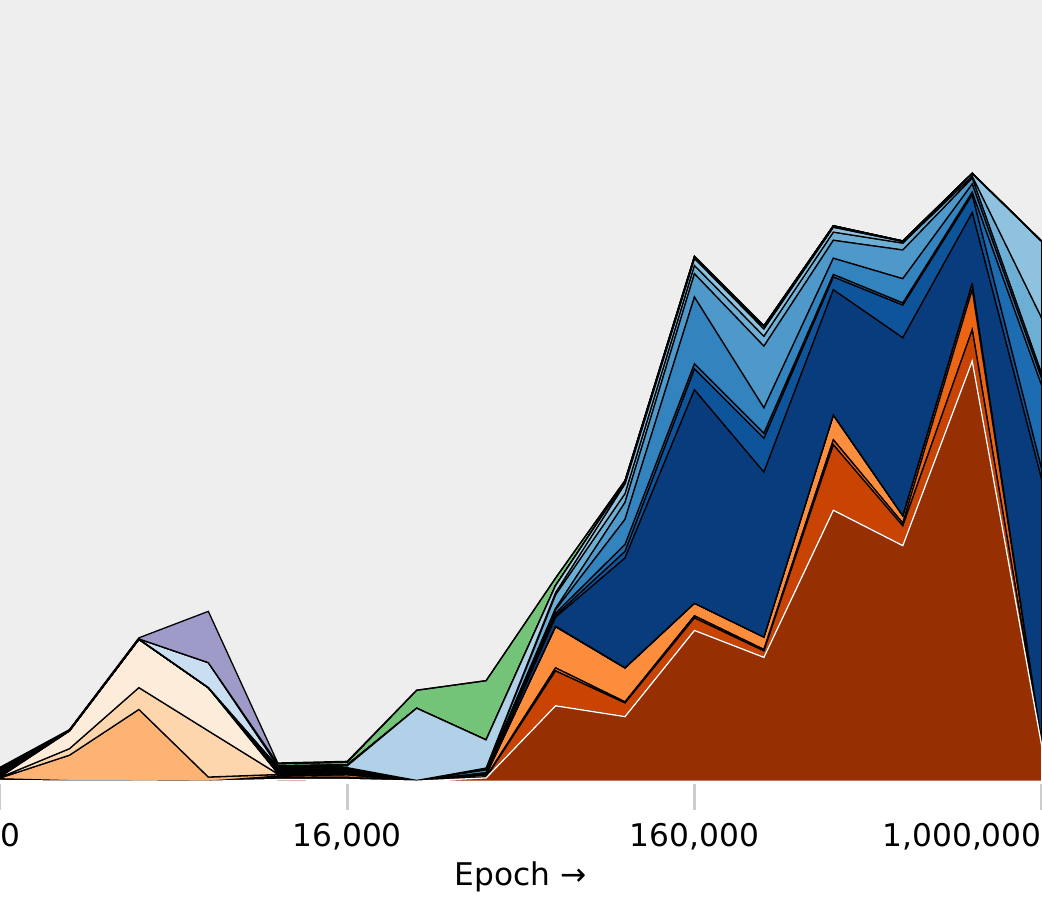} \\
  \small AlphaZero training steps \normalsize
  \caption{Training seed 1.}
\end{subfigure}
\hfill
\begin{subfigure}{.49\textwidth}
  \centering
  \includegraphics[clip,trim=0 15 0 0,width=\linewidth]{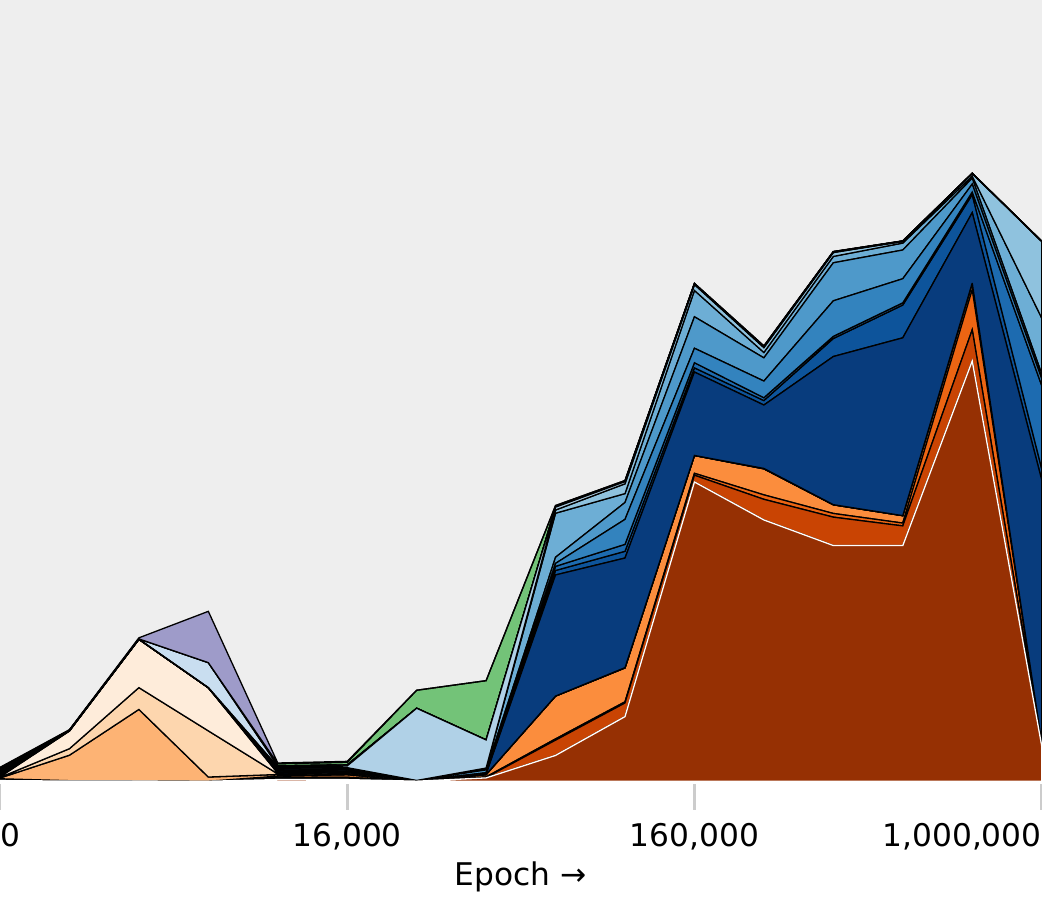}  \\
  \small AlphaZero training steps \normalsize
  \caption{Training seed 2.}
\end{subfigure}
\vspace{2mm}

\begin{subfigure}{.49\textwidth}
  \centering
  \includegraphics[clip,trim=0 15 0 0,width=\linewidth]{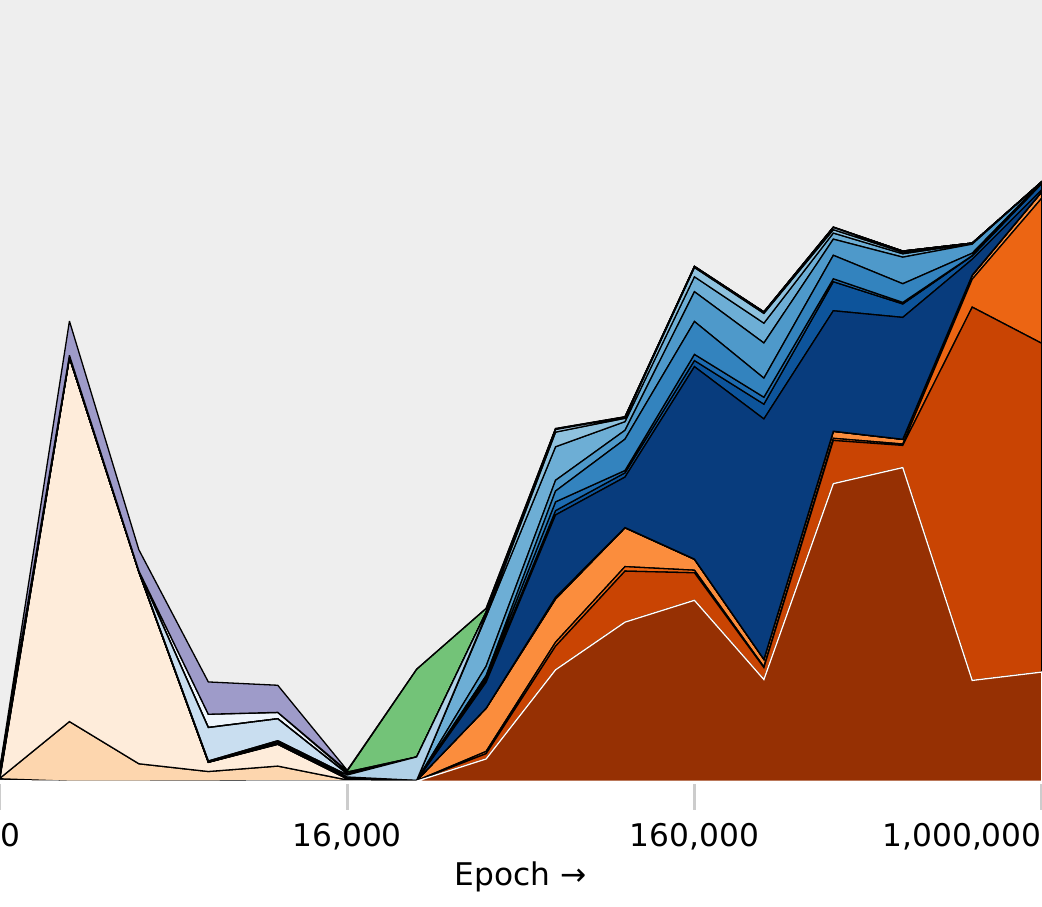} \\
  \small AlphaZero training steps \normalsize 
  \caption{Training seed 3.}
\end{subfigure}
\hfill
\begin{subfigure}{.49\textwidth}
  \centering
  \includegraphics[clip,trim=0 15 0 0,width=\linewidth]{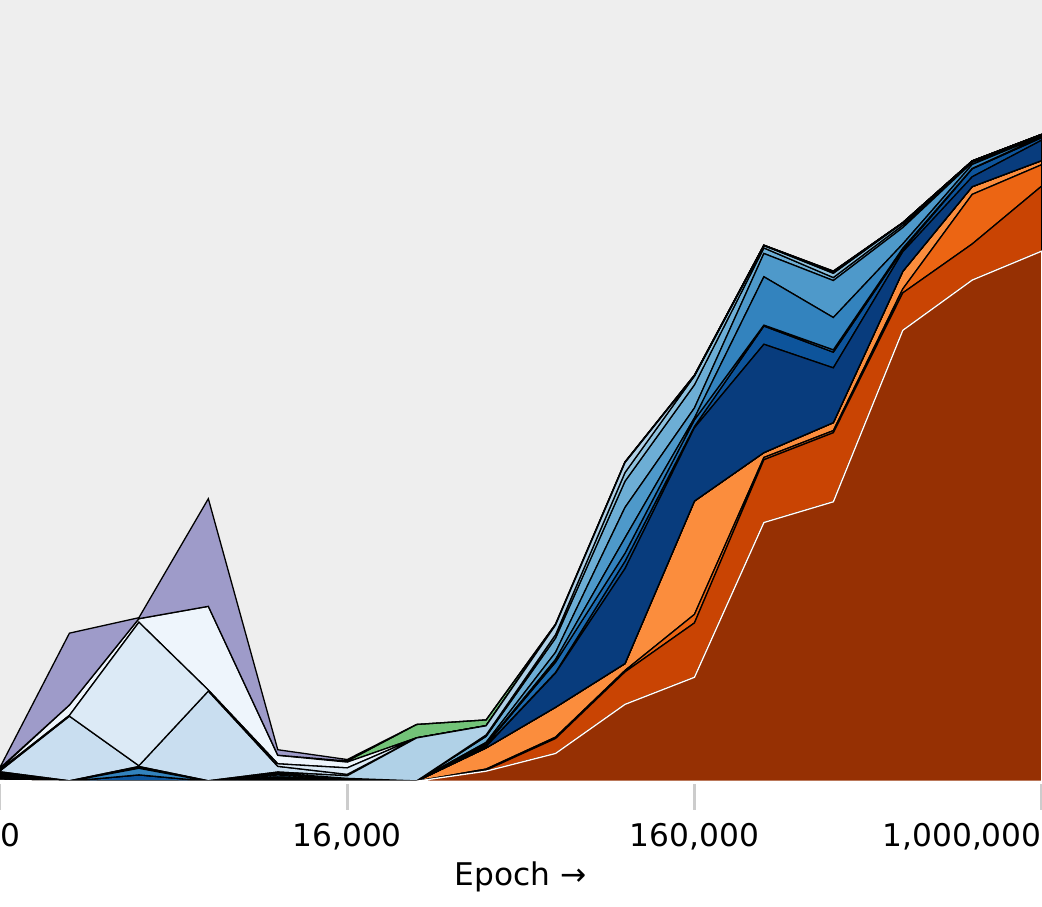} \\
  \small AlphaZero training steps \normalsize 
  \caption{Training seed 4.}
\end{subfigure}
\caption{Top 20 move sequences with the highest joint probability after 1. e4 e5 2. \symknight f3 \symknight c6 3. \symbishop b5.}
\label{fig:move_seq_01}
\end{figure}

\begin{figure}[h!]
\vspace{-10mm}
\centering
\begin{subfigure}[t]{.4\textwidth}
  \centering
  \includegraphics[width=1.0\linewidth]{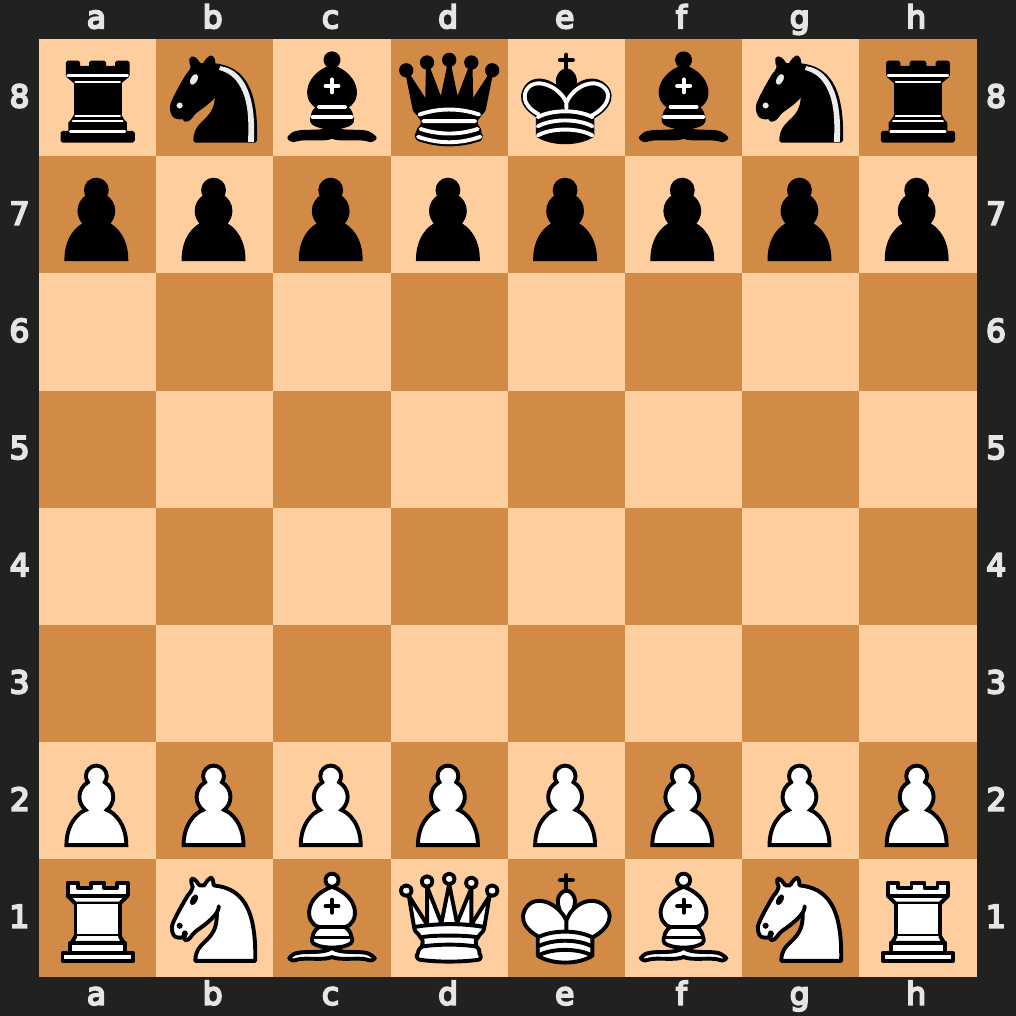}  
  \caption{Sequence start position.}
\end{subfigure}
\hspace{0.14\linewidth}
\begin{subfigure}[t]{.34\textwidth}
  \centering
  \includegraphics[width=1.0\linewidth]{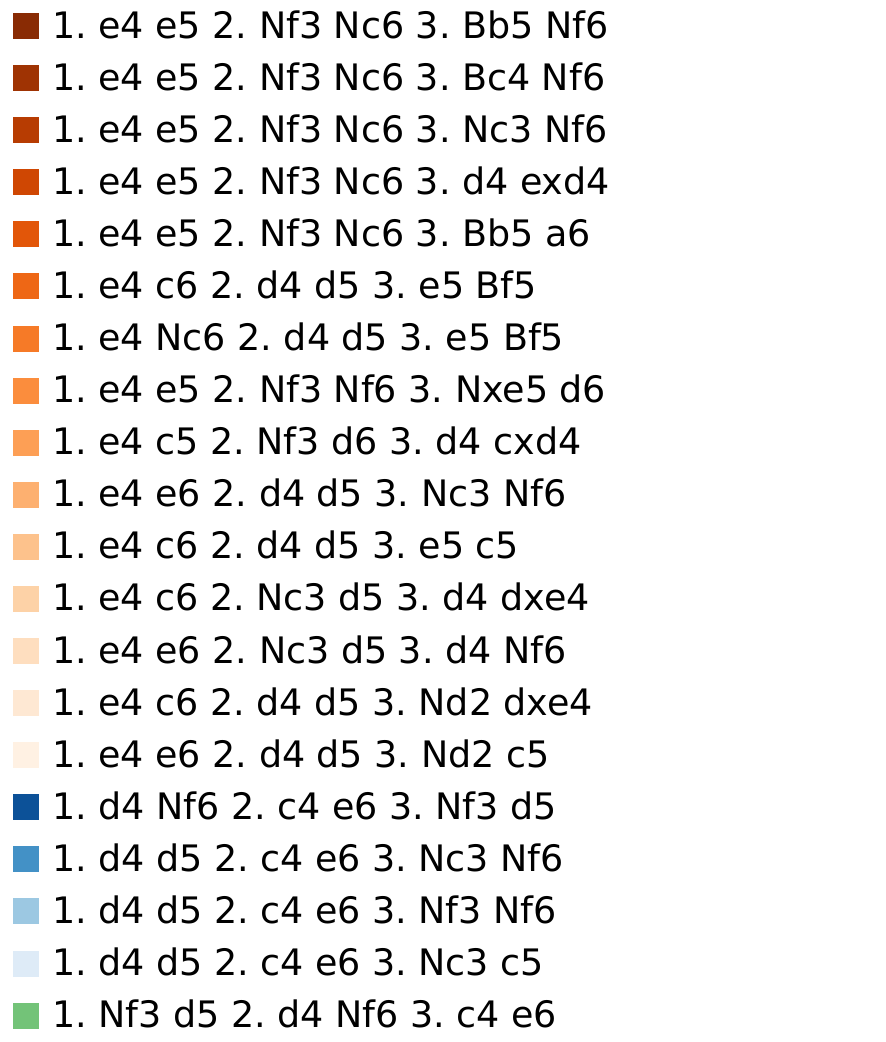}  
  \caption{Legend.}
\end{subfigure}
\vspace{2mm}

\begin{subfigure}{.49\textwidth}
  \centering
  \includegraphics[clip,trim=0 15 0 0,width=\linewidth]{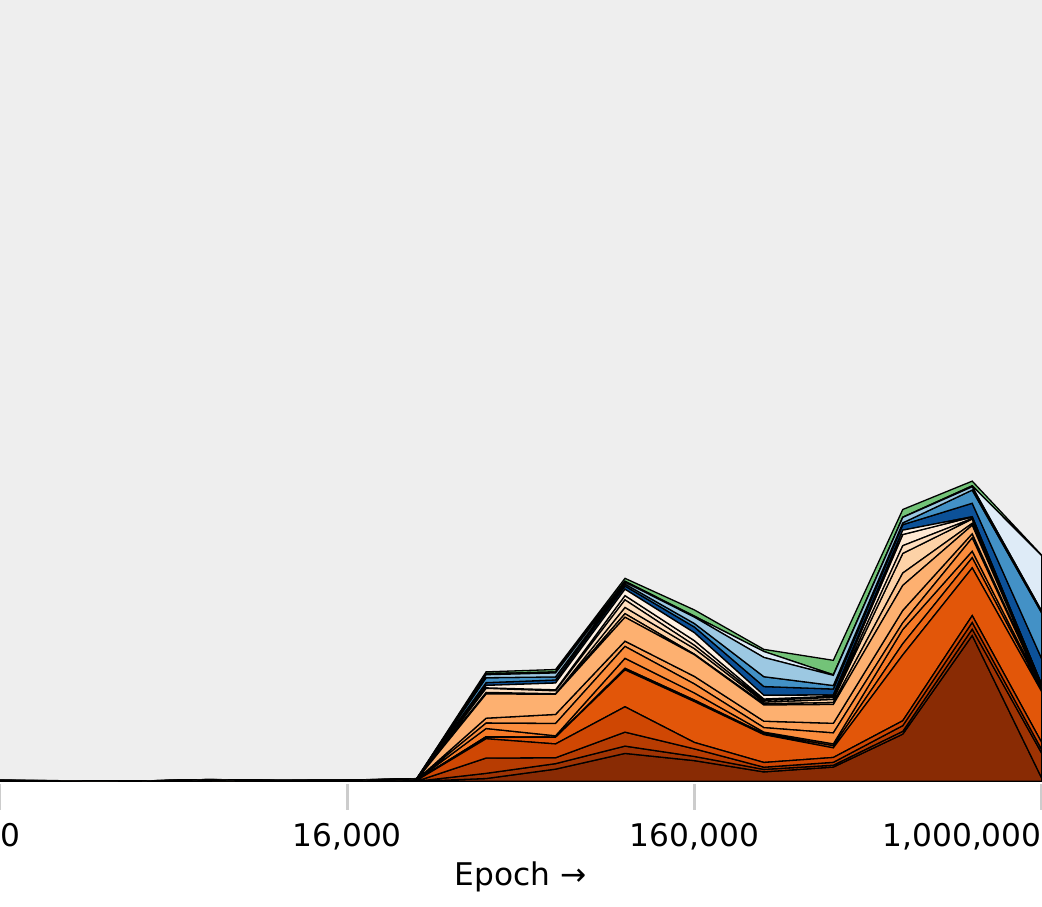} \\ \small AlphaZero training steps \normalsize
  \caption{Training seed 1.}
\end{subfigure}
\hfill
\begin{subfigure}{.49\textwidth}
  \centering
  \includegraphics[clip,trim=0 15 0 0,width=\linewidth]{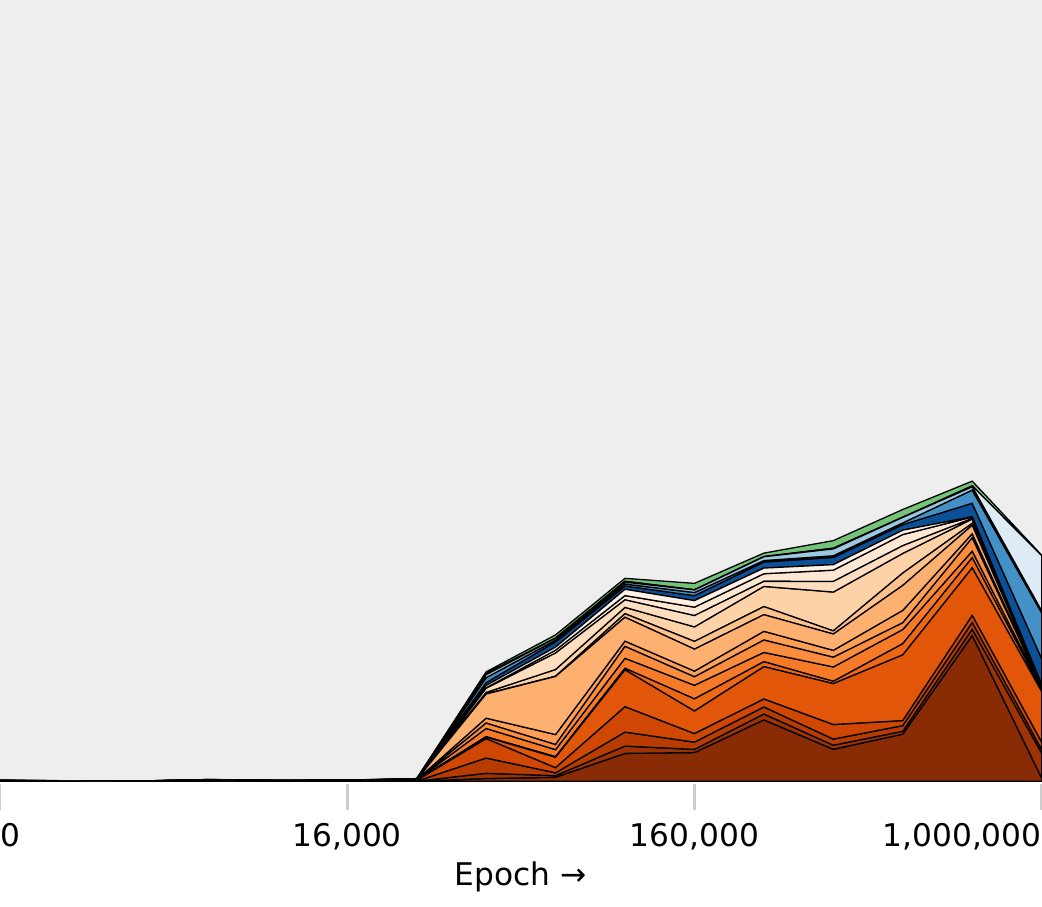} \\ \small AlphaZero training steps \normalsize
  \caption{Training seed 2.}
\end{subfigure}
\vspace{2mm}

\begin{subfigure}{.49\textwidth}
  \centering
  \includegraphics[clip,trim=0 15 0 0,width=\linewidth]{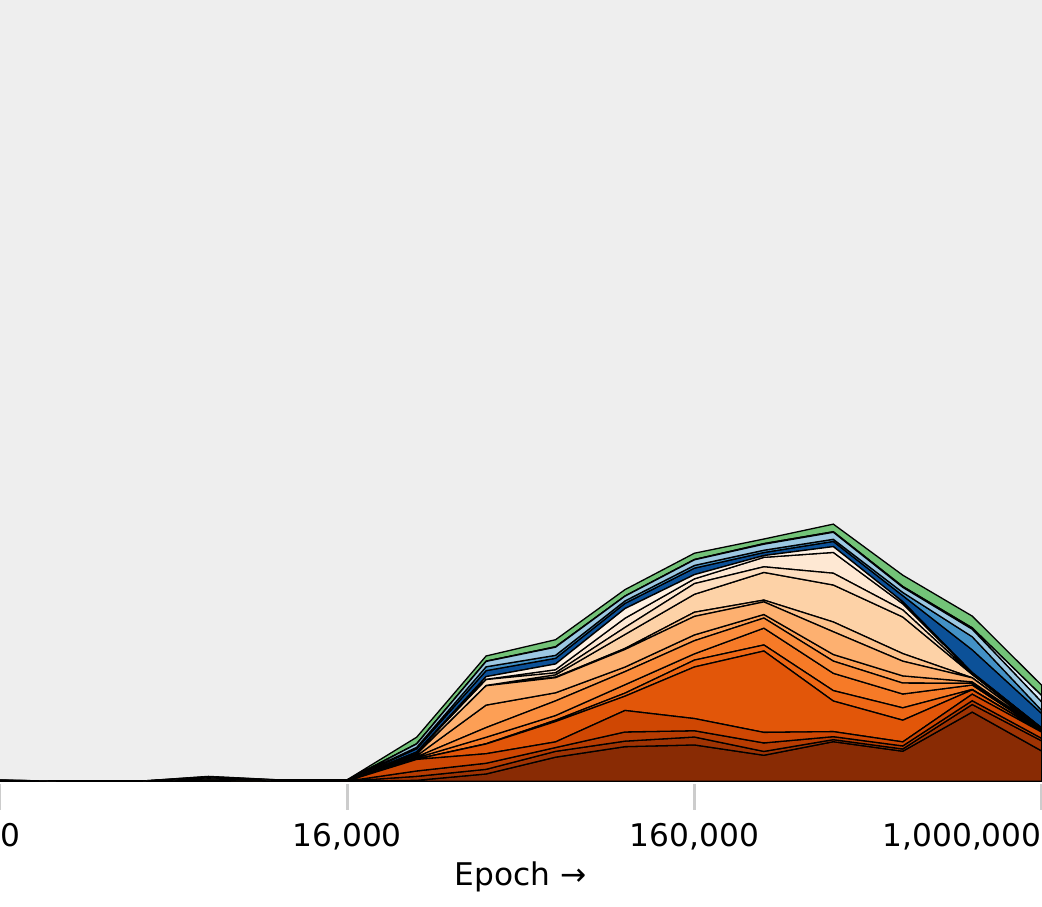} \\ \small AlphaZero training steps \normalsize 
  \caption{Training seed 3.}
\end{subfigure}
\hfill
\begin{subfigure}{.49\textwidth}
  \centering
  \includegraphics[clip,trim=0 15 0 0,width=\linewidth]{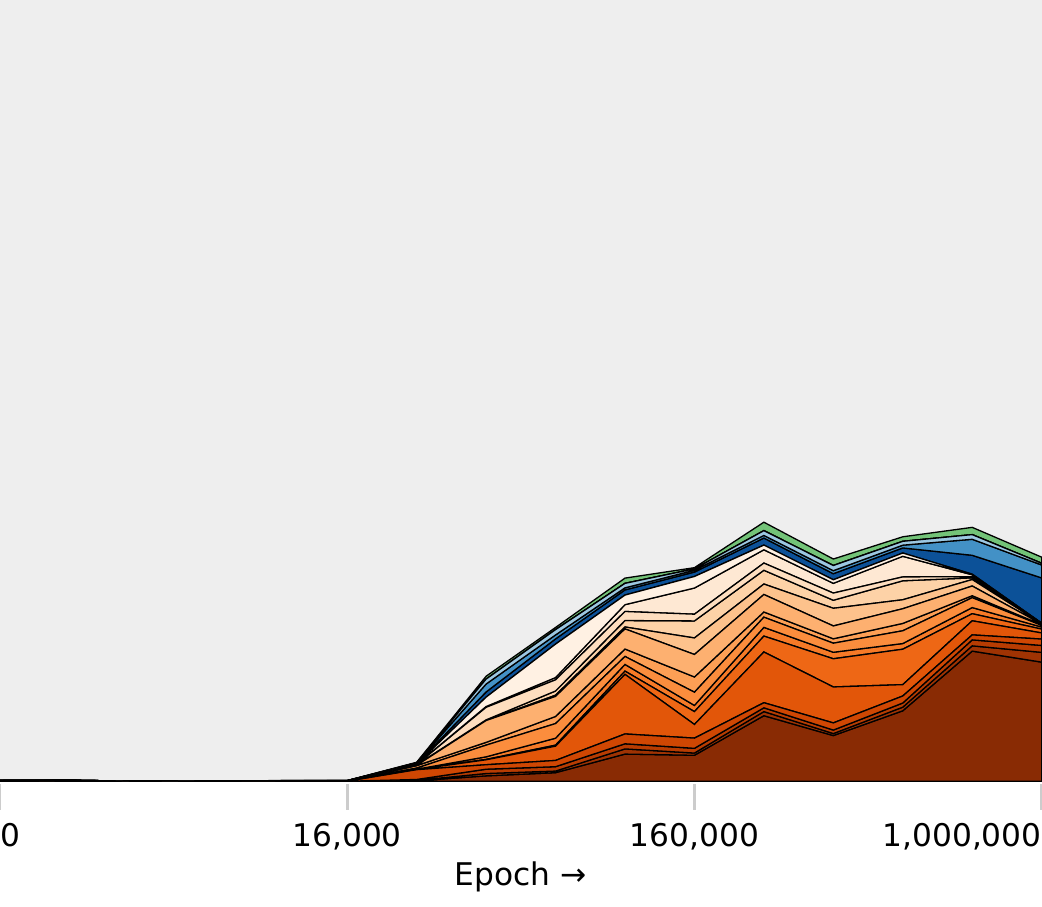} \\ \small AlphaZero training steps \normalsize
  \caption{Training seed 4.}
\end{subfigure}
\caption{Top 20 move sequences with the highest joint probability at the start position.} 
\label{fig:move_seq_05}
\end{figure}

\begin{figure}[h!]
\vspace{-10mm}
\centering
\begin{subfigure}[t]{.4\textwidth}
  \centering
  \includegraphics[width=1.0\linewidth]{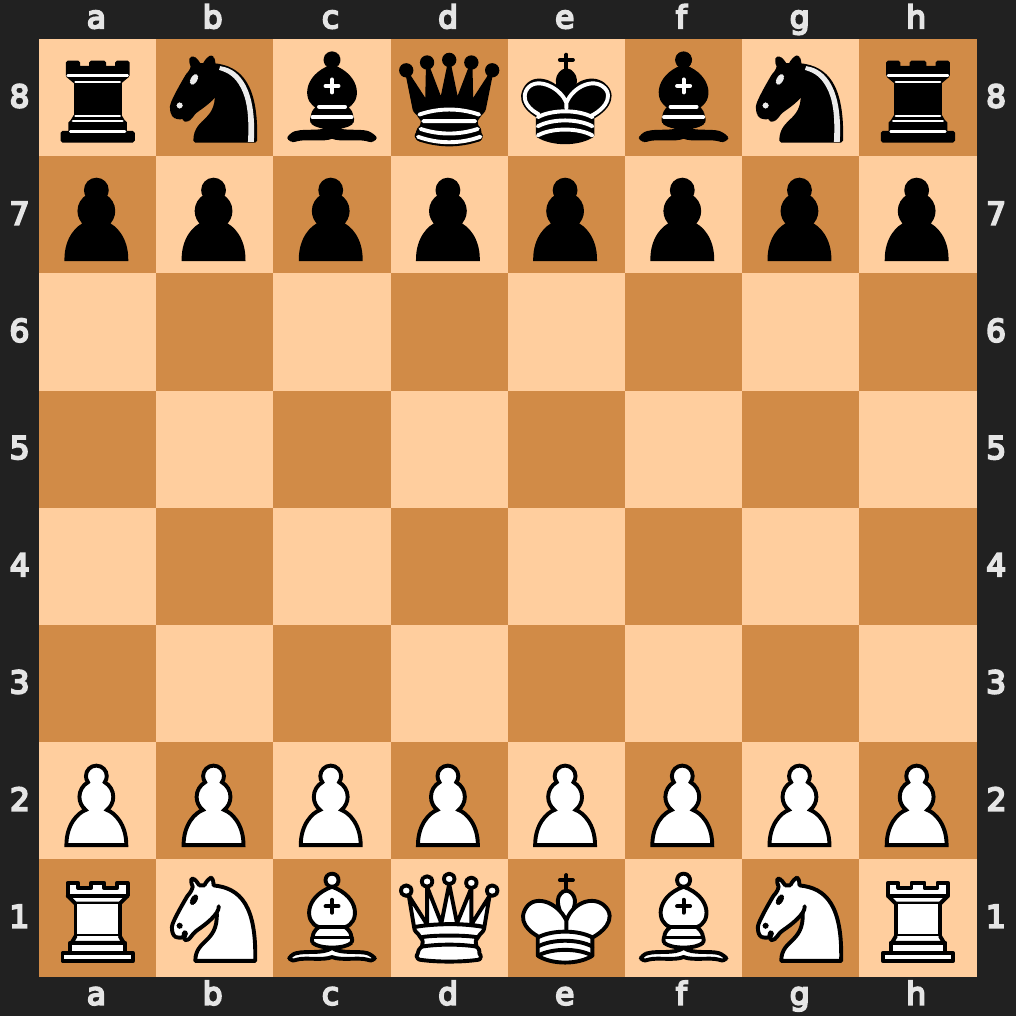} 
  \caption{Sequence start position.}
\end{subfigure}
\hspace{0.14\linewidth}
\begin{subfigure}[t]{.34\textwidth}
  \centering
  \includegraphics[width=1.0\linewidth]{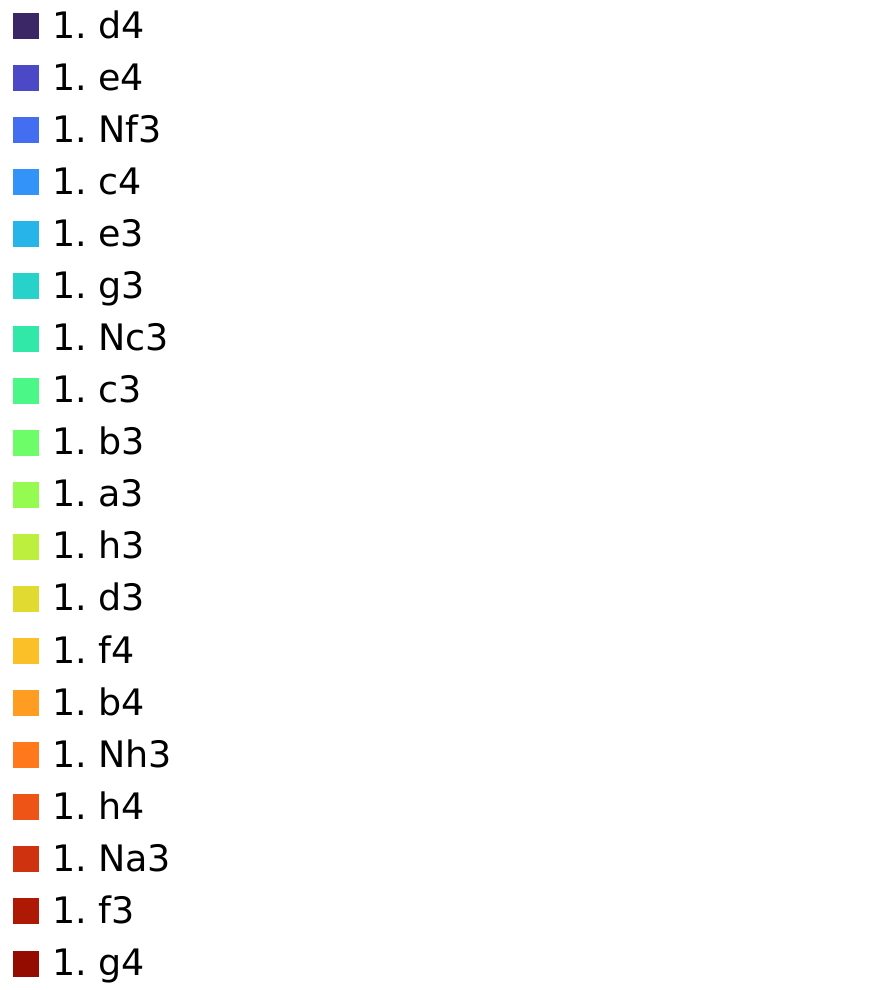}  
  \caption{Legend.}
\end{subfigure}
\vspace{2mm}

\begin{subfigure}{.49\textwidth}
  \centering
  \includegraphics[clip,trim=0 15 0 0,width=\linewidth]{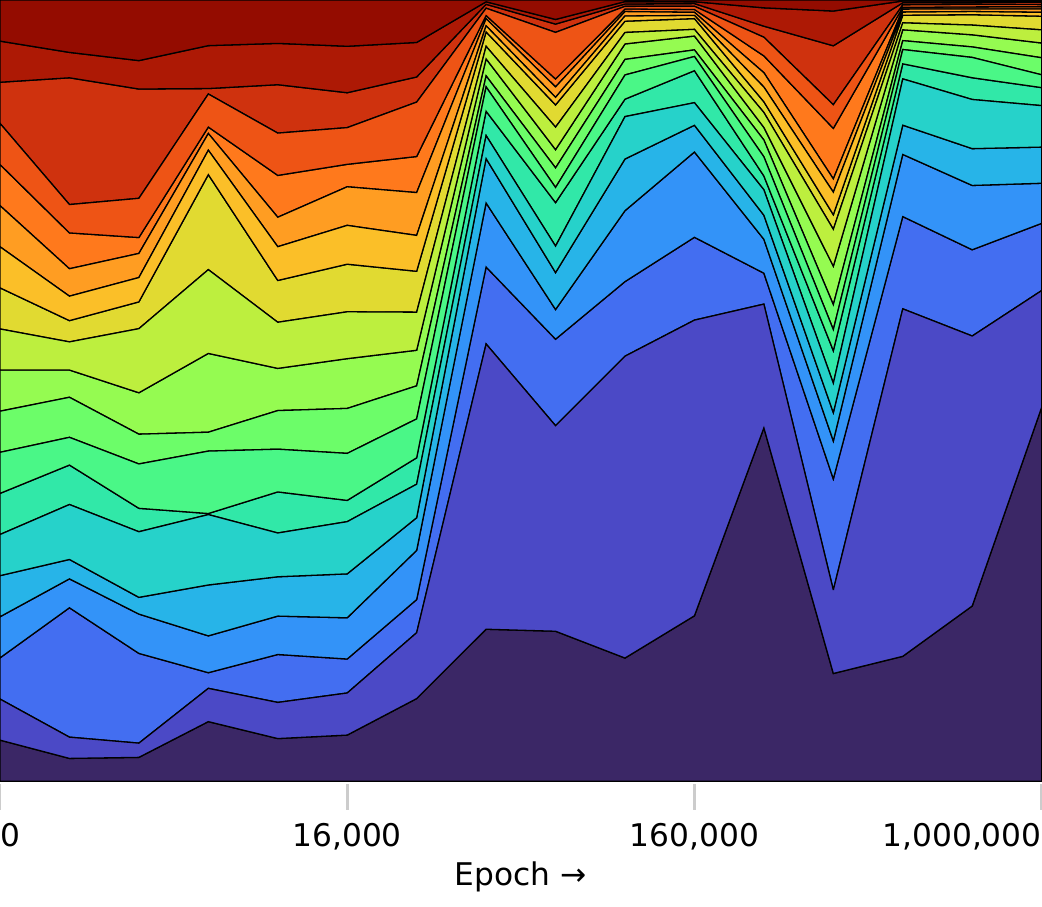} \\ \small AlphaZero training steps \normalsize
  \caption{Training seed 1.}
\end{subfigure}
\hfill
\begin{subfigure}{.49\textwidth}
  \centering
  \includegraphics[clip,trim=0 15 0 0,width=\linewidth]{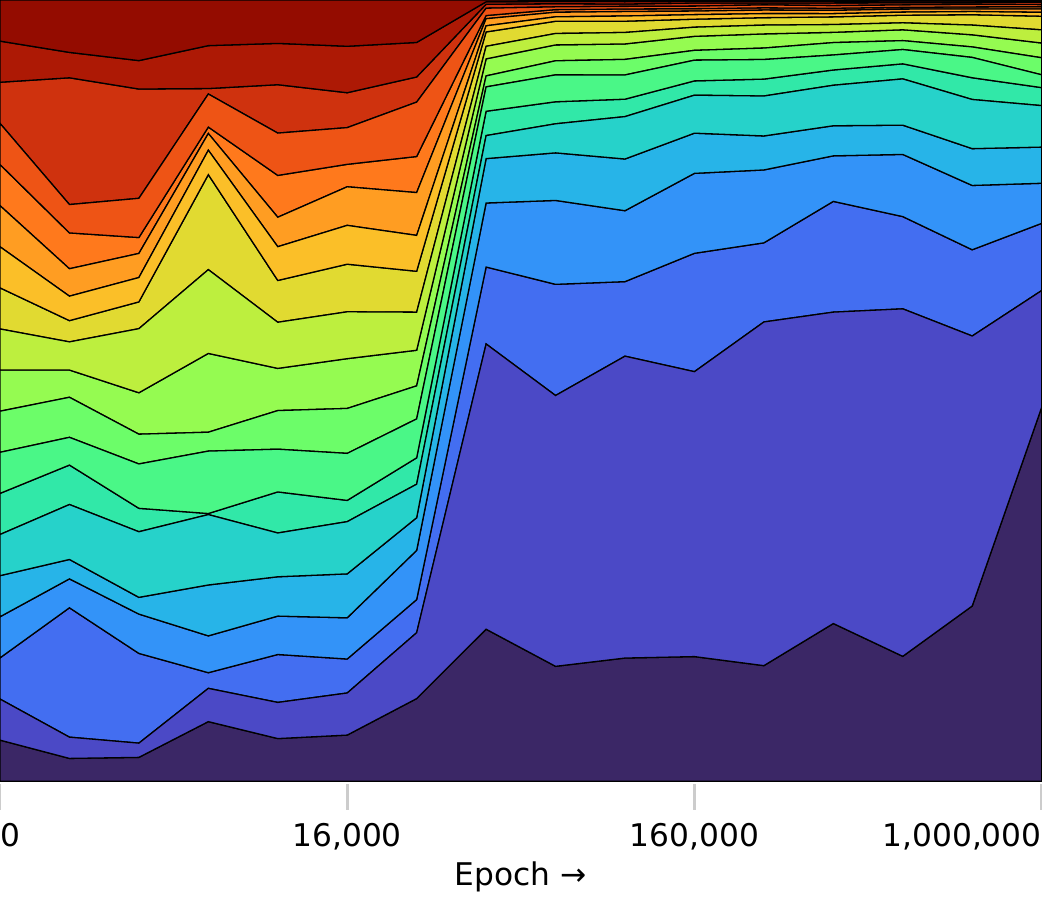} \\ \small AlphaZero training steps \normalsize
  \caption{Training seed 2.}
\end{subfigure}
\vspace{2mm}

\begin{subfigure}{.49\textwidth}
  \centering
  \includegraphics[clip,trim=0 15 0 0,width=\linewidth]{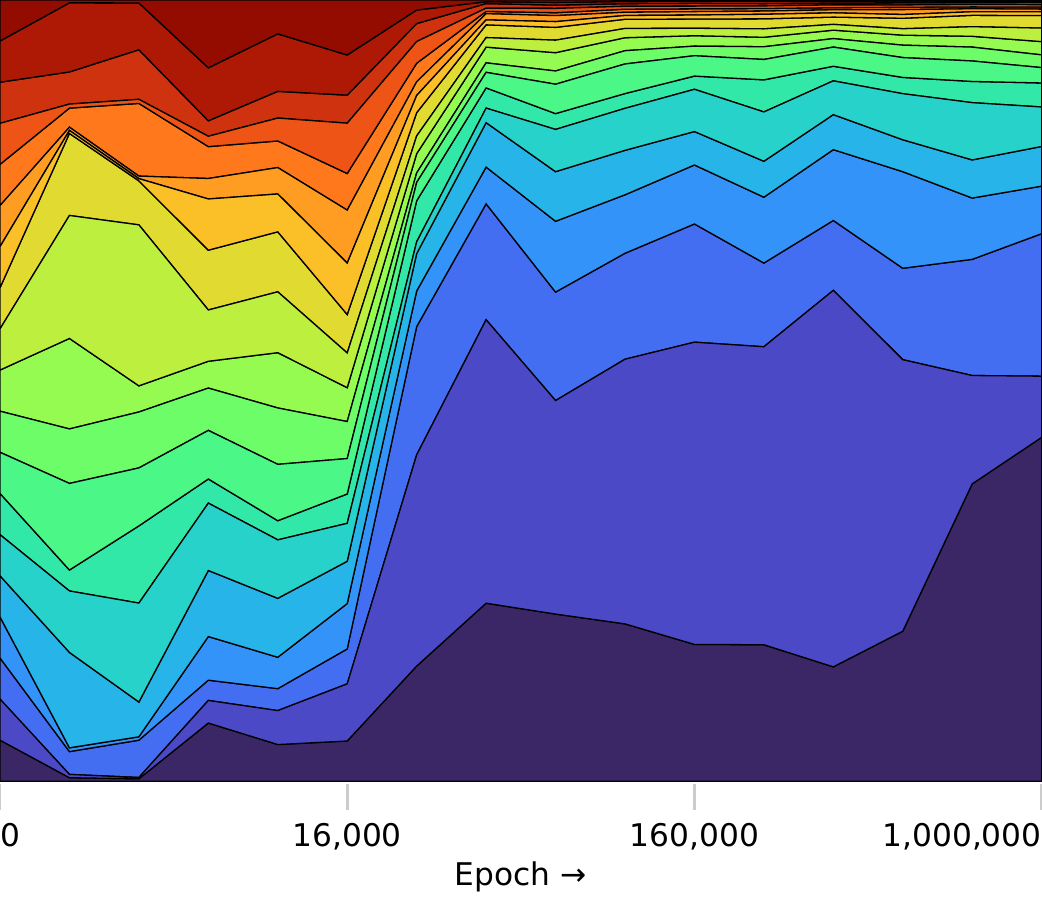} \\ \small AlphaZero training steps \normalsize  
  \caption{Training seed 3.}
\end{subfigure}
\hfill
\begin{subfigure}{.49\textwidth}
  \centering
  \includegraphics[clip,trim=0 15 0 0,width=\linewidth]{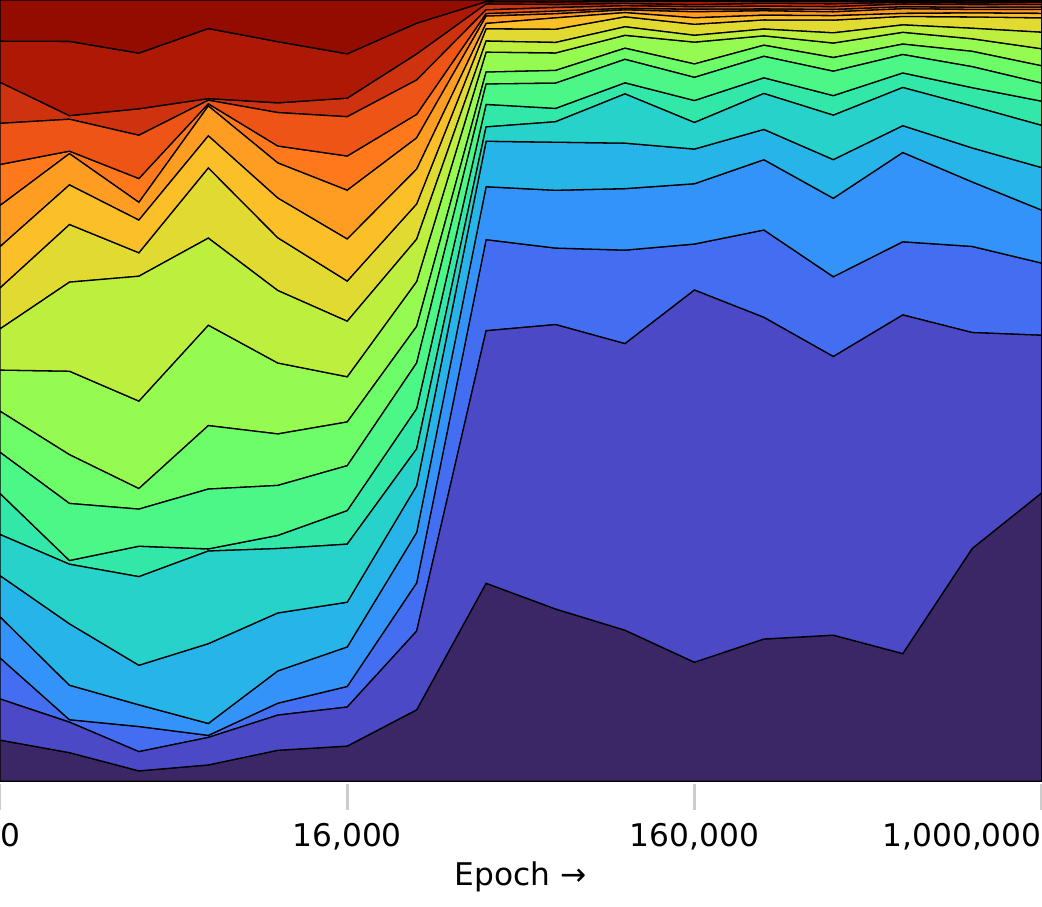} \\ \small AlphaZero training steps \normalsize 
  \caption{Training seed 4.}
\end{subfigure}
\caption{Top 19 moves with the highest probability at the start position.} 
\label{fig:move_seq_06}
\end{figure}

Figures \ref{fig:move_seq_01} to \ref{fig:move_seq_06} show the distribution of move sequences with the highest probabilities as predicted by the policy network. Every figure provides the sequence start position (a) and the probability distribution for different initial training seeds (c)-(f). The $x$ axis shows the training iteration for the given seed. The white area on the plots show the cumulative probability of all move sequences that are not included in (b). Only moves that were made at least once by human players were considered.

\begin{figure}[h!]
\vspace{-10mm}
\centering
\begin{subfigure}[t]{.4\textwidth}
  \centering
  \includegraphics[width=1.0\linewidth]{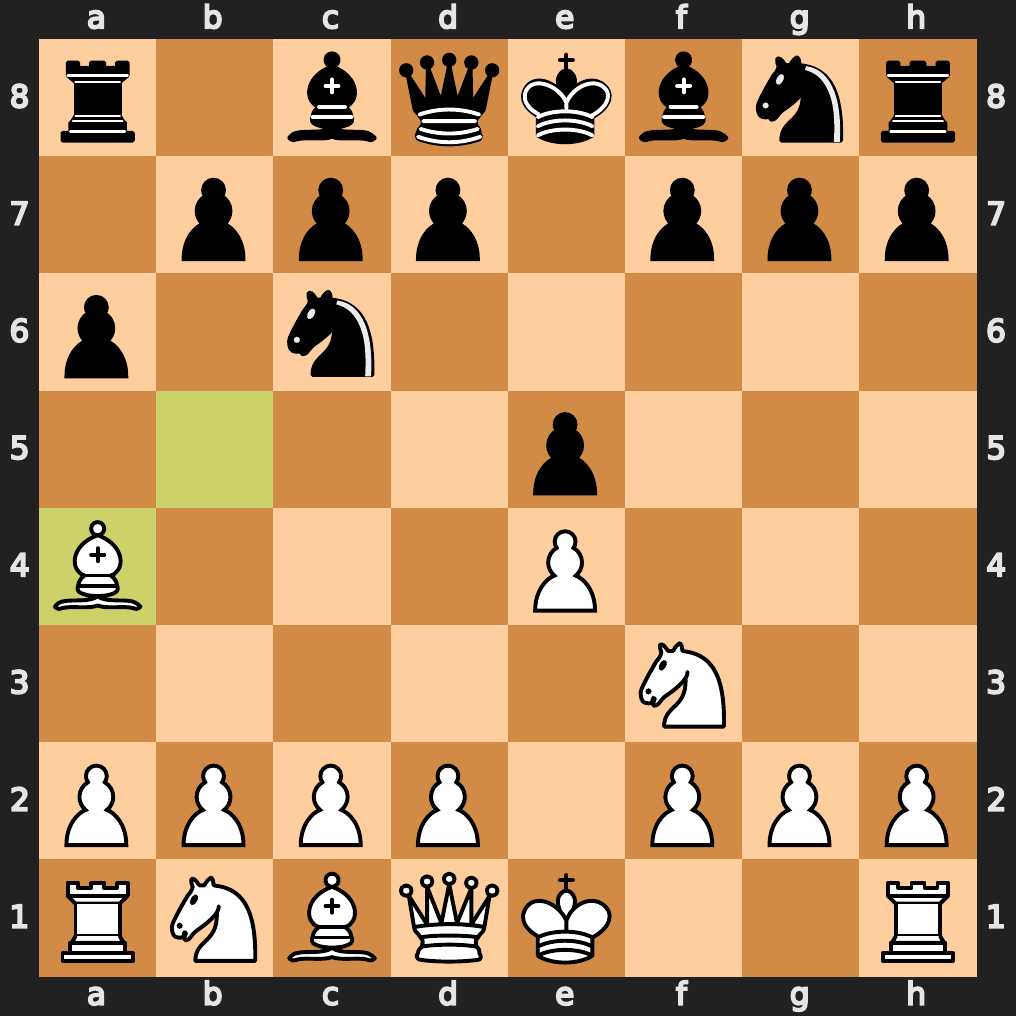}  
  \caption{Sequence start position.}
\end{subfigure}
\hspace{0.14\linewidth}
\begin{subfigure}[t]{.34\textwidth}
  \centering
  \includegraphics[width=1.0\linewidth]{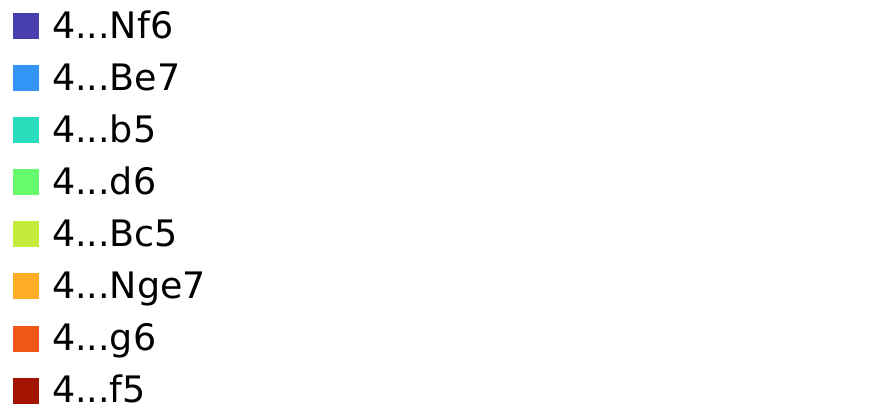}  
  \caption{Legend.}
\end{subfigure}
\vspace{2mm}

\begin{subfigure}{.49\textwidth}
  \centering
  \includegraphics[clip,trim=0 15 0 0,width=\linewidth]{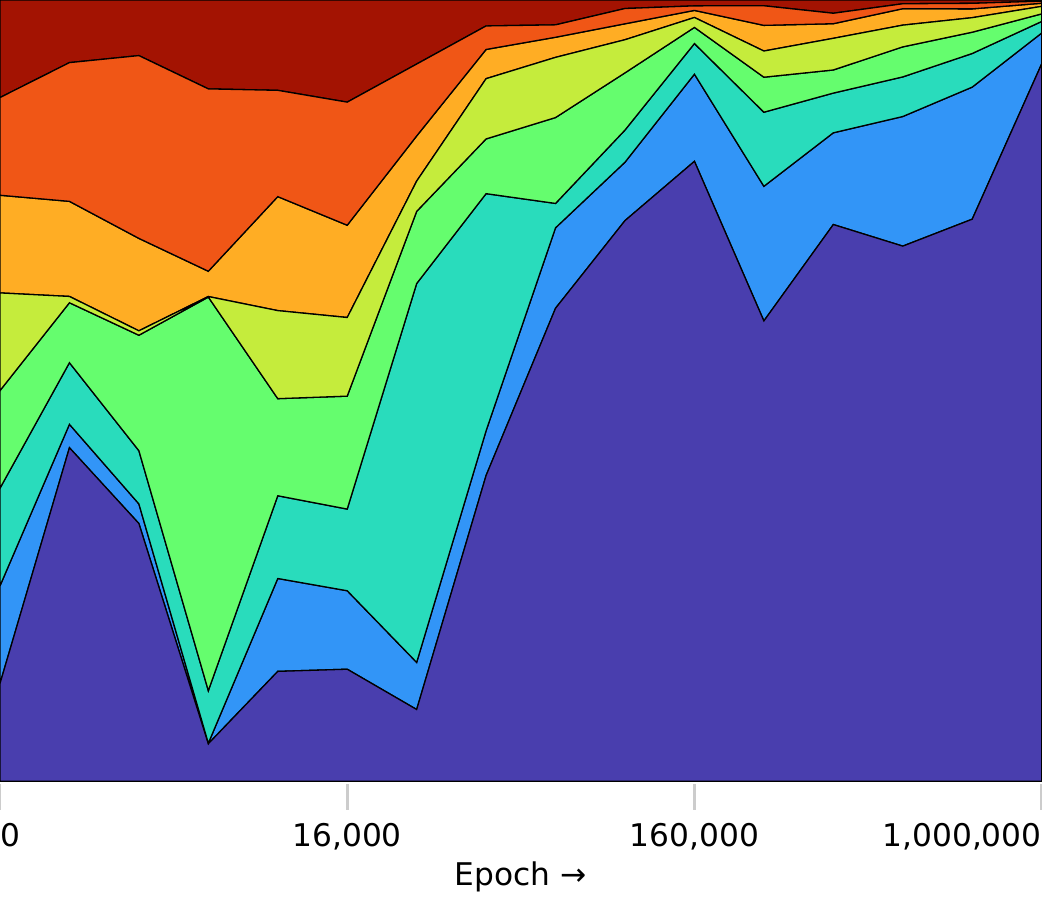} \\
  \small AlphaZero training steps \normalsize
  \caption{Training seed 1.}
\end{subfigure}
\hfill
\begin{subfigure}{.49\textwidth}
  \centering
  \includegraphics[clip,trim=0 15 0 0,width=\linewidth]{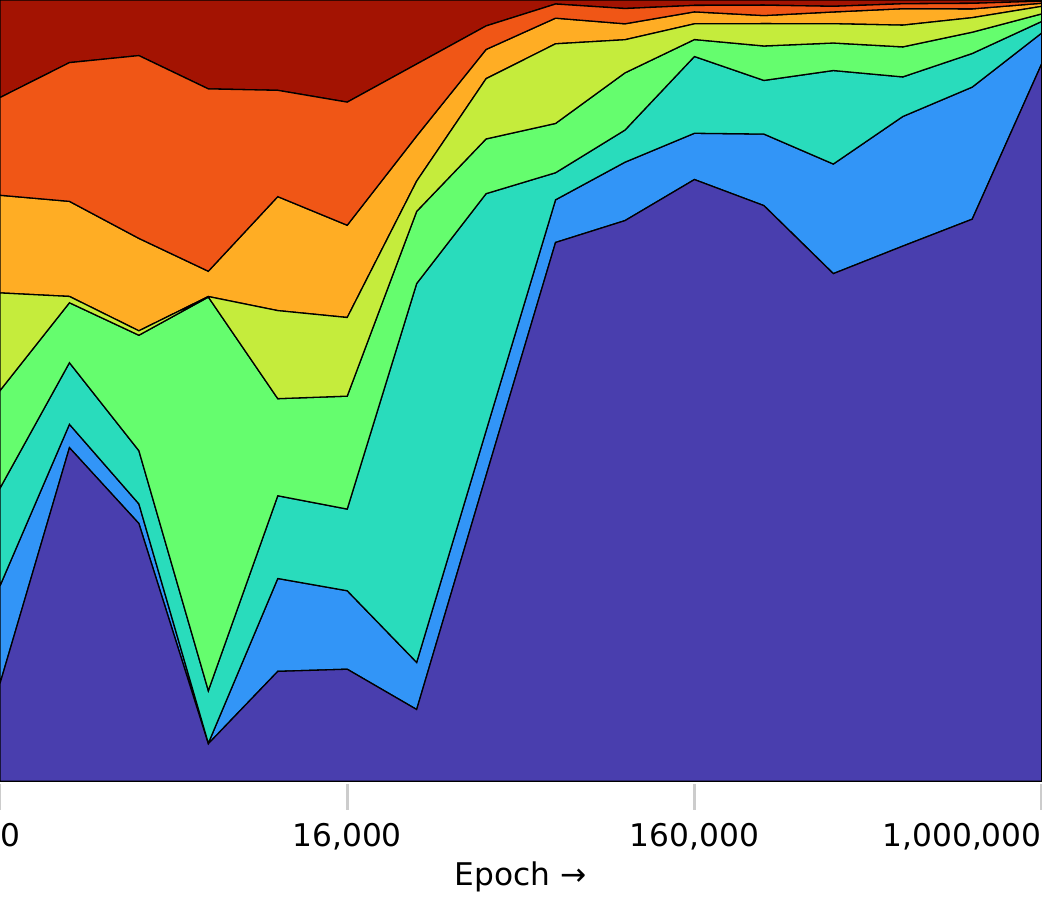} \\
  \small AlphaZero training steps \normalsize
  \caption{Training seed 2.}
\end{subfigure}
\vspace{2mm}

\begin{subfigure}{.49\textwidth}
  \centering
  \includegraphics[clip,trim=0 15 0 0,width=\linewidth]{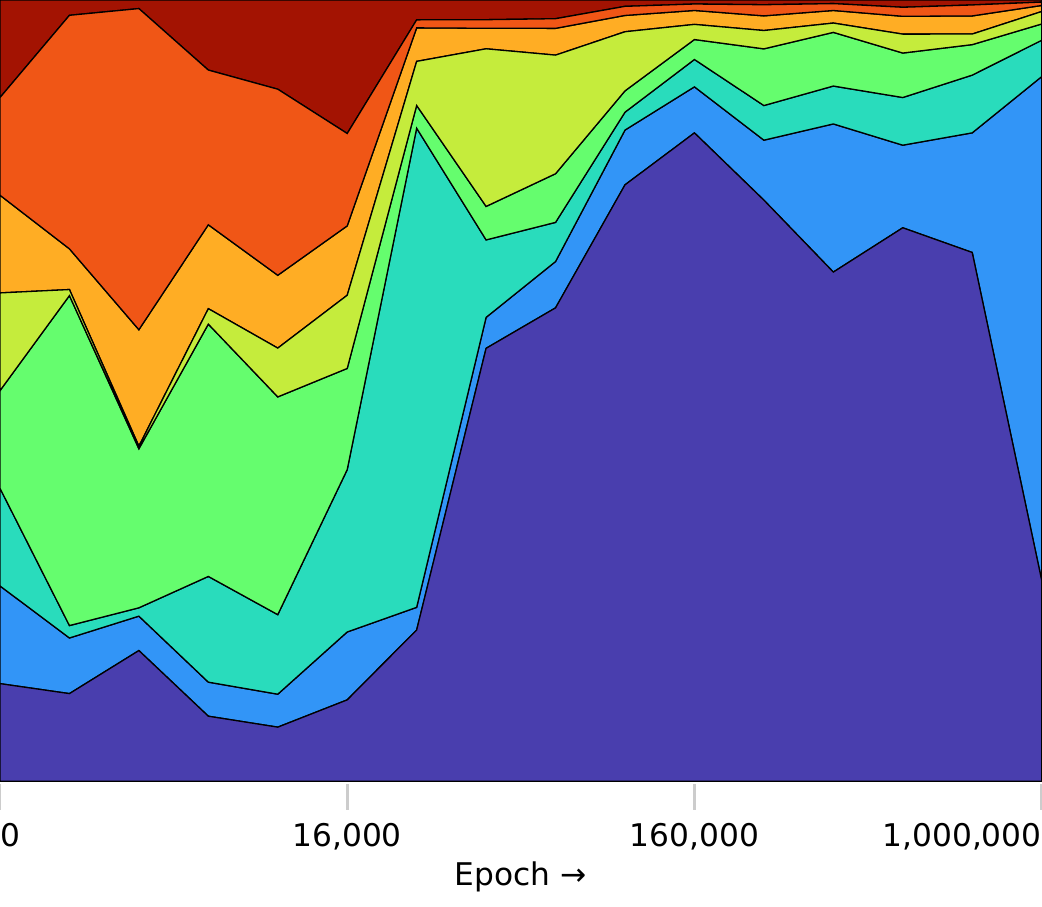} \\
  \small AlphaZero training steps \normalsize
  \caption{Training seed 3.}
\end{subfigure}
\hfill
\begin{subfigure}{.49\textwidth}
  \centering
  \includegraphics[clip,trim=0 15 0 0,width=\linewidth]{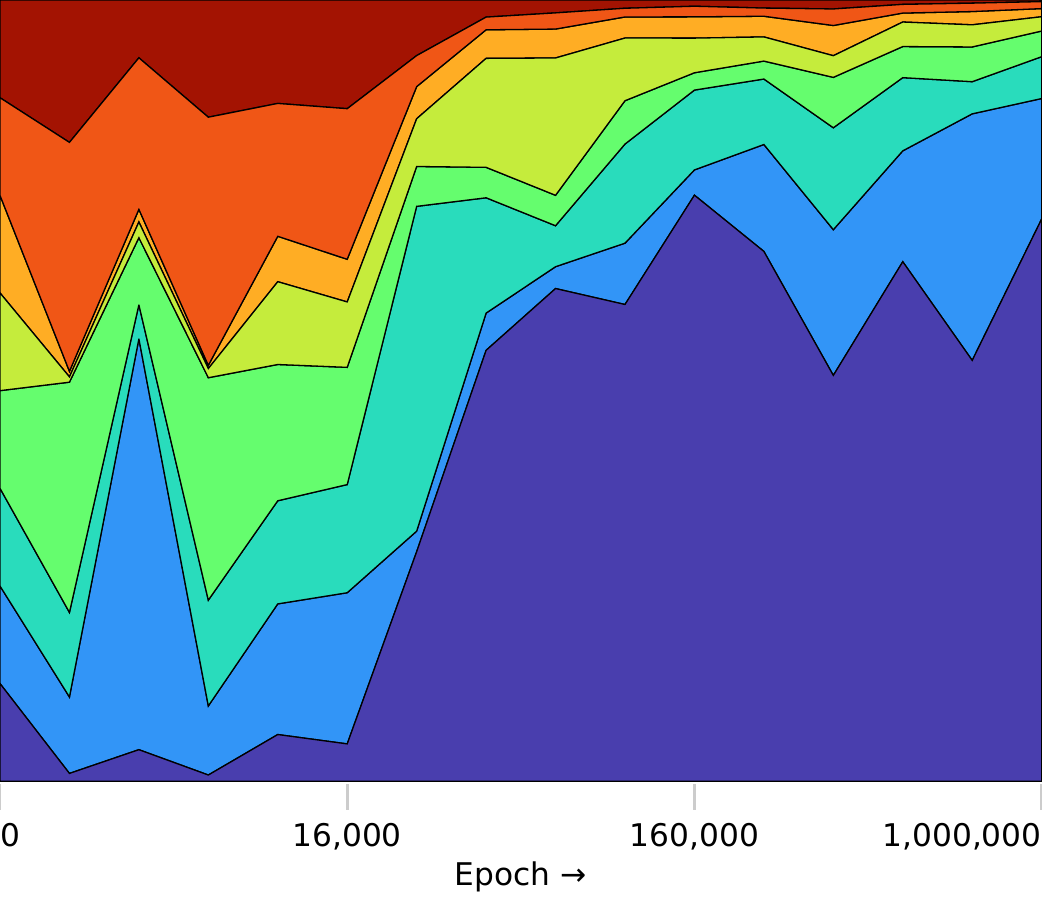} \\
  \small AlphaZero training steps \normalsize
  \caption{Training seed 4.}
\end{subfigure}
\caption{Top 8 moves with the highest probability after 1. e4 e5 2. \symknight f3 \symknight c6 3. \symbishop b5 a6 4. \symbishop a4.}
\end{figure}

\begin{figure}[h!]
\vspace{-10mm}
\centering
\begin{subfigure}[t]{.4\textwidth}
  \centering
  \includegraphics[width=1.0\linewidth]{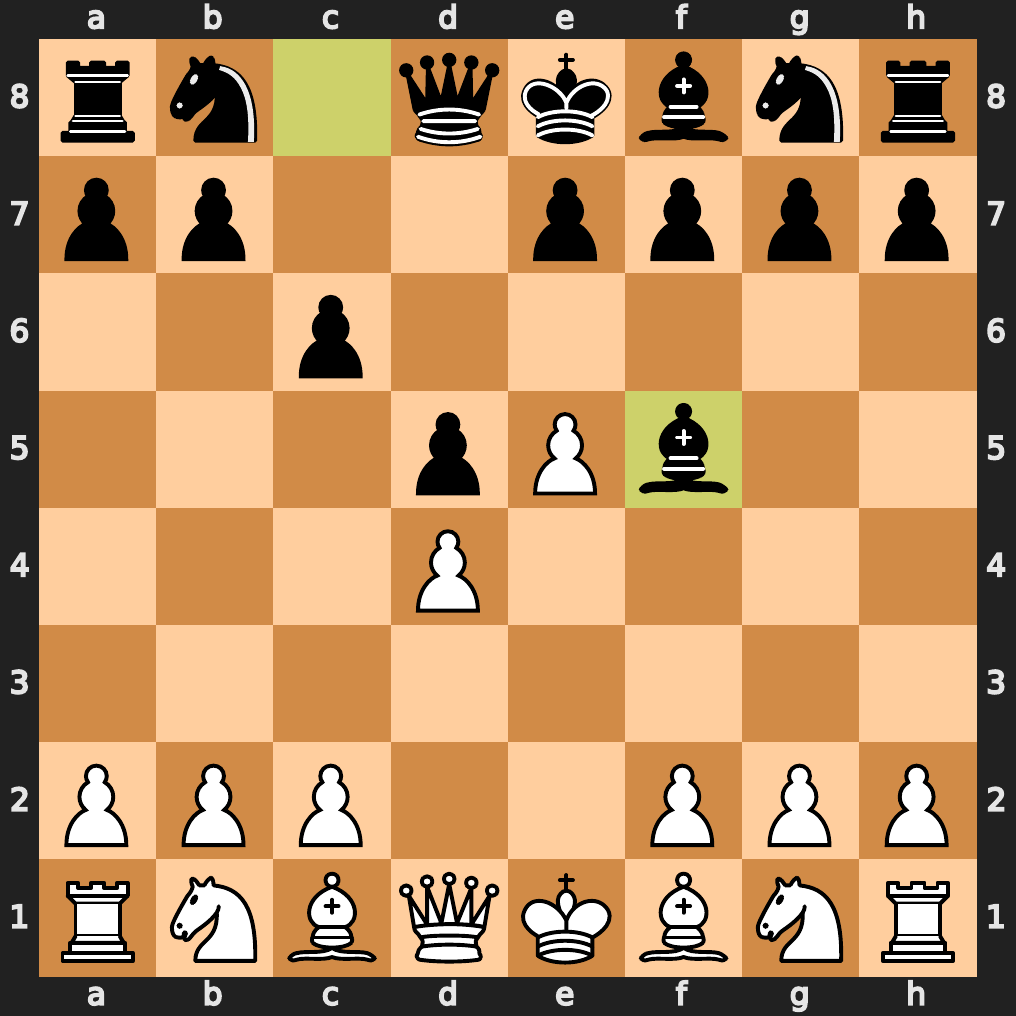}  
  \caption{Sequence start position.}
\end{subfigure}
\hspace{0.14\linewidth}
\begin{subfigure}[t]{.34\textwidth}
  \centering
  \includegraphics[width=1.0\linewidth]{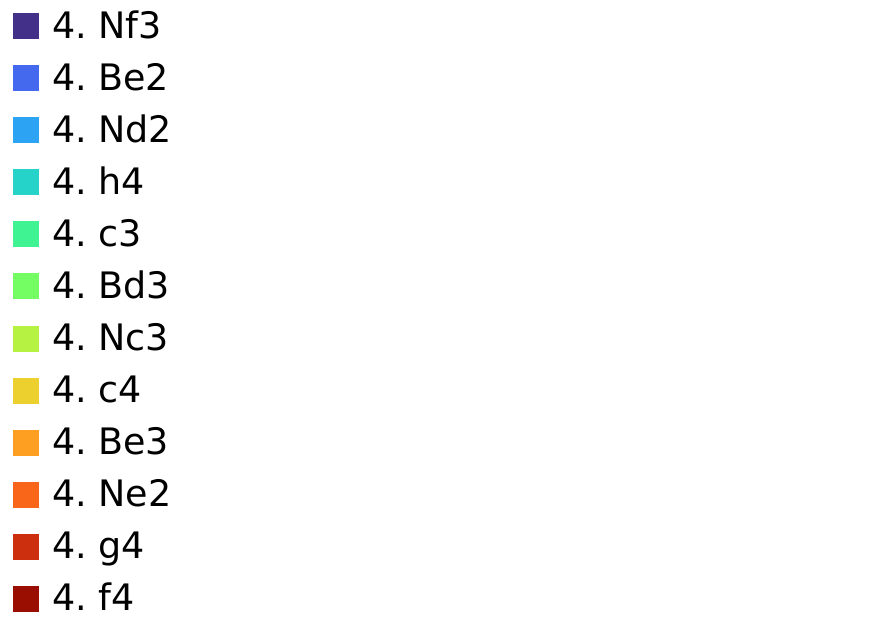}  
  \caption{Legend.}
\end{subfigure}
\vspace{2mm}

\begin{subfigure}{.49\textwidth}
  \centering
  \includegraphics[clip,trim=0 15 0 0,width=\linewidth]{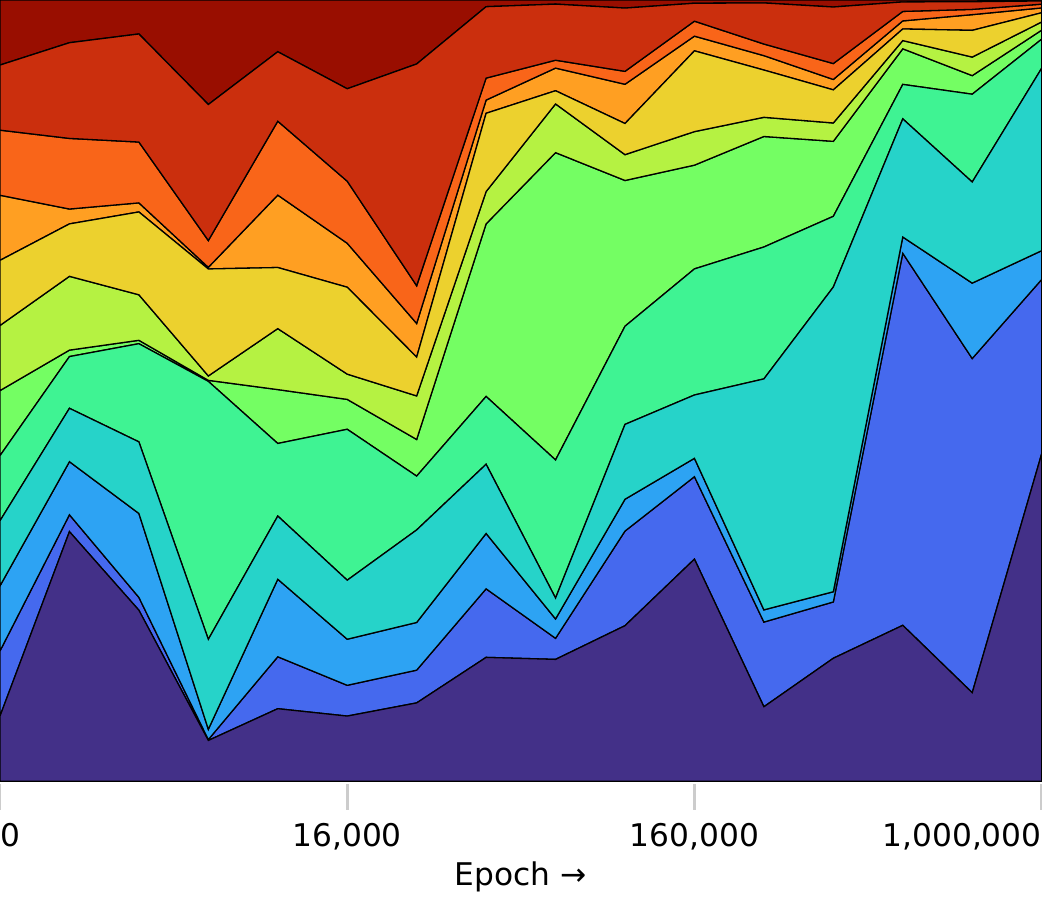} \\
  \small AlphaZero training steps \normalsize
  \caption{Training seed 1.}
\end{subfigure}
\hfill
\begin{subfigure}{.49\textwidth}
  \centering
  \includegraphics[clip,trim=0 15 0 0,width=\linewidth]{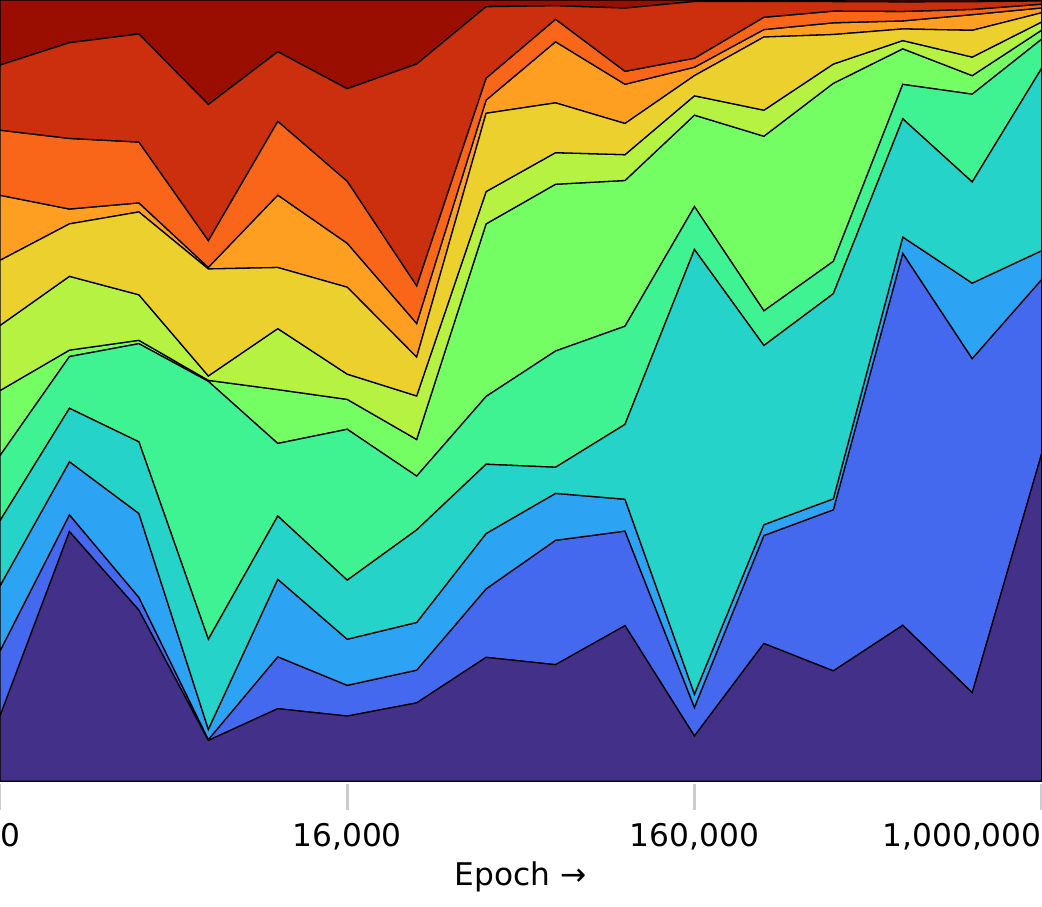} \\
  \small AlphaZero training steps \normalsize
  \caption{Training seed 2.}
\end{subfigure}
\vspace{2mm}

\begin{subfigure}{.49\textwidth}
  \centering
  \includegraphics[clip,trim=0 15 0 0,width=\linewidth]{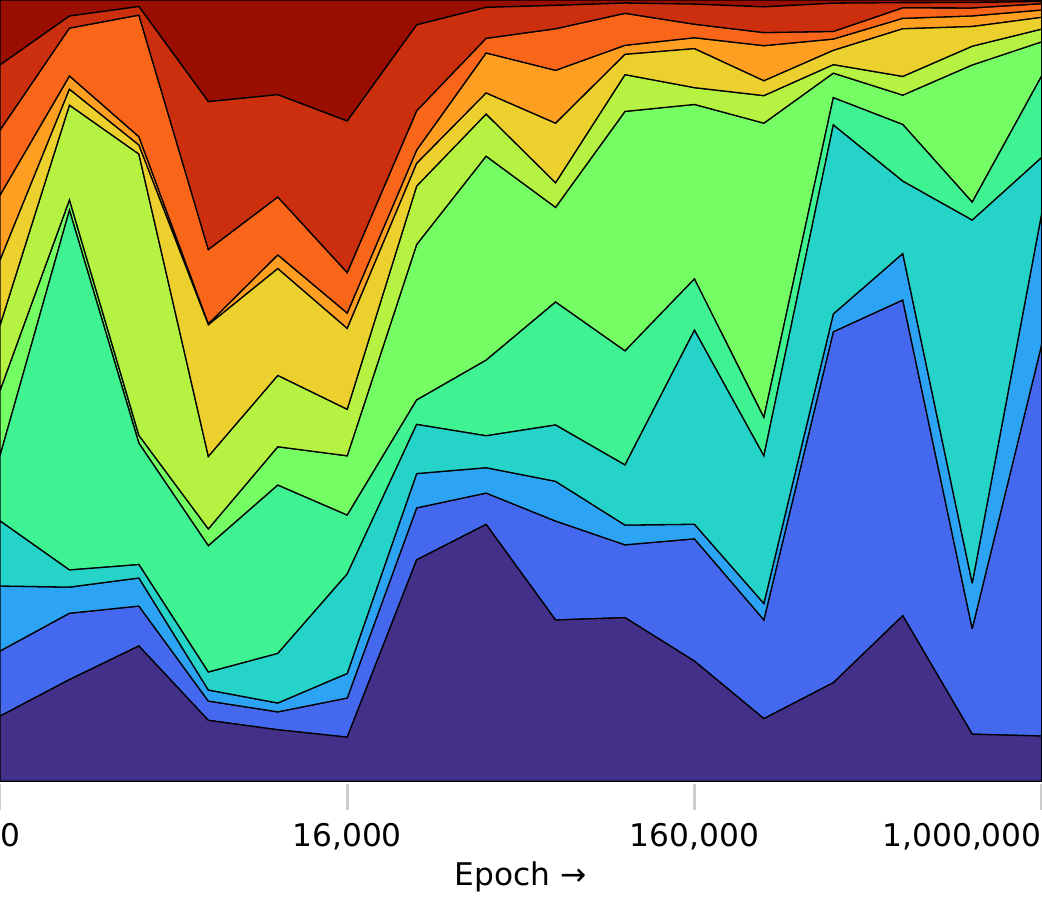} \\
  \small AlphaZero training steps \normalsize
  \caption{Training seed 3.}
\end{subfigure}
\hfill
\begin{subfigure}{.49\textwidth}
  \centering
  \includegraphics[clip,trim=0 15 0 0,width=\linewidth]{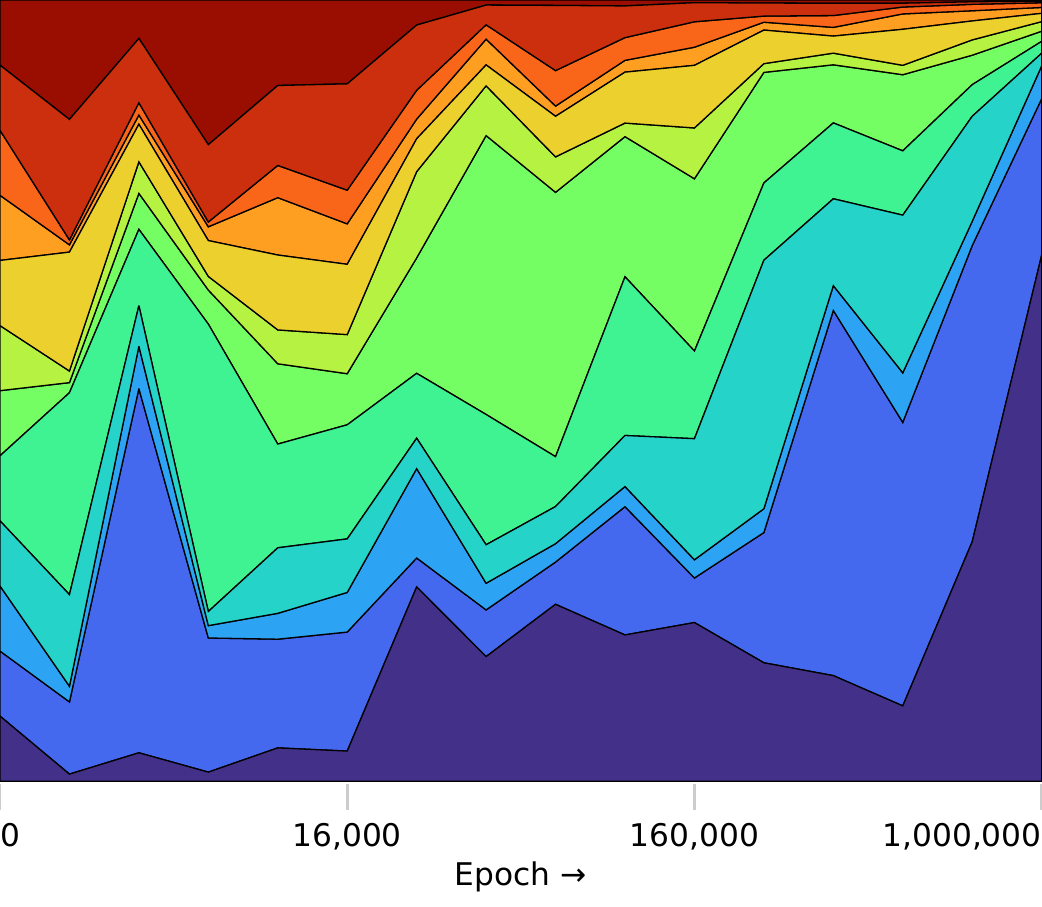} \\
  \small AlphaZero training steps \normalsize 
  \caption{Training seed 4.}
\end{subfigure}
\caption{Top 12 moves with the highest probability after 1. e4 c6 2. d4 d5 3. e5 \symbishop f5.} 
\end{figure}

\begin{figure}[h!]
\vspace{-10mm}
\centering
\begin{subfigure}[t]{.4\textwidth}
  \centering
  \includegraphics[width=1.0\linewidth]{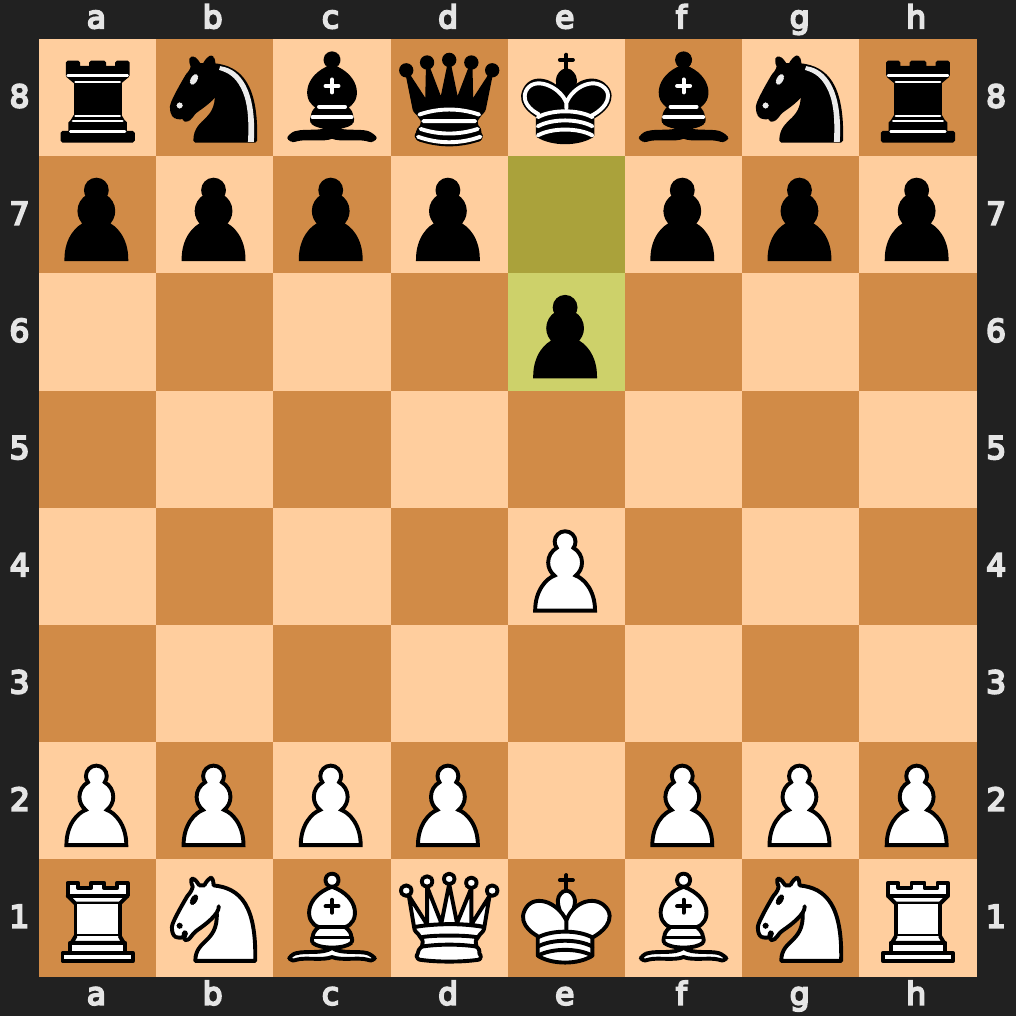}  
  \caption{Sequence start position.}
\end{subfigure}
\hspace{0.14\linewidth}
\begin{subfigure}[t]{.34\textwidth}
  \centering
  \includegraphics[width=1.0\linewidth]{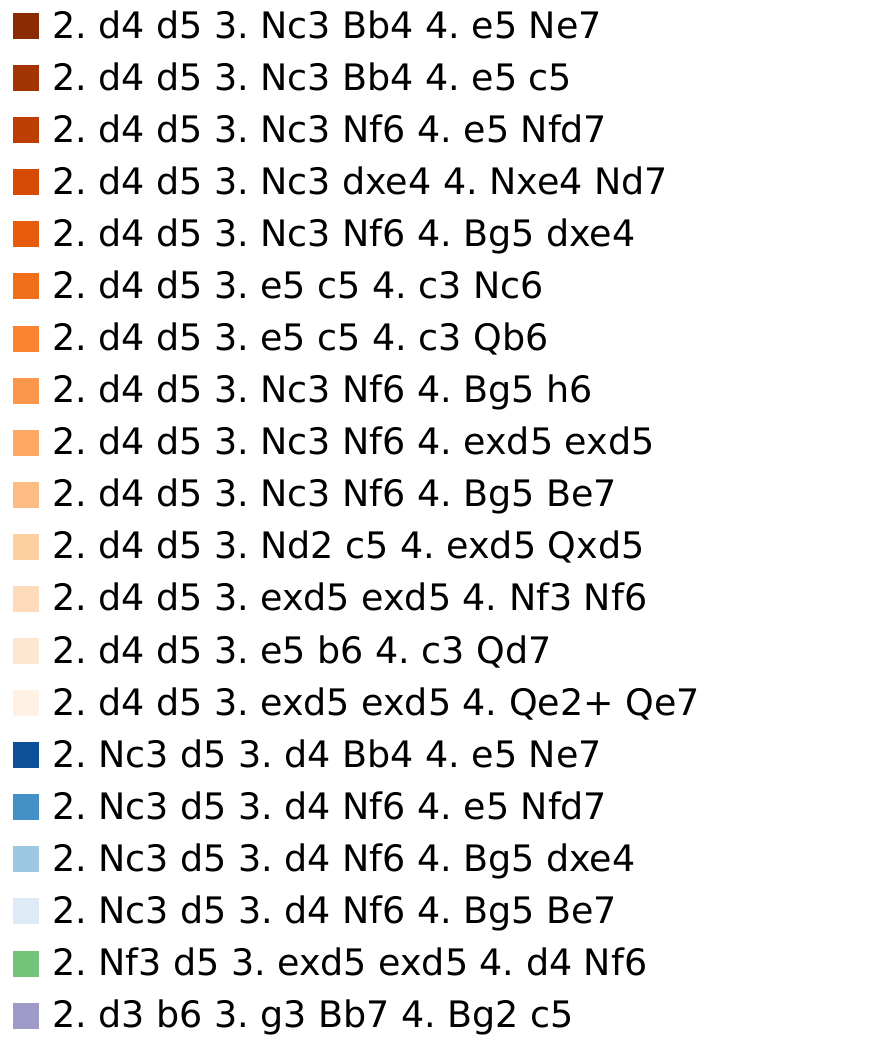}  
  \caption{Legend.}
\end{subfigure}
\vspace{2mm}

\begin{subfigure}{.49\textwidth}
  \centering
  \includegraphics[clip,trim=0 15 0 0,width=\linewidth]{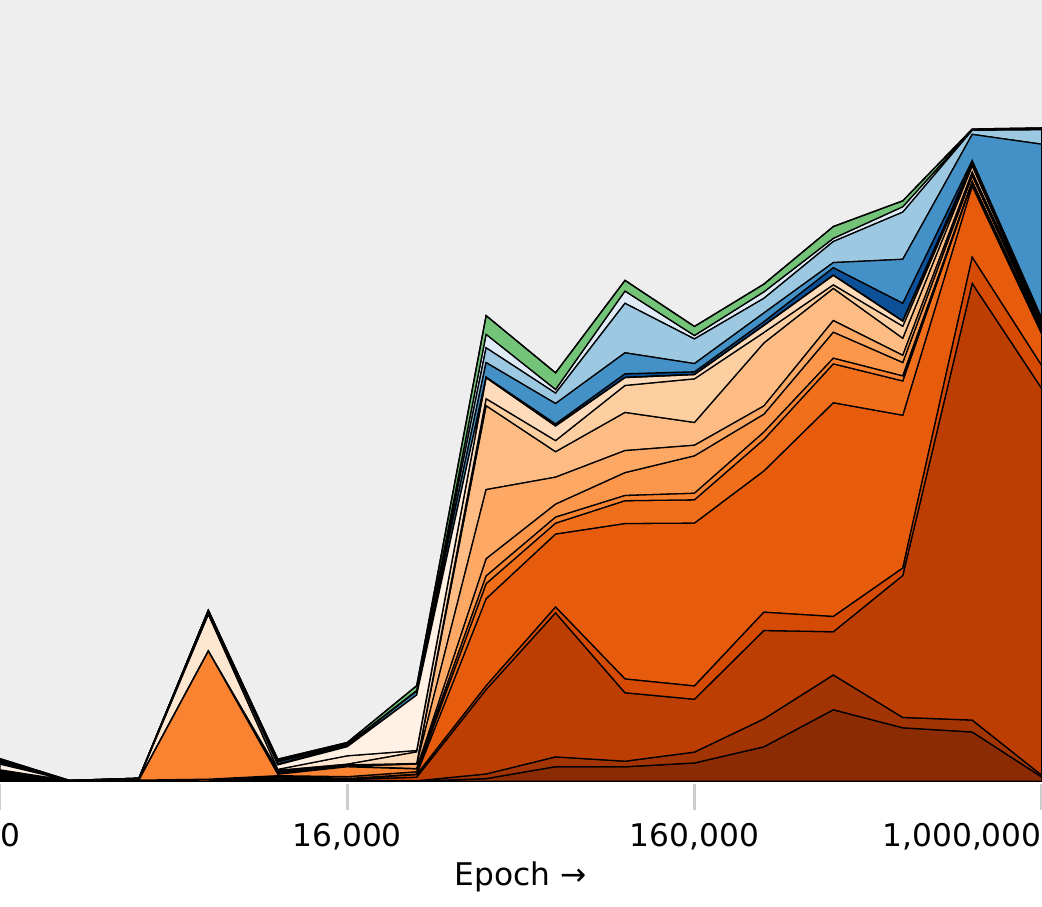} \\
  \small AlphaZero training steps \normalsize 
  \caption{Training seed 1.}
\end{subfigure}
\hfill
\begin{subfigure}{.49\textwidth}
  \centering
  \includegraphics[clip,trim=0 15 0 0,width=\linewidth]{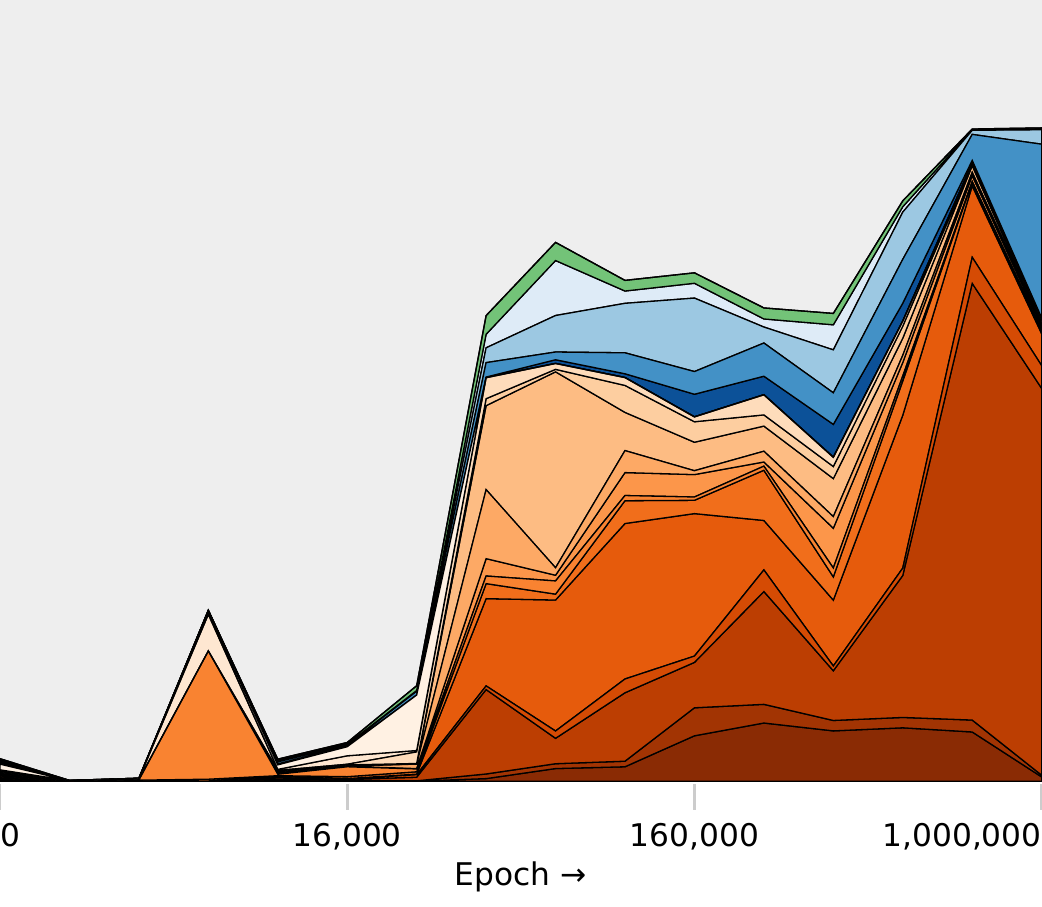} \\
  \small AlphaZero training steps \normalsize
  \caption{Training seed 2.}
\end{subfigure}
\vspace{2mm}

\begin{subfigure}{.49\textwidth}
  \centering
  \includegraphics[clip,trim=0 15 0 0,width=\linewidth]{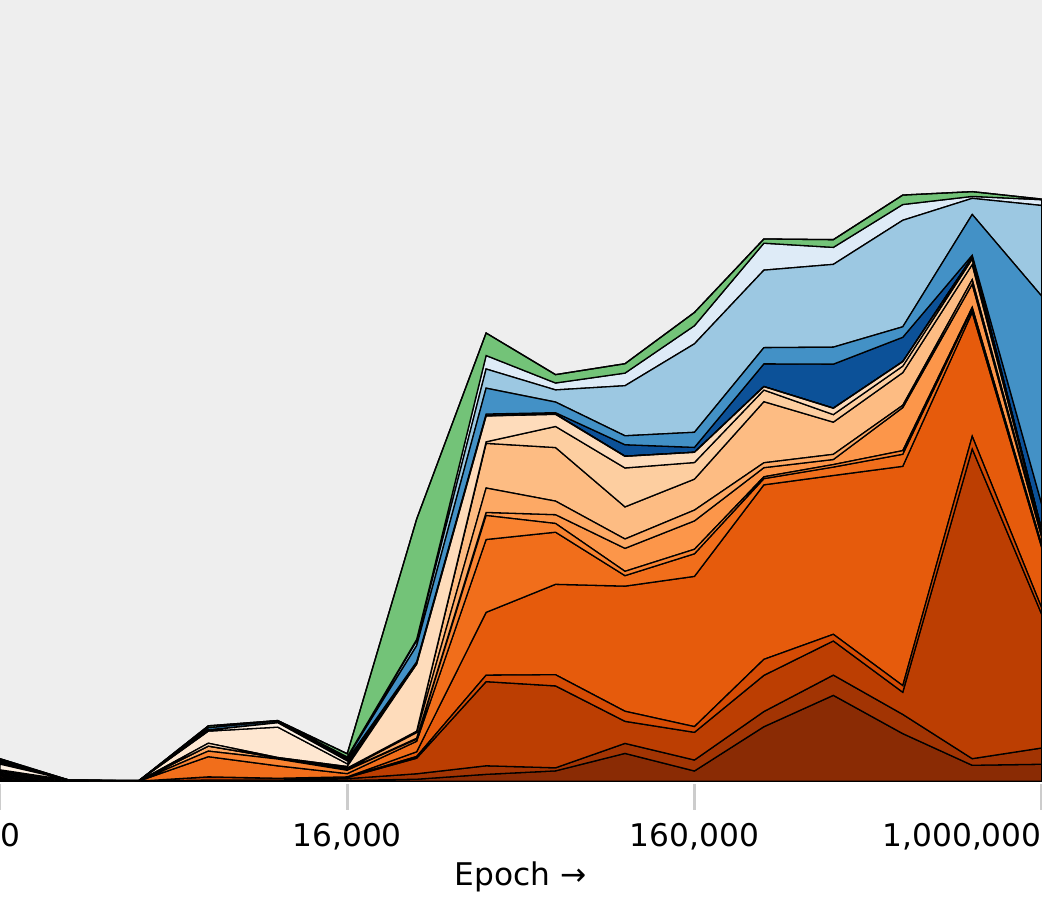} \\
  \small AlphaZero training steps \normalsize  
  \caption{Training seed 3.}
\end{subfigure}
\hfill
\begin{subfigure}{.49\textwidth}
  \centering
  \includegraphics[clip,trim=0 15 0 0,width=\linewidth]{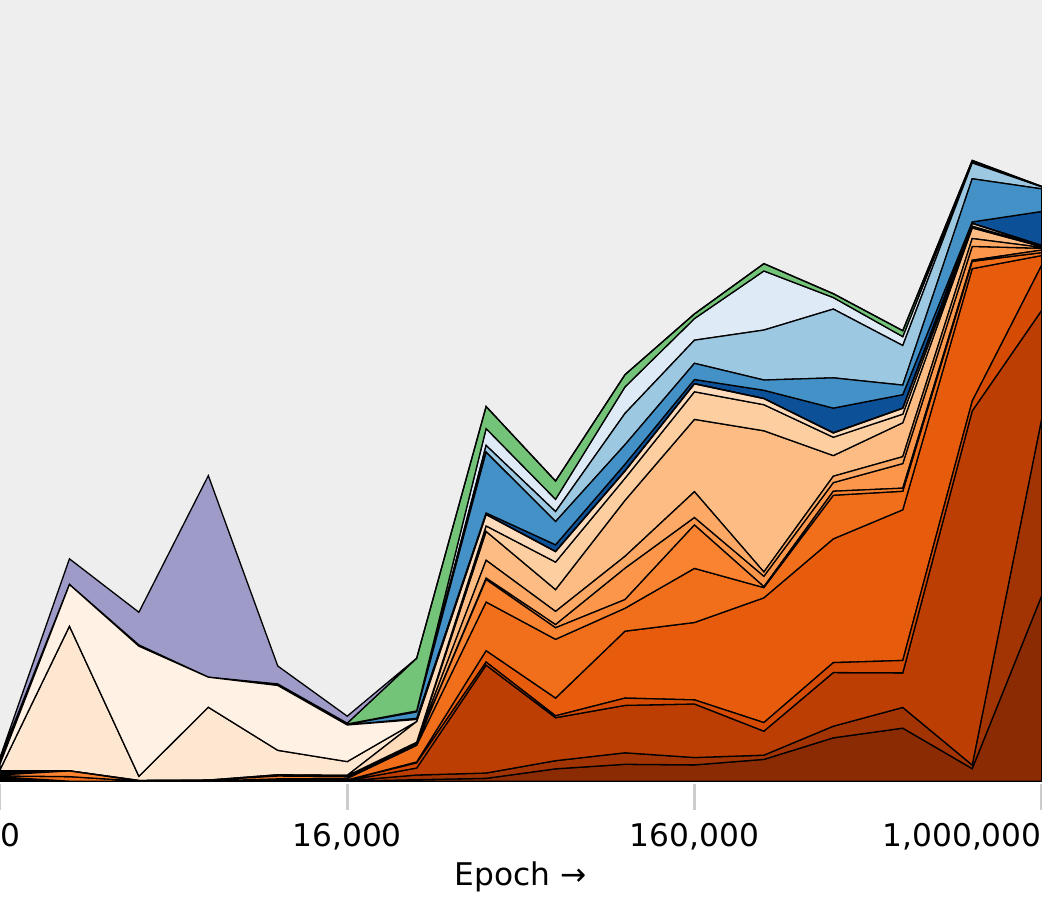} \\
  \small AlphaZero training steps \normalsize  
  \caption{Training seed 4.}
\end{subfigure}
\caption{Top 20 move sequences with the highest joint probability after 1. e4 e6.} 
\end{figure}

\FloatBarrier

\section{Outlier positions in score regression}

This section shows all the negative outlier positions shown in Figure~\ref{fig:residuals} (those highlighted in red with true value less than zero). Black's queen can be captured in every position.

\begin{figure}[h]
    \centering
    \includegraphics[width=0.95\textwidth]{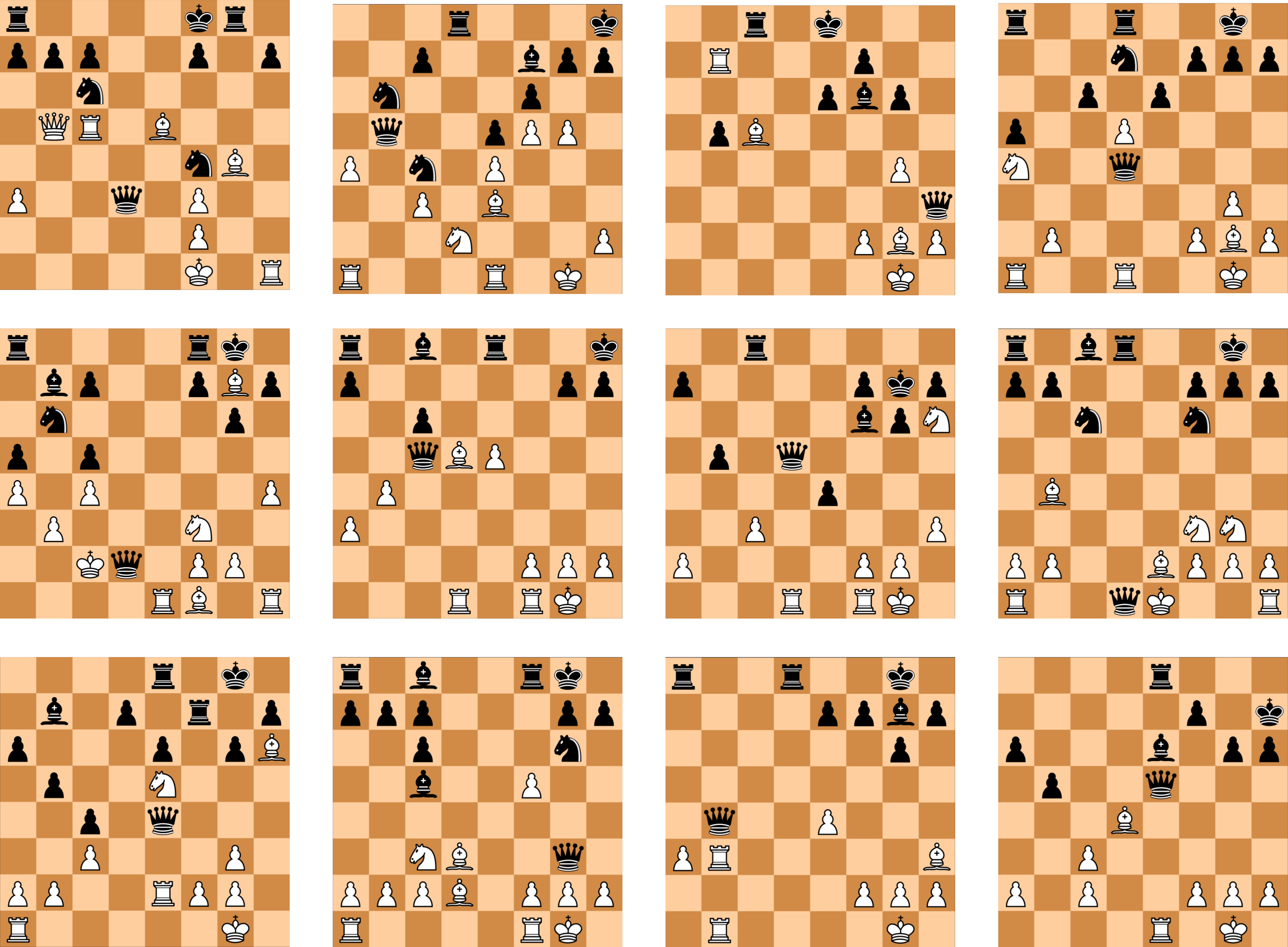}
    \caption{All negative outliers shown in Figure~\ref{fig:residuals}. Black's queen can be taken in every position.}
    \label{fig:supp_residuals}
\end{figure}

\FloatBarrier

\newpage

\section{Concept regression results for second seed}\label{sec:concept_regression_second_seed}

In order to ensure that our results for concept regression generalise beyond a single training run, we applied our concept regression methodology to a second trainingrun of AlphaZero. Our initial analysis was conducted on seed 3, and the following regression plots are obtained from seed 1.
They plots match those in
Figures \ref{fig:all_sf_concepts_1}
to \ref{fig:custom_pawn_concepts_2} in
Appendix \ref{sec:all_concepts_regression_supplement}.

\begin{figure}[h]
    \centering
    \includegraphics[width=\textwidth]{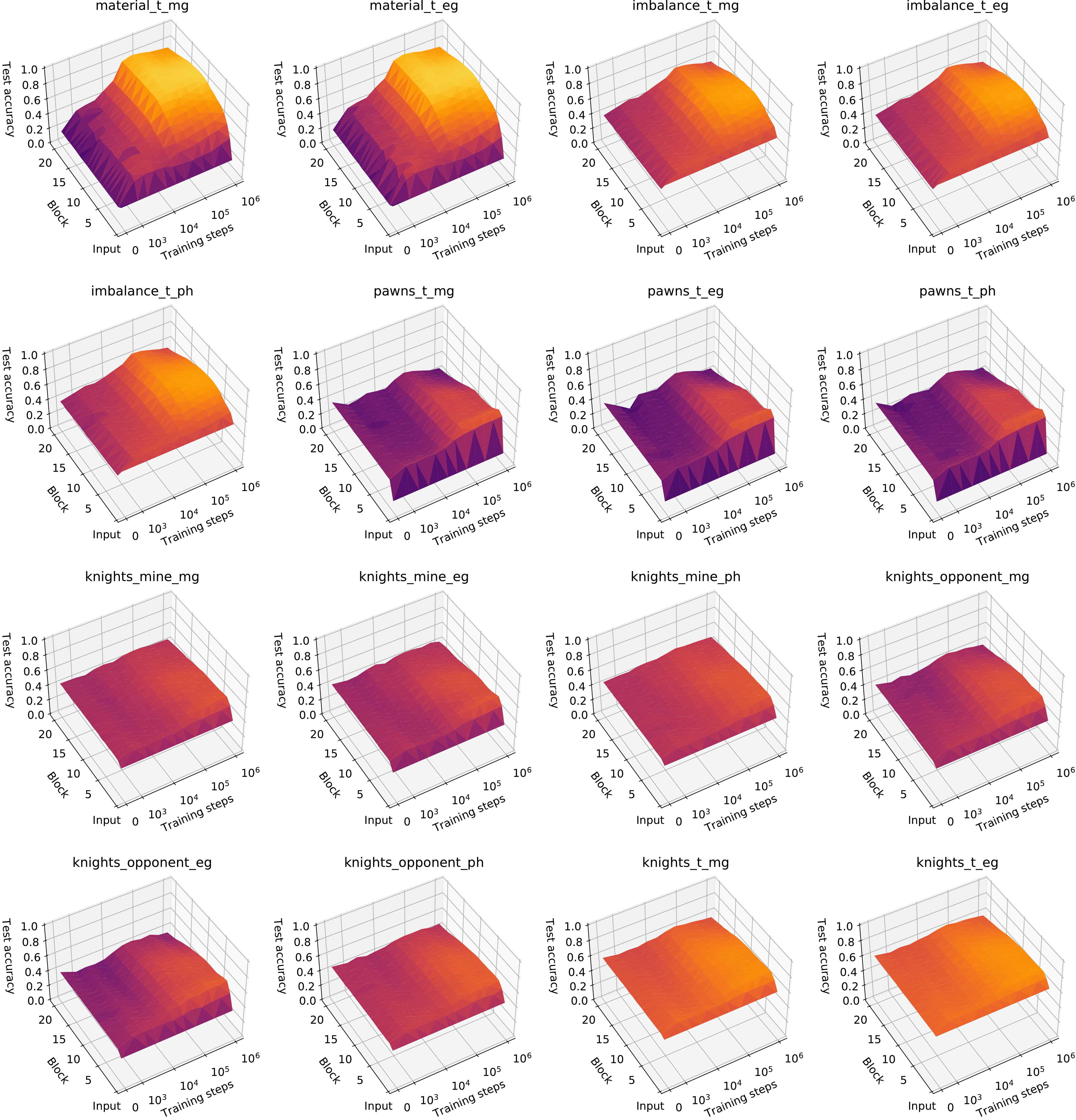}
    \caption{Regression results for Stockfish concepts from Table~\ref{tab:concepts}.}
    \label{fig:all_sf_concepts_1_seed2}
\end{figure}

\begin{figure}
    \centering
    \includegraphics[width=\textwidth]{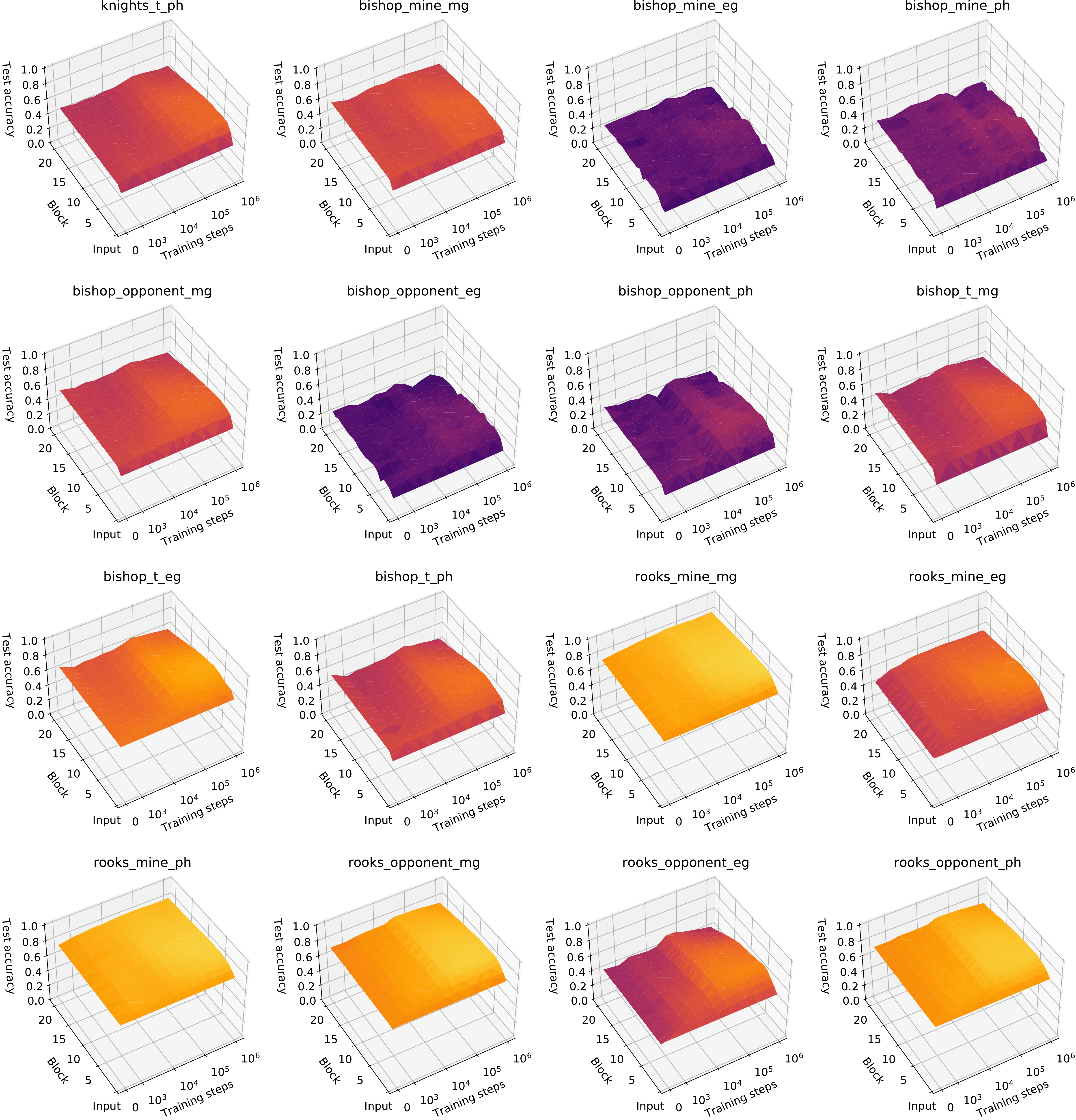}
    \caption{Regression results for Stockfish concepts from Table~\ref{tab:concepts}, continued.}
    \label{fig:all_sf_concepts_2_seed2}
\end{figure}

\begin{figure}
    \centering
    \includegraphics[width=\textwidth]{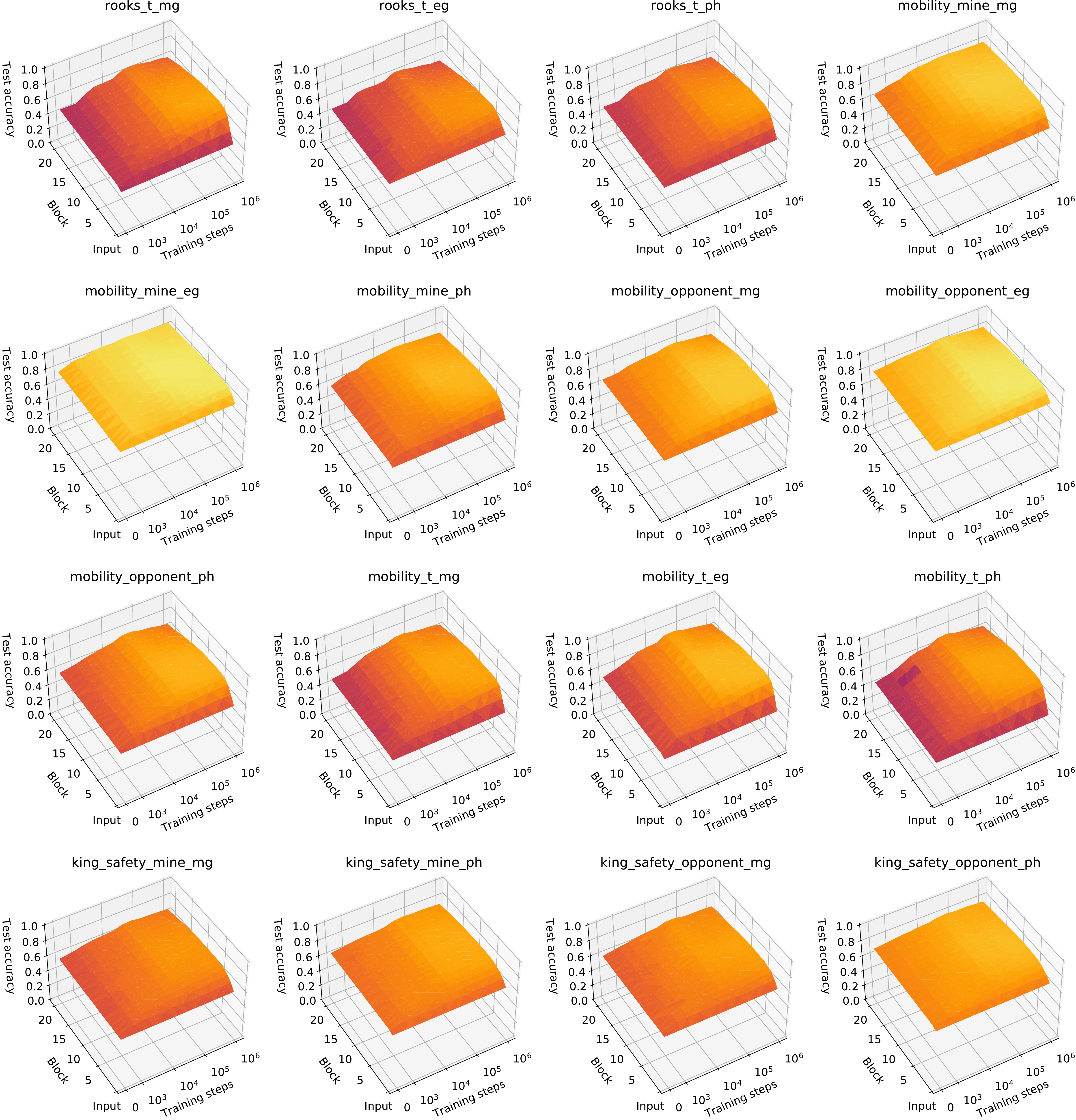}
    \caption{Regression results for Stockfish concepts from Table~\ref{tab:concepts}, continued.}
    \label{fig:all_sf_concepts_3_seed2}
\end{figure}

\begin{figure}
    \centering
    \includegraphics[width=\textwidth]{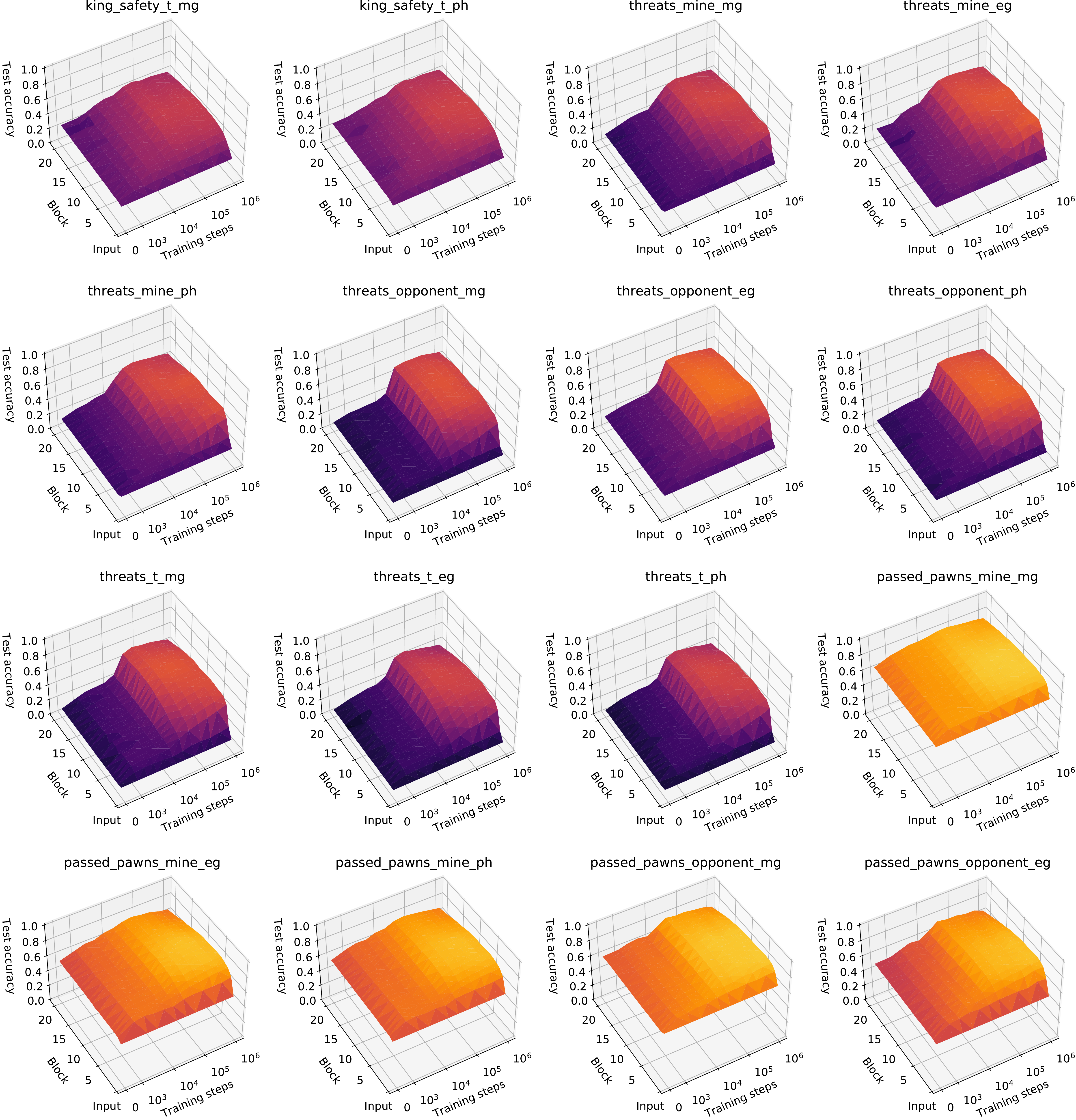}
    \caption{Regression results for Stockfish concepts from Table~\ref{tab:concepts}, continued.}
    \label{fig:all_sf_concepts_4_seed2}
\end{figure}

\begin{figure}
    \centering
    \includegraphics[width=\textwidth]{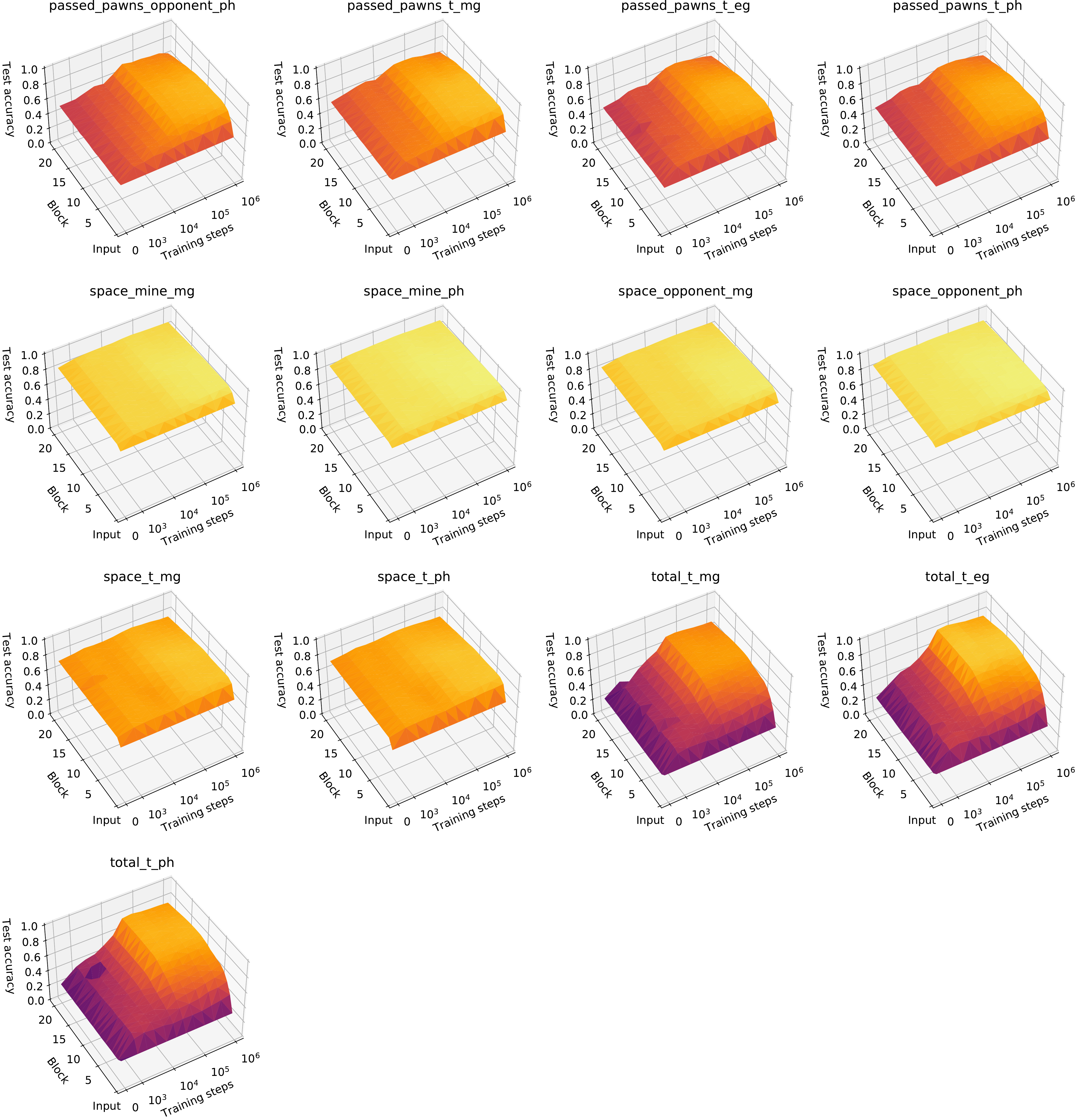}
    \caption{Regression results for Stockfish concepts from Table~\ref{tab:concepts}, continued.}
    \label{fig:all_sf_concepts_5_seed2}
\end{figure}

\begin{figure}
    \centering
    \includegraphics[width=\textwidth]{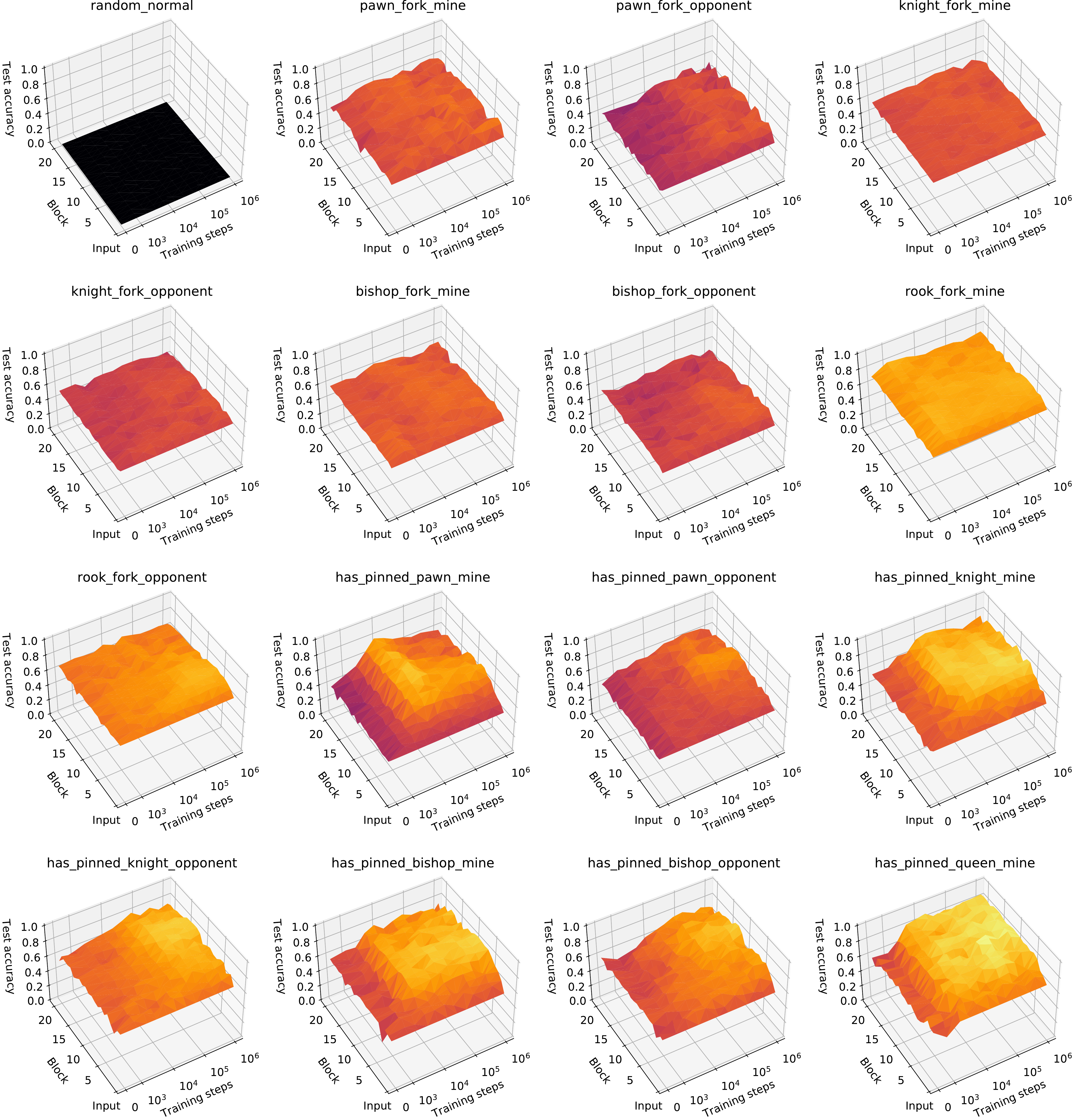}
    \caption{Regression results for custom concepts from Table~\ref{tab:concepts-custom}, excluding capture-related concepts.}
    \label{fig:custom_concepts_nocapture_1_seed2}
\end{figure}

\begin{figure}
    \centering
    \includegraphics[width=\textwidth]{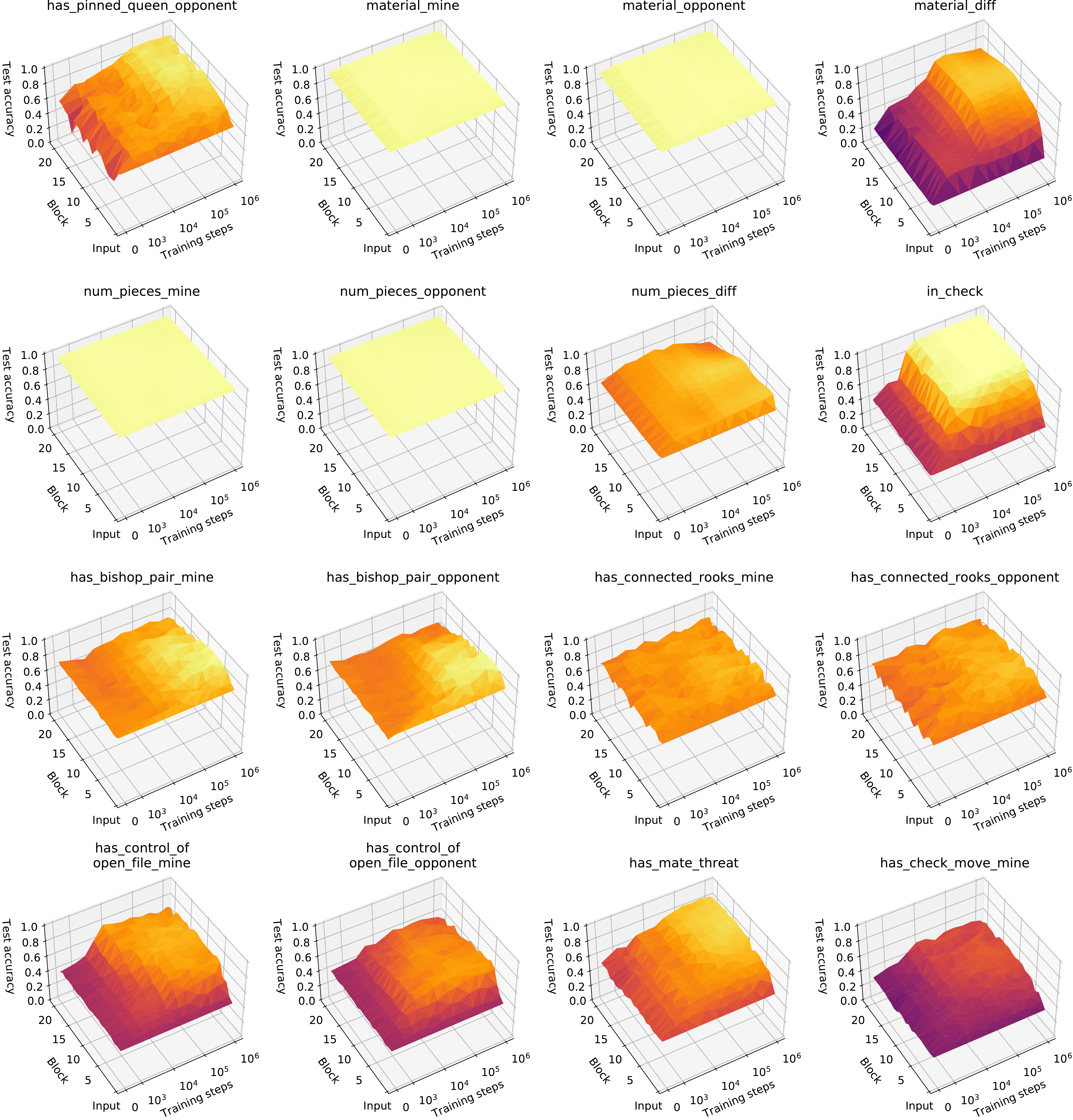}
    \caption{Regression results for custom concepts from Table~\ref{tab:concepts-custom}, excluding capture-related concepts, continued.}
    \label{fig:custom_concepts_nocapture_2_seed2}
\end{figure}

\begin{figure}
    \centering
    \includegraphics[width=\textwidth]{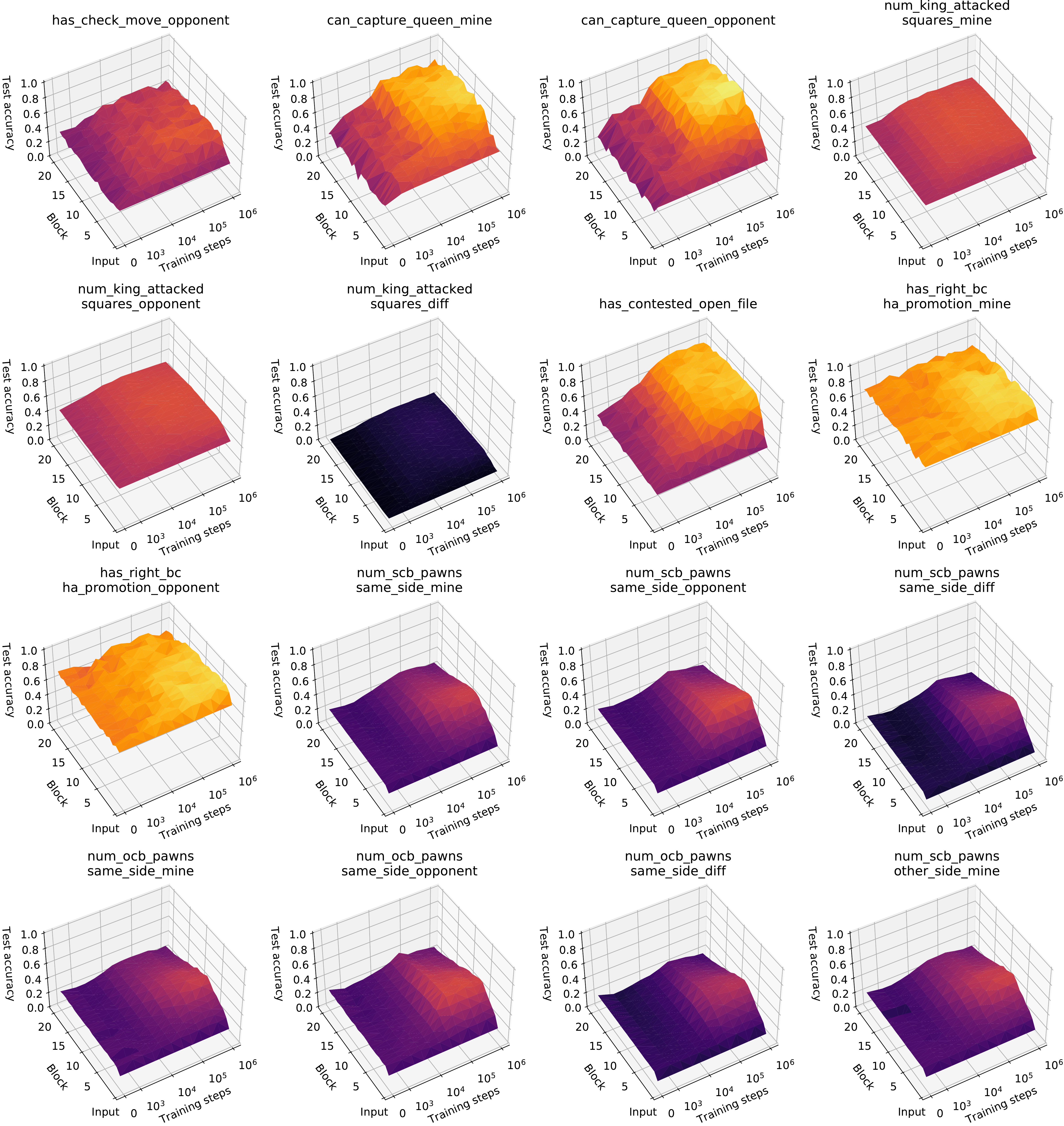}
    \caption{Regression results for custom concepts from Table~\ref{tab:concepts-custom}, excluding capture-related concepts, continued.}
    \label{fig:custom_concepts_nocapture_3_seed2}
\end{figure}

\begin{figure}
    \centering
    \includegraphics[width=\textwidth]{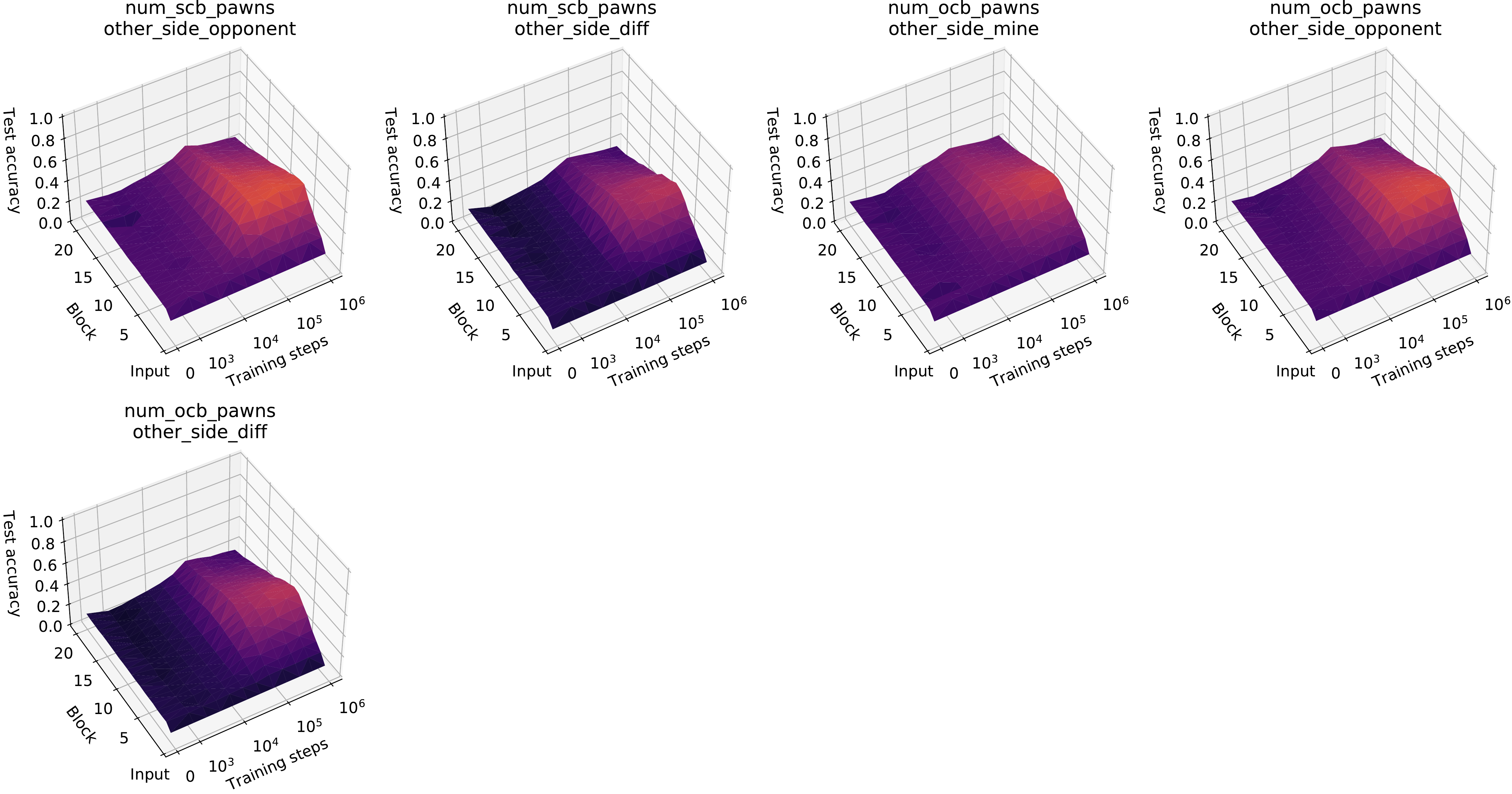}
    \caption{Regression results for custom concepts from Table~\ref{tab:concepts-custom}, excluding capture-related concepts, continued.}
    \label{fig:custom_concepts_nocapture_4_seed2}
\end{figure}

\begin{figure}
    \centering
    \includegraphics[width=\textwidth]{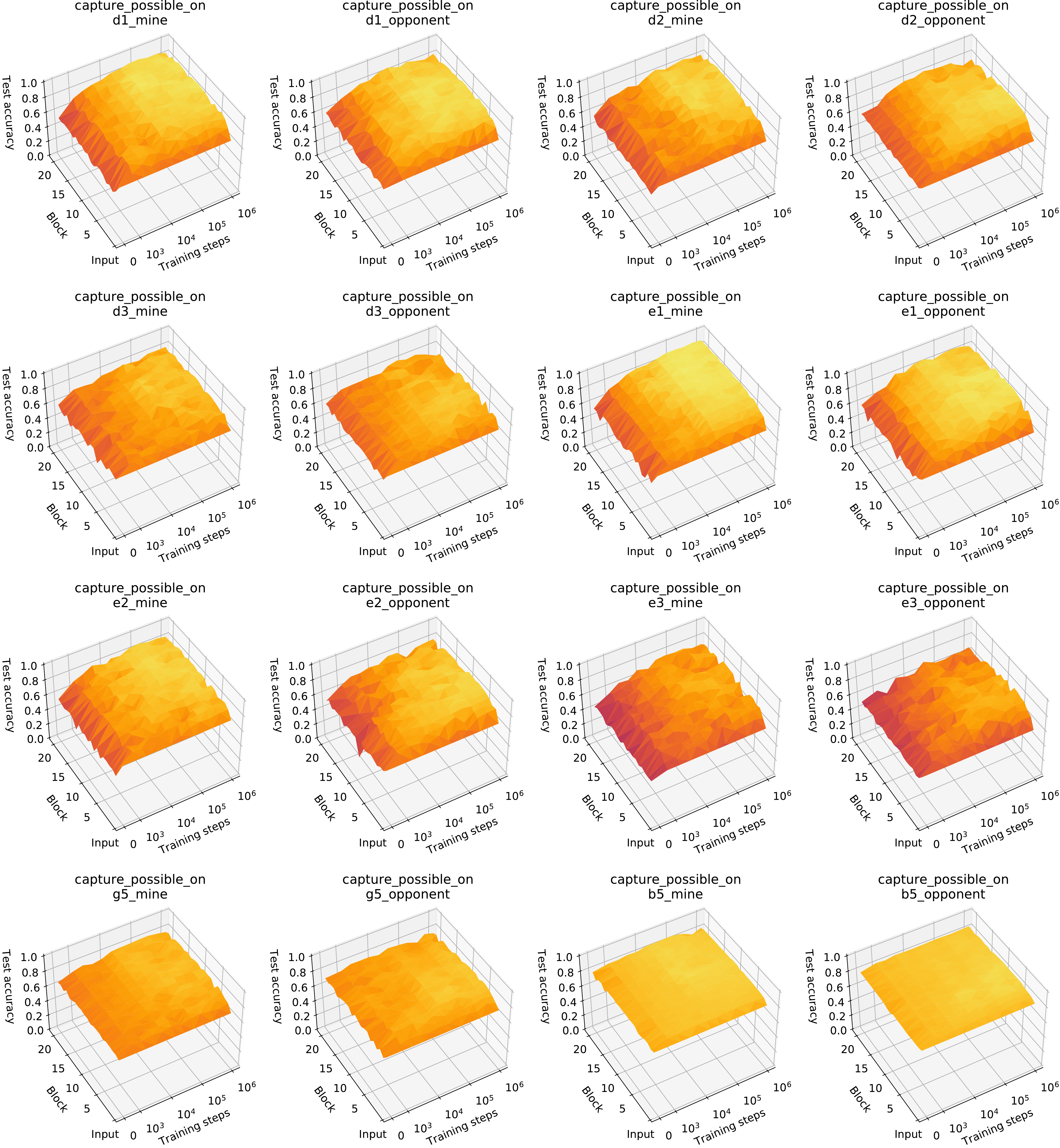}
    \caption{Regression results for custom concepts from Table~\ref{tab:concepts-custom} related to captures.}
    \label{fig:custom_concepts_capture_seed2}
\end{figure}

\begin{figure}
    \centering
    \includegraphics[width=\textwidth]{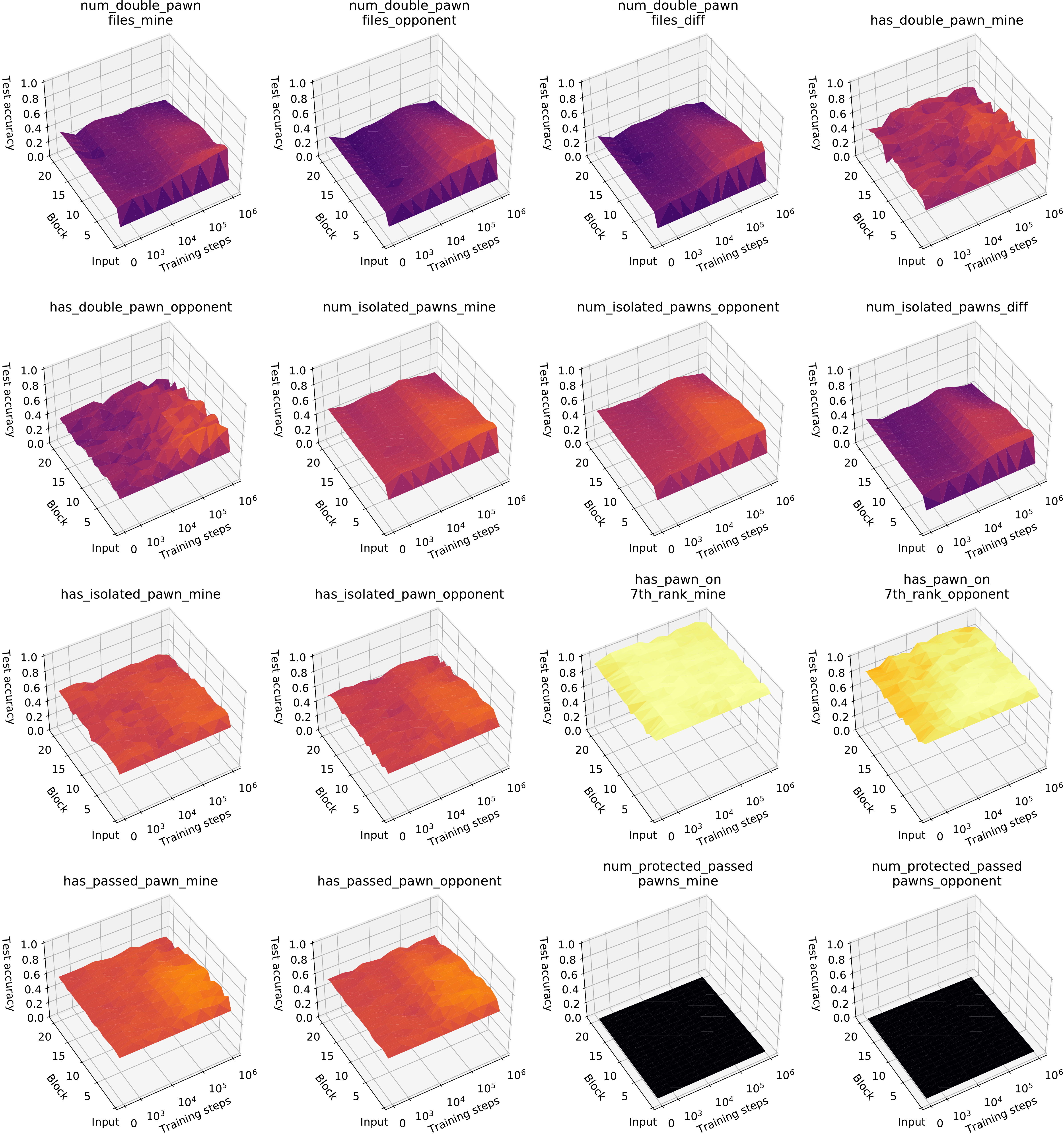}
    \caption{Regression results for custom pawn-related concepts from Table~\ref{tab:concepts-custom-pawns}.}
    \label{fig:custom_pawn_concepts_1_seed2}
\end{figure}

\begin{figure}
    \centering
    \includegraphics[width=\textwidth]{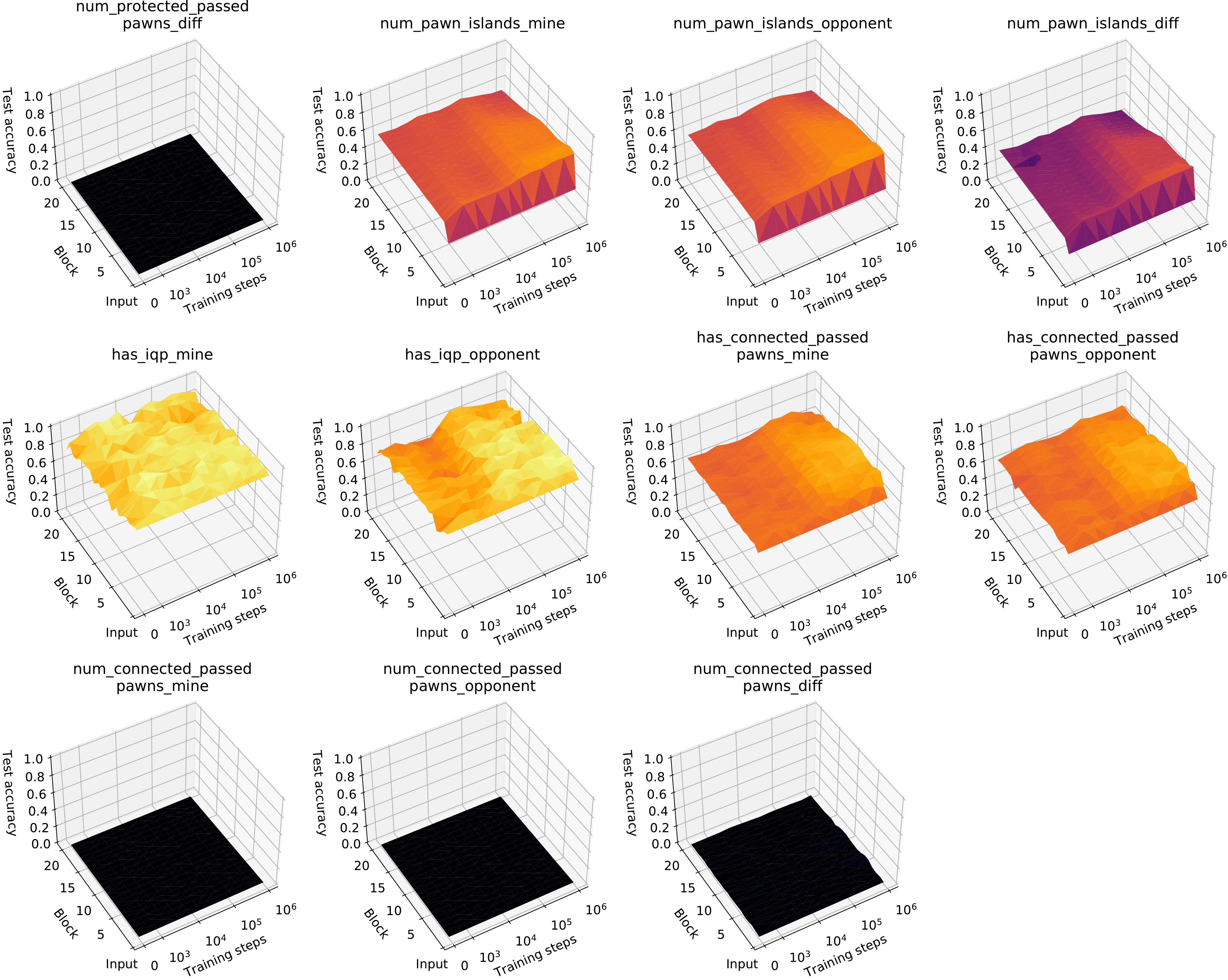}
    \caption{Regression results for custom pawn-related concepts from Table~\ref{tab:concepts-custom-pawns}, continued.}
    \label{fig:custom_pawn_concepts_2_seed2}
\end{figure}

\end{document}